\documentclass{article}

 \usepackage[preprint]{neurips_2026}
 \usepackage{algorithm}
\usepackage{algorithmic}
\usepackage{amsmath}
\usepackage{amssymb}
\usepackage{graphicx}
\usepackage{tablefootnote}
\usepackage{makecell}
\usepackage{threeparttable}
\graphicspath{ {./Figures/} }

\usepackage[utf8]{inputenc} 
\usepackage[T1]{fontenc}    
\usepackage{hyperref}       
\usepackage{url}            
\usepackage{booktabs}       
\usepackage{amsfonts}       
\usepackage{nicefrac}       
\usepackage{microtype}      
\usepackage{xcolor}         
\usepackage{tikz}
\title{Building The Ph(ysical)AI Layer Of Machine Intelligence}

\author{
  Ulbert J. Botero \\
  MIT Lincoln Laboratory\\
  \texttt{Joey.Botero@ll.mit.edu} \\
  \And
  Liam Smith \\
  MIT Lincoln Laboratory \\
  \texttt{Liam.Smih@ll.mit.edu} \\
  \AND
  Brooks Olney \\
  MIT Lincoln Laboratory \\
  \texttt{Brooks.Olney@ll.mit.edu} \\
  \And
  Pooya Khorrami \\
  MIT Lincoln Laboratory \\
  \texttt{Pooya.Khorrami@ll.mit.edu} \\
  \And
  Steven Kusiak \\
  MIT Lincoln Laboratory \\
  \texttt{Steven.Kusiak@ll.mit.edu} \\
  \And
  Watson Jia \\
  MIT Lincoln Laboratory \\
  \texttt{Watson.Jia@ll.mit.edu} \\
  \And
  Sage Trudeau \\
  MIT Lincoln Laboratory \\
  \texttt{Sage.Trudeau@ll.mit.edu} \\
  \And
  Daniel Capecci \\
  MIT Lincoln Laboratory \\
  \texttt{Daniel.Capecci@ll.mit.edu} \\
}

\begin{document}

\maketitle

\begin{abstract}
Foundation models achieve generalization through massive-scale training on diverse data, but have limitations with transfer to truly unseen domains without paired training data. We propose principle-driven foundation models that encode signal-theoretic principles (Fourier decomposition, energy conservation, symmetry) rather than learn untethered statistical correlations. We hypothesize that domains differ not in fundamental physics, but in learnable transformations in time, frequency, magnitude, or phase. Training exclusively on radio-frequency (RF) data with co-designed architecture and losses incorporating these principles, we achieve cross-modal transfer to audio, images, text, and video using only frozen representations learned from RF data, requiring no fine-tuning of the encoder on target domains. Our 1.99M parameter frozen encoder achieves \textbf{77.7}\% average accuracy (\textbf{91.9}\% top-3) across 15 diverse tasks via linear probing, with systematic variation: \textbf{84.5}\% on physically-grounded tasks (speaker recognition, seismology, RF fingerprinting) versus \textbf{70.0}\% on semantic tasks (music genre, language recognition). This reveals that principle-driven and scale-driven approaches offer complementary paths: physical principles enable efficient cross-modal transfer while naturally establishing the boundary between physical and semantic understanding.
\end{abstract}

\section{Introduction}

Foundation models achieve generalization through massive-scale self-supervised learning on diverse multi-modal data~\cite{radford2021learning, girdhar2023imagebind}. This scale-driven paradigm assumes that generalization emerges from learning correlations across domains via large amounts of paired training data.
While effective within their training distribution, these models face challenges with cross-modal transfer to domains absent from training data~\cite{fang2022data, miller2021accuracy, taori2020measuring}.

We propose a complementary approach: \textbf{principle-driven foundation models} that encode physical laws rather than learn statistical correlations. All structured data—temporal sequences, spatial images, graph networks—can be viewed as signals defined over some domain. Fourier's theorem generalizes beyond classical time-frequency analysis: graph Fourier transforms decompose signals on graphs~\cite{shuman2013emerging}, spherical harmonics on spheres, wavelets across scales. By grounding models in these general signal-theoretic principles, we hypthesize cross-modal generalization is possible. We display a promising step in this direction by achieving cross-modal transfer from a single training domain to diverse unseen domains without any encoder fine-tuning.

Our approach builds on two principles: (1) \textbf{Fourier decomposition}—signals decomposed into frequency components regardless of the domain, and (2) \textbf{symmetry learning}—representations should transform predictably under operations (translations, rotations, scalings). We hypothesize that \textbf{what distinguishes domains is not fundamentally different physics, but learnable transformations in time, frequency, magnitude, or phase}. If true, then learning transformation rules on one signal-rich domain should enable transfer to any domain respecting the same mathematical structure. 
We test this by training exclusively on radio-frequency (RF) data—contextually distant from our evaluation targets (images, speech, seismology, text), yet exceptionally rich in signal diversity—and then evaluating cross-modal transfer via linear probing on frozen representations.

The frozen encoder probing achieves 84.5\% average accuracy on physically-grounded tasks (instrument classification, speaker recognition, RF fingerprinting, modulation classification, seismic event detection) and 70.0\% on semantic tasks (music genre, language recognition, clothing classification) across time series (seismology, RF, speech), image 
(MNIST, FashionMNIST), and video. This systematic pattern supports our hypothesis: signal-theoretic learning captures physical structure but semantic content requires further abstraction. Critically, this is not a limitation but a feature—it reveals the boundary of our approach and suggests a hierarchical architecture for AI: physical foundation models provide the base layer upon which semantic reasoning can be built.

We emphasize that our goal differs from typical supervised learning benchmarks. We do not ask 'what is the best possible performance on task X?' but rather 'can training on RF alone enable meaningful transfer to task X?' 
Our contribution is demonstrating that physics-driven transfer is possible and competitive (within 3.2\% of CLIP on physical tasks with 76× fewer parameters), offering a resource-efficient path that complements scale-driven approaches.

Our contributions are: \begin{enumerate} \item  
We demonstrate that training on a single signal-rich domain (RF) enables zero-shot cross-modal generalization, achieving strong performance (91.9\% top-3 average accuracy) without encoder fine-tuning, establishing principle-driven design as a viable path alongside scale-driven approaches, particularly effective for physically-grounded tasks.\item We introduce PlanFormer with co-designed architecture (Parseval Focus, frequency-preserving pooling) and symmetry losses (IsoFICReg, LED) that embed signal-theoretic principles. \item We establish the boundary between physical and semantic understanding through systematic evaluation, showing that signal-theoretic learning captures physical structure (84.5\% top-1 average accuracy) but not semantic content (70.0\% top-1 average accuracy), revealing what can and cannot be learned from signal processing principles alone. \end{enumerate}

\section{Related Work}
\textbf{Multi-Modal Foundation Models:} Multi-modal foundation models~\cite{radford2021learning, girdhar2023imagebind} achieve generalization through large-scale training on diverse paired data, learning semantic correlations within their training distribution. In contrast, we train on a single domain (RF) and transfer to completely unseen domains, testing whether physical principles alone suffice for cross-modal generalization.

\textbf{Physics-Informed ML:} Physics-Informed Neural Networks~\cite{raissi2019physicsinformed} and Fourier Neural Operators~\cite{li2021fourier} incorporate physical constraints for domain-specific problems. We extend this to foundation models: encoding general signal-theoretic principles (Fourier decomposition, energy conservation, symmetry) that enable cross-modal transfer rather than domain-specific solutions.

\textbf{Symmetry and Equivariance:} Group equivariant networks~\cite{cohen2016group, weiler2019general} hard-code symmetries for specific domains. Others learn equivariant symmetries through targeted data augmentations and SSL criterions~\cite{yu2025selfsupervisedtransformationlearningequivariant, garrido2023self}. We build on the latter by learning symmetries through explicit equivariance objectives (LED) targeting fundamental symmetries anchored by fourier analysis.

\textbf{Self-Supervised Learning (SSL):} SSL methods learn representations through masked prediction~\cite{he2022masked}, contrastive learning~\cite{pmlr-v119-chen20j}, or redundancy reduction~\cite{bardes2021vicreg}. We extend VICReg with focal reweighting and coherent integration~\cite{richards2005fundamentals} for noise robustness, and complement invariance with explicit equivariance objectives (LED)—learning how representations should change predictably under transformations.

\textbf{Signal Processing in Deep Learning:} Recent work integrates signal processing: FNet~\cite{lee-thorp-etal-2022-fnet} uses Fourier transforms, time-frequency joint-embeddings~\cite{zhang2022self} processes dual domains. We enforce fundamental physical laws through co-designed architecture and losses: Parseval's theorem via consistency mechanisms, frequency-preserving pooling to avoid spectral bias~\cite{pmlr-v97-rahaman19a}, and explicit symmetry learning via equivariance objectives.

\section{Methodology}
\label{section:Methodology} Physical phenomena manifest as signals with general mathematical structure. A seismic wave, RF transmission, and visual scene differ semantically but share fundamental properties: frequency decomposition (Fourier), predictable transformations (symmetries), and causal structure. We hypothesize that domain differences arise from transformations in time, frequency, magnitude, or phase—transformations that manifest as learnable symmetries. A model that learns these transformation rules should generalize across domains sharing the same mathematical structure.

\textbf{RF Data as Training Domain:} We select RF data for its exceptional signal diversity: frequency content (kHz to GHz), temporal dynamics (modulations, transients, fading), and transformation types (Doppler shifts, multipath, channel effects). This rich diversity forces learning of general signal properties rather than domain-specific shortcuts. RF naturally requires joint time-frequency analysis, making it ideal for learning dual-domain representations. Critically, RF is distant from our evaluation domains (images, seismology, speech, histopathology), providing a stringent test of our hypothesis. Our RF fingerprinting datasets contain fine-grained hardware imperfections, forcing highly discriminative feature learning while retaining higher-order structure.

\subsection{Co-Designed Learning System}
\label{subsection:Co-DesignedLearningSystem}

Our approach prioritizes principled design, explicitly encoding established physical principles (Fourier decomposition, energy conservation, symmetry)—following a path similar to transformers~\cite{vaswani2023attentionneed} and CNNs~\cite{lecun-gradientbased-learning-applied-1998}, where architectural innovations preceded formal theoretical understanding.

Standard architectures lack inductive biases for symmetry losses to generate meaningful gradients. Learning frequency translation equivariance requires preserving high-frequency information—violated by standard pooling that creates spectral bias~\cite{pmlr-v97-rahaman19a}. 
As a result, the architecture and loss functions must be co-designed.

We design PlanFormer's components to provide computational substrate for our learning objectives: frequency-preserving pooling enables equivariance losses to receive gradients for high-frequency transformations; Parseval Focus enforces energy conservation (a physics-informed architectural constraint); the Noise Sink enables invariance learning in negative-SNR regimes. We combine complementary objectives—symmetry losses (IsoFICReg for invariance, LED for equivariance) that learn transformation rules, and reconstruction losses for instance-specific details—each enabled by specific architectural mechanisms.

\subsection{The Plan(cherel)Former Architecture}
\label{subsection:PlanFormer}

\subsubsection{Dual-Domain Architecture:}
\label{subsubsection:DualDomainArchitecture}
The encoder embeds Fourier's theorem through dual-domain processing operating simultaneously in time and frequency. Input signals are processed in parallel: the time-domain operates on the original signal and the frequency-domain on its Fourier transform. Unlike Fourier Neural Operators~\cite{li2021fourier} which apply asymmetric operations, PlanFormer applies symmetric sliding convolutions in both domains to learn complementary local features, relying on transformer mechanisms for global dependencies.

We represent signals as pairs of interleaved real-imaginary in-phase and quadrature components (IQ). Real-valued inputs (speech, images) undergo a Hilbert transformation~\cite{BendatPiersol2010Hilbert} to generate analytic representations, highlighting instantaneous frequency and phase changes. Within transformer blocks, we reshape tensors such that each token's embedding interleaves IQ pairs from consecutive sequence indices, ensuring a sequence length that aligns with physical complex-valued samples.

\textbf{Frequency-Preserving Pooling:} Standard pooling creates aliasing and discards high-frequency information, contributing to spectral bias~\cite{pmlr-v97-rahaman19a}. This is catastrophic for learning frequency translation equivariance—if high-frequency information is eliminated early, equivariance losses have no gradient signal for high-frequency transformations. We perform average pooling directly on complex-valued spectra, preserving the spectral envelope rather than truncating it. Both high- and low-frequency information is retained through down-sampling, with the spectrum compressed rather than band-limited.

\textbf{Convolutional Tokenization:} The encoder partitions inputs into non-overlapping windows, processing each in both domains to capture non-stationary spectral behavior. Parallel convolutional tokenizers extract local features, with cross-domain gated fusion at three points: after tokenization, after each transformer block, and after sequence pooling. Domain-specific aggregation respects physical meaning: time-domain outputs are concatenated to preserve temporal ordering; frequency-domain outputs are averaged to retain global periodic information. 

\subsubsection{Encoder: Energy Conservation using Multi-Head Parseval Focus}
\label{subsubsection:EnergyConservation}

Parseval's theorem establishes energy conservation between time and frequency domains~\cite{parseval1799}. To learn representations respecting this principle, we introduce Multi-Head Parseval Focus, which enforces cross-domain consistency through bidirectional attention and consistency regularization.

The mechanism operates as follows: time-domain tokens undergo FFT while frequency-domain tokens undergo IFFT, leveraging interleaved IQ representation for complex-valued transformations followed by domain-specific QKV projections. Within each domain, we compute the Scaled Covariance Focus:

\[Focus(Q, K, V) = softmax\left(\frac{Cov(Q, K)\cdot L_{sequence}}{\sqrt{d_k}\cdot K_{focus}}\right)V\]

where $Cov(Q,K) = (Q-\mu_Q)(K-\mu_K)^T$ captures functional relationships rather than instantaneous similarity. The focus factor $K_{focus} \in [1, L_{sequence}]$ is predicted from attention score statistics (mean token distinctiveness, proportion of highly distinctive tokens) via $f_\theta$, providing adaptive temperature $\tau = K_{focus}/L_{sequence}$. Low $K_{focus}$ sharpens attention when uncertain; high $K_{focus}$ maintains broad attention when confident (details in Appendix~\ref{appendix:DynamicFocusMechanism}).

Bidirectional cross-domain focus (time-as-query/frequency-as-key and vice-versa) captures complementary features. For physical consistency, both permutations should produce equivalent focus distributions. We compute the Jensen-Shannon Distance ($\text{JSD}_{time}(\mathbf{P}_{tf} || \tilde{\mathbf{P}}_{ft})$/$\text{JSD}_{freq}(\mathbf{P}_{ft} || \tilde{\mathbf{P}}_{tf})$, where $\tilde{\mathbf{P}}=\text{softmax}(\text{Score}^{\textbf{T}})$) between distributions, serving dual purposes: as a regularization loss (minimizing JSD encourages consistent cross-domain relationships) and as dynamic gating (down-weighting inconsistent relationships, amplifying consistent ones) alternative to softmax. In-domain and cross-domain focus representations are fused via gated mechanisms with diversity regularization ensuring complementary learning. All multi-head focus mechanisms employ head orthogonalization and regularization~\cite{bardes2021vicreg} to enforce specialization and reduce common-mode noise~\cite{ye2024differential}.

\begin{figure}
    \centering
    \includegraphics[width=\textwidth]{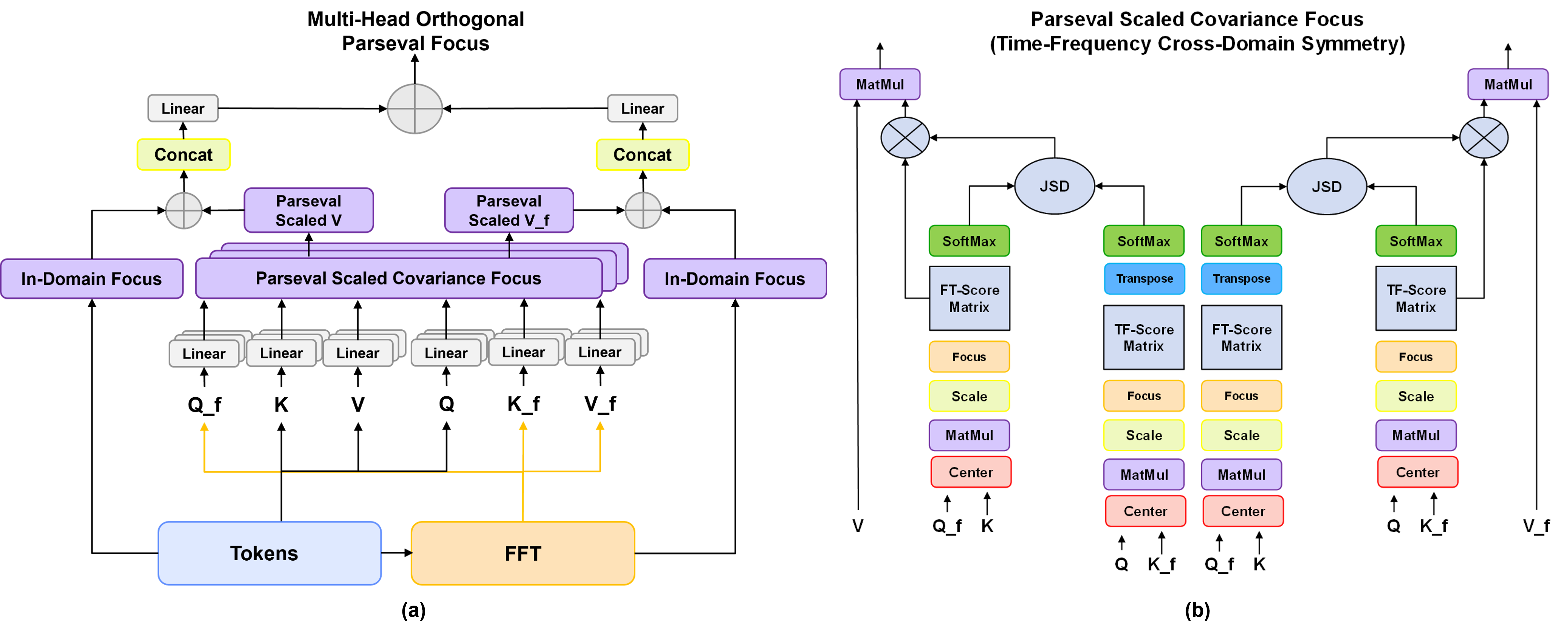}
    
\caption{\textbf{Multi-Head Parseval Focus architecture.} (a) Multi-Head Orthogonal Parseval Focus fuses in-domain (time-time, freq-freq) and cross-domain (time-freq, freq-time) focus via gated linear units for comprehensive signal analysis. (b) Parseval Scaled Covariance Focus: Cross-domain transformations (FFT/IFFT) with covariance-based attention. JSD between bidirectional distributions enforces Parseval consistent gating. Adaptive $K_{focus}$ enables dynamic sparsity control.}
\label{fig:parseval_focus}
\end{figure}

\subsubsection{Additional Encoder Mechanisms}
\label{subsubection:AdditionalEncoderBlocks}

\textbf{Causal Cross-Window Focus} computes focus between consecutive windows' convolutional representations to capture phase relationships spanning longer timescales.
\textbf{Noise Sink} explicitly estimates and removes noise during tokenization, regularized via Pearson correlation minimization and power matching constraints, enabling learning in negative-SNR regimes. \textbf{Attentional Pooling}~\cite{hassani2021escaping} produces fixed-size latent representations from variable length sequences, with time- and frequency-domain latents used independently or concatenated for downstream tasks.

\subsection{Decoder Architecture}
\label{subsection:Decoder}
The decoder employs a dual-domain UNet architecture~\cite{ronneberger2015u}, mirroring the encoder's time-frequency structure, and adding three key innovations enabling effective gradient generation: (1) \textbf{Instance-Specific Latent Conditioning} via attention-based FiLM~\cite{perez2017filmvisualreasoninggeneral} provides global context for instance-specific reconstruction guidance, (2) \textbf{Parseval Focus-Based Skip Connection Sinks} use Multi-Head Parseval Focus conditioned on the bottleneck latent to dynamically filter skip connections, creating adaptive filter banks that remove unwanted information (noise, interfering sources), and (3) \textbf{Frequency-Domain Upsampling} performs upsampling in frequency domain with spectral infilling from skip connections, preserving the spectral envelope and avoiding low-frequency bias. These mechanisms enable reconstruction and source separation losses to provide clean gradients, particularly for high-frequency features and in low-SNR regimes (details in Appendix~\ref{appendix:B_Decoder}).

\subsection{Training Methodology: Co-Designed Loss Functions}
\label{subsection:CoDesignedLosses}

\subsubsection{Isotropic Focal Invariance Covariance Regularization (IsoFICReg)}
\label{subsubsection:IsoFICReg} 
We extend VICReg~\cite{bardes2021vicreg} to enhance invariance learning under extreme noise while maintaining fine-grained discrimination. We leverage weak supervision from RF training data structure: multiple samples per emitter with unique hardware fingerprints.

\textbf{Latent Coherent Integration:} Within each batch, we identify all samples from the same emitter and compute pairwise N-choose-2 invariance losses. This is analogous to coherent integration in classical signal processing~\cite{richards2005fundamentals}—enforcing consistency across multiple noisy observations causes noise (uncorrelated) to become inconsistent while signal (correlated) becomes consistent, effectively improving SNR in latent space. 

\textbf{Focal Reweighting:} We Z-score standardize projections  
then apply regression-focused focal reweighting (extended from~\cite{lin2017focal}) to emphasize low-SNR samples over trivial high-SNR cases. We complement attraction-only invariance with repulsion loss for different emitter pairs, encouraging fine-grained discrimination.  
Focal reweighting is also applied to covariance regularization, emphasizing decorrelation along harder dimensions. The Z-score standardization paired with covariance based decorrelation encourages learning isotropic representations. Unlike the original VICReg, we omit the variance term due to Z-score standardization explicitly normalizing each dimension to unit variance.

\textbf{Dual-Domain Consistency:} We apply IsoFICReg to both time and frequency domain latents independently, and compute cross-domain invariance (time-frequency pairs from the same sample), enforcing Parseval-like consistency.

\subsubsection{Latent Equivariant Differences (LED)}
\label{subsubsection:LatentEquivarianceLearning}
While IsoFICReg learns invariances (what should remain the same), we explicitly learn equivariances (how representations should change predictably under transformations). This enables cross-modal generalization: by learning that physical transformations have predictable effects in latent space, the model learns transformation rules that are domain-invariant.

\begin{figure}
    \centering
    \includegraphics[width=\textwidth]{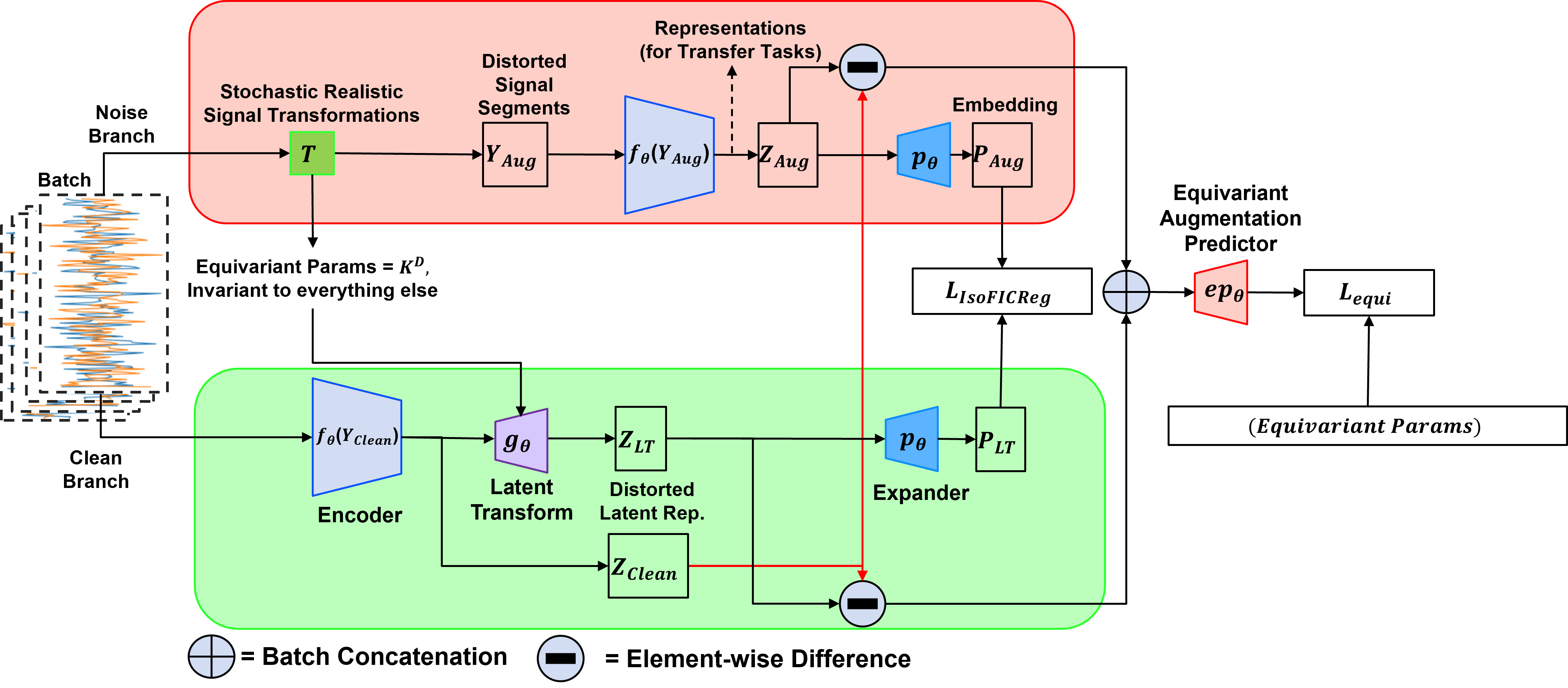}
    \caption{Time-Frequency Joint Embedding Predictive Architecture for Invariant and Equivariant Symmetry Learning. The output of the PlanFormer encoder produces time and frequency representations that both go through subsequent domain specific latent transformations and expander networks. Afterwards, the time-time, freq-freq, and time-freq latent losses are computed.}
    \label{fig:TimeFreqJEPASymmetry}
\end{figure}

We employ a two-branch scheme(Figure~\ref{fig:TimeFreqJEPASymmetry}): the clean branch processes unaugmented signals (producing $z_{clean}$), while the augmented branch processes signals with domain-invariant transformations—frequency shifts, time shifts, phase rotations, and AWGN (producing $z_{aug}$). For each transformation $T$ applied to the input (e.g., frequency shift by $\Delta$f), we:$(1)$ Apply $T$ to input $\rightarrow$ encode $\rightarrow$ $z_{aug}$ $(2)$ Encode clean input $\rightarrow$ apply latent transformer $T_{latent}$ $\rightarrow$ $z_{latent\_transform}$ and $(3)$ Enforce that $z_{aug}$ $\cong$ $z_{latent\_transform}$. The latent transformer applies transformations directly to encoded token representations using our Parseval Transformer architecture. Transformation parameters are linearly projected and prepended as conditioning tokens.

To verify that latent transformations correspond to input transformations, we  
compute differences over the pooled encoder outputs prior to IsoFICReg's loss over projected embeddings:

    \[\Delta_{latent} = z_{latent\_transform} - z_{clean}\]
    \[\Delta_{input} = z_{aug} - z_{clean}\]

These differences capture "how the representation changed" rather than "what the final representation is." We enforce dual regression:

    \[L_{Equi} = ||ep_{\theta}(\Delta_{latent})-\theta_{true}||^2 + ||ep_{\theta}(\Delta_{input})-\theta_{true}||^2 + ||\Delta_{latent}-\Delta_{input}||^2 \]
    
where $ep_{\theta}$ is an MLP projecting differences to match ground-truth transformation parameters $\theta_{true}$. The first two terms ensure each branch learns the correct transformation; the third ensures they learn the same transformation. We apply focal reweighting to emphasize difficult instances.

\subsubsection{Complementary Objectives}
\label{subsubsection:SourceSepReconstruction}
We complement global latent learning with instance-specific objectives. 
\textbf{Token-Level Source Separation} creates mixtures of two signals at controlled SINR (20-0dB) and separates them at the token level using a Parseval Transformer and ICA-inspired processing. Separated tokens are processed through the remaining encoder and decoder with reconstruction/invariance/equivariance targets as the original unmixed signals. \textbf{Dual-Domain Reconstruction} computes losses in both time and frequency domains with circular phase variants, using augmented signals (minus noise) as targets to provide explicit equivariance supervision. We add perceptual loss using an EMA-updated encoder. All reconstruction losses are focally reweighted to emphasize low-SNR instances.

\subsubsection{Overall Training Objectives}
\label{subsubsection:OverallTrainingCriterions}
Our complete training loss combines complementary objectives:

    \[L_{total} = L_{IsoFICReg} + L_{Equi} + 
              \gamma_{sep} \cdot L_{sep} + \gamma_{recon} \cdot L_{recon} + 
              L_{reg}\]

where $L_{reg}$ includes Parseval consistency, noise decorrelation, power matching, SNR regression, etc. (details in Appendix~\ref{appendix:C_AuxiliaryNetworks}). 

\section{Experiments}
\label{Section:Experiments}

\textbf{Training Data:} We train exclusively on RF fingerprinting datasets to test whether signal-theoretic principles learned on RF transfer to diverse domains. We combine five datasets (ORACLE~\cite{sankhe2019no}, POWDER~\cite{reus2020trust}, and 3 internally collected RF fingerprinting datasets we intend to release upon acceptance) providing 39 emitter classes with diversity across hardware families, modulation schemes (11 schemes), protocols (Wi-Fi, 4G, 5G), and channel conditions (over-the-air, temporal/spatial variation). This forces learning of general signal properties rather than dataset-specific patterns. Training details are in Appendix~\ref{appendix:D_TrainingDetails} and ~\ref{appendix:E_DatasetEvalDetails}.

\textbf{Evaluation Protocol:} We evaluate cross-modal transfer via linear probing on frozen encoder representations. The encoder remains completely frozen while lightweight linear classifiers are trained on labeled target-domain data via 5-fold cross-validation. This standard protocol~\cite{pmlr-v119-chen20j, he2020momentum, radford2021learning} measures representation quality through linear separability. We use 'frozen-encoder transfer' rather than 'zero-shot' to accurately reflect this methodology. Non-linear (RBF kernel) results are in Appendix~\ref{appendix:RBFResults}. We evaluate across time series, images, text, and video to test: (1) signal-theoretic principles enable cross-modal generalization, and (2) performance varies systematically with task type.

\textbf{Preprocessing:} Real-valued domains undergo Hilbert transformation~\cite{BendatPiersol2010Hilbert} to generate complex-valued (IQ) representations. \textbf{1D:} Resample to 5120 points minimum; text converts to byte streams. \textbf{2D:} Unwrap images in snake pattern (vertical/horizontal axes), process color channels independently, and concatenate. This tests whether signal-theoretic principles enable transfer despite discarding 2D spatial structure and no prior exposure to spatial data. \textbf{3D:} Unwrap frames as in 2D and concatenate successive frames. Full details in Appendix~\ref{appendix:E_DatasetEvalDetails}.

\begin{table*}[t]
\centering
\small
\begin{threeparttable}
\resizebox{\textwidth}{!}{\begin{tabular}{|c|c|c|c|c|c|c|c|c|c|}
\hline
\textbf{Domain} & \textbf{Task (Dataset)} & \textbf{Classes} & \textbf{Type} & \makecell{\textbf{Top-1(Lin)}\\\textbf{Random Init.}}  & \textbf{Top-1(Lin)} & \textbf{Top-3 (Lin)} & \makecell{\textbf{CLIP ViT-B/32~\cite{radford2021learning}}\\\textbf{Top-1(Lin)}} & \makecell{\textbf{DinoV3 ViT-S~\cite{Simeoni2025DINOv3}}\\\textbf{Top-1(Lin)}} & \textbf{Baseline}  \\
\hline
\multicolumn{10}{|c|}{\textit{1-D}} \\
\hline
RF* & Modulation Recognition & 8 & Physical & 57.8$\pm$0.3 & 95.5$\pm$0.8 & 99.2$\pm$0.2 & 97.0$\pm$0.3 & 92.2$\pm$0.1 & N/A \\
RF* & Fingerprinting (POWDER)~\cite{reus2020trust} & 4 & Physical & 67.9$\pm$2.7 & 87.6$\pm$0.3 & N/A & 55$\pm$0.2 & 66.3$\pm$0.3 & 93.0$^{SR}$ \\
Speech & BiLingual Speaker Recognition (TidyVoice)~\cite{farhadipour2026tidyvoice} & 50 & Physical & 60.5$\pm$ 5.4 & 90.1$\pm$2.6 & 97.7$\pm$1.0 & 97.8$\pm$0.4 & 96.8$\pm$0.8 & 93.8$\pm$1.9$^E$ \\
Speech & Language Recognition (TidyVoice)~\cite{farhadipour2026tidyvoice} & 29 & Semantic & 49.9$\pm$4.3 &69.8$\pm$4.1& 91.2$\pm$2.0 & 87.3$\pm$2.6 & 83.1$\pm$2.1 & 69.7$\pm$3.5$^E$ \\
Music & Instrument Family (TinySOL)~\cite{cella_2020_3685331} & 4 & Physical & 91.6$\pm$1.0 & 91.7$\pm$1.1 & N/A & 99.1$\pm$0.4 & 96.5$\pm$0.8 & 90.6$\pm$1.3$^E$ \\
Music & Individual Instrument (TinySOL)~\cite{cella_2020_3685331} & 14 & Phys+Sem & 74.8$\pm$1.5 & 80.5$\pm$1.5 & 95.5$\pm$0.9 &  97.5$\pm$0.5 & 89.9$\pm$1.1 & 90.0$\pm$1.3$^E$ \\
Music & Genre Recognition (GTZAN)~\cite{tzanetakis2002musical} & 10 & Semantic & 47.1$\pm$3.3 &64.1$\pm$3.2 & 89.4$\pm$2.1 & 82.1$\pm$2.4 & 70.2$\pm$1.2 & 62.5$\pm$2.3$^E$ \\
Seismology & Seismic Event Classification~\cite{SCSN1926, SCEDC2013} & 3 & Physical & 85.4$\pm$0.4 & 89.0$\pm$0.3 & N/A & 93.0$\pm$0.4 & 94.2$\pm$0.2 & 98.0$^{SRa}$ \\ 
Text & ArXiv Paper Sub-Discipline Classification~\cite{He2019LongDC} & 9 & Semantic & 24.2$\pm$0.7 & 36.9$\pm$0.4 & 71.8$\pm$0.3 & N/A & N/A & 80.47$^{SR}$ \\
Text & ArXiv Paper Field Classification~\cite{He2019LongDC} & 2 & Structural+Sem & 77.0$\pm$0.1 & 82.7$\pm$0.2 & N/A & N/A & N/A & N/A \\
\hline
\multicolumn{10}{|c|}{\textit{2-D}} \\
\hline

Vision & Digit (MNIST)~\cite{lecun2010mnist} & 10 & Phys+Sem & 91.8$\pm$0.1 & 79.2$\pm$0.1 & 94.5$\pm$0.05 & 98.6$\pm$0.1 & 93.4$\pm$0.1 & 99.2$^T$ \\
Vision & Clothing (FashionMNIST)~\cite{xiao2017/online} & 10 & Phys+Sem & 83.4$\pm$0.1 &77.0$\pm$0.01 & 96.1$\pm$0.04 & 90.4$\pm$0.1 & 83.3$\pm$0.1 & 94.9$^S$~\cite{fashionmnist_leaderboard} \\
Medical & Pathology (PathMNIST)~\cite{medmnistv2} & 9 & Physical & 67.1$\pm$0.2 &71.3$\pm$0.2 & 91.9$\pm$0.2 & 92.0$\pm$0.2 & 75.9$\pm$0.1 & 91.1$^{SR}$ \\
Vision & Fake Detection (CIFAKE)~\cite{bird2023cifakeimageclassificationexplainable} & 2 & Physical & 78.1$\pm$0.2 &82.0$\pm$0.1 & N/A & 94.73$\pm$0.1 & 77.1$\pm$0.3 & 92.98$^{SR}$ \\
\hline
\multicolumn{10}{|c|}{\textit{3-D}} \\
\hline
Video & Normal vs Abnormal Mitosis (Full Video)~\cite{delgado2024automatic,Delgado_Mitosis_Classification_2023} & 2 & Physical & 65.7$\pm$4.9 & 68.7$\pm$2.8 & N/A & 72.6$\pm$3.3 & 69.3$\pm$2.1 & 94$^{SR}$ \\

Video & Normal vs Abnormal Mitosis (1st Frame)~\cite{delgado2024automatic,Delgado_Mitosis_Classification_2023} & 2 & Physical & 67.4$\pm$1.7 & 64.1$\pm$1.1 & N/A & 65.8$\pm$1.7 & 63.5$\pm$2.1 & N/A \\
\hline

\multicolumn{4}{|r|}{\textbf{Average (All)}} & 68.2\% & 77.7\% & 91.9\% & 83.8\% & 83.7\% & 88.4\% \\
\multicolumn{4}{|r|}{\textbf{Average (1-D Only)}} & 63.6\% & 78.8\% & 90.8\% & 88.6\% & 86.2\% & 84.8\% \\
\multicolumn{4}{|r|}{\textbf{Average (Physical Only)}} & 71.8\% & 84.5\% & 96.3\% & 87.7\% & 83.6\% & 93.4\% \\
\multicolumn{4}{|r|}{\textbf{Average (Semantic/Mixed)}} & 64.0\% & 70.0\% &  89.8\% & 91.2\% & 84\% & 82.8\% \\
\multicolumn{4}{|r|}{\textbf{Number of Parameters(M = Million)}} & - & 1.99M &  - & 151.28M & 21M & N/A \\
\multicolumn{4}{|r|}{Floating Point Operations (\textbf{FLOPs})} & - & 93.6MFLOPs &  - & 14,780MFLOPs~\cite{ilharco_gabriel_2021_5143773} & 12,000MFLOPs$^{EF}$~\cite{Simeoni2025DINOv3} & N/A \\

\hline
\end{tabular}}
\end{threeparttable}
\caption{Values given as mean$\pm$std unless noted. Frozen-encoder transfer via linear probing (5-fold CV) on frozen representations. \textbf{Type}: Physical (structure-based), Semantic (meaning-based), Phys+Sem (mixed). *Within-domain generalization. \textbf{Baselines:} $^E$Expert features (MFCC) + linear SVM; $^S$Supervised neural network; $^{SR}$Source paper results; $^T$Typical literature performance. $^a$Meier+ 2019 (avg. 99.95\% local, 95.36\% teleseismic). Signal-theoretic learning captures physical structure effectively (84.5\%), with graceful degradation on semantic tasks (70.0\%). \textbf{CLIP/DinoV3 comparison:} PlanFormer achieves competitive performance on physical tasks (84.5\% vs 87.7\%/83.6\%) with 76$\times$/11$\times$ fewer parameters and 158$\times$/128$\times$ lower FLOPs; semantic gaps (70.0\% vs 91.2\%/84.0\%) validate physical/semantic boundary. $^{EF}$FLOPs for 256$\times$256, inputs used for evaluation 224$\times$224.}
\label{tab:frozenEncoder_results}
\end{table*}

\subsection{Main Results: Frozen-Encoder Cross-Modal Transfer}
Table~\ref{tab:frozenEncoder_results} presents results across 15 diverse tasks spanning time series, text, images, and video. Our learned representations, trained exclusively on RF data, are performant on physical tasks (84.5\% top-1, 96.3\% top-3 average) and match or exceed expert-crafted features (MFCC) on semantic tasks: 69.8\%(ours) vs. 69.7\%(MFCC) for language recognition and 64.1\%(ours) vs. 62.5\%(MFCC) for music genre classification. This demonstrates that signal-theoretic learning captures features comparable to decades of domain-specific signal processing research, while maintaining generality across diverse domains (images, text, video) where MFCC features do not apply.

\textbf{Efficiency vs Scale-Driven Models.} 
We compare to CLIP ViT-B/32~\cite{radford2021learning} (151M params) and DinoV3 ViT-S~\cite{Simeoni2025DINOv3} (21M params). PlanFormer achieves competitive performance on physical tasks (84.5\% vs 87.7\%/83.6\%) with 76$\times$/11$\times$ fewer parameters and 158$\times$/128$\times$ lower FLOPs. Semantic gaps (70.0\% vs 91.2\%/84.0\%) validate our physical/semantic boundary hypothesis. PlanFormer outperforms both on RF fingerprinting (87.6\% vs 55.0\%/66.3\%), demonstrating domain-aligned pretraining advantages. 

\begin{figure}
    \centering
    \includegraphics[width=\textwidth]{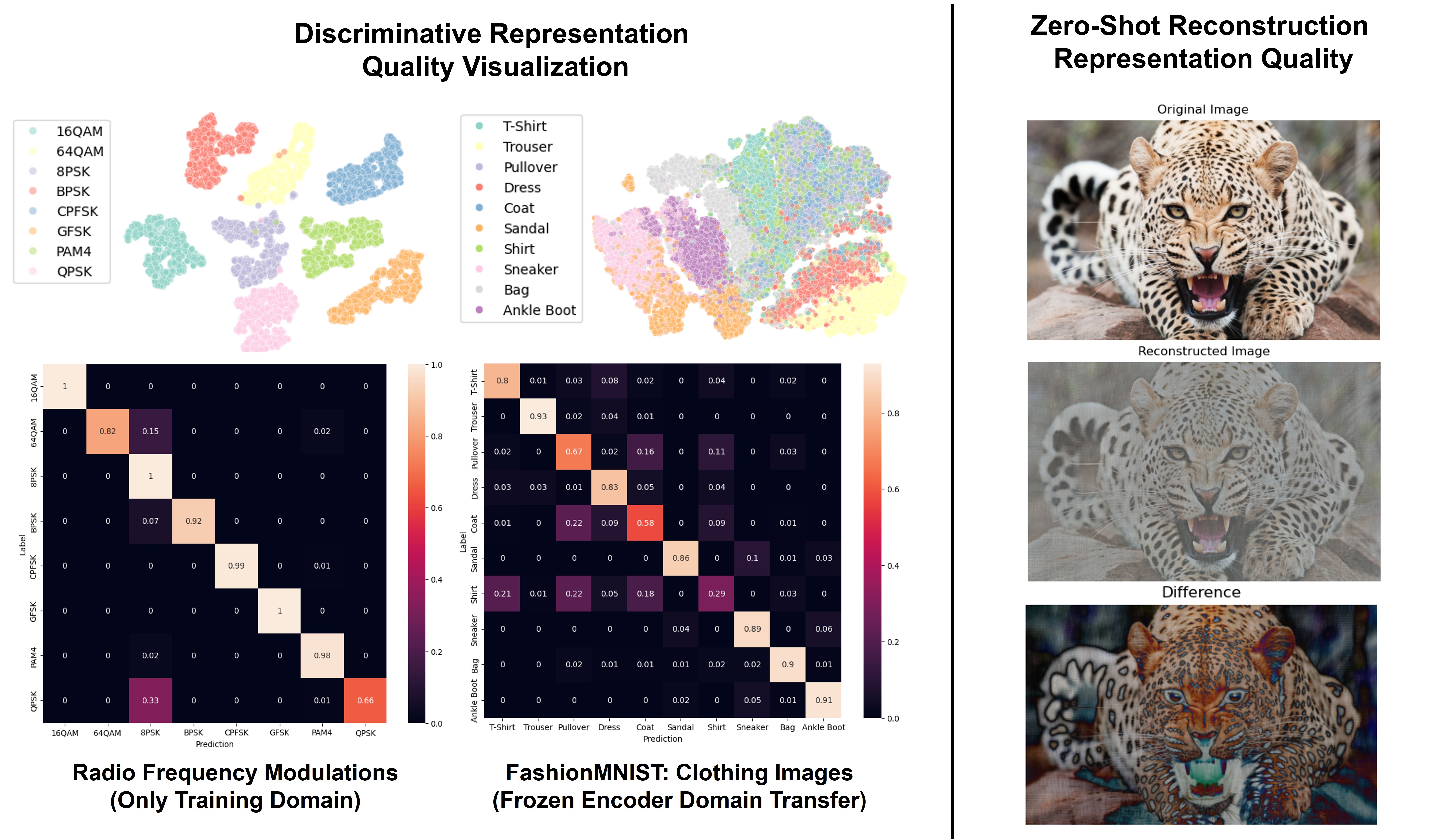}
\caption{\textbf{Learned representation quality.} RF modulation t-SNE and confusion matrix show tight clustering for physical task. FashionMNIST shows structural clustering with within-category confusion. Zero-shot image reconstruction:  
reconstructions (middle) preserve structural coherence and spatial relationships from originals (top)
.}
\label{fig:multimodal_viz}
\end{figure}

\textbf{Representation Quality.} High top-3 accuracy (91.9\% average) reveals top-1 errors are within-category confusions rather than fundamental misunderstanding. Zero-shot image reconstructions (Figure~\ref{fig:multimodal_viz}) provides visual confirmation: despite training only on RF with 1D unwrapping, reconstructions preserve structural coherence and spatial relationships. These complementary perspectives demonstrate learned representations capture transferable physical features rather than domain-specific patterns.

\subsection{Probing the Physical-Semantic Boundary}%
\label{subsubsection:PhysicalSemanticBoundary}
We design controlled experiments testing our hypothesis, each demonstrating strong performance on physical tasks and graceful degradation on semantic tasks with interpretable failure modes. This occurs despite zero prior exposure to these domains—only physically-anchored representations governing signal dynamics. Our strict frozen-encoder protocol—training exclusively on RF with zero target-domain exposure—provides unambiguous evidence of cross-modal transfer through signal-theoretic principles, contrasting with approaches requiring large-scale diverse datasets.

\textbf{Instrument Classification: Human Semantics on Physical Mechanisms.} Instruments are physically manufactured to produce intentional frequency responses, but humans introduce nonlinear effects through excitation—notes, tempo, pitch, playing style. Using TinySOL~\cite{cella_2020_3685331} (14 instruments, 4 families), we achieve 80.5\% instrument-level and 91.7\% family-level accuracy. Top-3 accuracy (95.5\%) compensates for human-level semantics. Confusion occurs within families or across instruments with overlapping physical features (timbre, pitch), demonstrating learned physical structure with predictable semantic degradation—without prior musical data exposure.

\textbf{Bilingual Speaker Recognition: Human as Physical Mechanism.} We view the human as the physical mechanism and language as the semantic layer. Using TidyVoice~\cite{farhadipour2026tidyvoice, farhadipour2026tidyvoice2026challengeevaluation} (bilingual speakers across 29 languages), we achieve 90.1\% speaker recognition (physical: voice characteristics) but 69.8\% language classification (semantic: linguistic content). Critically, speaker recognition accuracy is consistent across languages—the model recognizes the same speaker whether speaking English or Mandarin—demonstrating that it learned voice characteristics independent of linguistic content. Top-3 accuracy (97.7\% for speaker, 91.2\% for language) shows the underlying physical encoding captures fundamental information meaningfully.

\textbf{Semantic Stress Tests: Music Genre \& Academic Discipline Text Classification.} 
As the most stringent tests of our hypothesis, we evaluate two tasks driven primarily by semantics: music genre classification (GTZAN~\cite{tzanetakis2002musical,gtzan_kaggle}) and academic discipline classification from arXiv papers~\cite{He2019LongDC}. These tasks probe different aspects of the physical-semantic boundary:

\textbf{Music Genre:} Genres share substantial physical structure—instruments, tempo, harmonic patterns—but differ in cultural context, production style, and artistic intent. We achieve 64.1\% top-1 accuracy (vs. 10\% random chance) and 89.4\% top-3 accuracy. 

\textbf{Academic Discipline:} Text classification provides the most extreme test: (1) drastic signal origin shift (text byte streams vs. RF), (2) discrimination driven by semantic content rather than physical structure. We achieve 36.9\% top-1 accuracy (vs. 11.1\% random chance) and 71.8\% top-3 accuracy. Binary classification between structurally distinct fields (math vs. computer science) achieves 82.7\%, demonstrating that even raw text contains structural cues.
Together, these validate graceful degradation as tasks become more semantic.

\textbf{Image Classification: Physical Structure vs. Semantic Context.} We evaluate on MNIST~\cite{lecun2010mnist} (digits, structural with human handwriting variation) and FashionMNIST~\cite{xiao2017/online} (clothing, physical structure with semantic labels). Despite disadvantages—(1) unwrapping images loses spatial correlations, (2) no prior exposure to spatial data—we achieve 79.2\% on MNIST and 77.0\% on FashionMNIST. Top-3 accuracy (94.5\% MNIST, 96.1\% FashionMNIST) shows structural features capture foundational information.

\textbf{Additional Image Tasks.} PathMNIST (9-class histopathology): 71.3\% accuracy demonstrates that signal-theoretic principles provide useful features for medical imaging despite our 1D unwrapping approach being suboptimal. CIFAKE (binary real vs. synthetic): 82.0\% accuracy suggests our model captures physical signatures of natural images 
that synthetic images approximate imperfectly. 

 \textbf{Video Dynamics Classification: Spatio-Temporal Extension.} 
We evaluate 3D extensibility using Mitosis Classification~\cite{delgado2024automatic,Delgado_Mitosis_Classification_2023}, differentiating normal vs. abnormal cell division. This dataset was chosen to mitigate frame-level shortcuts—the task label is a temporal process, not a static property. To validate spatio-temporal learning, we compare full video vs. first-frame-only processing. The first frame alone should not discriminate between processes, requiring full video for accurate classification.

We achieve 68.7\% (full video) vs. 64.1\% (first frame, not used in average metrics), a 4.6\% difference validating that the model leverages spatio-temporal information for encoding physical dynamics.

\textbf{Cross-Protocol RF Fingerprinting: Within-Domain Generalization}

To validate that our model learned strong RF representations, we evaluate on POWDER transmissions~\cite{reus2020trust} with an unseen protocol (5G) and different channel conditions (different collection day). The 5G protocol differs from training protocols (Wi-Fi, 4G) in modulation schemes, frame structures, and frequency bands. Despite compounded distribution shifts (protocols \& channels), linear probing achieves 87.6\% accuracy. This complements cross-modal results: transfer from RF to images/text/video (cross-modal) and from Wi-Fi/4G to 5G (within-domain)—both require learning robust physical structure.

\section{Conclusion}
\label{Section:Conclusion}
Across this series of diverse experiments, we provide strong evidence for our hypothesis: signal-theoretic principles enable robust cross-modal transfer. We achieve competitive performance on physical tasks (84.5\%, within 3.2\% of CLIP's 87.7\%) despite 76× fewer parameters and RF-only pretraining, while graceful degradation on semantic tasks (70.0\% vs CLIP's 91.2\%) empirically validates the physical-semantic boundary. This establishes a clear division of labor: physics-driven foundation models capture causal structure and domain-invariant properties upon which semantic reasoning can be built. 

\section*{Acknowledgment}
DISTRIBUTION STATEMENT A. Approved for public release. Distribution is unlimited.
 
This material is based upon work supported by the Under Secretary of War for Research and Engineering under Air Force Contract No. FA8702-15-D-0001 or FA8702-25-D-B002. Any opinions, findings, conclusions or recommendations expressed in this material are those of the author(s) and do not necessarily reflect the views of the Under Secretary of War for Research and Engineering.
 
© 2026 Massachusetts Institute of Technology.
 
Delivered to the U.S. Government with Unlimited Rights, as defined in DFARS Part 252.227-7013 or 7014 (Feb 2014). Notwithstanding any copyright notice, U.S. Government rights in this work are defined by DFARS 252.227-7013 or DFARS 252.227-7014 as detailed above. Use of this work other than as specifically authorized by the U.S. Government may violate any copyrights that exist in this work.

\section{References}
\bibliographystyle{unsrt} 
\bibliography{references} 

\begin{thebibliography}{10}

\bibitem{radford2021learning}
Alec Radford, Jong~Wook Kim, Chris Hallacy, Aditya Ramesh, Gabriel Goh, Sandhini Agarwal, Girish Sastry, Amanda Askell, Pamela Mishkin, Jack Clark, et~al.
\newblock Learning transferable visual models from natural language supervision.
\newblock In {\em International conference on machine learning}, pages 8748--8763. PmLR, 2021.

\bibitem{girdhar2023imagebind}
Rohit Girdhar, Alaaeldin El-Nouby, Zhuang Liu, Mannat Singh, Kalyan~Vasudev Alwala, Armand Joulin, and Ishan Misra.
\newblock Imagebind: One embedding space to bind them all.
\newblock In {\em Proceedings of the IEEE/CVF conference on computer vision and pattern recognition}, pages 15180--15190, 2023.

\bibitem{fang2022data}
Alex Fang, Gabriel Ilharco, Mitchell Wortsman, Yuhao Wan, Vaishaal Shankar, Achal Dave, and Ludwig Schmidt.
\newblock Data determines distributional robustness in contrastive language image pre-training (clip).
\newblock In {\em International conference on machine learning}, pages 6216--6234. PMLR, 2022.

\bibitem{miller2021accuracy}
John~P Miller, Rohan Taori, Aditi Raghunathan, Shiori Sagawa, Pang~Wei Koh, Vaishaal Shankar, Percy Liang, Yair Carmon, and Ludwig Schmidt.
\newblock Accuracy on the line: on the strong correlation between out-of-distribution and in-distribution generalization.
\newblock In {\em International conference on machine learning}, pages 7721--7735. PMLR, 2021.

\bibitem{taori2020measuring}
Rohan Taori, Achal Dave, Vaishaal Shankar, Nicholas Carlini, Benjamin Recht, and Ludwig Schmidt.
\newblock Measuring robustness to natural distribution shifts in image classification.
\newblock {\em Advances in Neural Information Processing Systems}, 33:18583--18599, 2020.

\bibitem{shuman2013emerging}
David~I. Shuman, Sunil~K. Narang, Pascal Frossard, Antonio Ortega, and Pierre Vandergheynst.
\newblock The emerging field of signal processing on graphs: Extending high-dimensional data analysis to networks and other irregular domains.
\newblock {\em IEEE Signal Processing Magazine}, 30(3):83--98, 2013.

\bibitem{raissi2019physicsinformed}
Maziar Raissi, Paris Perdikaris, and George~E. Karniadakis.
\newblock Physics-informed neural networks: A deep learning framework for solving forward and inverse problems involving nonlinear partial differential equations.
\newblock {\em Journal of Computational Physics}, 378:686--707, 2019.

\bibitem{li2021fourier}
Zongyi Li, Nikola Kovachki, Kamyar Azizzadenesheli, Burigede Liu, Kaushik Bhattacharya, Andrew Stuart, and Anima Anandkumar.
\newblock Fourier neural operator for parametric partial differential equations.
\newblock In {\em International Conference on Learning Representations}, 2021.

\bibitem{cohen2016group}
Taco Cohen and Max Welling.
\newblock Group equivariant convolutional networks.
\newblock In {\em Proceedings of the 33rd International Conference on Machine Learning}, volume~48 of {\em Proceedings of Machine Learning Research}, pages 2990--2999. PMLR, 2016.

\bibitem{weiler2019general}
Maurice Weiler and Gabriele Cesa.
\newblock General e(2)-equivariant steerable cnns.
\newblock In {\em Advances in Neural Information Processing Systems 32 (NeurIPS 2019)}, pages 14357--14368, 2019.

\bibitem{yu2025selfsupervisedtransformationlearningequivariant}
Jaemyung Yu, Jaehyun Choi, Dong-Jae Lee, HyeongGwon Hong, and Junmo Kim.
\newblock Self-supervised transformation learning for equivariant representations, 2025.

\bibitem{garrido2023self}
Quentin Garrido, Laurent Najman, and Yann Lecun.
\newblock Self-supervised learning of split invariant equivariant representations.
\newblock {\em arXiv preprint arXiv:2302.10283}, 2023.

\bibitem{he2022masked}
Kaiming He, Xinlei Chen, Saining Xie, Yanghao Li, Piotr Doll{\'a}r, and Ross Girshick.
\newblock Masked autoencoders are scalable vision learners.
\newblock In {\em Proceedings of the IEEE/CVF conference on computer vision and pattern recognition}, pages 16000--16009, 2022.

\bibitem{pmlr-v119-chen20j}
Ting Chen, Simon Kornblith, Mohammad Norouzi, and Geoffrey Hinton.
\newblock A simple framework for contrastive learning of visual representations.
\newblock In {\em Proceedings of the 37th International Conference on Machine Learning}, volume 119 of {\em Proceedings of Machine Learning Research}, pages 1597--1607. PMLR, 2020.

\bibitem{bardes2021vicreg}
Adrien Bardes, Jean Ponce, and Yann LeCun.
\newblock Vicreg: Variance-invariance-covariance regularization for self-supervised learning.
\newblock {\em arXiv preprint arXiv:2105.04906}, 2021.

\bibitem{richards2005fundamentals}
Mark~A Richards.
\newblock {\em Fundamentals of Radar Signal Processing}.
\newblock McGraw-Hill, New York, 2005.

\bibitem{lee-thorp-etal-2022-fnet}
James Lee-Thorp, Joshua Ainslie, Ilya Eckstein, and Santiago Ontanon.
\newblock {FN}et: Mixing tokens with {F}ourier transforms.
\newblock In {\em Proceedings of the 2022 Conference of the North American Chapter of the Association for Computational Linguistics: Human Language Technologies}, pages 4296--4313, Seattle, United States, July 2022. Association for Computational Linguistics.

\bibitem{zhang2022self}
Xiang Zhang, Ziyuan Zhao, Theodoros Tsiligkaridis, and Marinka Zitnik.
\newblock Self-supervised contrastive pre-training for time series via time-frequency consistency.
\newblock {\em Advances in neural information processing systems}, 35:3988--4003, 2022.

\bibitem{pmlr-v97-rahaman19a}
Nasim Rahaman, Aristide Baratin, Devansh Arpit, Felix Draxler, Min Lin, Fred Hamprecht, Yoshua Bengio, and Aaron Courville.
\newblock On the spectral bias of neural networks.
\newblock In Kamalika Chaudhuri and Ruslan Salakhutdinov, editors, {\em Proceedings of the 36th International Conference on Machine Learning}, volume~97 of {\em Proceedings of Machine Learning Research}, pages 5301--5310. PMLR, 09--15 Jun 2019.

\bibitem{vaswani2023attentionneed}
Ashish Vaswani, Noam Shazeer, Niki Parmar, Jakob Uszkoreit, Llion Jones, Aidan~N. Gomez, Lukasz Kaiser, and Illia Polosukhin.
\newblock Attention is all you need, 2023.

\bibitem{lecun-gradientbased-learning-applied-1998}
Yann LeCun, L\'{e}on Bottou, Yoshua Bengio, and Patrick Haffner.
\newblock Gradient-based learning applied to document recognition.
\newblock {\em Proceedings of the IEEE}, 86(11):2278--2324, 1998.

\bibitem{BendatPiersol2010Hilbert}
Julius~S. Bendat and Allan~G. Piersol.
\newblock {\em The Hilbert Transform}, chapter~13, pages 473--503.
\newblock Wiley, Hoboken, NJ, USA, 2010.

\bibitem{parseval1799}
Marc-Antoine Parseval~des Ch{\^e}nes.
\newblock M{\'e}moire sur les s{\'e}ries et sur l'int{\'e}gration compl{\`e}te d'une {\'e}quation aux diff{\'e}rences partielles lin{\'e}aire du second ordre, {\`a} coefficients constants.
\newblock {\em Paris}, 1799.

\bibitem{ye2024differential}
Tianzhu Ye, Li~Dong, Yuqing Xia, Yutao Sun, Yi~Zhu, Gao Huang, and Furu Wei.
\newblock Differential transformer.
\newblock {\em arXiv preprint arXiv:2410.05258}, 2024.

\bibitem{hassani2021escaping}
Ali Hassani, Steven Walton, Nikhil Shah, Abulikemu Abuduweili, Jiachen Li, and Humphrey Shi.
\newblock Escaping the big data paradigm with compact transformers.
\newblock {\em arXiv preprint arXiv:2104.05704}, 2021.

\bibitem{ronneberger2015u}
Olaf Ronneberger, Philipp Fischer, and Thomas Brox.
\newblock U-net: Convolutional networks for biomedical image segmentation.
\newblock In {\em International Conference on Medical image computing and computer-assisted intervention}, pages 234--241. Springer, 2015.

\bibitem{perez2017filmvisualreasoninggeneral}
Ethan Perez, Florian Strub, Harm de~Vries, Vincent Dumoulin, and Aaron Courville.
\newblock Film: Visual reasoning with a general conditioning layer, 2017.

\bibitem{lin2017focal}
Tsung-Yi Lin, Priya Goyal, Ross Girshick, Kaiming He, and Piotr Doll{\'a}r.
\newblock Focal loss for dense object detection.
\newblock In {\em Proceedings of the IEEE international conference on computer vision}, pages 2980--2988, 2017.

\bibitem{sankhe2019no}
Kunal Sankhe, Mauro Belgiovine, Fan Zhou, Luca Angioloni, Frank Restuccia, Salvatore D’Oro, Tommaso Melodia, Stratis Ioannidis, and Kaushik Chowdhury.
\newblock No radio left behind: Radio fingerprinting through deep learning of physical-layer hardware impairments.
\newblock {\em IEEE Transactions on Cognitive Communications and Networking}, 6(1):165--178, 2019.

\bibitem{reus2020trust}
Guillem Reus-Muns, Dheryta Jaisinghani, Kunal Sankhe, and Kaushik~R Chowdhury.
\newblock Trust in 5g open rans through machine learning: Rf fingerprinting on the powder pawr platform.
\newblock In {\em GLOBECOM 2020-2020 IEEE Global Communications Conference}, pages 1--6. IEEE, 2020.

\bibitem{he2020momentum}
Kaiming He, Haoqi Fan, Yuxin Wu, Saining Xie, and Ross Girshick.
\newblock Momentum contrast for unsupervised visual representation learning.
\newblock In {\em Proceedings of the IEEE/CVF Conference on Computer Vision and Pattern Recognition (CVPR)}, pages 9729--9738, 2020.

\bibitem{Simeoni2025DINOv3}
Oriane Sim{\'e}oni, Huy~V Vo, Maximilian Seitzer, Federico Baldassarre, Maxime Oquab, Cijo Jose, Vasil Khalidov, Marc Szafraniec, Seungeun Yi, Micha{\"e}l Ramamonjisoa, et~al.
\newblock Dinov3.
\newblock {\em arXiv preprint arXiv:2508.10104}, 2025.

\bibitem{farhadipour2026tidyvoice}
Aref Farhadipour, Jan Marquenie, Srikanth Madikeri, and Eleanor Chodroff.
\newblock Tidyvoice: A curated multilingual dataset for speaker verification derived from common voice.
\newblock {\em arXiv preprint arXiv:2601.16358}, 2026.

\bibitem{cella_2020_3685331}
Carmine-Emanuele Cella, Daniele Ghisi, Vincent Lostanlen, Fabien Lévy, Joshua Fineberg, and Yan Maresz.
\newblock Tinysol: an audio dataset of isolated musical notes, January 2020.

\bibitem{tzanetakis2002musical}
George Tzanetakis and Perry Cook.
\newblock Musical genre classification of audio signals.
\newblock {\em IEEE Transactions on Audio and Speech Processing}, 10(5):293--302, 2002.

\bibitem{SCSN1926}
{California Institute of Technology (Caltech)} and {United States Geological Survey (USGS)}.
\newblock Southern california seismic network, 1926.

\bibitem{SCEDC2013}
{Southern California Earthquake Data Center}.
\newblock Southern california earthquake data center, 2013.

\bibitem{He2019LongDC}
Jun He, Liqun Wang, Liu Liu, Jiao Feng, and Hao Wu.
\newblock Long document classification from local word glimpses via recurrent attention learning.
\newblock {\em IEEE Access}, 7:40707--40718, 2019.

\bibitem{lecun2010mnist}
Yann LeCun, Corinna Cortes, Chris Burges, et~al.
\newblock Mnist handwritten digit database, 2010.

\bibitem{xiao2017/online}
Han Xiao, Kashif Rasul, and Roland Vollgraf.
\newblock Fashion-mnist: a novel image dataset for benchmarking machine learning algorithms, 2017.

\bibitem{fashionmnist_leaderboard}
Zalando Research.
\newblock Fashion-mnist github repository and benchmark leaderboard.
\newblock \url{https://github.com/zalandoresearch/fashion-mnist}, 2017.
\newblock Accessed: 2024-05-22.

\bibitem{medmnistv2}
Jiancheng Yang, Rui Shi, Donglai Wei, Zequan Liu, Lin Zhao, Bilian Ke, Hanspeter Pfister, and Bingbing Ni.
\newblock Medmnist v2-a large-scale lightweight benchmark for 2d and 3d biomedical image classification.
\newblock {\em Scientific Data}, 10(1):41, 2023.

\bibitem{bird2023cifakeimageclassificationexplainable}
Jordan~J. Bird and Ahmad Lotfi.
\newblock Cifake: Image classification and explainable identification of ai-generated synthetic images, 2023.

\bibitem{delgado2024automatic}
Pablo Delgado-Rodriguez, Rodrigo~Morales S{\'a}nchez, Elouan Roum{\'e}as-No{\"e}l, Fran{\c{c}}ois Paris, and Arrate Munoz-Barrutia.
\newblock Automatic classification of normal and abnormal cell division using deep learning.
\newblock {\em Scientific Reports}, 14(1):14241, 2024.

\bibitem{Delgado_Mitosis_Classification_2023}
Pablo Delgado, Nicolas Gaggion, Lucas Mansilla, Diego~H. Milone, and Enzo Ferrante.
\newblock {Mitosis Classification}, 6 2023.

\bibitem{ilharco_gabriel_2021_5143773}
Gabriel Ilharco, Mitchell Wortsman, Ross Wightman, Cade Gordon, Nicholas Carlini, Rohan Taori, Achal Dave, Vaishaal Shankar, Hongseok Namkoong, John Miller, Hannaneh Hajishirzi, Ali Farhadi, and Ludwig Schmidt.
\newblock Openclip, July 2021.

\bibitem{farhadipour2026tidyvoice2026challengeevaluation}
Aref Farhadipour, Jan Marquenie, Srikanth Madikeri, Teodora Vukovic, Volker Dellwo, Kathy Reid, Francis~M. Tyers, Ingo Siegert, and Eleanor Chodroff.
\newblock Tidyvoice 2026 challenge evaluation plan, 2026.

\bibitem{gtzan_kaggle}
Andrada Olteanu.
\newblock Gtzan dataset - music genre classification, 2020.

\bibitem{krull2019noise2void}
Alexander Krull, Tim-Oliver Buchholz, and Florian Jug.
\newblock Noise2void-learning denoising from single noisy images.
\newblock In {\em Proceedings of the IEEE/CVF conference on computer vision and pattern recognition}, pages 2129--2137, 2019.

\bibitem{Shannon1949}
Claude~E. Shannon.
\newblock Communication in the presence of noise.
\newblock {\em Proceedings of the {IRE}}, 37(1):10--21, 1949.

\bibitem{zhang2019making}
Richard Zhang.
\newblock Making convolutional networks shift-invariant again.
\newblock In {\em International conference on machine learning}, pages 7324--7334. PMLR, 2019.

\bibitem{gu2024efficient}
Hanqing Gu, Lisheng Su, Yuxia Wang, Weifeng Zhang, and Chuan Ran.
\newblock Efficient channel-temporal attention for boosting rf fingerprinting.
\newblock {\em IEEE Open Journal of Signal Processing}, 5:478--492, 2024.

\bibitem{zhu2025transformers}
Jiachen Zhu, Xinlei Chen, Kaiming He, Yann LeCun, and Zhuang Liu.
\newblock Transformers without normalization.
\newblock In {\em Proceedings of the computer vision and pattern recognition conference}, pages 14901--14911, 2025.

\bibitem{kingma2015adam}
Diederik~P Kingma and Jimmy Ba.
\newblock Adam: {A} method for stochastic optimization.
\newblock In {\em International Conference on Learning Representations ({ICLR})}, 2015.

\bibitem{dao2022flashattention}
Tri Dao, Daniel~Y. Fu, Stefano Ermon, Atri Rudra, and Christopher R{\'e}.
\newblock Flash{A}ttention: {F}ast and {M}emory-{E}fficient {E}xact {A}ttention with {IO}-{A}wareness.
\newblock {\em arXiv preprint arXiv:2205.14135}, 2022.

\bibitem{scikit-learn}
F.~Pedregosa and et~al.
\newblock Scikit-learn: Machine learning in {P}ython.
\newblock {\em Journal of Machine Learning Research}, 12:2825--2830, 2011.

\bibitem{platt1999probabilistic}
John Platt.
\newblock Probabilistic outputs for support vector machines and comparisons to regularized likelihood methods.
\newblock In {\em Advances in large margin classifiers}, volume~10, pages 61--74. Cambridge, MA, 1999.

\bibitem{lyons2020python_speech_features}
James Lyons et~al.
\newblock jameslyons/python\_speech\_features: release v0.6.1, January 2020.

\bibitem{ericsson2021well}
Linus Ericsson, Henry Gouk, and Timothy~M Hospedales.
\newblock How well do self-supervised models transfer?
\newblock In {\em Proceedings of the IEEE/CVF conference on computer vision and pattern recognition}, pages 5414--5423, 2021.

\end{thebibliography}

\appendix

\section{Appendix A: Encoder Architecture}
\label{appendix:A_Encoder_Architecture}

\subsection{Planformer Encoder Architecture}
\label{appendix:PlanformerEncoder}

\textbf{Overview}: The PlanFormer encoder is designed to embed signal-theoretic principles through dual-domain processing that operates simultaneously in time and frequency domains. This appendix provides complete architectural specifications for reproducibility.

\subsection{High-Level Architecture}
\label{appendix:HighLevelEncoder}
The encoder consists of the following stages:
\begin{enumerate}
    \item \textbf{Complex-valued preprocessing} - Convert input to IQ representation
    \item \textbf{Dual-domain tokenization} - Parallel processing in time and frequency
    \item \textbf{Parseval Transformer blocks} - Physics-informed attention mechanisms that we term "Focus" based on our dynamic attention sharpening machinery
    \item \textbf{Attentional pooling} - Fixed-size latent representations per domain via pooling over variable token sequence lengths
\end{enumerate}

\textbf{Key architectural parameters:}
\begin{itemize}
    \item \textbf{Input length:} 5120 total samples (minimum), 1024 samples per window, 5 windows total
    \item \textbf{Embedding dimension:} 128 after IQ interleaving for complex tokenization
    \item \textbf{Number of transformer blocks:} 1
    \item \textbf{Number of attention heads:} 8
    \item \textbf{Time-domain latent size:} 128
    \item \textbf{Frequency-domain latent size:} 128
\end{itemize}

\subsection{Complex-Valued Preprocessing}
\label{appendix:ComplexValuedPreprocessing}

\subsubsection{Real-Valued Signal Conversion}
\label{appendix:Real2Complex}

For real-valued input signals (speech, images, seismology), we generate analytic representations via Hilbert transform:

\[x_{analytic}(t) = x(t) + j·H[x(t)]\]

where H[·] denotes the Hilbert transform. This enables seamless extension of our architecture originally developed for complex valued domains to real valued domains. Further exposing the real valued signals' instantaneous frequency and phase information.

\subsubsection{IQ Representation}
\label{appendix:IQRepresentation}
Signals are represented as interleaved real-imaginary (IQ) pairs:

\[x_{IQ} = [Re(x[0]), Im(x[0]), Re(x[1]), Im(x[1]), ..., Re(x[N-1]), Im(x[N-1])]\]

This preserves phase-magnitude relationships while leveraging optimized real-valued computation.

Rationale: Complex-valued operations are essential for frequency-domain processing and phase-aware learning, but most deep learning frameworks optimize for real-valued tensors. The standard approach treats real and imaginary components 
as separate channels in the input space for convolutional processing. However, this construction overlooks a critical issue: while I and Q components retain their physical relationship in the input space, this relationship breaks in deeper 
layers where the channel dimension represents learned feature maps rather than physical components.

Consider the standard approach: in deeper layers, tensors have shape [C, N] (channels × sequence length) or [C, H, W] for images, where C represents learned features, not physical I/Q pairs. Operations along the channel dimension (e.g., 
batch normalization, channel-wise attention) treat all channels equivalently, destroying the I-Q coupling that is fundamental to complex-valued signals.

We instead propose to interleave I and Q components along the sequence dimension: $[I_0, Q_0, I_1, Q_1, ..., I_{N-1}, Q_{N-1}]$. This ensures that as we progress deeper into the network, the physical relationship between components is always present and maintained—adjacent elements in the sequence dimension are always an I-Q pair, regardless of how many feature channels exist. Operations that respect local structure (e.g., convolutions with small kernels, local attention) naturally preserve I-Q coupling.

Furthermore, this construction enables seamless transitions between real and complex-valued representations via de-interleaving and re-interleaving at strategic points in the processing chain. For example, during frequency-domain pooling, we de-interleave to obtain explicit complex values, apply FFT, perform pooling on the complex-valued spectrum (respecting its Hermitian symmetry for real signals), and re-interleave for subsequent processing. This is essential for learning equivariant symmetries to transformations like frequency translation, where respecting the asymmetric structure of complex-valued spectra is critical.

\textbf{Computational Benefits:} This approach also reduces sequence length in token space by half (N/2 complex samples instead of N real samples), reducing attention's quadratic cost from O(N²) to O((N/2)²) = O(N²/4), while doubling embedding dimension. For typical sequence lengths (N > 1000) and embedding dimensions (d < 512), this trade-off is favorable.

\subsection{Convolutional Tokenization}
\label{appendix:ConvTokenization}

Our encoder architecture takes large inspiration from \cite{hassani2021escaping} with the utilization of a convolutional tokenizer to learn tokens that are then utilized within a transformer block at a compressed sequence length, followed by a linear attention pooling over the sequence for the final latent. We build on top of this starting point and combine it with additional inspiration from \cite{zhang2022self} to unify time-frequency symmetry to produce our final architecture. While these works serve as inspirations we have advanced them considerably in the following ways.

\subsubsection{Windowing and Dual-Domain Processing}
\label{appendix:WindowingDualDomainTokenization}

Input signals are partitioned into non-overlapping windows of size W:

Windows: $[x[0:W], x[W:2W], ..., x[(N_w-1)W:N_w·W]]$
where $N_w = [N/W]$ is the number of windows.

Each window is processed in both time and frequency domains:

Time domain: Direct processing of windowed signal
Frequency domain: FFT of windowed signal
Window size: [1024 complex samples (2048 interleaved real valued samples)]

\textbf{Rationale:} Time-varying phenomena are ubiquitous in time series analysis—whether intentional (frequency hopping, chirps, transient events) or unintentional (time-varying noise, non-stationary channels). Effective representation learning requires balancing local and global processing to capture both instantaneous spectral characteristics and long-range dependencies.

\textbf{Why Windowed Processing?} Computing the FFT over the entire sequence produces a global frequency representation that averages spectral content across time, obscuring time-varying behavior. For example, a signal that shifts from 1 kHz to 2 kHz midway through would show energy at both frequencies in the global FFT, but the \textit{when} of this transition is lost. Windowed processing—
computing local spectra over short time segments—preserves temporal localization of spectral features, enabling the network to learn representations that respect time-varying dynamics.

\textbf{Computational and Architectural Benefits:} Windowed processing provides 
several advantages:

\begin{enumerate}
    \item \textbf{Translational equivariance:} Convolution's translational equivariance property minimizes the need for overlapping windows. A feature learned at position $t$ in one window transfers to position $t$ in other windows, reducing redundancy and parameter requirements.
    
    \item \textbf{Computational efficiency:} Attention mechanisms scale quadratically with sequence length ($O(N^2)$). By processing windows of length $W$, attention cost per window is $O(W^2)$, and total cost across $N_w = N/W$ windows is $O(N_w \cdot W^2) = O(N \cdot W)$—linear in sequence length. For $W \ll N$, this provides substantial savings.
    
    \item \textbf{Reconstruction efficiency:} In the decoder, operating on windowed representations enables localized reconstruction with manageable memory footprints, particularly important for long sequences ($N > 10,000$).
\end{enumerate}

\textbf{Addressing Long-Range Dependencies:} The primary drawback of windowed processing is that convolutions operate in isolation within each window, potentially missing long-range dependencies such as phase coherence across extended temporal spans (e.g., carrier phase in RF signals, pitch contours in speech). We address this through two complementary mechanisms:

\begin{enumerate}
    \item \textbf{Causal Cross-Window Focus} (Section~\ref{appendix:CausalCrossWindowFocus}): Explicit attention between consecutive windows' tokenized representations models inter-window dependencies. Domain-specific positional encodings enable the network to learn causal phase relationships (time domain) and time-varying spectral evolution (frequency domain).
    
    \item \textbf{Parseval Transformer} (Section~\ref{appendix:ParsevalTransformer}): After tokenization, all window tokens are processed jointly through transformer blocks, enabling global attention across the entire sequence. This captures long-range dependencies that span multiple windows while maintaining the benefits of localized spectral analysis.
\end{enumerate}

In summary, windowed processing provides the best of both worlds: local spectral analysis for time-varying phenomena and global dependency modeling through cross-window attention and transformers. This design is essential for learning representations that respect both instantaneous signal characteristics and extended temporal structure.

\textbf{Relation to Short-Time Fourier Transform (STFT):} Our windowed processing is conceptually similar to STFT, which computes local spectra over sliding windows to create time-frequency representations (spectrograms). However, unlike STFT which uses fixed, hand-crafted windows (e.g., Hamming, Hann), our approach learns optimal window-level representations through convolutional tokenization and transformer processing. This enables the network to adapt to signal-specific time-frequency characteristics rather than relying on predetermined window shapes.

\subsubsection{Tokenization Components}
\label{appendix:TokenizationComponents}

\begin{figure}
    \centering
    \includegraphics[width=\textwidth, angle = -90]{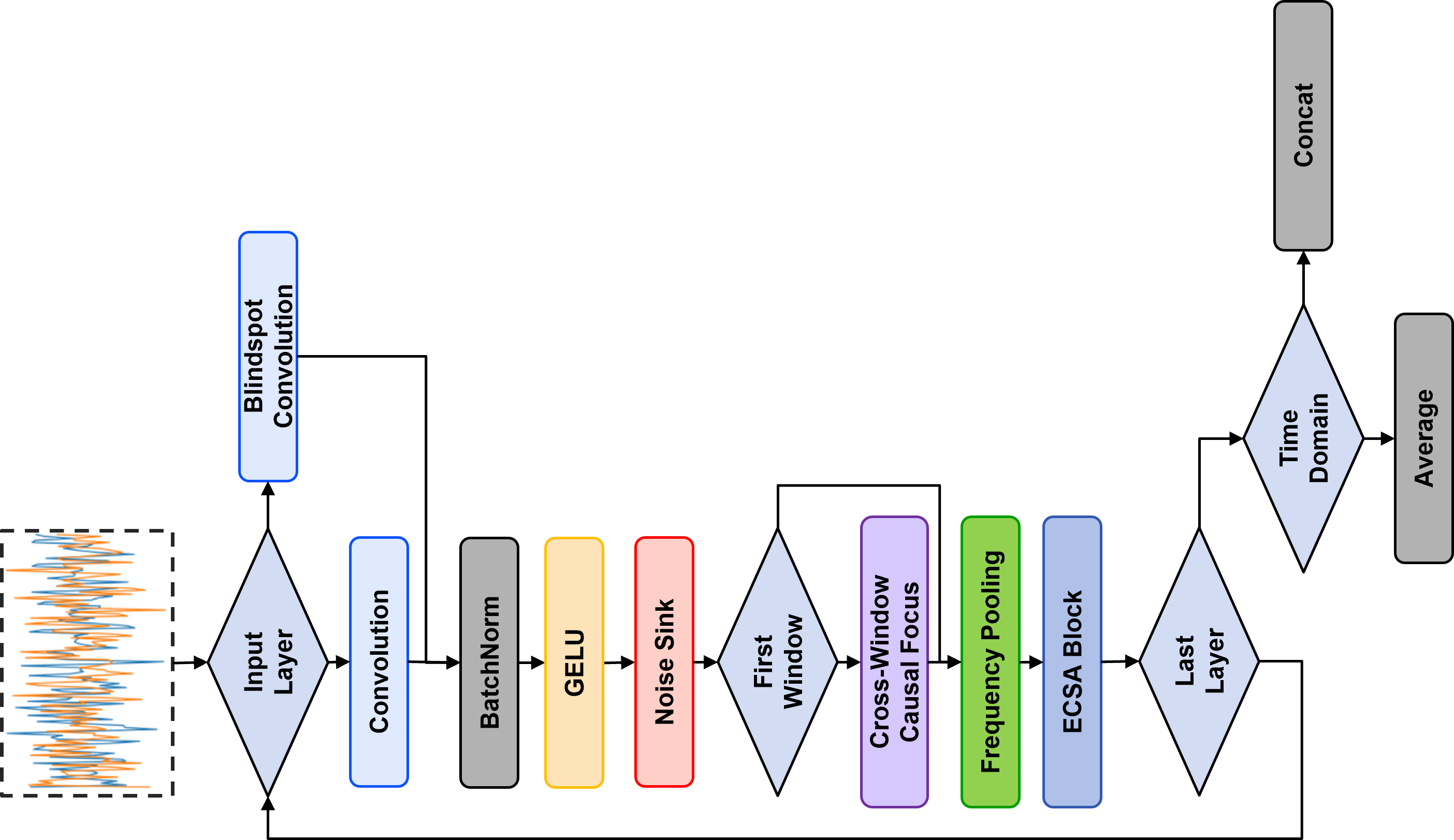}
    \caption{Full Convolutional Tokenizer Block Diagram}
    \label{fig:AppendixConvTokenizer}
\end{figure}

The full convolutional tokenizer and its constituent blocks are represented in Figure~\ref{fig:AppendixConvTokenizer}. We will specifically describe the unique contributions of our work below:

\paragraph{Blindspot Convolution}

The first convolutional layer uses a blindspot kernel~\cite{krull2019noise2void} 
(center element zeroed) for processing additive noise in the input space:

\begin{equation}
K_{blindspot}[i] = \begin{cases}
K[i] & \text{if } i \neq \text{center} \\
0 & \text{if } i = \text{center}
\end{cases}
\end{equation}

\textbf{Input Layer Blindspot Convolution Parameters:}
\begin{itemize}
    \item \textbf{Kernel size:} 5
    \item \textbf{Stride:} 1
    \item \textbf{Output channels:} 16
\end{itemize}
  
\textbf{Rationale:} Noise is ubiquitous across physical signals, but additive 
noise sources are typically locally uncorrelated relative to the structured signal 
of interest. We leverage this inductive bias through a blindspot kernel: the 
zeroed center element forces the network to infer signal structure from context 
rather than learning identity mappings, creating an implicit denoising prior. 
This approach, originally developed for self-supervised denoising in images~\cite{krull2019noise2void}, 
is extensible to all signal types.

\textbf{Addressing the Capacity Trade-off:} A key drawback of blindspot kernels 
is reduced representational capacity compared to full kernels. We address this 
through our dual-branch training architecture (Section~\ref{subsubsection:IsoFICReg}): 
the noise-augmented branch uses blindspot convolutions, the clean branch uses 
full-capacity kernels, and invariance losses encourage the blindspot branch to 
approximate full-kernel representations. This provides both denoising priors and 
full representational capacity.

\begin{figure}
    \centering
    \includegraphics[width=0.5\linewidth]{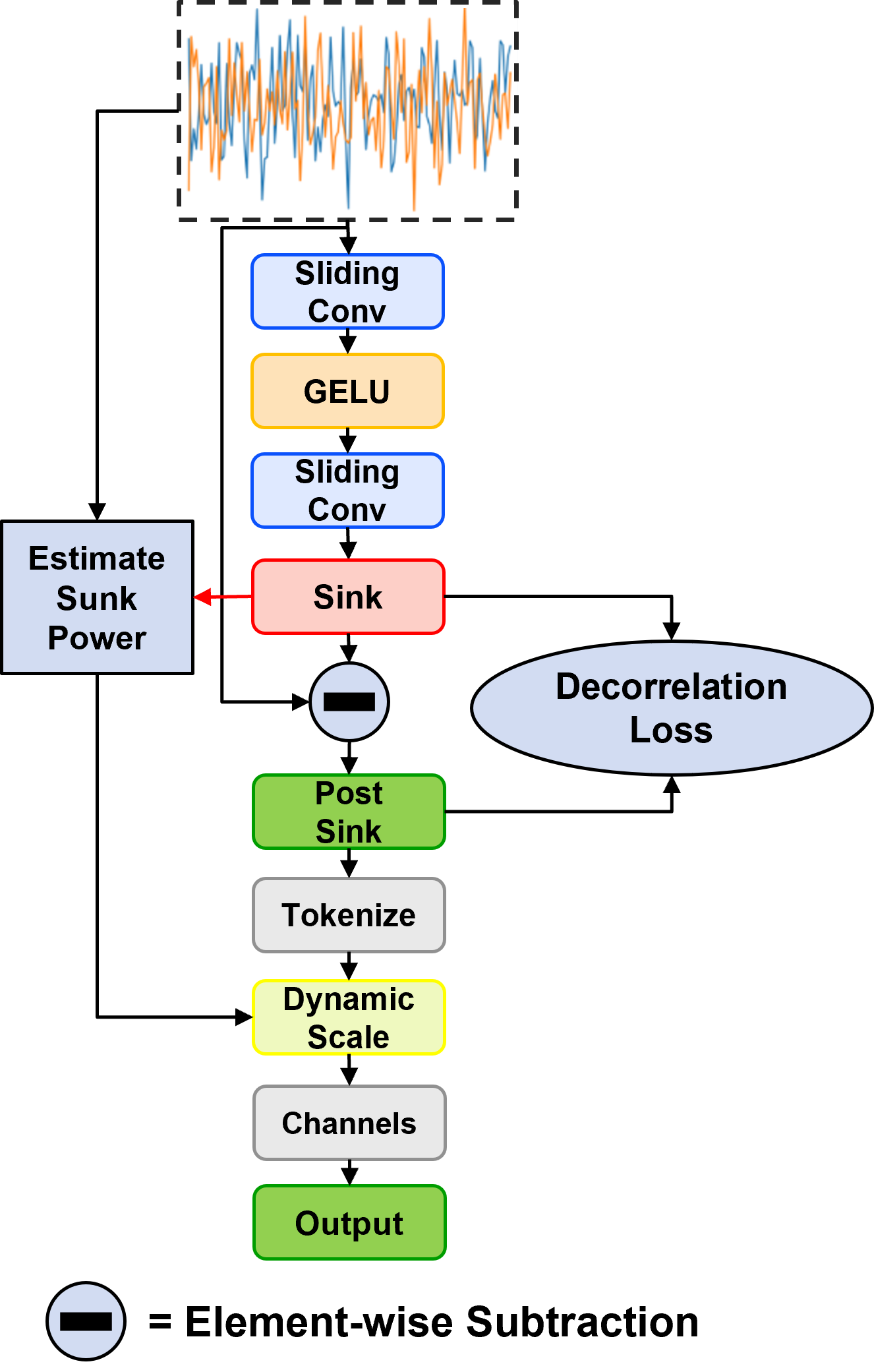}
    \caption{Construction of the Noise Sink: (1)Noise Estimation (2)Noise Subtraction (3)Regularization via Decorrelation (4) Dynamic Representation Power Compensation}
    \label{fig:AppendixNoiseSinkConstruction}
\end{figure}

\paragraph{Noise Sink Module}
\label{appendix:NoiseSinkModule}

To complement the implicit denoising priors of Blindspot Convolution and Latent 
Coherent Integration (Section~\ref{subsubsection:IsoFICReg}), we develop the 
Noise Sink Module to explicitly estimate and remove uncorrelated noise from the 
representation. Figure~\ref{fig:AppendixNoiseSinkConstruction} illustrates the 
four-stage construction: (1) noise estimation, (2) noise subtraction, (3) 
regularization via decorrelation, and (4) dynamic power compensation.

\subparagraph{Architecture Overview:}

The Noise Sink operates on intermediate representations $x \in \mathbb{R}^{B \times C \times L}$ 
(batch $\times$ channels $\times$ length) and produces cleaned representations 
$x_{clean} \in \mathbb{R}^{B \times C \times L}$:

\begin{algorithm}[H]
\caption{Noise Sink Module}
\begin{algorithmic}[1]
\STATE \textbf{Input:} $x \in \mathbb{R}^{B \times C \times L}$ (potentially noisy representation)
\STATE \textbf{Noise Estimation:} $n_{est} = \text{ResidualConv}(x)$
\STATE \textbf{Noise Subtraction:} $x_{sub} = x - n_{est}$
\STATE \textbf{Decorrelation Loss:} $L_{decorr} = \text{PearsonCorr}(n_{est}, x_{sub})$
\STATE \textbf{Token Reshaping:} $x_{tokens} = \text{ReshapeToToken}(x_{sub})$
\STATE \textbf{Power Estimation:} $P_{noise}, P_{signal} = \text{ConvPowerPooler}(n_{est}, x)$
\STATE \textbf{Power Ratio:} $r = P_{noise} / P_{signal}$
\STATE \textbf{Affine Parameters:} $\gamma, \beta = \text{AffineMLP}(r)$
\STATE \textbf{Compensation:} $x_{tokens,comp} = \gamma \cdot x_{tokens} + \beta$
\STATE \textbf{Token Normalization:} $x_{tokens,norm} = \text{RMSNorm}(x_{tokens,comp})$
\STATE \textbf{Reshape Back:} $x_{clean} = \text{ReshapeToChannels}(x_{tokens,norm})$
\STATE \textbf{Output:} $x_{clean} \in \mathbb{R}^{B \times C \times L}$, $n_{est}$, $L_{decorr}$, $P_{noise}$
\end{algorithmic}
\end{algorithm}

\subparagraph{Stage 1: Noise Estimation}

We leverage two inductive biases: (1) additive noise is locally uncorrelated 
with structured signals, and (2) multi-layer networks with non-linear activations 
can approximate arbitrary functions. The noise estimator consists of two convolutional 
layers with a GELU activation:

\begin{equation}
n_{est} = \text{Conv}_2(\text{GELU}(\text{Conv}_1(x)))
\end{equation}

\begin{table}[h]
\centering
\small
\begin{tabular}{|l|c|c|c|c|c|}
\hline
\textbf{Layer} & \textbf{In Channels} & \textbf{Out Channels} & \textbf{Kernel} & \textbf{Stride} & \textbf{Padding} \\
\hline
Conv$_1$ & $C$ & $C/4$ & 5 & 1 & same \\
Activation & \multicolumn{5}{c|}{GELU} \\
Conv$_2$ & $C/4$ & $C$ & 5 & 1 & same \\
\hline
\end{tabular}
\caption{Noise Estimator Architecture. Bias is disabled for all layers.}
\label{tab:noise_estimator}
\end{table}

\textbf{Rationale:} The bottleneck structure ($C \rightarrow C/4 \rightarrow C$) 
forces the network to learn a compressed representation of noise patterns, preventing 
it from learning identity mappings that would simply copy the input.

\subparagraph{Stage 2: Noise Subtraction}

The estimated noise is subtracted from the input:
\begin{equation}
x_{sub} = x - n_{est}
\end{equation}

\subparagraph{Stage 3: Regularization via Pearson Decorrelation}

To ensure the Noise Sink removes noise rather than signal, we compute the Pearson 
correlation coefficient between the estimated noise and cleaned signal:

\begin{align}
\text{sunk}_{flat} &= n_{est}.flatten(1) \in \mathbb{R}^{B \times (C \cdot L)} \\
\text{post}_{flat} &= x_{sub}.flatten(1) \in \mathbb{R}^{B \times (C \cdot L)} \\
\text{sunk}_{centered} &= \text{sunk}_{flat} - \mu_{sunk} \\
\text{post}_{centered} &= \text{post}_{flat} - \mu_{post} \\
\text{sunk}_{norm} &= \text{sunk}_{centered} / \sigma_{sunk} \\
\text{post}_{norm} &= \text{post}_{centered} / \sigma_{post} \\
\rho &= \frac{1}{N} \sum_{i=1}^{N} \text{sunk}_{\text{norm}}[i] \cdot \text{post}_{\text{norm}}[i] \\
L_{decorr} &= |\rho|
\end{align}

where $\mu$ and $\sigma$ are computed per sample over the flattened dimension. $N = C \cdot L$ is the total number of elements in the flattened representation. This normalization ensures $\rho \in [-1, 1]$, representing a valid Pearson correlation coefficient.

\textbf{Rationale:} If $n_{est}$ is truly noise, it should be uncorrelated with 
the cleaned signal $x_{sub}$ (Pearson $\rho \approx 0$). High correlation indicates 
the sink is removing signal content.

\subparagraph{Stage 4: Dynamic Power Compensation}

A critical challenge arises in low or negative-SNR regimes: after noise subtraction, 
signal power may be severely attenuated. We introduce dynamic power compensation 
based on the ratio of sunk noise power to original signal power.

\textbf{Token Reshaping:}

We convert the convolutional representation to token representation with a fixed token width of 2. Where $L/2$ is the number of tokens and $C \cdot 2$ is the token dimension. This groups consecutive IQ pairs per channel into tokens, preserving the complex-valued structure (each token contains one complete complex sample across all channels).

\textbf{Convolutional Power Pooling:}

Rather than computing power directly, we use an attention-based convolutional pooler that learns to weight different time steps:

Attention Logits:
\[\text{a} = \text{Conv}_{1x1}(x) \in \mathbb{R}^{B \times 1 \times L/2}\]
Attention Weights (over sequence dimension):
\[\alpha = \text{softmax}(a, dim = -1) \in \mathbb{R}^{(B \times 1 \times L/2)}\]
Weighted Pooling:
\[\text{pooled} = \text{bmm}(\alpha, x^{T}) \in \mathbb{R}^{(B\times 1 \times 2c)} \rightarrow \mathbb{R}^{(B \times 2C)}\]

Where $x^{T}$ denotes x.transpose(1,2) $\in \mathbb{R}^{(B \times L/2 \times 2C)}$, and bmm performs batch matrix multiplication: $[B \times 1 \times L/2] @ [B \times L/2 \times 2C] \rightarrow [B \times 1 \times 2C]$ 

\textbf{ConvPowerPooler Architecture:}
\begin{itemize}
    \item \textbf{Attention Conv:} 1×1 convolution: $C \rightarrow 1$
    \item \textbf{Softmax:} Over time dimension
    \item \textbf{Weighted pooling:} Batch matrix multiplication
    \item \textbf{Power:} Sum of squares over channels
\end{itemize}

\textbf{Rationale:} The learned attention weights allow the network to focus on 
informative time steps when estimating power, rather than uniform averaging. This 
is particularly useful for signals with time-varying SNR or transient events.

\textbf{Power Ratio Computation:}

We compute power for both the estimated noise and original signal:

\begin{align}
P_{noise} &= \text{ConvPowerPooler}(n_{est}) \in \mathbb{R}^{B \times L/2} \\
P_{signal} &= \text{ConvPowerPooler}(x) \in \mathbb{R}^{B \times L/2} \\
r &= \frac{P_{noise}}{P_{signal}} \in \mathbb{R}^{B \times L/2}
\end{align}

The ratio is clamped to $[0, 2]$ to handle extreme cases:
\begin{equation}
r = \text{clamp}(r, 0.0, 2.0)
\end{equation}

\textbf{Affine MLP (Hypernetwork):}

The power ratio (a scalar per token) is fed to an MLP that predicts affine 
parameters for all tokens:

\begin{equation}
[\gamma; \beta] = \text{AffineMLP}(r) \in \mathbb{R}^{B \times 2 \cdot (L/2)}
\end{equation}

\begin{table}[h]
\centering
\small
\begin{tabular}{|l|c|c|}
\hline
\textbf{Layer} & \textbf{Input Dim} & \textbf{Output Dim} \\
\hline
Linear$_1$ & 1 & $C \cdot 2 \cdot 2$ \\
ReLU & \multicolumn{2}{c|}{-} \\
Linear$_2$ & $C \cdot 2 \cdot 2$ & $2 \cdot (C \cdot 2 \cdot 2)$ \\
\hline
\end{tabular}
\caption{Affine MLP Architecture. Input is a scalar power ratio per token; 
output is concatenated $[\gamma; \beta]$ where $\gamma, \beta \in \mathbb{R}^{L/2}$ 
is broadcasted over the token dimension to apply the same scale and shift per element/token across all channels.}
\label{tab:affine_mlp}
\end{table}

The affine parameters are applied to each token individually and broadcasted across all channels per token. This is with the assumption that noise is localized at the same elements/tokens in a sequence across all feature channels. For example, noise in feature map 1 element 0 will also be noise at feature map 2 element 0. 

\textbf{Token Normalization:}

After compensation, tokens are normalized using RMS normalization:

\begin{equation}
x_{tokens,norm} = \text{RMSNorm}(x_{tokens,comp})
\end{equation}

\textbf{Reshape Back to Convolutional:}

Finally, we reverse the tokenization to convert the signals from $\mathbb{R}^{B \times L/2 \times 2C}$ back to $\mathbb{R}^{B \times C \times L}$ for subsequent convolutional processing.

\textbf{Rationale:} The token-level conditioning provides a 
localized compensation strategy based on the local SNR of the token. In low-SNR tokens/samples (high $r$), the network learns to amplify specific tokens rather than all of them (useful for time-varying noise); in high-SNR samples/tokens (low $r$), it preserves the representation with $\gamma \approx 1$. The RMS normalization after compensation ensures stable magnitudes regardless of the affine transformation. 

\subparagraph{Complete Noise Sink Summary:}

The Noise Sink provides explicit, regularized denoising with token-adaptive compensation. Key design principles:
\begin{itemize}
    \item \textbf{Explicit estimation:} Convolutional bottleneck approximates noise functions
    \item \textbf{Pearson decorrelation:} Ensures noise (not signal) is removed
    \item \textbf{Learned power pooling:} Attention-weighted power estimation
    \item \textbf{Token-level compensation:} Per-Token affine transformation based on SNR ratio
    \item \textbf{Token normalization:} RMS normalization stabilizes representations
    \item \textbf{Hierarchical:} Multiple sinks throughout encoder provide progressive denoising
\end{itemize}

\subparagraph{Additional Regularization Losses:}

Beyond the decorrelation loss computed within the module, the Noise Sink is further regularized through hierarchical losses computed across all sinks in the encoder:

\textbf{(a) Hierarchical Power Matching:} Total estimated noise power across all sinks matches injected noise power:
\begin{equation}
L_{power} = \left| \sum_{i=1}^{N_{sinks}} P_{noise}^{(i)} - P_{injected} \right|
\end{equation}

\textbf{(b) SNR Regression:} Reconstructed signal's SNR matches target SNR:
\begin{equation}
\text{SNR}_{est} = \log(||x_{recon}||^2) - \log\left(\sum_{i=1}^{N_{sinks}} P_{noise}^{(i)}\right)
\end{equation}
\begin{equation}
L_{SNR} = \text{MSE}_{focal}(\text{SNR}_{est}, \text{SNR}_{GT})
\end{equation}

\begin{figure}
    \centering
    \includegraphics[width=\textwidth]{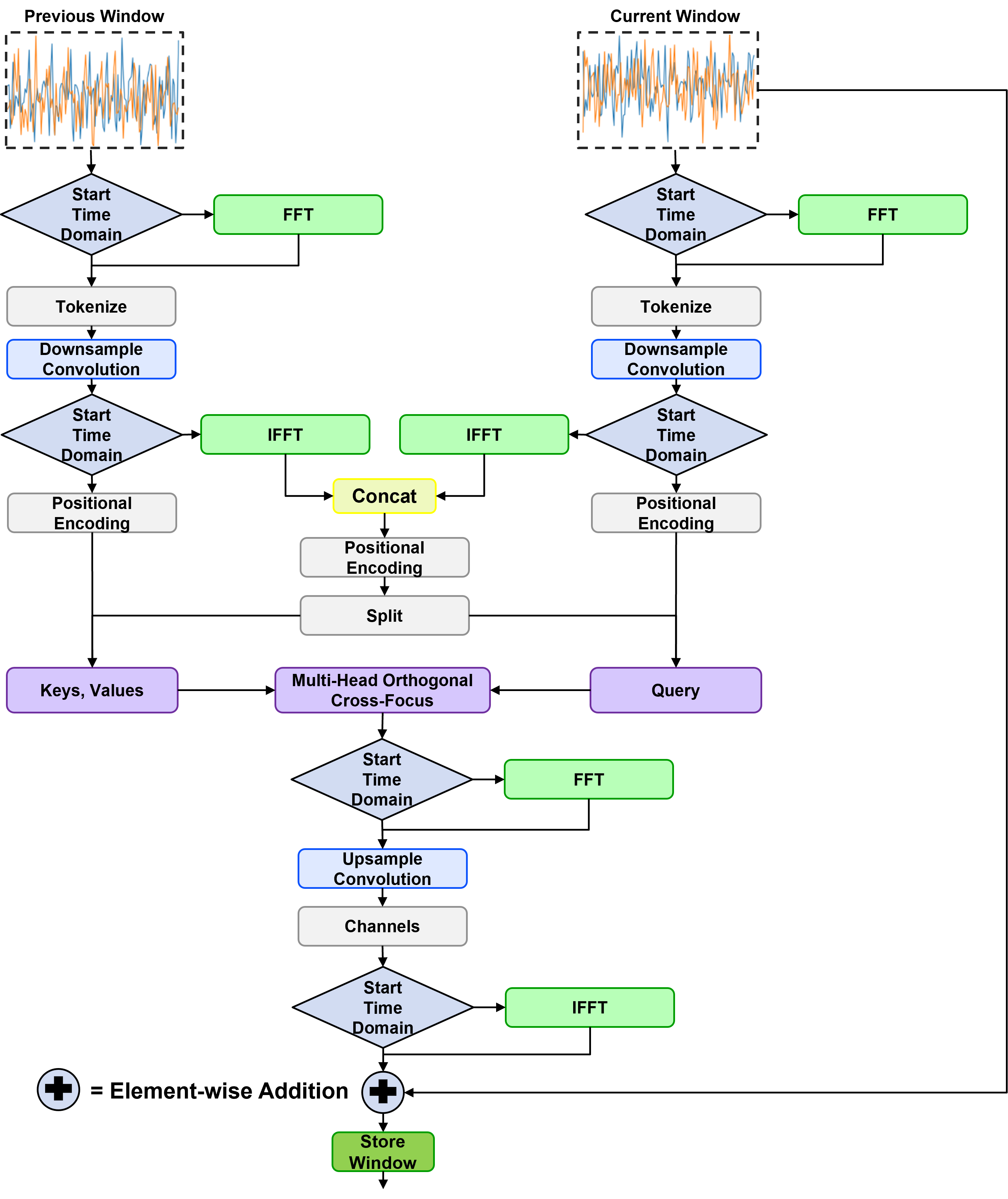}
    \caption{Causal Cross Window Focus Block}
    \label{fig:CausalCrossWindowFocus}
\end{figure}

\paragraph{Causal Cross-Window Focus:}
\label{appendix:CausalCrossWindowFocus}

The Causal Cross-Window Focus mechanism addresses a fundamental challenge in windowed sequence processing: modeling long-range dependencies across temporal boundaries while maintaining causal constraints. Our approach leverages domain-specific positional encodings and cross-attention between consecutive windows to capture phase relationships that would otherwise be lost in naive windowed processing.

Figure~\ref{fig:CausalCrossWindowFocus} illustrates the complete data flow through the mechanism. A key contribution lies in our adaptive treatment of time-domain versus frequency-domain representations, where positional encoding strategies are tailored to preserve the physical meaning of tokens in each domain.

\subparagraph{Main Algorithm:}
Algorithm~\ref{alg:ar_attention} presents the complete Causal Cross-Window Focus procedure. The mechanism operates on consecutive windows of a signal, using the previous window to provide contextual information for processing the current window through a cross-attention mechanism.

\begin{algorithm}[H]
\caption{Causal Cross-Window Focus (AR\_attention)}
\label{alg:ar_attention}
\begin{algorithmic}[1]
\REQUIRE Current window $\mathbf{x} \in \mathbb{R}^{B \times C \times L}$, Previous window $\mathbf{x}_{\text{prev}} \in \mathbb{R}^{B \times C \times L}$, Block index $b \in \{0, 1, \ldots, N_{\text{blocks}}-1\}$, Domain flag $\text{is\_freq} \in \{\text{True}, \text{False}\}$
\ENSURE Transformed representation $\mathbf{x}_{\text{out}} \in \mathbb{R}^{B \times C \times L}$, Orthogonality regularization loss $\mathcal{L}_{\text{orth}} \in \mathbb{R}$
\STATE $\mathbf{x}_{\text{res}} \leftarrow \mathbf{x}$ \COMMENT{Store residual connection}
\STATE
\STATE \COMMENT{\textbf{Domain Transformation}}
\IF{$\text{is\_freq} = \text{False}$}
    \STATE $\mathbf{x} \leftarrow \text{FFT}(\mathbf{x})$ \COMMENT{Transform to frequency domain}
    \STATE $\mathbf{x}_{\text{prev}} \leftarrow \text{FFT}(\mathbf{x}_{\text{prev}})$
\ENDIF
\STATE
\STATE \COMMENT{\textbf{Step 1: Tokenization \& Learned Spectral Compression}}
\STATE $\mathbf{T}, B, C, L \leftarrow \textsc{ReshapeToToken}(\mathbf{x})$ \COMMENT{$\mathbf{T} \in \mathbb{R}^{B \times L/2 \times 2C}$}
\STATE $\mathbf{T}_{\text{prev}}, \_, \_, \_ \leftarrow \textsc{ReshapeToToken}(\mathbf{x}_{\text{prev}})$
\STATE
\STATE $s \leftarrow \begin{cases} 16 & \text{if } b = 0 \\ 4 & \text{otherwise} \end{cases}$ \COMMENT{Adaptive downsampling stride}
\STATE
\STATE $\mathbf{T}_{\text{down}} \leftarrow \text{Conv1D}^{(b)}_{\text{down}}(\mathbf{T})$ \COMMENT{$\in \mathbb{R}^{B \times (L/2)/s \times 2C}$}
\STATE $\mathbf{T}_{\text{prev,down}} \leftarrow \text{Conv1D}^{(b)}_{\text{down}}(\mathbf{T}_{\text{prev}})$
\STATE
\STATE \COMMENT{\textbf{Step 2: Domain-Specific Positional Encoding}}
\IF{$\text{is\_freq} = \text{False}$}
    \STATE $\mathbf{T}_{\text{down}}, \mathbf{T}_{\text{prev,down}} \leftarrow \textsc{TimeDomainPosEnc}(\mathbf{T}_{\text{down}}, \mathbf{T}_{\text{prev,down}}, B, C, L, b)$
    \STATE \hspace{5cm} \COMMENT{Algorithm~\ref{alg:time_pos_enc}}
\ELSE
    \STATE $\mathbf{T}_{\text{down}}, \mathbf{T}_{\text{prev,down}} \leftarrow \textsc{FreqDomainPosEnc}(\mathbf{T}_{\text{down}}, \mathbf{T}_{\text{prev,down}}, b)$
    \STATE \hspace{5cm} \COMMENT{Algorithm~\ref{alg:freq_pos_enc}}
\ENDIF
\STATE
\STATE \COMMENT{\textbf{Step 3: Multi-Head Orthogonal Cross-Focus Attention}}
\STATE $\mathbf{T}_{\text{out}}, \mathcal{L}_{\text{orth}} \leftarrow \text{MultiHeadFocus}^{(b)}(\mathbf{Q}=\mathbf{T}_{\text{down}}, \mathbf{K}=\mathbf{T}_{\text{prev,down}}, \mathbf{V}=\mathbf{T}_{\text{prev,down}})$
\STATE
\STATE \COMMENT{\textbf{Step 4: Spectral Upsampling \& Reconstruction}}
\IF{$\text{is\_freq} = \text{False}$}
    \STATE $\mathbf{T}_{\text{out}} \leftarrow \textsc{FFT}(\mathbf{T}_{\text{out}})$ 
\ENDIF
\STATE
\STATE $\mathbf{T}_{\text{up}} \leftarrow \text{Conv1D}^{(b)}_{\text{up}}(\mathbf{T}_{\text{out}})$ \COMMENT{Upsample to original token count}
\STATE $\mathbf{x}_{\text{out}} \leftarrow \textsc{ReshapeToChannels}(\mathbf{T}_{\text{up}}, B, C, L)$
\STATE
\IF{$\text{is\_freq} = \text{False}$}
    \STATE $\mathbf{x}_{\text{out}} \leftarrow \text{IFFT}(\mathbf{x}_{\text{out}})$ \COMMENT{Return to time domain}
\ENDIF
\STATE
\STATE $\mathbf{x}_{\text{out}} \leftarrow \mathbf{x}_{\text{out}} + \mathbf{x}_{\text{res}}$ \COMMENT{Residual connection}
\STATE
\RETURN $\mathbf{x}_{\text{out}}, \mathcal{L}_{\text{orth}}$
\end{algorithmic}
\end{algorithm}

\subparagraph{Domain-Specific Positional Encoding}

A critical contribution in our approach is the use of domain-specific positional encoding strategies. The physical interpretation of sequence position differs fundamentally between time and frequency domains, necessitating distinct encoding approaches.

\subparagraph{Time Domain Positional Encoding}

For time-domain processing, temporal continuity across window boundaries is paramount. Algorithm~\ref{alg:time_pos_enc} describes our causal concatenation strategy, which preserves phase relationships by encoding windows in their natural temporal order.

\begin{algorithm}[H]
\caption{Time Domain Positional Encoding}
\label{alg:time_pos_enc}
\begin{algorithmic}[1]
\REQUIRE Downsampled tokens $\mathbf{T}_{\text{curr}} \in \mathbb{R}^{B \times N \times D}$, $\mathbf{T}_{\text{prev}} \in \mathbb{R}^{B \times N \times D}$
\REQUIRE Block index $b$
\ENSURE Position-encoded tokens $\mathbf{T}'_{\text{curr}}, \mathbf{T}'_{\text{prev}}$
\STATE \COMMENT{Convert to time domain for causal concatenation}
\STATE $\mathbf{T}_{\text{curr}} \leftarrow \text{IFFT}(\mathbf{T}_{\text{curr}})$ \COMMENT{Transform to time domain}
\STATE $\mathbf{T}_{\text{prev}} \leftarrow \text{IFFT}(\mathbf{T}_{\text{prev}})$
\STATE
\STATE \COMMENT{Causal concatenation preserves temporal ordering}
\STATE $\mathbf{T}_{\text{concat}} \leftarrow \text{Concat}([\mathbf{T}_{\text{prev}}, \mathbf{T}_{\text{curr}}], \text{dim}=1)$ \COMMENT{$\in \mathbb{R}^{B \times 2N \times D}$}
\STATE
\STATE $\mathbf{PE} \leftarrow \textsc{SinusoidalPosEnc}(2N, D)$ \COMMENT{Standard sinusoidal encoding}
\STATE $\mathbf{T}_{\text{concat}} \leftarrow \mathbf{T}_{\text{concat}} + \mathbf{PE}$
\STATE
\STATE \COMMENT{Split and normalize}
\STATE $\mathbf{T}'_{\text{curr}} \leftarrow \text{RMSNorm}^{(b)}(\mathbf{T}_{\text{concat}}[:, N:, :])$ \COMMENT{Current window tokens}
\STATE $\mathbf{T}'_{\text{prev}} \leftarrow \text{RMSNorm}^{(b)}(\mathbf{T}_{\text{concat}}[:, :N, :])$ \COMMENT{Previous window tokens}
\STATE
\RETURN $\mathbf{T}'_{\text{curr}}, \mathbf{T}'_{\text{prev}}$
\end{algorithmic}
\end{algorithm}

\textbf{Rationale:} In the time domain, the relative temporal position between tokens in consecutive windows carries critical phase information. By transforming frequency-domain representations back to time, concatenating causally (previous before current), and then applying positional encoding to the unified sequence, we ensure that the model can learn continuous phase relationships across window boundaries. This is essential for signals where phase coherence matters, such as in communications or audio processing.

\subparagraph{Frequency Domain Positional Encoding}

For frequency-domain processing, spectral bin relationships within each window are more important than temporal ordering across windows. Algorithm~\ref{alg:freq_pos_enc} presents our independent encoding strategy.

\begin{algorithm}[H]
\caption{Frequency Domain Positional Encoding}
\label{alg:freq_pos_enc}
\begin{algorithmic}[1]
\REQUIRE Downsampled tokens $\mathbf{T}_{\text{curr}} \in \mathbb{R}^{B \times N \times D}$, $\mathbf{T}_{\text{prev}} \in \mathbb{R}^{B \times N \times D}$
\REQUIRE Block index $b$
\ENSURE Position-encoded tokens $\mathbf{T}'_{\text{curr}}, \mathbf{T}'_{\text{prev}}$
\STATE
\STATE \COMMENT{Independent encoding for spectral bin relationships}
\STATE $\mathbf{PE} \leftarrow \textsc{SinusoidalPosEnc}(N, D)$
\STATE
\STATE $\mathbf{T}_{\text{curr}} \leftarrow \mathbf{T}_{\text{curr}} + \mathbf{PE}$
\STATE $\mathbf{T}_{\text{prev}} \leftarrow \mathbf{T}_{\text{prev}} + \mathbf{PE}$
\STATE
\STATE $\mathbf{T}'_{\text{curr}} \leftarrow \text{RMSNorm}^{(b)}(\mathbf{T}_{\text{curr}})$
\STATE $\mathbf{T}'_{\text{prev}} \leftarrow \text{RMSNorm}^{(b)}(\mathbf{T}_{\text{prev}})$
\STATE
\RETURN $\mathbf{T}'_{\text{curr}}, \mathbf{T}'_{\text{prev}}$
\end{algorithmic}
\end{algorithm}

\textbf{Rationale:} In the frequency domain, each position corresponds to a spectral bin rather than a temporal instant. The relationship between corresponding bins across windows (e.g., how the 100 Hz component evolves) is more meaningful than the temporal ordering of windows. Independent positional encoding allows the cross-attention mechanism to learn spectral covariance patterns—how energy in specific frequency bands correlates across time—without imposing artificial temporal structure on what is fundamentally a spectral representation.

\subparagraph{Mathematical Formulation}

We now provide explicit mathematical definitions for each transformation in the Causal Cross-Window Focus mechanism.

\textbf{Tokenization.} Given a convolutional feature map $\mathbf{X} \in \mathbb{R}^{B \times C \times L}$ where adjacent channel pairs represent complex-valued features (real and imaginary components), we reshape to token representation:
\begin{equation}
\mathbf{T} = \text{Reshape}(\mathbf{X}) \in \mathbb{R}^{B \times (L/2) \times (2C)}
\end{equation}
This operation groups complex pairs along the sequence dimension, creating tokens that span all channels.

\textbf{Learned Spectral Compression.} 

To avoid aliasing artifacts common in naive downsampling, we perform learned 
spectral compression in the frequency domain before temporal decimation. 
Specifically:

1. We reshape the frequency-domain representation $[B, F, T]$ (where $F$ represents 
   frequency bins) into token format $[B, T/2, 2F]$ where each token encodes a 
   complex-valued frequency representation.

2. We apply a learned $1×1$ convolution with stride s in token space, which 
   effectively learns to compress the frequency spectrum by selecting which 
   frequency bins to preserve before downsampling.

3. For time-domain processing, we convert back via IFFT, ensuring the temporal 
   signal has reduced bandwidth appropriate for the lower sampling rate.

This approach implements a learned, adaptive low-pass filter in the frequency 
domain, preventing aliasing while preserving task-relevant spectral components.

\begin{equation}
\mathbf{T}_{\downarrow} = \text{Conv}^{(b)}_{1 \times 1, s}(\mathbf{T}) \in \mathbb{R}^{B \times N_{\text{down}} \times (2C)}
\end{equation}
where $N_{\text{down}} = (L/2)/s$ with adaptive stride:
\begin{equation}
s = \begin{cases} 
16 & \text{if } b = 0 \text{ (early layers, aggressive compression)} \\
4 & \text{if } b > 0 \text{ (later layers, preserve detail)}
\end{cases}
\end{equation}

\textbf{Cross-Window Focus.} Multi-head orthogonal cross-focus with $H=4$ heads:
\begin{align}
\text{head}_h &= \text{Focus}(\mathbf{Q}_h, \mathbf{K}_h, \mathbf{V}_h) \\
&= \text{softmax}\left(\frac{Cov(\mathbf{Q}_h,\mathbf{K}_h) \cdot L_{sequence}}{\sqrt{d_k}\cdot K_{focus}}\right) \mathbf{V}_h \label{eq:attention}
\end{align}
where $Cov(Q,K) = (Q-\mu_Q)(K-\mu_K)^T$.

The focus factor $K_{focus} \in [1, L_{sequence}]$ is computed from attention score statistics. First, we compute per-token variance:
\[\sigma^2_i = Var_j(Cov(Q, K)_{:,i}), \quad i \in [1, L_{sequence}]\]

where $\sigma^2_i$ is the variance of attention scores for token $i$ across all queries. We then compute per-token SNR estimates relative to cumulative variance:
\[\text{SNR}_i = \frac{\sigma^2_i}{\sum_{j \neq i} \sigma^2_j + \epsilon}\]

From the SNR distribution, we extract summary statistics:
\begin{align}
\text{mean\_snr} &= \frac{1}{L_{sequence}} \sum_{i=1}^{L_{sequence}} \text{SNR}_i \\
\text{high\_snr\_prop} &= \frac{1}{L_{sequence}} \sum_{i=1}^{L_{sequence}} \mathbb{1}[\text{SNR}_i > \text{median}(\{\text{SNR}_j\}_{j=1}^{L_{sequence}})]
\end{align}

These statistics are concatenated and fed to a learned network:
\[\mathbf{k}_{\text{input}} = [\text{mean\_snr}, \text{high\_snr\_prop}]\]
\[K_{focus} = 1+(L_{sequence}-1) \cdot \sigma(f_\theta(\mathbf{k}_{\text{input}}))\]

where $f_\theta$ is a lightweight MLP and $\sigma(\cdot)$ is the sigmoid function.

\begin{align}
\mathbf{Q}_h &= \mathbf{T}_{\text{curr}} \mathbf{W}^Q_h, \quad \mathbf{Q}_h \in \mathbb{R}^{B \times N \times d_k} \\
\mathbf{K}_h &= \mathbf{T}_{\text{prev}} \mathbf{W}^K_h, \quad \mathbf{K}_h \in \mathbb{R}^{B \times N \times d_k} \\
\mathbf{V}_h &= \mathbf{T}_{\text{prev}} \mathbf{W}^V_h, \quad \mathbf{V}_h \in \mathbb{R}^{B \times N \times d_v}
\end{align}
with $d_k = d_v = (2C)/H$. The multi-head outputs are concatenated and projected:
\begin{equation}
\mathbf{T}_{\text{out}} = \text{Concat}(\text{head}_1, \ldots, \text{head}_H) \mathbf{W}^O
\end{equation}

\textbf{Orthogonality Regularization.} To encourage diverse attention patterns across heads, we apply orthogonality regularization as discussed in Section~\ref{appendix:head_orth}. 

\textbf{Spectral Upsampling and Reconstruction.} The attended representation is upsampled back to the original token count:
\begin{equation}
\mathbf{T}_{\text{up}} = \text{Conv}^{(b)}_{1 \times 1, 1/s}(\mathbf{T}_{\text{out}}) \in \mathbb{R}^{B \times (L/2) \times (2C)}
\end{equation}
followed by reshaping and optional inverse FFT:
\begin{equation}
\mathbf{X}_{\text{out}} = \begin{cases}
\text{IFFT}(\textsc{ReshapeToChannels}(\mathbf{T}_{\text{up}})) & \text{if } \text{is\_freq} = \text{False} \\
\textsc{ReshapeToChannels}(\mathbf{T}_{\text{up}}) & \text{if } \text{is\_freq} = \text{True}
\end{cases}
\end{equation}

\textbf{Residual Connection.} Finally, we add the input as a residual connection:
\begin{equation}
\mathbf{X}_{\text{final}} = \mathbf{X}_{\text{out}} + \mathbf{X}_{\text{input}}
\end{equation}

\textbf{Tensor Shape Transformations}

Table~\ref{tab:tensor_shapes} tracks tensor dimensions through the complete forward pass, aiding implementation and debugging.

\begin{table}[h]
\centering
\caption{Tensor Shape Transformations in Causal Cross-Window Focus}
\label{tab:tensor_shapes}
\begin{tabular}{p{5.5cm}p{4cm}p{5cm}}
\toprule
\textbf{Operation} & \textbf{Shape} & \textbf{Description} \\
\midrule
Input $\mathbf{x}$ & $\mathbb{R}^{B \times C \times L}$ & Convolutional representation \\
\midrule
After FFT (if time-domain) & $\mathbb{R}^{B \times C \times L}$ & Frequency domain (complex as real pairs) \\
\midrule
After \textsc{ReshapeToToken} & $\mathbb{R}^{B \times L/2 \times 2C}$ & Token representation \\
\midrule
After Downsample Conv & $\mathbb{R}^{B \times (L/2)/s \times 2C}$ & Compressed tokens ($s \in \{4, 16\}$) \\
\midrule
After Pos. Encoding (time) & $\mathbb{R}^{B \times (L/2)/s \times 2C}$ & Causally concatenated, then split \\
\midrule
After Pos. Encoding (freq) & $\mathbb{R}^{B \times (L/2)/s \times 2C}$ & Independently encoded \\
\midrule
After Cross-Attention & $\mathbb{R}^{B \times (L/2)/s \times 2C}$ & Attended representation \\
\midrule
After Freq. Upsample (time) & $\mathbb{R}^{B \times (L/2)/s \times 2C}$ & Back to frequency domain \\
\midrule
After Upsample Conv & $\mathbb{R}^{B \times L/2 \times 2C}$ & Restored token count \\
\midrule
After \textsc{ReshapeToChannels} & $\mathbb{R}^{B \times C \times L}$ & Back to convolutional shape \\
\midrule
After IFFT (if time-domain) & $\mathbb{R}^{B \times C \times L}$ & Time domain output \\
\midrule
After Residual & $\mathbb{R}^{B \times C \times L}$ & Final output \\
\bottomrule
\end{tabular}
\end{table}

\subparagraph{Design Rationale}

We now explain the key design decisions that distinguish our approach from standard attention mechanisms.

\textbf{Why domain-specific positional encoding?} 
The fundamental insight is that position has different physical meanings in time versus frequency domains. In time-domain processing, causal concatenation of windows preserves temporal continuity, allowing the model to learn phase relationships across window boundaries. This is critical for signals where phase coherence matters—for example, in communications systems where carrier phase must be tracked across packet boundaries, or in audio where transient events may span multiple windows.

In contrast, frequency-domain processing benefits from independent encoding per window. Here, each token position corresponds to a spectral bin (e.g., the 100 Hz component), and the cross-focus mechanism learns how energy in specific frequency bands evolves across time. Concatenating windows would impose artificial temporal ordering on what is fundamentally a spectral covariance problem. Independent encoding allows the model to learn patterns like "when the 100 Hz component is strong in the previous window, the 200 Hz component tends to be strong in the current window," without conflating this with temporal position.

\textbf{Why learned spectral compression?}
Traditional downsampling via strided convolution or pooling can introduce aliasing artifacts, especially when applied in the time or spatial domain. Our learned $1 \times 1$ convolution (with stride) performs adaptive compression across all channels simultaneously, learning which spectral components to preserve at each layer depth due to its explicit application in the frequency domain. Early layers (block 0) use aggressive compression ($s=16$) to mitigate quadratic attention costs, while later layers ($s=4$) preserve more detail for fine-grained pattern recognition at a reduced computational burden for attention.

\textbf{Why cross-focus instead of self-focus?}
Using the previous window as Keys and Values while the current window provides Queries enforces causal information flow, essential for autoregressive processing. This design has several advantages:
\begin{enumerate}
    \item \textbf{Causality:} Information flows strictly from past to present, enabling online/streaming processing.
    \item \textbf{Efficiency:} We avoid the $\mathcal{O}(L^2)$ complexity of full-sequence self-focus, instead computing $\mathcal{O}((L/s)^2)$ focus within compressed windows.
    \item \textbf{Interpretability:} focus weights reveal how the model uses past context to inform current predictions, facilitating analysis of learned temporal dependencies.
\end{enumerate}

\paragraph{Spectral Compression Via Frequency Domain Pooling}
\label{appendix:FrequencyPooling}

Pooling operations are ubiquitous in modern deep learning architectures, serving to reduce computational costs while learning abstract feature representations. The most common approaches—max pooling and average pooling—are typically applied in the time or spatial domain. However, these conventional methods fundamentally violate the Nyquist-Shannon sampling theorem~\cite{Shannon1949}, leading to aliasing artifacts where high-frequency components fold back into lower frequency bands during downsampling. While recent work has introduced anti-aliasing filters prior to pooling~\cite{zhang2019making}, these approaches either dilute high-frequency content through low-pass filtering or explicitly discard it, fundamentally limiting the representational capacity of the network.

\subparagraph{Motivation: The High-Frequency Information Problem}

The discarding or corruption of high-frequency information during pooling has two critical consequences:

\textbf{(1) Low-Frequency Learning Bias.} It is well-documented that deep neural networks exhibit a spectral bias toward learning low-frequency functions faster than high-frequency counterparts~\cite{pmlr-v97-rahaman19a}. We argue this is not merely an empirical observation but a deterministic consequence of progressive downsampling. As one descends deeper into a convolutional architecture with repeated pooling operations, more high-frequency detail is either aliased or removed via anti-aliasing filters. This progressively leaves only low-frequency information in the learned representations. Notably, early layers in deep convolutional networks—which undergo minimal downsampling—are well-known to capture high-frequency features such as edges, corners, and fine textures. This is not coincidental: these layers retain the spectral bandwidth necessary to represent such features.

\textbf{(2) Violation of Frequency Translation Equivariance.} For architectures designed to learn equivariant representations with respect to frequency translations (e.g., shift-invariance in the spectral domain), it is essential to retain complete spectral fidelity across both low and high frequencies, including both magnitude and phase information. Traditional pooling methods fundamentally break this symmetry by selectively removing or corrupting portions of the spectrum. This is particularly problematic for complex-valued signals, where the asymmetric Hermitian symmetry of the spectrum must be preserved to maintain physical interpretability.

\subparagraph{Proposed Method: Frequency Domain Average Pooling}

We propose a simple yet effective solution based on the principle that \textit{spectral structure} is more important for signal representation than \textit{absolute frequency location}. Specifically, when downsampling in the PlanFormer architecture, we apply average pooling directly in the frequency domain on complex-valued spectra.

The procedure is as follows:
\begin{enumerate}
    \item \textbf{De-interleave:} Convert the real-valued interleaved representation (where adjacent elements per channel represent real and imaginary components) to explicit complex format.
    \item \textbf{FFT (if time-domain):} If the input is in the time domain, transform to frequency domain via FFT.
    \item \textbf{Average Pool:} Apply average pooling along the frequency axis, reducing the sequence length by a factor of $r$ (typically $r \in \{2, 4\}$).
    \item \textbf{IFFT (if time-domain):} If the original input was time-domain, transform back via IFFT.
    \item \textbf{Re-Interleave:} Convert the complex-valued sequence back to a real-valued interleaved sequence for subsequent real-valued processing blocks.
\end{enumerate}

This produces a representation that retains the \textit{spectral envelope} of the higher-resolution signal but at reduced sequence length. 
Crucially, this operation preserves both magnitude and phase information in a coarsened form, effectively creating a coarser binning of the original spectrum. The network's non-linearities and representational capacity can then learn to compensate for relative spectral envelope structure rather than requiring absolute frequency component locations. This approach:
\begin{itemize}
    \item Explicitly retains both high and low-frequency components (in compressed form)
    \item Respects the asymmetric Hermitian symmetry of complex-valued signals
    \item Avoids aliasing by operating directly in the frequency domain
    \item Facilitates learning of frequency translation equivariance
\end{itemize}

\textbf{Implementation Note:} Frequency Preserving Pooling contrasts with the earlier mentioned Learned Spectral Compression by not using any learned parameters or layers during this pooling stage. This only utilizes a parametric complex valued FFT(IFFT to transform back to time domain if originating from time domain) and non-trainable average pooling kernel. This is done in an effort to minimize learnable parameters to only those that are absolutely necessary.

\subparagraph{Implications and Benefits}

\textbf{U-Net Style Architectures.} Frequency-domain pooling is particularly powerful when combined with decoder pipelines in a U-Net fashion. Skip connections can be fused in the frequency domain, where higher-resolution spectra from earlier layers infill and sharpen the reconstruction during upsampling. This facilitates explicit high-frequency detail retention and reconstruction, enhancing the fidelity of generated outputs. For fine-grained tasks where high-frequency details provide discriminative signals (e.g., texture classification, signal modulation recognition), this retention is essential.

\textbf{Extreme Compression Ratios.} By leveraging frequency-domain pooling for spectral compression, we achieve downsampling from input space to token space by a factor of $64\times$ through three sequential pooling stages, each with stride $s=4$ (i.e., $4^3 = 64$). This extreme compression is essential for enabling our cross-window focus mechanism and Parseval Focus Mechanisms within the Parseval Transformer~\ref{subsubsection:EnergyConservation} over very long sequences, circumventing the quadratic complexity of attention mechanisms while preserving spectral fidelity. Critically, each pooling stage operates in the frequency domain, ensuring that the spectral envelope—including high-frequency components—is retained throughout the compression pipeline.

\textbf{Sample Rate Equivariance.} When combined with our Multi-Head Orthongal Parseval Focus mechanism (Section~\ref{subsubsection:EnergyConservation}) for time-frequency energy conservation and function learning, frequency-domain pooling enables learning a functional representation of time versus frequency across a structurally representative spectrum. This facilitates learning sample rate equivariance, such that the trained architecture gracefully extends to arbitrary sample rates without requiring sample-rate-specific retraining or fine-tuning during deployment. We demonstrate this capability through zero-shot deployment of the architecture across various domains with the encoder frozen, despite training only on a single domain at a single sample rate (RF signals at 7.69 megasamples per second).

\textbf{Frequency Translation Equivariance.} Most importantly, spectral envelope retention is essential for learning full frequency translation equivariance, which allows natural scaling across domains and frequencies. Traditional pooling methods directly handicap this capability by corrupting the spectral structure, leading to poor gradient flow and degraded performance. By preserving the complete spectral envelope—including high-frequency components—our approach enables the network to learn truly equivariant representations with respect to frequency shifts.

\subparagraph{Theoretical Justification}

From a signal processing perspective, frequency-domain average pooling can be understood as a form of \textit{spectral decimation} that preserves the overall energy distribution while reducing resolution. Unlike time-domain downsampling, which must obey the Nyquist criterion to avoid aliasing, frequency-domain pooling operates on the already-decomposed spectral representation. The averaging operation acts as a form of spectral smoothing that preserves the envelope while discarding fine-grained frequency resolution—a fundamentally different trade-off than discarding high-frequency content entirely.

\subparagraph{Comparison to Existing Methods}

Table~\ref{tab:pooling_comparison} contrasts our frequency-domain pooling with conventional approaches.

\begin{table}[h]
\centering
\caption{Comparison of Pooling Methods}
\label{tab:pooling_comparison}
\begin{tabular}{p{3.5cm}p{2.5cm}p{2.5cm}p{2.5cm}p{2.5cm}}
\toprule
\textbf{Method} & \textbf{Domain} & \textbf{Aliasing} & \textbf{High-Freq Retention} & \textbf{Hermitian Symmetry} \\
\midrule
Max Pooling & Time/Spatial & Yes & No & No \\
Average Pooling & Time/Spatial & Yes & No & No \\
Anti-Aliased Pooling~\cite{zhang2019making} & Time/Spatial & Reduced & Partial & No \\
\textbf{Freq-Domain Pooling (Ours)} & \textbf{Frequency} & \textbf{No} & \textbf{Yes (compressed)} & \textbf{Yes} \\
\bottomrule
\end{tabular}
\end{table}

\paragraph{Convolutional Attention}
\label{appendix:ECSA}

The convolution block concludes with an Efficient Channel-Spatial(Temporal for 1d) Attention (ECSA) block~\cite{gu2024efficient} to provide a final attention in convolutional space. 

\paragraph{Domain-Specific Token Aggregation}
\label{appendix:domain_token_aggregation}

A critical design choice in our tokenization strategy is the domain-specific aggregation of these windowed representations. Rather than applying a uniform aggregation strategy across domains, we respect the physical structure and interpretation of each domain.

\subparagraph{Time Domain Aggregation: Causal Concatenation}

For the time domain, temporal ordering is paramount. We aggregate tokenized windows by concatenating them along the sequence dimension to preserve the causal temporal structure:
\begin{equation}
\mathbf{x}_{\text{time}} = \text{Concat}([\mathbf{x}_{\text{win}_1}, \mathbf{x}_{\text{win}_2}, \ldots, \mathbf{x}_{\text{win}_{N_w}}], \text{dim}=2) \in \mathbb{R}^{B \times C \times (N_w \cdot L)}
\end{equation}
where $N_w$ is the number of windows. This produces a unified temporal representation that maintains the sequential ordering of events across the entire signal. Each window's features are placed in their correct temporal position, enabling downstream modules to learn long-range temporal dependencies while respecting causality.

\subparagraph{Frequency Domain Aggregation: Latent Spectral Averaging}

For the frequency domain, the physical interpretation is fundamentally different. Each window's frequency representation captures the spectral content of that temporal segment. To obtain a global spectral representation, we average the tokenized frequency representations across all windows:
\begin{equation}
\mathbf{x}_{\text{freq}} = \frac{1}{N_w} \sum_{i=1}^{N_w} \mathbf{x}_{\text{win}_i} \in \mathbb{R}^{B \times C \times L}
\end{equation}

Critically, this averaging occurs in the \textit{learned latent space} after each window has undergone local representation learning through the convolutional tokenization pipeline (convolution, cross-window focus, frequency-domain pooling, and ECSA blocks). This is fundamentally different from naive spectral averaging in the input space.

\subparagraph{Rationale: Physics-Informed Design}

These domain-specific aggregation strategies are anchored in the physical meaning of each representation rather than heuristic choices:

\textbf{Time Domain.} In the temporal domain, position has absolute meaning—an event at time $t_1$ is fundamentally different from the same event at time $t_2$. Concatenation preserves this temporal ordering, enabling the model to learn causal relationships, temporal dynamics, and sequential patterns. Averaging would destroy this critical temporal structure.

\textbf{Frequency Domain.} In the spectral domain, each frequency bin represents a specific oscillatory component (e.g., the 100 Hz component). However, unlike the time domain where position is absolute, the \textit{presence and strength} of spectral components may vary across time.

\textbf{Latent Space vs. Input Space Averaging.} This distinction is critical: averaging in latent space is fundamentally different from naive spectral averaging in the input space. Averaging raw frequency-domain inputs (e.g., windowed FFTs) produces a time-averaged spectrum that obscures time-varying spectral phenomena. In contrast, our latent space averaging operates on learned feature representations where each window's tokenizer has already encoded local spectral patterns, transient events, and time-varying structure into the feature channels $C$. The averaging operation then aggregates these learned representations, preserving information about time-varying phenomena in the channel dimension while producing a compact global spectral summary in the sequence dimension.

For example, consider a frequency-hopping signal where the carrier frequency shifts across windows. Input-space spectral averaging would produce a blurred spectrum showing energy across all hopped frequencies with no temporal structure. In contrast, our approach allows the per-window tokenizers to learn representations that encode "frequency hop at 100 Hz in this window" and "frequency hop at 200 Hz in that window" in distinct feature channels. The subsequent averaging produces a latent representation where different channels capture different hopping patterns, preserving the time-varying spectral structure that would otherwise be lost.

This design enables the frequency-domain encoder to learn both stationary spectral structure (consistent patterns across windows) and time-varying spectral dynamics (patterns that evolve across windows), without requiring an artificially long sequence that would conflate temporal evolution with spectral content. For signals with time-varying spectral characteristics—such as frequency hopping in communications, chirps in radar, or vibrato in audio—this latent space averaging is essential for preserving discriminative information while maintaining computational efficiency.

\subparagraph{Output Shapes and Downstream Processing}

The domain-specific aggregation produces tokenized representations with distinct shapes:
\begin{align}
\text{Time Domain:} \quad &\mathbf{x}_{\text{time}} \in \mathbb{R}^{B \times C \times (N_w \cdot L)} \\
\text{Frequency Domain:} \quad &\mathbf{x}_{\text{freq}} \in \mathbb{R}^{B \times C \times L}
\end{align}

\subsection{Learned Domain Transformation \& Cross-Domain Fusion}
\label{appendix:CrossDomainFusion}

Upon completion of the convolutional tokenization pipeline, we prepare the domain-specific representations for processing by the Parseval Transformer blocks. This preparation involves two critical steps: (1) transforming convolutional feature maps into token representations suitable for transformer processing~\cite{hassani2021escaping}, and (2) enabling cross-domain information flow through learned transformations and fusion mechanisms. The latter builds on the fundamental signal processing principle that comprehensive signal analysis requires joint time-frequency analysis rather than isolated domain-specific processing.

\subsubsection{Complex-Aware Tokenization}

We transform convolutional feature maps into token representations in a manner that respects the complex-valued structure of the input space. Recall that the convolutional tokenizer outputs feature maps $\mathbf{x} \in \mathbb{R}^{B \times C \times L}$, where $L$ represents an interleaved real-valued sequence of in-phase (I) and quadrature (Q) components. 

To preserve the complex-valued structure during tokenization, we reshape such that the sequence length $N = L/2$ corresponds to the number of complex samples, while the embedding dimension doubles to $2C$ to accommodate both I and Q components per channel:
\begin{equation}
\mathbf{x} \in \mathbb{R}^{B \times C \times L} \rightarrow \mathbf{T} \in \mathbb{R}^{B \times N \times 2C}, \quad \text{where } N = L/2
\end{equation}

This tokenization strategy provides dual benefits:
\begin{enumerate}
    \item \textbf{Complex-valued structure preservation:} Each token position corresponds to a complex sample (I/Q pair), enabling the transformer to model relationships between complex samples rather than treating I and Q components as independent entities.
    \item \textbf{Computational efficiency:} Halving the sequence length reduces the quadratic complexity of self-attention from $\mathcal{O}(L^2)$ to $\mathcal{O}((L/2)^2) = \mathcal{O}(L^2/4)$, providing a $4\times$ reduction in attention computation cost.
\end{enumerate}

\subsubsection{Cross-Domain Information Fusion}

Once tokenized, we leverage the complementary nature of time and frequency domain representations. Comprehensive signal analysis requires joint time-frequency processing—a principle long established in signal processing through techniques such as the Short-Time Fourier Transform (STFT) and wavelet analysis. These classical methods apply fixed, parametric transformations to obtain time-frequency representations.

We propose a learned alternative: our architecture functions as an \textit{end-to-end learnable and differentiable time-frequency transform in latent feature space}. Rather than fixed transformations, we employ learned domain transformations and gated fusion layers to facilitate adaptive cross-domain information flow. This enables the network to learn domain-specific features while ensuring that information learned in one domain informs processing in the other.

\subsubsection{Learned Domain Transformation}
\label{appendix:LearnedDomainTransforms}

At the conclusion of tokenization, we have two sets of domain-specific tokens with different shapes due to the aggregation strategies described in Section~\ref{appendix:domain_token_aggregation}:
\begin{align}
\text{Time Domain:} \quad &\mathbf{T}_t \in \mathbb{R}^{B \times N_t \times 2C} \\
\text{Frequency Domain:} \quad &\mathbf{T}_f \in \mathbb{R}^{B \times N_f \times 2C}
\end{align}
where $N_t = N_w \cdot (L/2)$ due to concatenation across windows, and $N_f = L/2$ due to averaging across windows.

Furthermore, domain-specific processing results in latent features that occupy different subspaces of the representation space. To enable effective cross-domain fusion, we must first align representations to compatible shapes and map them to a common latent subspace. We achieve this through learned $1 \times 1$ convolutions that simultaneously transform the latent subspace and adjust sequence lengths.

\paragraph{Time-to-Frequency Transformation.} To inject time-domain context into frequency processing:
\begin{equation}
\mathbf{T}_{t \rightarrow f} = \text{Conv}_{1 \times 1}(\mathbf{T}_t) \in \mathbb{R}^{B \times N_f \times 2C}
\end{equation}
The $1 \times 1$ convolution operates along the sequence dimension in token space (regular channel dimension for convolution), learning to compress the longer time-domain sequence ($N_t$ tokens) to match the frequency-domain length ($N_f$ tokens) while transforming the latent representation to the frequency-domain subspace.

\paragraph{Frequency-to-Time Transformation.} Conversely, to inject frequency-domain context into time processing:
\begin{equation}
\mathbf{T}_{f \rightarrow t} = \text{Conv}_{1 \times 1}(\mathbf{T}_f) \in \mathbb{R}^{B \times N_t \times 2C}
\end{equation}
Here, the convolution expands the compact frequency representation to match the time-domain sequence length while transforming to the time-domain latent subspace.

These learned transformations are fundamentally different from fixed domain transforms (e.g., FFT/IFFT). Rather than converting between time and frequency representations of the signal itself, they learn to map between latent feature subspaces, identifying which frequency-domain features are most relevant for time-domain processing and vice versa.

\subsubsection{Gated Linear Unit Fusion}
\label{appendix:GLUFusion}

Once domain transformations align the shapes, we employ Gated Linear Units (GLUs) to perform adaptive feature fusion. The gating mechanism enables learned, dynamic fusion during inference, allowing the network to selectively emphasize or suppress cross-domain information based on the input signal characteristics.

The fusion operates at the token level across the embedding dimension, providing fine-grained control over which features from each domain are combined:

\paragraph{Time-Domain Fusion.}
\begin{equation}
\mathbf{T}_{\text{fused-t}} = \text{GLU}(\text{Concat}([\mathbf{T}_t, \mathbf{T}_{f \rightarrow t}], \text{dim}=2)) \in \mathbb{R}^{B \times N_t \times 2C}
\end{equation}
where the concatenation produces $\mathbb{R}^{B \times N_t \times 4C}$ and the GLU projects back to $\mathbb{R}^{B \times N_t \times 2C}$.

\paragraph{Frequency-Domain Fusion.}
\begin{equation}
\mathbf{T}_{\text{fused-f}} = \text{GLU}(\text{Concat}([\mathbf{T}_f, \mathbf{T}_{t \rightarrow f}], \text{dim}=2)) \in \mathbb{R}^{B \times N_f \times 2C}
\end{equation}

Algorithm~\ref{alg:glu_fusion} details the GLU fusion procedure.

\begin{algorithm}[H]
\caption{Gated Linear Unit Fusion}
\label{alg:glu_fusion}
\begin{algorithmic}[1]
\REQUIRE Concatenated features $\mathbf{X} \in \mathbb{R}^{B \times N \times D_{\text{in}}}$ where $D_{\text{in}} = 4C$
\REQUIRE Target dimension $D_{\text{out}} = 2C$
\ENSURE Fused representation $\mathbf{X}_{\text{fused}} \in \mathbb{R}^{B \times N \times D_{\text{out}}}$
\STATE
\STATE \COMMENT{Normalize input features}
\STATE $\mathbf{X}_{\text{norm}} \leftarrow \text{RMSNorm}(\mathbf{X})$
\STATE
\STATE \COMMENT{Compute value and gate projections}
\STATE $\mathbf{V} \leftarrow \mathbf{X}_{\text{norm}} \mathbf{W}_v + \mathbf{b}_v$ \COMMENT{Value: $\mathbb{R}^{B \times N \times D_{\text{out}}}$}
\STATE $\mathbf{G} \leftarrow \mathbf{X}_{\text{norm}} \mathbf{W}_g + \mathbf{b}_g$ \COMMENT{Gate: $\mathbb{R}^{B \times N \times D_{\text{out}}}$}
\STATE
\STATE \COMMENT{Apply sigmoid activation to gate}
\STATE $\mathbf{G}_{\text{act}} \leftarrow \sigma(\mathbf{G})$ \COMMENT{Element-wise sigmoid}
\STATE
\STATE \COMMENT{Gated fusion}
\STATE $\mathbf{X}_{\text{fused}} \leftarrow \mathbf{V} \odot \mathbf{G}_{\text{act}}$ \COMMENT{Element-wise multiplication}
\STATE
\RETURN $\mathbf{X}_{\text{fused}}$
\end{algorithmic}
\end{algorithm}

The GLU mechanism can be expressed mathematically as:
\begin{equation}
\text{GLU}(\mathbf{X}) = (\mathbf{X} \mathbf{W}_v + \mathbf{b}_v) \odot \sigma(\mathbf{X} \mathbf{W}_g + \mathbf{b}_g)
\end{equation}
where $\mathbf{W}_v, \mathbf{W}_g \in \mathbb{R}^{D_{\text{in}} \times D_{\text{out}}}$ are learnable projection matrices, $\mathbf{b}_v, \mathbf{b}_g \in \mathbb{R}^{D_{\text{out}}}$ are bias terms, $\sigma(\cdot)$ is the sigmoid activation, and $\odot$ denotes element-wise multiplication.

The gating mechanism provides several advantages:
\begin{itemize}
    \item \textbf{Adaptive fusion:} The sigmoid gate learns to dynamically weight the contribution of each feature dimension, enabling signal-dependent fusion strategies.
    \item \textbf{Gradient flow:} The multiplicative gating provides direct gradient paths, facilitating stable training.
    \item \textbf{Interpretability:} Gate activations can be analyzed to understand which cross-domain features are emphasized for different signal types.
\end{itemize}

\subsubsection{Multi-Stage Fusion Architecture}

We apply this learned transformation and fusion process strategically at three key points in the architecture:
\begin{enumerate}
    \item \textbf{Post-tokenization:} After convolutional tokenization, before transformer processing
    \item \textbf{Inter-block:} Between each Parseval Transformer block (currently one block in our architecture)
    \item \textbf{Post-pooling:} After attentional sequence pooling, before final domain-specific embeddings
\end{enumerate}

This multi-stage fusion ensures continuous cross-domain information flow throughout the network hierarchy, enabling the model to refine its joint time-frequency representation at multiple levels of abstraction.

\subsubsection{Interpretation as Learned Time-Frequency Analysis}

At its conclusion, this process can be viewed as a \textit{learned latent spectrogram or wavelet transform}. Unlike classical STFT or wavelet analysis with fixed basis functions, our approach learns:
\begin{itemize}
    \item \textbf{Adaptive basis functions:} The convolutional tokenizers learn domain-specific feature extractors tailored to the data distribution.
    \item \textbf{Cross-domain mappings:} The $1 \times 1$ convolutions learn which time-domain features correspond to which frequency-domain features.
    \item \textbf{Fusion strategies:} The GLU gates learn how to combine time and frequency information for optimal task performance.
\end{itemize}

This end-to-end differentiable time-frequency analysis adapts to the specific characteristics of the signal domain and task, providing a more flexible and powerful alternative to fixed parametric transforms.

\subsection{The Parseval Transformer}
\label{appendix:ParsevalTransformer}
Once the tokens have been properly prepared we next utilize them within the our custom transformer architecture that we term the Parseval Transformer. Standard transformer block construction is fairly well-refined throughout the literature, and we leave it largely unmodified. Our main contributions are to the attention mechanism within the standard transformer block. First, we discuss our proposed evolution of Scaled Dot-Product Attention to the new Scaled Covariance Focus. Then we highlight how this is used both across and within time/frequency domains for a truly comprehensive learned focus mechanism that we call Parseval Focus. With these contributions,motivated by Parseval's theorem, we sought to anchor Transformer design with known physical symmetries and a more functional understanding of the time and frequency domains.

\textbf{Implementation Note:} Each domain specific branch of the PlanFormer has an independent Parseval Transformer. When computing the Multi-Head Parseval Focus mechanism it is applied within the context of that domain specific Parseval Transformer's domain specific tokens which have a complementary parametric transformation applied (e.g. Time-Domain Parseval Transformer$\rightarrow$ FFT(time-domain tokens) and Frequency-Domain Parseval Transformer$\rightarrow$ IFFT(frequency-domain tokens)). This produces the cross-domain series of tokens that match exactly in length/dimensionality which are needed for valid downstream cross-domain probability distribution comparisons. It is afterwards where each domain-specific Parseval Transformer output is fused with the other through the earlier mentioned learned domain transforms and cross domain gated fusions(Appendix~\ref{appendix:LearnedDomainTransforms}).

\subsubsection{Scaled Covariance Focus:}
\label{appendix:ScaledCovarianceFocus}
\paragraph{From Dot-Product to Covariance.}
\label{appendix:DotProcutToCovariance}
The standard attention mechanism is implemented via a learned dot product across the projections of the sequence tokens. This simple but effective computation allows a model to learn the similarity of one token to all other tokens in a sequence and weight them appropriately as a probability distribution via the softmax operation. While historically very effective, this computation is very discrete in nature. We argue this discrete computation and analysis is limiting, especially in relation to time-series information which naturally has higher-order information encoded within it. An appropriate example is how trends within time series are a function of past behavior as opposed to only how similar a present segment of time is relative to a past segment.

For time-series data, we argue that functional relationships—how tokens co-vary—are more informative than discrete similarity. We replace dot-product attention with covariance-based attention:

\[Attention(Q,K,V) = softmax\left(\frac{Cov(Q, K)}{\sqrt{d_k}}\right)V\]

Where

\[Cov(Q,K) = (Q-\mu_Q)(K-\mu_K)^T\]

This only requires centering the query and key matrices prior to computing their inner product, but now the resultant computation represents the functional relationship of a token relative to the positive, negative, or neutral change of another. This captures whether tokens exhibit positive, negative, or neutral co-variation, encoding higher-order temporal dynamics (trends, correlations) rather than instantaneous similarity. This higher-order information is extremely useful for generalizing time-series analysis and learning fundamental symmetries of the data.

\paragraph{Dynamic Focus Mechanism.}
\label{appendix:DynamicFocusMechanism}
We introduce a Focus mechanism inspired by neurocognitive load in low-SNR environments. In low-SNR scenarios, multiple stimuli may compete for attention, but many are distractions rather than the signal of interest. Think of a static television or a needle in a haystack. In these situations, effective attention requires sharpening the region one attends to dynamically.

The focus factor $K_{focus}$ is computed from attention score statistics through the following process:

\textbf{Step 1: Per-Token Variance.} Compute variance of attention scores for each token across all queries:
\[\sigma^2_i = Var_j(Cov(Q, K)_{:,i}), \quad i \in [1, L_{sequence}]\]

\textbf{Step 2: Cumulative Variance.} Sum variances across all tokens:
\[\sigma^2_{\text{cum}} = \sum_{i=1}^{L_{sequence}} \sigma^2_i\]

\textbf{Step 3: Per-Token SNR Estimation.} Compute each token's variance relative to the total:
\[\text{SNR}_i = \frac{\sigma^2_i}{\sigma^2_{\text{cum}} - \sigma^2_i + \epsilon}\]

where $\epsilon = 10^{-8}$ prevents division by zero. High $\text{SNR}_i$ indicates token $i$ has high variance relative to others (distinctive), while low $\text{SNR}_i$ indicates low relative variance (not distinctive).

\textbf{Step 4: Summary Statistics.} Extract two features from the SNR distribution:
\begin{align}
\text{mean\_snr} &= \frac{1}{L_{sequence}} \sum_{i=1}^{L_{sequence}} \text{SNR}_i \\
\text{high\_snr\_prop} &= \frac{1}{L_{sequence}} \sum_{i=1}^{L_{sequence}} \mathbb{1}[\text{SNR}_i > \text{median}(\{\text{SNR}_j\}_{j=1}^{L_{sequence}})]
\end{align}

where $\text{mean\_snr}$ captures the average distinctiveness across tokens, and $\text{high\_snr\_prop}$ captures the proportion of tokens with above-median distinctiveness.

\textbf{Step 5: Learned Focus Prediction.} Concatenate statistics and predict focus factor:
\[\mathbf{k}_{\text{input}} = [\text{mean\_snr}, \text{high\_snr\_prop}] \in \mathbb{R}^2\]
\[K_{focus} = 1 + (L_{sequence}-1) \cdot \sigma(f_\theta(\mathbf{k}_{\text{input}}))\]

where $f_\theta: \mathbb{R}^2 \rightarrow \mathbb{R}$ is a lightweight MLP (2 hidden layers, 64 units each, ReLU in between) and $\sigma(\cdot)$ is the sigmoid function.

The effective temperature is $\tau = K_{focus}/L_{sequence} \in [1/L_{sequence}, 1]$. The network learns to map attention score distributions to appropriate temperature scaling: when attention is uncertain (high mean\_snr, many tokens competing), $K_{focus}$ is low ($\tau$ small) to sharpen the distribution; when attention is confident (low mean\_snr, few clear winners), $K_{focus}$ is high ($\tau \approx 1$) to maintain standard attention. This adaptive mechanism enables dynamic sparsity control based on the confidence of the attention distribution.

The complete Focus mechanism is:
\[Focus(Q, K, V) = softmax\left(\frac{Cov(Q, K)\cdot L_{sequence}}{\sqrt{d_k}\cdot K_{focus}}\right)V\]

\subsubsection{Head Orthogonalization Regularization}
\label{appendix:head_orth}

\paragraph{Motivation and Approach}

The multi-head attention mechanism is designed to allow different heads to attend to distinct aspects of the input representations~\cite{vaswani2023attentionneed}. However, without explicit enforcement, heads often learn redundant attention patterns. Recent work on Differential Transformers~\cite{ye2024differential} has demonstrated that common-mode noise is prevalent across attention heads, reducing the effective capacity of the multi-head mechanism.

To address this, we introduce a head orthogonalization regularization loss applied during training of every Multi-Head Scaled Covariance Self-Focus layer. Our loss explicitly encourages head specialization by minimizing the covariance between attention distributions of different heads while maintaining sufficient variance within each head.

\paragraph{Mathematical Formulation}

Given attention weights $\mathbf{A} \in \mathbb{R}^{B \times H \times L \times L}$ where $B$ is batch size, $H$ is the number of heads, and $L$ is sequence length, we compute the head orthogonalization loss as follows.

First, we reshape and transpose the attention weights:
\begin{equation}
\tilde{A} = \text{reshape}(\text{permute}(A, [0, 2, 1, 3]), [BL, H, L])
\end{equation}
where $\text{permute}(A, [0, 2, 1, 3])$ reorders dimensions from $(B, H, L_q, L_k)$ to $(B, L_q, H, L_k)$, grouping batch and query-sequence dimensions before reshaping to $\mathbb{R}^{BL \times H \times L}$.
We then center each attention head by subtracting its mean across the sequence dimension:
\begin{equation}
\bar{\mathbf{A}} = \tilde{\mathbf{A}} - \mathbb{E}_{l}[\tilde{\mathbf{A}}]
\end{equation}
where $\mathbb{E}_{l}[\cdot]$ denotes the mean over the sequence length dimension. This centering operation transforms the subsequent inner product into a covariance computation, allowing us to measure statistical dependence between attention heads.

The similarity matrix between heads is computed as:
\begin{equation}
\mathbf{S} = |\bar{\mathbf{A}} \bar{\mathbf{A}}^T| \in \mathbb{R}^{BL \times H \times H}
\end{equation}

We decompose $\mathbf{S}$ into diagonal and off-diagonal components:
\begin{align}
\mathbf{D} &= \text{diag}(\mathbf{S}) \\
\mathbf{S}_{\text{off}} &= \mathbf{S} - \mathbf{D}
\end{align}

The head orthogonalization loss consists of two terms:
\begin{equation}
\mathcal{L}_{\text{orth}} = \mathbb{E}[|\mathbf{S}_{\text{off}}|] + \mathbb{E}[\text{ReLU}(1 - \sqrt{\mathbf{D} + \epsilon})]
\label{eq:orth_loss}
\end{equation}
where $\epsilon = 10^{-4}$ is a small constant for numerical stability. The first term minimizes covariance between different heads, encouraging orthogonalization. The second term penalizes heads with low variance (standard deviation below 1) to prevent degeneracy, ensuring each head maintains meaningful attention patterns rather than collapsing to uniform distributions.

\paragraph{Preventing Projection Degeneration with SoftAbsFloor}

A critical challenge with orthogonalization-based regularization is the risk of over-regularization, which can cause query and key projections to collapse toward zero, resulting in uniform attention distributions. To mitigate this, we apply a \texttt{SoftAbsFloor} activation function to the outputs of the $\mathbf{Q}$ and $\mathbf{K}$ projection layers:

\begin{equation}
\text{SoftAbsFloor}(x; \epsilon) = x + \text{sign}(x) \cdot \epsilon \cdot \sigma\left(-\frac{|x|}{\epsilon}\right)
\label{eq:soft_abs_floor}
\end{equation}

where $\sigma(\cdot)$ is the sigmoid function and $\epsilon = 10^{-4}$ controls the floor magnitude. This function behaves approximately as the identity for $|x| \gg \epsilon$ and smoothly approaches $\pm\epsilon$ for $|x| \approx 0$. The smooth transition preserves gradient flow while preventing complete collapse of the projections, maintaining the benefits of orthogonalization without the pathological degeneration that can occur with hard constraints.

\textbf{Behavior Analysis.} For large magnitudes, $\sigma(-|x|/\epsilon) \approx 0$, so $\text{SoftAbsFloor}(x) \approx x$. For small magnitudes, $\sigma(-|x|/\epsilon) \approx 1$, providing an effective floor at $\pm\epsilon$. The transition sharpness is controlled by $1/\epsilon$, creating a nearly hard floor while maintaining full differentiability.

\paragraph{Implementation Details}

The head orthogonalization loss (Eq.~\ref{eq:orth_loss}) is computed at each attention layer and aggregated across layers. The SoftAbsFloor activation (Eq.~\ref{eq:soft_abs_floor}) is applied element-wise to query and key projections before the attention computation:

\begin{align}
\mathbf{Q}' &= \text{SoftAbsFloor}(\mathbf{W}_Q \mathbf{X}) \\
\mathbf{K}' &= \text{SoftAbsFloor}(\mathbf{W}_K \mathbf{X})
\end{align}

where $\mathbf{W}_Q, \mathbf{W}_K$ are the learned projection matrices and $\mathbf{X}$ is the input. The value projection $\mathbf{V}$ does not require this activation, as it is not involved in the attention score computation and thus not subject to the same collapse dynamics.

\subsection{Multi-Head Parseval Focus Mechanism}
\label{appendix:MHPF}

\subsubsection{Theoretical Foundation}
\label{appendix:MHPF_TheoreticalFoundation}

We now extend the Scaled Covariance Focus mechanism to leverage a fundamental principle from signal processing: Parseval's theorem~\cite{parseval1799}. Parseval's theorem establishes that energy is conserved between time and frequency domains:

\begin{equation}
\sum_{n} |x[n]|^2 = \frac{1}{N} \sum_{k} |X[k]|^2
\end{equation}

This energy conservation represents a fundamental invariance symmetry—the total information content of a signal is preserved across domain transformations. We extend this principle to learned representations: \textit{physically consistent features should exhibit predictable, symmetric relationships across time and frequency domains}.

The Multi-Head Parseval Focus mechanism enforces this consistency through a novel cross-domain attention architecture with Parseval-based regularization. Rather than treating time and frequency representations independently, we explicitly model their bidirectional relationships and penalize physically inconsistent attention patterns.

\subsubsection{Architecture Overview}
\label{appendix:MHPF_ArchitectureOverview}

The Multi-Head Parseval Focus mechanism integrates three complementary attention operations:

\begin{enumerate}
    \item \textbf{In-Domain Self-Focus:} Captures domain-specific patterns (temporal ordering in time, spectral periodicity in frequency)
    \item \textbf{Cross-Domain Parseval Focus:} Models bidirectional time-frequency relationships with energy conservation constraints
    \item \textbf{Strategic Fusion:} Combines in-domain and cross-domain representations via gated mechanisms
\end{enumerate}

\paragraph{Domain Preparation}

Given time-domain tokens $\mathbf{T}_t \in \mathbb{R}^{B \times N_t \times d}$ and frequency-domain tokens $\mathbf{T}_f \in \mathbb{R}^{B \times N_f \times d}$ from the dual-domain tokenization pipeline, we first prepare both representations in both domains:

\begin{align}
\mathbf{T}_t^{(freq)} &= \text{FFT}(\mathbf{T}_t) \quad \text{(time tokens in frequency domain)} \\
\mathbf{T}_f^{(time)} &= \text{IFFT}(\mathbf{T}_f) \quad \text{(frequency tokens in time domain)}
\end{align}

These transformations leverage our interleaved IQ representation: we de-interleave the embedding dimension to obtain complex-valued tokens, apply FFT/IFFT, and re-interleave for subsequent real-valued processing.

\subsubsection{In-Domain Self-Focus}
\label{appendix:MHPF_InDomainSelfFocus}

Before computing cross-domain relationships, we capture domain-specific patterns through Scaled Covariance Self-Focus (Section~\ref{appendix:ScaledCovarianceFocus}) within each domain:

\textbf{Time-Domain Self-Focus:}
\begin{equation}
\mathbf{O}_{\text{self}}^{(t)} = \text{Focus}(\mathbf{Q}_t^{(self)}, \mathbf{K}_t^{(self)}, \mathbf{V}_t^{(self)})
\end{equation}
where $\mathbf{Q}_t^{(self)}, \mathbf{K}_t^{(self)}, \mathbf{V}_t^{(self)}$ are projections of $\mathbf{T}_f^{(time)}$ (frequency tokens transformed to time domain).

\textbf{Frequency-Domain Self-Focus:}
\begin{equation}
\mathbf{O}_{\text{self}}^{(f)} = \text{Focus}(\mathbf{Q}_f^{(self)}, \mathbf{K}_f^{(self)}, \mathbf{V}_f^{(self)})
\end{equation}
where $\mathbf{Q}_f^{(self)}, \mathbf{K}_f^{(self)}, \mathbf{V}_f^{(self)}$ are projections of $\mathbf{T}_t^{(freq)}$ (time tokens transformed to frequency domain).

\textbf{Rationale:} In-domain self-focus captures patterns that are naturally expressed within a single domain: temporal ordering, causal dependencies, and sequential structure in time; spectral periodicity, harmonic relationships, and frequency correlations in frequency. These domain-specific patterns are complementary to cross-domain relationships and essential for comprehensive signal analysis.

\subsubsection{Cross-Domain Parseval Focus}
\label{appendix:MHPF_CrossDomainFocus}

The core innovation of Multi-Head Parseval Focus is the bidirectional cross-domain attention mechanism with energy conservation constraints.

\paragraph{Bidirectional Cross-Domain Attention}

We compute attention in both directions:

\textbf{Time-Query, Frequency-Key (T$\rightarrow$F):}
\begin{align}
\mathbf{Q}_{tf} &= \text{SoftAbsFloor}(\mathbf{W}_Q^{(tf)} \mathbf{T}_t) \\
\mathbf{K}_{tf} &= \text{SoftAbsFloor}(\mathbf{W}_K^{(tf)} \mathbf{T}_f) \\
\mathbf{V}_{tf} &= \mathbf{W}_V^{(tf)} \mathbf{T}_f \\
\mathbf{S}_{tf} &= \frac{\text{Cov}(\mathbf{Q}_{tf}, \mathbf{K}_{tf})}{\sqrt{d_k} \cdot K_{\text{focus}}^{(tf)}}
\end{align}

\textbf{Frequency-Query, Time-Key (F$\rightarrow$T):}
\begin{align}
\mathbf{Q}_{ft} &= \text{SoftAbsFloor}(\mathbf{W}_Q^{(ft)} \mathbf{T}_f) \\
\mathbf{K}_{ft} &= \text{SoftAbsFloor}(\mathbf{W}_K^{(ft)} \mathbf{T}_t) \\
\mathbf{V}_{ft} &= \mathbf{W}_V^{(ft)} \mathbf{T}_t \\
\mathbf{S}_{ft} &= \frac{\text{Cov}(\mathbf{Q}_{ft}, \mathbf{K}_{ft})}{\sqrt{d_k} \cdot K_{\text{focus}}^{(ft)}}
\end{align}

where $K_{\text{focus}}$ is computed per Section~\ref{appendix:DynamicFocusMechanism}.

\paragraph{Parseval Consistency via Jensen-Shannon Distance}

For physically consistent representations, both cross-domain permutations should produce equivalent attention distributions. We enforce this through Jensen-Shannon Distance (JSD), a symmetric, bounded divergence measure.

We compute attention distributions with consistent normalization:
\begin{align}
\mathbf{P}_{tf} = \text{softmax}_{axis=-1}(S_{tf}) \in \mathbb{R}^{B \times H \times N_t \times N_f} \\
\mathbf{P}_{ft} = \text{softmax}_{axis=-1}(S_{ft}) \in \mathbb{R}^{B \times H \times N_f \times N_t} \\
\end{align}

To align dimensions for comparison, we transpose the score matrices \emph{before} applying softmax:

\begin{align}
\tilde{P}_{ft} = \text{softmax}_{axis=-1}(S_{ft}^T) \in \mathbb{R}^{B \times H \times N_t \times N_f}  \\
\tilde{P}_{tf} = \text{softmax}_{axis=-1}(S_{tf}^T) \in \mathbb{R}^{B \times H \times N_f \times N_t}  \\ 
\end{align}

Now both $P_{tf}$ and $\tilde{P}_{ft}$ are valid probability distributions over the same event space (time tokens attending to frequency tokens), and similarly for the reverse direction. We compute bidirectional Jensen-Shannon Distance:

\textbf{Time-to-Frequency Direction:}
\begin{align}
\mathbf{M}_{tf} = \frac{1}{2}(P_{tf} + \tilde{P}_{ft})  \\
\mathbf{KL}_{tf \to M} = \sum_{i,j} P_{tf}[i,j] \log \frac{P_{tf}[i,j] + \epsilon}{M_{tf}[i,j] + \epsilon}  \\
\mathbf{KL}_{\tilde{ft} \to M} = \sum_{i,j} \tilde{P}_{ft}[i,j] \log \frac{\tilde{P}_{ft}[i,j] + \epsilon}{M_{tf}[i,j] + \epsilon}  \\
\mathbf{JSD}_{time} = \frac{1}{2}(KL_{tf \to M} + KL_{\tilde{ft} \to M})  \\
\end{align}

\textbf{Frequency-to-Time Direction:}
\begin{align}
\mathbf{M}_{ft} = \frac{1}{2}(P_{ft} + \tilde{P}_{tf})  \\
\mathbf{JSD}_{freq} = \frac{1}{2}(KL_{ft \to M} + KL_{\tilde{tf} \to M})  \\
\end{align}

The JSD reweighting factors are computed as:
\begin{align}
\mathbf{w}_{time} = (1 - JSD_{time})  \\
\mathbf{w}_{freq} = (1 - JSD_{freq})  \\
\end{align}
    
\textbf{Rationale:} By transposing before softmax normalization, we ensure both distributions represent the same conditional probability structure (e.g., $P(freq|time)$), making the JSD mathematically well-defined. High JSD indicates inconsistent cross-domain relationships, which we down-weight; low JSD indicates consistent relationships, which we amplify.

\textbf{Dual Role of JSD:} The JSD serves two complementary purposes:

\textbf{(1) Regularization Loss:}
\begin{equation}
\mathcal{L}_{\text{Parseval}} = \mathbb{E}[\text{JSD}_{time}(\mathbf{P}_{tf} || \tilde{\mathbf{P}}_{ft})] + \mathbb{E}[\text{JSD}_{freq}(\mathbf{P}_{ft} || \tilde{\mathbf{P}}_{tf})]
\end{equation}
Minimizing this bidirectional loss encourages the model to learn cross-domain relationships that are consistent regardless of which domain serves as query or key—a direct analog of Parseval's energy conservation. The tilde notation ($\tilde{\mathbf{P}}$) indicates distributions computed by transposing scores before softmax normalization, ensuring both distributions share the same probability space for valid divergence computation.

\textbf{(2) Dynamic Reweighting:}
\begin{align}
w_{\text{Parseval}}^{time} &= 1 - \text{JSD}_{time} \in [0, 1] \\
w_{\text{Parseval}}^{freq} &= 1 - \text{JSD}_{freq} \in [0, 1] \\
\mathbf{S}_{tf}^{\text{reweighted}} &= \mathbf{S}_{tf} \cdot w_{\text{Parseval}}^{time} \\
\mathbf{S}_{ft}^{\text{reweighted}} &= \mathbf{S}_{ft} \cdot w_{\text{Parseval}}^{freq}
\end{align}

This reweighting amplifies attention patterns that exhibit Parseval consistency (low JSD) and suppresses physically inconsistent patterns (high JSD). Each direction (time-to-frequency and frequency-to-time) is weighted independently based on its respective consistency. The mechanism is fully differentiable and learned end-to-end.

\textbf{Physical Interpretation:} From a signal processing perspective, Parseval consistency ensures that the attention mechanism respects the fundamental duality between time and frequency domains. If a time-domain feature strongly attends to a frequency-domain feature, the reverse relationship should hold with equivalent strength—just as energy in a time-domain signal component corresponds to energy in its frequency-domain counterpart.

\paragraph{Cross-Domain Focus Output}

After JSD reweighting, we compute the final cross-domain attended representations:

\begin{align}
\mathbf{O}_{\text{cross}}^{(t)} &= \mathbf{S}_{tf}^{\text{reweighted}}\cdot\mathbf{V}_{tf} \in \mathbb{R}^{B \times H \times N_t \times d_k} \\
\mathbf{O}_{\text{cross}}^{(f)} &= \mathbf{S}_{ft}^{\text{reweighted}}\cdot\mathbf{V}_{ft} \in \mathbb{R}^{B \times H \times N_f \times d_k}
\end{align}

Notably, we utilize the domain specific parseval reweighted score matrices as is rather than wrapping them in a softmax normalization. By only utilizing parseval reweighting this provides a path to mitigate the over-smoothing potential of the softmax normalization procedure and instead provides a physically meaningful gating to the attention score matrix that is then multiplied against V.

\subsubsection{Strategic Fusion and Diversity Regularization}
\label{appendix:MHPF_Fusion}

To produce comprehensive signal representations, we fuse in-domain self-focus with cross-domain Parseval focus.

\paragraph{Domain-Specific Fusion}

For each domain, we concatenate self-focus and cross-focus outputs and apply gated fusion:

\textbf{Time Domain:}
\begin{align}
\mathbf{O}_{\text{concat}}^{(t)} &= \text{Concat}([\mathbf{O}_{\text{self}}^{(t)}, \mathbf{O}_{\text{cross}}^{(t)}], \text{dim}=-1) \in \mathbb{R}^{B \times N_t \times 2d} \\
\mathbf{O}_{\text{fused}}^{(t)} &= \text{GLU}(\mathbf{O}_{\text{concat}}^{(t)}) \in \mathbb{R}^{B \times N_t \times d}
\end{align}

\textbf{Frequency Domain:}
\begin{align}
\mathbf{O}_{\text{concat}}^{(f)} &= \text{Concat}([\mathbf{O}_{\text{self}}^{(f)}, \mathbf{O}_{\text{cross}}^{(f)}], \text{dim}=-1) \in \mathbb{R}^{B \times N_f \times 2d} \\
\mathbf{O}_{\text{fused}}^{(f)} &= \text{GLU}(\mathbf{O}_{\text{concat}}^{(f)}) \in \mathbb{R}^{B \times N_f \times d}
\end{align}

\paragraph{Diversity Regularization}

A critical risk in this architecture is that self-focus and cross-focus mechanisms could learn redundant representations, wasting capacity. We enforce diversity through an orthogonality regularization loss:

\begin{align}
\bar{\mathbf{O}}_{\text{self}}^{(t)} &= \frac{\mathbf{O}_{\text{self}}^{(t)}}{\|\mathbf{O}_{\text{self}}^{(t)}\|_2 + \epsilon} \quad \text{(normalize)} \\
\bar{\mathbf{O}}_{\text{cross}}^{(t)} &= \frac{\mathbf{O}_{\text{cross}}^{(t)}}{\|\mathbf{O}_{\text{cross}}^{(t)}\|_2 + \epsilon} \\
\mathcal{L}_{\text{diversity}}^{(t)} &= \left(\sum_{i} \bar{\mathbf{O}}_{\text{self}}^{(t)}[i] \cdot \bar{\mathbf{O}}_{\text{cross}}^{(t)}[i]\right)^2
\end{align}

Similarly for frequency domain. The total diversity loss is:
\begin{equation}
\mathcal{L}_{\text{diversity}} = \frac{1}{2}\left(\mathcal{L}_{\text{diversity}}^{(t)} + \mathcal{L}_{\text{diversity}}^{(f)}\right) + \lambda_{\text{norm}} \mathcal{L}_{\text{norm}}
\end{equation}

where $\mathcal{L}_{\text{norm}}$ penalizes representations with insufficient magnitude:
\begin{equation}
\mathcal{L}_{\text{norm}} = \sum_{X \in \{\mathbf{O}_{\text{self}}^{(t)}, \mathbf{O}_{\text{cross}}^{(t)}, \mathbf{O}_{\text{self}}^{(f)}, \mathbf{O}_{\text{cross}}^{(f)}\}} \text{ReLU}(\sqrt{d} - \|X\|_2)
\end{equation}

This prevents the trivial solution where representations collapse to zero to satisfy orthogonality.

\paragraph{Final Cross-Domain Fusion}

Finally, we fuse the time and frequency domain representations:
\begin{align}
\mathbf{O}_{\text{final}} &= \text{GLU}(\text{Concat}([\mathbf{O}_{\text{fused}}^{(t)}, \mathbf{O}_{\text{fused}}^{(f)}], \text{dim}=-1))
\end{align}

\subsubsection{Complete Multi-Head Parseval Focus Algorithm}

Algorithm~\ref{alg:mhpf} summarizes the complete forward pass.

\begin{algorithm}[H]
\caption{Multi-Head Parseval Focus}
\label{alg:mhpf}
\begin{algorithmic}[1]
\REQUIRE Time tokens $\mathbf{T}_t$, Frequency tokens $\mathbf{T}_f$
\ENSURE Fused output $\mathbf{O}_{\text{final}}$, Losses $\{\mathcal{L}_{\text{orth}}, \mathcal{L}_{\text{Parseval}}, \mathcal{L}_{\text{diversity}}\}$
\STATE
\STATE \COMMENT{\textbf{Domain Preparation}}
\STATE $\mathbf{T}_t^{(freq)} \leftarrow \text{FFT}(\mathbf{T}_t)$
\STATE $\mathbf{T}_f^{(time)} \leftarrow \text{IFFT}(\mathbf{T}_f)$
\STATE
\STATE \COMMENT{\textbf{In-Domain Self-Focus}}
\STATE $\mathbf{O}_{\text{self}}^{(t)}, \mathcal{L}_{\text{orth}}^{(t)} \leftarrow \text{ScaledCovarianceSelfFocus}(\mathbf{T}_f^{(time)})$
\STATE $\mathbf{O}_{\text{self}}^{(f)}, \mathcal{L}_{\text{orth}}^{(f)} \leftarrow \text{ScaledCovarianceSelfFocus}(\mathbf{T}_t^{(freq)})$
\STATE
\STATE \COMMENT{\textbf{Cross-Domain Parseval Focus}}
\STATE $\mathbf{O}_{\text{cross}}^{(t)}, \mathbf{O}_{\text{cross}}^{(f)}, \mathcal{L}_{\text{orth}}^{(cross)}, \mathcal{L}_{\text{Parseval}} \leftarrow$
\STATE \hspace{2cm} $\text{ScaledCovarianceParsevalAttention}(\mathbf{T}_t, \mathbf{T}_f)$
\STATE
\STATE \COMMENT{\textbf{Domain-Specific Fusion}}
\STATE $\mathbf{O}_{\text{fused}}^{(t)} \leftarrow \text{GLU}(\text{Concat}([\mathbf{O}_{\text{self}}^{(t)}, \mathbf{O}_{\text{cross}}^{(t)}]))$
\STATE $\mathbf{O}_{\text{fused}}^{(f)} \leftarrow \text{GLU}(\text{Concat}([\mathbf{O}_{\text{self}}^{(f)}, \mathbf{O}_{\text{cross}}^{(f)}]))$
\STATE
\STATE \COMMENT{\textbf{Diversity Regularization}}
\STATE $\mathcal{L}_{\text{diversity}} \leftarrow \text{ComputeDiversityLoss}(\mathbf{O}_{\text{self}}^{(t)}, \mathbf{O}_{\text{cross}}^{(t)}, \mathbf{O}_{\text{self}}^{(f)}, \mathbf{O}_{\text{cross}}^{(f)})$
\STATE
\STATE \COMMENT{\textbf{Final Cross-Domain Fusion}}
\STATE $\mathbf{O}_{\text{final}} \leftarrow \text{GLU}(\text{Concat}([\mathbf{O}_{\text{fused}}^{(t)}, \mathbf{O}_{\text{fused}}^{(f)}]))$
\STATE
\STATE $\mathcal{L}_{\text{orth}} \leftarrow \mathcal{L}_{\text{orth}}^{(t)} + \mathcal{L}_{\text{orth}}^{(f)} + \mathcal{L}_{\text{orth}}^{(cross)}$
\STATE
\RETURN $\mathbf{O}_{\text{final}}, \{\mathcal{L}_{\text{orth}}, \mathcal{L}_{\text{Parseval}}, \mathcal{L}_{\text{diversity}}\}$
\end{algorithmic}
\end{algorithm}

\subsubsection{Summary of Regularization Losses}

Table~\ref{tab:parseval_losses} summarizes all regularization losses in the Multi-Head Parseval Focus mechanism.

\begin{table}[h]
\centering
\caption{Multi-Head Parseval Focus Regularization Losses}
\label{tab:parseval_losses}
\begin{tabular}{p{4cm}p{7cm}p{3cm}}
\toprule
\textbf{Loss} & \textbf{Purpose} & \textbf{Weight} \\
\midrule
$\mathcal{L}_{\text{orth}}$ & Encourage diverse attention heads (Section~\ref{appendix:head_orth}) & $\lambda_{\text{orth}}$ \\
\midrule
$\mathcal{L}_{\text{Parseval}}$ & Enforce cross-domain consistency via JSD & $\lambda_{\text{Parseval}}$ \\
\midrule
$\mathcal{L}_{\text{diversity}}$ & Prevent redundancy between self-focus and cross-focus & $\lambda_{\text{diversity}}$ \\
\bottomrule
\end{tabular}
\end{table}

\subsubsection{Design Rationale and Physical Grounding}

\paragraph{Why Bidirectional Cross-Domain Attention?}

Parseval's theorem establishes a symmetric relationship between time and frequency domains. Our bidirectional attention architecture directly models this symmetry: if time-domain feature $i$ relates to frequency-domain feature $j$, the reverse relationship should hold with equivalent strength. The JSD regularization enforces this consistency.

\paragraph{Why JSD Instead of KL Divergence?}

KL divergence is asymmetric: $\text{KL}(P||Q) \neq \text{KL}(Q||P)$. For Parseval consistency, we require a symmetric measure. JSD is the symmetrized, bounded variant of KL divergence:
\begin{equation}
\text{JSD}(P||Q) = \frac{1}{2}\text{KL}(P||M) + \frac{1}{2}\text{KL}(Q||M), \quad M = \frac{1}{2}(P + Q)
\end{equation}
satisfying $\text{JSD}(P||Q) = \text{JSD}(Q||P)$ and $0 \leq \text{JSD} \leq 1$.

\paragraph{Why In-Domain Self-Focus + Cross-Domain Focus?}

Comprehensive signal analysis requires both:
\begin{itemize}
    \item \textbf{Domain-specific patterns:} Temporal ordering, causality (time); spectral periodicity, harmonics (frequency)
    \item \textbf{Cross-domain relationships:} How time-domain events manifest in frequency, and vice versa
\end{itemize}

Self-focus captures the former, cross-focus the latter. The diversity regularization ensures they learn complementary (not redundant) representations.

\paragraph{Connection to Classical Signal Processing}

Multi-Head Parseval Focus can be viewed as a learned, adaptive generalization of classical time-frequency analysis:
\begin{itemize}
    \item \textbf{STFT/Spectrogram:} Fixed windows, fixed basis functions
    \item \textbf{Wavelet Transform:} Fixed mother wavelet, fixed time-frequency resolution trade-off
    \item \textbf{Parseval Focus:} Learned windows (tokenization), learned basis functions (projections), learned time-frequency relationships (cross-domain attention), adaptive resolution (dynamic focus)
\end{itemize}

The Parseval consistency constraint ensures the learned transform respects fundamental physical principles while adapting to task-specific requirements.

\begin{figure}
    \centering
    \includegraphics[width=\textwidth, angle = -90]{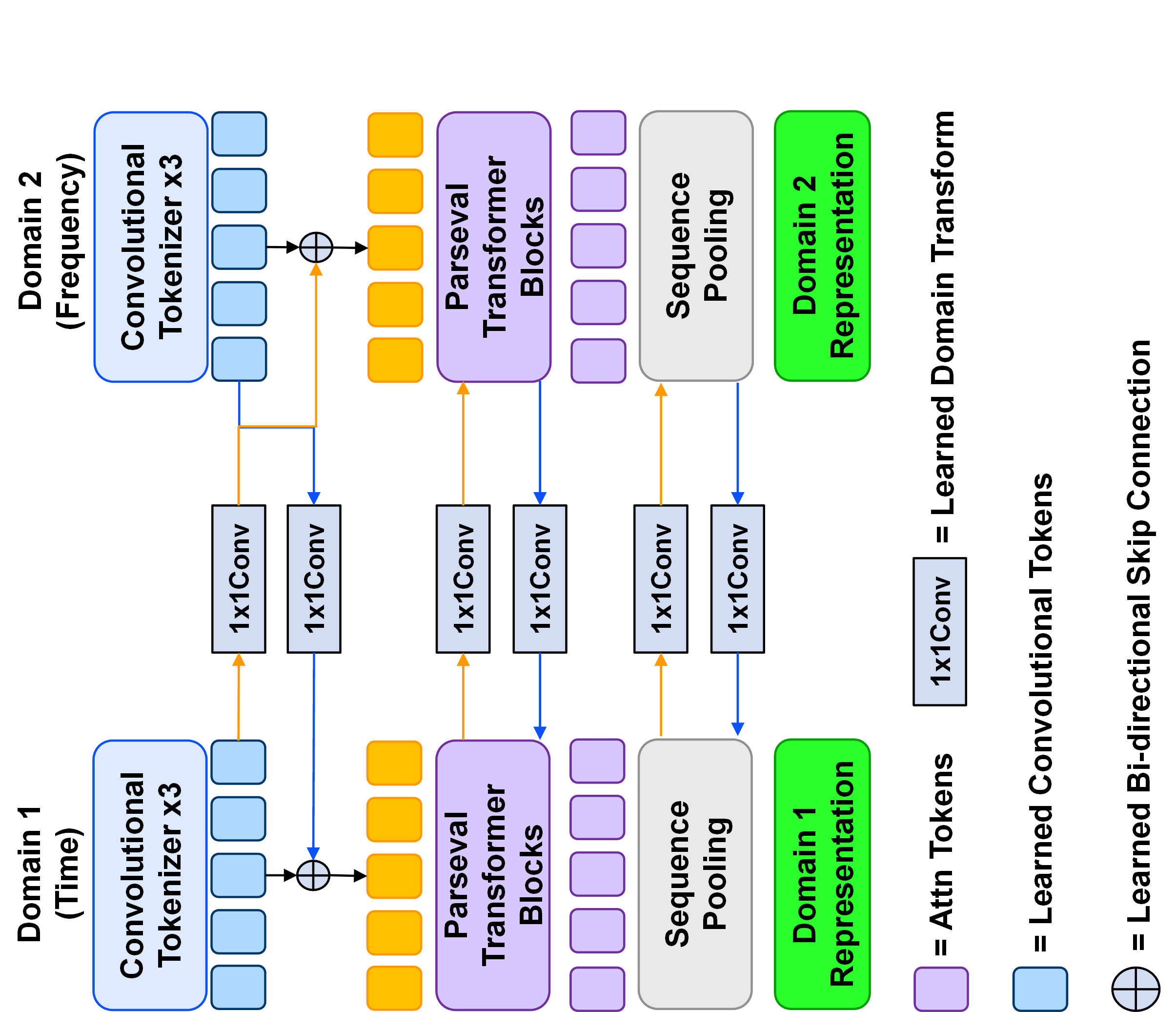}
    \caption{High Level Block Diagram of PlanFormer Encoder: Starting with domain specific transformations that are then tokenized via learned convolutions$\rightarrow$turned into tokens$\rightarrow$domain specific tokens receive cross-domain context injection and then are processed via the Parseval Transformer$\rightarrow$ post Parseval Transformer tokens recieve cross domain context injection prior to sequence pooling $\rightarrow$ one last round of cross-domain information injection before final encodings are produced per domain. }
    \label{fig:HighLevelPlanFormer}
\end{figure}

\subsection{Attentional Sequence Pooling}
\label{appendix:SequencePooling}

Following the Parseval Transformer blocks and final cross-domain fusion, we must aggregate the variable-length token sequences into fixed-size latent representations suitable for downstream tasks. We employ attentional sequence pooling~\cite{hassani2021escaping}, which learns to weight tokens by their global relevance rather than applying uniform pooling.

\subsubsection{Linear Attention Pooling Mechanism}

Given token sequences from each domain after the final cross-domain fusion:
\begin{align}
\mathbf{T}_t^{\text{final}} &\in \mathbb{R}^{B \times N_t \times d} \quad \text{(time domain)} \\
\mathbf{T}_f^{\text{final}} &\in \mathbb{R}^{B \times N_f \times d} \quad \text{(frequency domain)}
\end{align}

we compute domain-specific pooled representations through learned attention weights.

\paragraph{Pooling Procedure (per domain):}

\begin{algorithm}[H]
\caption{Attentional Sequence Pooling}
\label{alg:seq_pooling}
\begin{algorithmic}[1]
\REQUIRE Token sequence $\mathbf{T} \in \mathbb{R}^{B \times N \times d}$
\ENSURE Pooled representation $\mathbf{z} \in \mathbb{R}^{B \times d}$
\STATE
\STATE \COMMENT{\textbf{Step 1: Normalize}}
\STATE $\mathbf{T}_{\text{norm}} \leftarrow \text{RMSNorm}(\mathbf{T})$
\STATE
\STATE \COMMENT{\textbf{Step 2: Compute attention logits}}
\STATE $\mathbf{a}_{\text{logits}} \leftarrow \mathbf{W}_a \mathbf{T}_{\text{norm}} + \mathbf{b}_a$ \COMMENT{$\in \mathbb{R}^{B \times N \times 1}$}
\STATE
\STATE \COMMENT{\textbf{Step 3: Softmax over sequence dimension}}
\STATE $\mathbf{a} \leftarrow \text{softmax}(\mathbf{a}_{\text{logits}}, \text{dim}=1)$ \COMMENT{$\in \mathbb{R}^{B \times N \times 1}$}
\STATE
\STATE \COMMENT{\textbf{Step 4: Weighted aggregation}}
\STATE $\mathbf{z} \leftarrow \sum_{i=1}^{N} \mathbf{a}[i] \cdot \mathbf{T}[i]$ \COMMENT{$= \mathbf{a}^T \mathbf{T} \in \mathbb{R}^{B \times d}$}
\STATE
\RETURN $\mathbf{z}$
\end{algorithmic}
\end{algorithm}

Mathematically, for each domain:
\begin{align}
\mathbf{T}_{\text{norm}} &= \text{RMSNorm}(\mathbf{T}) \\
\mathbf{a}_{\text{logits}} &= \mathbf{W}_a \mathbf{T}_{\text{norm}} + \mathbf{b}_a \in \mathbb{R}^{B \times N \times 1} \\
\mathbf{a} &= \text{softmax}(\mathbf{a}_{\text{logits}}, \text{dim}=1) \\
\mathbf{z} &= \sum_{i=1}^{N} a_i \mathbf{T}_i = \mathbf{a}^T \mathbf{T} \in \mathbb{R}^{B \times d}
\end{align}

where $\mathbf{W}_a \in \mathbb{R}^{1 \times d}$ and $\mathbf{b}_a \in \mathbb{R}$ are learned parameters.

\textbf{Computational Efficiency:} This linear attention mechanism scales as $O(Nd)$ compared to $O(N^2d)$ for standard self-attention, making it suitable for long sequences. The pooler learns a \textbf{global token importance distribution} over the $N$ tokens, which is applied uniformly across all $d$ features. Specifically, $\mathbf{W}_a \in \mathbb{R}^{1 \times d}$ projects each token to a scalar attention logit, producing a single attention distribution $\mathbf{a} \in \mathbb{R}^{N \times 1}$ that determines which tokens are most relevant for the overall representation. The resulting latent $\mathbf{z} \in \mathbb{R}^d$ is computed as $\mathbf{z} = \sum_{i=1}^{N} a_i \mathbf{T}_i$, where each dimension of $\mathbf{z}$ is a weighted combination of the corresponding feature across all tokens.

\subsubsection{Domain-Specific Pooled Representations}

Applying this procedure to both domains yields:
\begin{align}
\mathbf{z}_t &= \text{AttentionalPool}(\mathbf{T}_t^{\text{final}}) \in \mathbb{R}^{B \times d} \\
\mathbf{z}_f &= \text{AttentionalPool}(\mathbf{T}_f^{\text{final}}) \in \mathbb{R}^{B \times d}
\end{align}

These fixed-size representations encode:
\begin{itemize}
    \item $\mathbf{z}_t$: Global temporal structure, causal dependencies, time-domain events
    \item $\mathbf{z}_f$: Global spectral structure, frequency content, harmonic relationships
\end{itemize}

\subsubsection{Semantic Interpretation: Relative Globality}

A key insight is that "global" is relative to the input sequence length:

\textbf{Short Sequences ($N \sim 100$):} The pooled representation captures local, fine-grained descriptions. For example, in speech processing, this might encode phoneme-level information.

\textbf{Long Sequences ($N \sim 1000$):} The pooled representation summarizes global structure. In the same speech example, this would encode sentence-level semantics or speaker characteristics.

This scale-adaptive behavior emerges naturally from the windowed tokenization and attentional pooling: the network learns to aggregate information at the appropriate level of abstraction for the given sequence length.

\subsection{Dual-Domain Latent Representation}
\label{appendix:DualDomainRepresentation}

Figure~\ref{fig:HighLevelPlanFormer} illustrates the complete encoder pipeline, from dual-domain preprocessing through tokenization, Parseval Transformer processing, and final sequence pooling.

At the conclusion of encoder processing, we obtain a pair of complementary fixed-size latent representations:
\begin{equation}
\mathbf{z} = \{\mathbf{z}_t, \mathbf{z}_f\} \quad \text{where } \mathbf{z}_t, \mathbf{z}_f \in \mathbb{R}^{d}
\end{equation}

\subsubsection{Flexible Downstream Usage}

These dual-domain representations provide a general signal embedding with flexible usage patterns:

\textbf{(1) Independent Usage:}
\begin{itemize}
    \item Use $\mathbf{z}_t$ alone for time-domain-specific tasks (e.g., temporal event detection)
    \item Use $\mathbf{z}_f$ alone for frequency-domain-specific tasks (e.g., spectral classification)
\end{itemize}

\textbf{(2) Concatenated Usage:}
\begin{equation}
\mathbf{z}_{\text{concat}} = [\mathbf{z}_t; \mathbf{z}_f] \in \mathbb{R}^{2d}
\end{equation}
Provides comprehensive time-frequency representation for tasks requiring both perspectives.

\textbf{(3) Task-Specific Fusion:}
Downstream tasks can apply learned fusion (e.g., attention-weighted combination, gated fusion) tailored to task requirements.

\subsubsection{General Embedding Properties}

The dual-domain latent representation exhibits several desirable properties:

\textbf{Task Agnostic:} No task-specific architectural modifications required. The same encoder produces embeddings suitable for:
\begin{itemize}
    \item Classification (e.g., modulation recognition, speaker identification)
    \item Regression (e.g., SNR estimation, parameter prediction)
    \item Retrieval (e.g., signal similarity search)
    \item Generation (e.g., signal reconstruction, denoising)
\end{itemize}

\textbf{Domain Agnostic:} Through frequency-domain pooling (Section~\ref{appendix:ConvTokenization}) and Parseval Focus (Section~\ref{appendix:MHPF}), the encoder learns representations that generalize across signal types (RF, audio, seismic) without domain-specific retraining.

\textbf{Scale Adaptive:} The relative globality property enables the same architecture to process signals at multiple scales, from fine-grained local analysis to coarse-grained global summarization.

\subsubsection{Extension to Variable-Length Sequences}
\label{appendix:VariableLengthProcessing}

While the encoder architecture assumes fixed-size input windows (5120 samples = 5 windows × 1024 samples/window) during training, deployment often requires processing signals of arbitrary length. We extend to variable-length processing through a simple but effective strategy that leverages the encoder's natural structure.

\paragraph{Variable-Length Processing Strategy}

For an input signal of arbitrary length $L_{\text{input}}$:

\textbf{Step 1: Segment into Fixed-Size Windows.} Partition the input into non-overlapping segments of the training window size (5120 samples):
\begin{equation}
\mathbf{x}_{\text{input}} \rightarrow \{\mathbf{x}_1, \mathbf{x}_2, \ldots, \mathbf{x}_M\} \quad \text{where } M = \lceil L_{\text{input}} / 5120 \rceil
\end{equation}

If the final segment is shorter than 5120 samples, it can be resampled to 5120 depending on the application.

\textbf{Step 2: Process Segments Through Encoder.} Each segment is processed independently through the complete encoder pipeline (tokenization, Parseval Transformer, cross-domain fusion), producing token-level representations:
\begin{align}
\mathbf{x}_i &\rightarrow \{\mathbf{T}_t^{(i)}, \mathbf{T}_f^{(i)}\} \quad \text{for } i = 1, \ldots, M
\end{align}

where $\mathbf{T}_t^{(i)} \in \mathbb{R}^{N_t \times d}$ and $\mathbf{T}_f^{(i)} \in \mathbb{R}^{N_f \times d}$.

\textbf{Step 3: Domain-Specific Aggregation.} Aggregate tokens across segments using domain-appropriate strategies:

\textit{Time Domain (Concatenation):}
\begin{equation}
\mathbf{T}_t^{\text{concat}} = \text{Concat}([\mathbf{T}_t^{(1)}, \mathbf{T}_t^{(2)}, \ldots, \mathbf{T}_t^{(M)}], \text{dim}=0) \in \mathbb{R}^{(M \cdot N_t) \times d}
\end{equation}

This preserves temporal ordering across the entire input signal, creating a unified sequence representation.

\textit{Frequency Domain (Averaging):}
\begin{equation}
\mathbf{T}_f^{\text{avg}} = \frac{1}{M} \sum_{i=1}^{M} \mathbf{T}_f^{(i)} \in \mathbb{R}^{N_f \times d}
\end{equation}

This produces a time-averaged spectral representation, consistent with the latent space averaging strategy described in Section~\ref{appendix:domain_token_aggregation}.

\textbf{Step 4: Attentional Pooling.} Apply the sequence pooler to the aggregated token representations:
\begin{align}
\mathbf{z}_t &= \text{AttentionalPool}(\mathbf{T}_t^{\text{concat}}) \in \mathbb{R}^{d} \\
\mathbf{z}_f &= \text{AttentionalPool}(\mathbf{T}_f^{\text{avg}}) \in \mathbb{R}^{d}
\end{align}

\textbf{Step 5: Final Cross-Domain Fusion.} Apply the post-pooling cross-domain fusion (Section~\ref{appendix:DualDomainRepresentation}) to produce the final dual-domain representation.

\paragraph{Rationale and Benefits}

This strategy provides several advantages:

\textbf{(1) Consistency with Training:} Each segment is processed through the encoder with the same input size used during training, ensuring consistent feature extraction without distribution shift.

\textbf{(2) Leveraging Translational Equivariance:} The use of non-overlapping windows exploits the translational equivariance properties learned during training. Our architecture is explicitly designed to learn shift-invariant representations through:
\begin{itemize}
    \item \textbf{Convolutional tokenization:} Convolution's inherent translational equivariance ensures features detected at position $t$ in one window transfer to position $t$ in other windows
    \item \textbf{Co-designed training losses:} Our training objectives (Section~\ref{subsubsection:LatentEquivarianceLearning}) explicitly encourage translational equivariance in time, ensuring features shift predictably regardless of their position in the sequence
\end{itemize}

This means a feature learned at the beginning of segment $i$ should produce equivalent representations when it appears at the end of segment $i-1$ or the beginning of segment $i+1$. Non-overlapping windows leverage this learned symmetry: if the model has successfully learned translational equivariance, segment boundaries become arbitrary reference points rather than discontinuities in the representation space.

\textbf{(3) Domain-Appropriate Aggregation:} Time-domain concatenation preserves causal structure across the full signal, while frequency-domain averaging produces a global spectral summary—consistent with the physical interpretation of each domain.

\textbf{(4) Scalability:} The approach scales to arbitrarily long signals without architectural modifications or retraining. The attentional pooler naturally adapts to the increased sequence length in the time domain ($M \cdot N_t$ tokens) while the frequency domain remains fixed ($N_f$ tokens).

\textbf{(5) Memory Efficiency:} By processing segments independently through the encoder, peak memory usage is bounded by the single-segment processing cost, avoiding the quadratic memory growth of processing the entire signal as one sequence.

\paragraph{Implementation Note}

Algorithm~\ref{alg:variable_length} summarizes the complete procedure.

\begin{algorithm}[H]
\caption{Variable-Length Signal Processing}
\label{alg:variable_length}
\begin{algorithmic}[1]
\REQUIRE Input signal $\mathbf{x}_{\text{input}}$ of arbitrary length $L_{\text{input}}$
\REQUIRE Segment size $L_{\text{seg}} = 5120$ (training window size)
\ENSURE Dual-domain latent representation $\{\mathbf{z}_t, \mathbf{z}_f\}$
\STATE
\STATE \COMMENT{\textbf{Segment input signal}}
\STATE $M \leftarrow \lceil L_{\text{input}} / L_{\text{seg}} \rceil$
\STATE $\{\mathbf{x}_1, \ldots, \mathbf{x}_M\} \leftarrow \text{Segment}(\mathbf{x}_{\text{input}}, L_{\text{seg}})$
\STATE
\STATE \COMMENT{\textbf{Process each segment through encoder}}
\FOR{$i = 1$ to $M$}
    \STATE $\mathbf{T}_t^{(i)}, \mathbf{T}_f^{(i)} \leftarrow \text{Encoder}(\mathbf{x}_i)$ \COMMENT{Token-level representations}
\ENDFOR
\STATE
\STATE \COMMENT{\textbf{Domain-specific aggregation}}
\STATE $\mathbf{T}_t^{\text{concat}} \leftarrow \text{Concat}([\mathbf{T}_t^{(1)}, \ldots, \mathbf{T}_t^{(M)}], \text{dim}=0)$ \COMMENT{Time: concatenate}
\STATE $\mathbf{T}_f^{\text{avg}} \leftarrow \frac{1}{M} \sum_{i=1}^{M} \mathbf{T}_f^{(i)}$ \COMMENT{Frequency: average}
\STATE
\STATE \COMMENT{\textbf{Attentional pooling}}
\STATE $\mathbf{z}_t^{\text{pre}} \leftarrow \text{AttentionalPool}(\mathbf{T}_t^{\text{concat}})$
\STATE $\mathbf{z}_f^{\text{pre}} \leftarrow \text{AttentionalPool}(\mathbf{T}_f^{\text{avg}})$
\STATE
\STATE \COMMENT{\textbf{Post-pooling cross-domain fusion}}
\STATE $\mathbf{z}_t \leftarrow \text{GLU}([\mathbf{z}_t^{\text{pre}}; \text{Transform}_{f \rightarrow t}(\mathbf{z}_f^{\text{pre}})])$
\STATE $\mathbf{z}_f \leftarrow \text{GLU}([\mathbf{z}_f^{\text{pre}}; \text{Transform}_{t \rightarrow f}(\mathbf{z}_t^{\text{pre}})])$
\STATE
\RETURN $\{\mathbf{z}_t, \mathbf{z}_f\}$
\end{algorithmic}
\end{algorithm}

\subsubsection{Encoder Output Summary}

Table~\ref{tab:encoder_outputs} summarizes the encoder's output representations and their typical usage.

\begin{table}[h]
\centering
\caption{PlanFormer Encoder Output Representations}
\label{tab:encoder_outputs}
\begin{tabular}{p{3.5cm}p{3cm}p{7cm}}
\toprule
\textbf{Representation} & \textbf{Shape} & \textbf{Typical Usage} \\
\midrule
Time-domain tokens $\mathbf{T}_t^{\text{final}}$ & $\mathbb{R}^{B \times N_t \times d}$ & Sequence-to-sequence tasks (e.g., denoising, forecasting) \\
\midrule
Frequency-domain tokens $\mathbf{T}_f^{\text{final}}$ & $\mathbb{R}^{B \times N_f \times d}$ & Spectral analysis, frequency-domain generation \\
\midrule
Time-domain latent $\mathbf{z}_t$ & $\mathbb{R}^{B \times d}$ & Time-domain classification, temporal retrieval \\
\midrule
Frequency-domain latent $\mathbf{z}_f$ & $\mathbb{R}^{B \times d}$ & Spectral classification, frequency-based retrieval \\
\midrule
Dual-domain latent $[\mathbf{z}_t; \mathbf{z}_f]$ & $\mathbb{R}^{B \times 2d}$ & Comprehensive signal understanding, multi-task learning \\
\bottomrule
\end{tabular}
\end{table}

\textbf{Design Philosophy:} By providing both token-level and pooled representations in both domains, the encoder maximizes flexibility for downstream applications. Token-level representations preserve spatial/temporal structure for generation tasks, while pooled representations provide compact embeddings for discriminative tasks. The dual-domain structure ensures that both time-domain and frequency-domain perspectives are available, enabling the downstream task to leverage whichever view (or combination) is most appropriate.

\section{Appendix B: Decoder Architecture}
\label{appendix:B_Decoder}

\subsection{Theoretical Foundation and Design Philosophy}
\label{appendix:DecoderFoundation}

\textbf{Design Philosophy: Architecture as Gradient Substrate.} The decoder architecture serves a specific purpose: providing a computational substrate that enables symmetry losses to generate meaningful gradients throughout the encoder. Unlike the encoder—whose design directly embodies signal-theoretic principles—the decoder's design is pragmatic: each component addresses specific gradient flow challenges that arise when training with our co-designed loss functions (IsoFICReg, LED, reconstruction, source separation).

We do not claim these decoder components are optimal or represent the best possible design. Rather, they are sufficient solutions that fill critical gaps: enabling learning in negative-SNR regimes (skip sinks), preserving high-frequency gradients (frequency-domain upsampling), and providing instance-specific guidance (dynamic FiLM). The decoder's value lies in its functional role—supporting the encoder's learning of physical principles—rather than as an architectural contribution in its own right.

\textbf{Encoder-Decoder Role Distinction.} The encoder must remain 
task-agnostic, avoiding excessive a priori guidance that could cause 
information loss. It encodes the complete physical structure—"what it is 
and how it is"—producing information-maximal bottleneck representations 
without knowledge of downstream tasks.

The decoder operates with explicit targets derived from the tokens and bottleneck 
latent. It interprets the encoder's information-rich representation to 
reconstruct specific targets, using the latent as guidance for "what to 
pull out and reconstruct." This asymmetry is critical: if the encoder 
produces well-structured representations capturing physical structure, 
the decoder can reconstruct any target sharing that structure—enabling 
zero-shot generalization to unseen domains.

This role distinction justifies design choices that would be inappropriate 
during encoding. For example, Skip Connection Sinks 
(Section~\ref{appendix:SkipSinks}) use three-way constraint 
regularization against explicit targets—acceptable during decoding where 
targets are known, but inappropriate during encoding where task-agnostic 
information preservation is paramount.

\textbf{Key Capability: Zero-Shot Reconstruction Generalization.} 
This encoder-decoder separation enables a critical capability: zero-shot 
reconstruction of signals from previously unseen domains. Because the 
encoder captures domain-invariant physical structure rather than 
domain-specific semantics, and the decoder interprets this structure 
rather than memorizing reconstruction patterns, the system generalizes 
to new domains without retraining. 

This is particularly valuable for inverse problems (denoising, source 
separation, inpainting) where mixed information is encoded but specific 
targets must be reconstructed. The bottleneck latent encodes \textit{what 
physical structure exists}, not semantic categories—the decoder then 
extracts the relevant structure for reconstruction. This explains how 
RF-trained models can denoise audio or reconstruct image sources: the physical transformations (noise, mixing, 
occlusion) are domain-invariant, even when the signal content differs.

\textbf{Decoder Components Overview.} The decoder introduces three primary 
innovations beyond the encoder's mechanisms:
\begin{enumerate}
    \item \textbf{Attention-Based Dynamic FiLM} (Section~\ref{appendix:DynamicFiLM}): 
    Bidirectional latent conditioning via cross-focus for instance-specific guidance
    \item \textbf{Parseval Focus-Based Skip Sinks} (Section~\ref{appendix:SkipSinks}): 
    Target-guided dynamic filtering of skip connections
    \item \textbf{Sliding Window Self-Focus Activation} (Section~\ref{appendix:SlidingWindowActivation}): 
    Complex-valued adaptive modulation for final reconstruction
\end{enumerate}

The remaining components (frequency-domain upsampling, causal cross-window 
focus, transformer refinement) adapt encoder mechanisms for the decoder's 
reconstruction objectives.

\subsection{Design Rationale: Balancing Compression, Attention Cost, and Reconstruction}
\label{appendix:DecoderRationale}

The decoder architecture employs a U-Net structure with carefully designed skip connections that 
balance three competing constraints: (1) \textbf{attention computational cost}, which drives 
aggressive sequence compression, (2) \textbf{reconstruction fidelity}, which requires high-resolution 
information, and (3) \textbf{preventing trivial information leakage}, which ensures the bottleneck 
learns meaningful representations.

\subsubsection{Compression Driven by Attention Cost}

The encoder achieves \textbf{64x sequence compression} from input to bottleneck:
\begin{itemize}
    \item Input: 5120 samples (complex-valued IQ)
    \item After 3 pooling stages within convolutinoal tokenization (4x, 4x, 4x): 80 tokens
    \item Final latent: 128-dimensional vector after attentional pooling
\end{itemize}

This aggressive compression is driven by the \textbf{quadratic cost of self-attention}: 
$\mathcal{O}(N^2 \cdot d)$ where $N$ is sequence length. Processing 5120 tokens with multiple attention/focus computations per 
Parseval Transformer block would be computationally prohibitive. By compressing to 80 tokens, 
we achieve $(80/5120)^2 = 0.02\%$ of the attention cost, enabling more efficient and real-time processing 
with Multi-Head Parseval Focus (Section~\ref{appendix:MHPF}).

The compressed bottleneck latent dimensionality (128) matches the embedding dimension per token 
due to attentional pooling across all 80 tokens—each dimension aggregates information from the 
full sequence via learned attention weights (Section~\ref{appendix:SequencePooling}).

\subsubsection{Skip Connection Design: Preventing Trivial Leakage}

Reconstructing from 64x compression without skip connections is architecturally infeasible—the 
bottleneck would need to encode every high-resolution detail, or we would need to process much 
longer sequences through the transformer (defeating the compression goal). However, naively 
including all skip connections would enable trivial information leakage, allowing the decoder 
to bypass the bottleneck entirely.

Our solution: \textbf{selective skip connections from compressed intermediate features only}.

\paragraph{Skip Connection Hierarchy}

We include only two skip connections, both from already-compressed encoder features:

\begin{table}[h]
\centering
\caption{Skip Connection Design}
\label{tab:skip_hierarchy}
\begin{tabular}{lccc}
\toprule
\textbf{Encoder Stage} & \textbf{Compression} & \textbf{Skip Connection} & \textbf{Rationale} \\
\midrule
Input & 1x & \textcolor{red}{Omitted} & Raw samples (trivial leakage) \\
Conv Tokenization & 1x & \textcolor{red}{Omitted} & Linear features (trivial leakage) \\
After 1st Pooling & 4x & \textcolor{green}{\textbf{Included}} & Compressed local patterns \\
After 2nd Pooling & 16x & \textcolor{green}{\textbf{Included}} & Compressed mid-level features \\
After 3rd Pooling & 64x & \textcolor{red}{Omitted} & Redundant with token representation \\
Bottleneck Tokens & 64x & N/A & Global guidance (via FiLM conditioning) \\
\bottomrule
\end{tabular}
\end{table}

\textbf{Key Design Choices:}
\begin{enumerate}
    \item \textbf{Omit top-level skips}: No connection from input or first convolutional layer, 
    preventing the decoder from simply copying input details
    \item \textbf{First skip at 4x compression}: Already learned features from frequency-preserving 
    pooling (Section~\ref{appendix:FrequencyPooling}), not raw samples
    \item \textbf{Omit pre-transformer skip}: The 64x compressed tokens undergo Parseval Transformer 
    processing and their pooled representation are used for bottleneck latent guidance via FiLM conditioning 
    (Section~\ref{appendix:DynamicFiLM}). A skip connection here would be redundant with 
    the token-based upsampling path and would bypass the learned transformer representations.
\end{enumerate}

\subsubsection{Role Division: Bottleneck Latent vs. Skip Connections}

The decoder architecture creates a deliberate division of labor:

\textbf{Bottleneck Latent (via Dynamic FiLM):}
\begin{itemize}
    \item \textbf{Global structural guidance}: Targeted content, object identity, overall frequency 
    distribution
    \item \textbf{Task-specific modulation}: Conditions all decoder layers via attention-based FiLM 
    (Section~\ref{appendix:DynamicFiLM})
    \item \textbf{Cross-modal transfer}: Learned from RF, transfers to images/audio/text (validated 
    by linear probing using bottleneck only)
    \item \textbf{Transformer-processed representation}: Captures long-range dependencies and 
    cross-domain relationships via Parseval Focus
\end{itemize}

\textbf{Skip Connections (from 4x and 16x compressed features):}
\begin{itemize}
    \item \textbf{High-resolution details}: Local textures, sharp edges, fine-grained frequency components
    \item \textbf{Gradient flow}: Provides dense gradients to early encoder layers during training
    \item \textbf{Frequency infilling}: Parseval Focus-based skip sinks (Section~\ref{appendix:SkipSinks}) 
    dynamically filter skip connections, removing noise and interference while preserving signal structure
    \item \textbf{Compressed representations}: Already 4x and 16x downsampled, forcing encoder to 
    learn meaningful features rather than pass through raw inputs
\end{itemize}

\subsubsection{Empirical Validation: Random Weight Baseline}

\begin{figure}[h]
    \centering
    \includegraphics[width=\textwidth]{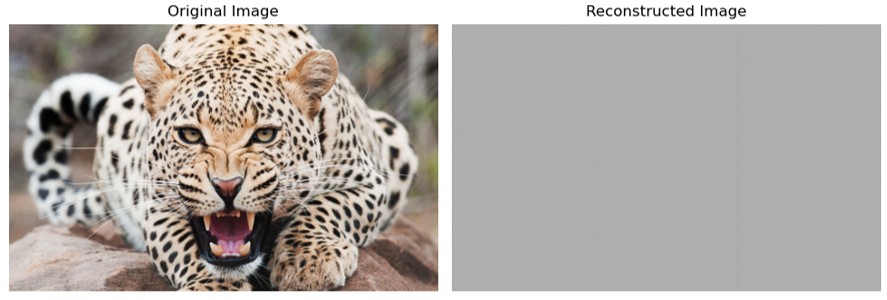}
    \caption{\textbf{Random weight baseline for reconstruction.} Same architecture as trained model, 
    but with random weight initialization. \textbf{Left:} Original image. \textbf{Right:} Reconstruction 
    with random weights—complete failure producing uniform gray field. This validates that learned representations (not 
    architectural biases or skip connections) drive reconstruction quality. Compare to 
    Figure~\ref{fig:image_reconstructions} showing coherent reconstructions with trained weights.}
    \label{fig:random_reconstruction}
\end{figure}

To validate that learned representations (not architectural inductive biases) drive reconstruction 
quality, we evaluate the same architecture with random weight initialization. 
Figure~\ref{fig:random_reconstruction} shows that random weights produce near-complete reconstruction 
failure—uniform gray fields with no discernible structure. This demonstrates:

\begin{enumerate}
    \item \textbf{Learned weights are essential}: The architecture alone (skip connections + U-Net 
    structure) cannot reconstruct without learned representations
    \item \textbf{Bottleneck encodes meaningful structure}: Random bottleneck latents provide no 
    guidance, resulting in collapsed reconstructions
    \item \textbf{Skip connections are not shortcuts}: Even with skip connections, random weights 
    fail to reconstruct, proving skips don't bypass learning
\end{enumerate}

\subsubsection{Why This Design Validates Transfer Learning}

Critically, our transfer learning evaluation (Table~\ref{tab:frozenEncoder_results}) uses \textbf{only 
the bottleneck latent} for linear probing—skip connections are not involved. The strong classification 
performance (77.7\% average, 91.9\% top-3) validates that the bottleneck latent alone captures 
transferable physical structure. Skip connections contribute to reconstruction quality during training 
but do not confound our transfer learning claims.

\textbf{Reconstruction as Training Signal:} The reconstruction objective (with skip connections) 
serves as a dense training signal that encourages the encoder to learn physically meaningful features. 
However, the learned representations must also satisfy discriminative objectives (IsoFICReg, LED) 
that operate solely on the bottleneck latent. The combination ensures the bottleneck learns compressed, 
transferable representations rather than relying on skip connection shortcuts. The random weight 
baseline (Figure~\ref{fig:random_reconstruction}) confirms that skip connections alone cannot 
reconstruct—learned bottleneck representations are essential.

\subsubsection{Interpretation of Reconstruction Results}

The zero-shot reconstruction quality (Section E.5, Figure~\ref{fig:image_reconstructions}) should be 
interpreted as evidence that:
\begin{enumerate}
    \item \textbf{The encoder-decoder system} (bottleneck + skip connections + learned weights) 
    captures transferable physical structure across modalities
    \item \textbf{The bottleneck latent} provides global structural guidance (validated by linear 
    probing on bottleneck alone, and by random weight failure)
    \item \textbf{Skip connections} preserve high-resolution details from compressed features (4x, 16x) 
    necessary for pixel-level reconstruction fidelity
    \item \textbf{Learned weights are essential}: Random initialization fails completely 
    (Figure~\ref{fig:random_reconstruction}), proving architectural biases alone are insufficient
\end{enumerate}

The reconstruction quality is not solely attributable to the bottleneck latent, nor solely to skip 
connections, nor to architectural inductive biases—all three components (learned bottleneck, 
compressed skip connections, learned decoder weights) are necessary and complementary. However, 
the transfer learning results (linear probing) validate that the bottleneck latent alone captures 
sufficient physical structure for cross-modal classification tasks.

\subsection{Decoder Components}
\label{appendix:DecoderComponents}

\subsection{Core Decoder Components}
\label{appendix:DecoderComponents}
Having established the architectural rationale, we now detail the specific decoder components.
The decoder maintains the encoder's dual-domain architecture with parallel 
time and frequency branches. Like the encoder, the decoder processes 
reconstruction in non-overlapping windows to address time-varying phenomena 
(noise fluctuations, local frequency trends). Since frequency-domain tokens 
represent global spectral aggregation, the same frequency tokens serve as 
context for each time-domain window, with temporally local frequency content 
injected through skip connections.

Each decoder layer consists of the following components, applied symmetrically 
to both branches:

\subsubsection{Convolutional Processing and Frequency-Domain Upsampling}

\textbf{Convolutional Processing:} 1D convolutions with progressively 
decreasing channel dimensions (reversing the encoder's expansion from 
Section~\ref{appendix:ConvTokenization}) refine local features. As the decoder progresses from abstract latent representations toward concrete signal reconstructions, channel dimensionality decreases while sequence length increases.

\textbf{Frequency-Domain Upsampling:} Transposed convolution applied in the frequency domain progressively increases sequence length while preserving spectral structure, consistent with the encoder's frequency-preserving pooling strategy (Section~\ref{appendix:ConvTokenization}).

\textbf{Synergy with Skip Connections:} Skip connection spectrums infill high-fidelity frequency components during upsampling, facilitating finer-grained reconstruction fidelity than discrete time-domain sample interpolation. 

\subsubsection{Attention-Based Dynamic FiLM Conditioning}
\label{appendix:DynamicFiLM}

\textbf{Motivation:} Traditional Feature-wise Linear Modulation (FiLM)~\cite{perez2017filmvisualreasoninggeneral} applies channel-wise affine transformations conditioned on an external signal. However, standard FiLM has critical limitations:
\begin{enumerate}
    \item \textbf{One-way conditioning:} The conditioning signal generates modulation parameters without considering the features being modulated
    \item \textbf{Global rigidity:} Channel-wise modulation is spatially/temporally global, failing to respect equivariant symmetries learned during training
\end{enumerate}

\textbf{Innovation:} We extend FiLM by combining it with Dynamic TanH activation~\cite{zhu2025transformers} construction, where modulation parameter generation occurs via Scaled Covariance Cross-Focus. (Section~\ref{appendix:ScaledCovarianceFocus}) between the bottleneck latent and the sequence of features being modulated. This creates a bidirectional relationship: the latent guides modulation based on the reconstruction target, but the features' characteristics influence how that guidance is applied.

\paragraph{Modulation Parameter Generation}

The cross-focus mechanism computes attention between the bottleneck latent (as query) and the decoder features (as key/value), producing modulation parameters that are adaptive to both the reconstruction target and the current feature state.

\textbf{Gamma ($\gamma$):} Generated per element in the sequence, applied inside the TanH nonlinearity for maximal expressivity. This pointwise modulation respects local equivariant symmetries learned during training.

\textbf{Alpha ($\alpha$) and Beta ($\beta$):} Generated channel-wise for parameter efficiency, applied as standard affine parameters outside the nonlinearity for stability.

\paragraph{Residual Formulation}

To ensure stability and preserve information flow, modulation is applied as a series of residual connections:
\begin{equation}
\mathbf{x}_{\text{modulated}} = \alpha \cdot \tanh(\gamma \cdot \mathbf{x} + \mathbf{x}) + \beta + \mathbf{x}
\label{eq:dynamic_film}
\end{equation}

This nested residual structure ensures that even if modulation parameters collapse, the identity pathway preserves gradient flow.

\textbf{Rationale:} By conditioning modulation on the bottleneck latent via cross-focus, the decoder can adaptively emphasize or suppress features based on the reconstruction objective. For denoising, the modulation might amplify signal-like features and suppress noise-like features; for source separation, it might emphasize features corresponding to the target source. Critically, this adaptation occurs dynamically based on the latent encoding of the desired output, enabling the same decoder architecture to handle diverse reconstruction tasks.

\subsubsection{Parseval Focus-Based Skip Connection Sinks}
\label{appendix:SkipSinks}

\textbf{Motivation:} In UNet architectures, skip connections facilitate high-resolution details but present a vulnerability: they hold intermediately processed representations before full encoding is complete. This is problematic for tasks where multiple signal components are present during encoding but only specific components are desired at reconstruction output (e.g., denoising: signal + noise $\rightarrow$ signal only; source separation: speaker A + speaker B $\rightarrow$ speaker A only).

\textbf{Challenge:} A naive approach would learn fixed transformations after skip fusion to attenuate unwanted components. However, this rigid prior does not scale to arbitrary additive noise sources or mixture scenarios where the unwanted component varies dynamically.

\textbf{Solution:} We introduce a unified mechanism combining: (1) Noise Sink mechanism (Section~\ref{appendix:ConvTokenization}) to clean skip connections, extended with conditional guidance toward the reconstruction target, (2) Parseval-consistent cross-domain focus (Section~\ref{appendix:MHPF}) to ensure physical consistency, and (3) Dynamic power compensation for SNR-adaptive fusion.

\textbf{Key Insight:} The bottleneck latent encodes \textit{which physical structure to reconstruct}. By conditioning the skip sink on this latent, we enable dynamic filtering: the same skip connection can be processed differently depending on the reconstruction objective. For denoising, the sink removes noise; for source separation, it removes interfering sources; for signal restoration, it removes artifacts—all using the same architecture, guided by the latent encoding of the desired output structure.

\paragraph{Theoretical Foundation: Dynamic Filter Banks}

For dynamic noise sources that vary in time and frequency, it is essential to learn dynamic filter behaviors (low-pass, high-pass, bandpass, notch). To do so effectively, we need access to the full context of signals in both time and frequency domains.

\paragraph{Implementation: Parseval Focus with Dynamic Filtering}

The skip sink implementation varies by domain but follows a unified procedure:

\textbf{Domain-Specific Preprocessing:}
\begin{itemize}
    \item \textbf{Time-domain skips:} Processed via non-overlapping sliding 
    windows, leveraging the inductive bias that additive noise is locally 
    uncorrelated with signal
    \item \textbf{Frequency-domain skips:} Processed with full spectral context 
    (no windowing), using learned convolutional downsampling to produce 
    manageable spectral bins
\end{itemize}

\textbf{Parseval Focus for Dynamic Noise Removal:}

Both time and frequency representations undergo complementary parametric 
domain transforms (FFT/IFFT) to produce dual-domain token sequences. The 
bottleneck latent is prepended to each domain's sequence as a conditioning 
token, and Multi-Head Parseval Focus (Section~\ref{appendix:MHPF}) is 
computed over these augmented sequences.

This joint time-frequency processing enables learning dynamic filters 
(low-pass, high-pass, bandpass, notch) conditioned on three critical factors:
\begin{enumerate}
    \item \textbf{What to remove:} Encoded in bottleneck latent 
    (e.g., "remove noise, keep signal")
    \item \textbf{Where to remove:} Time localization via windowed processing
    \item \textbf{At what frequencies:} Spectral localization via 
    frequency-domain processing
\end{enumerate}

Parseval Focus is ideal because it jointly considers time and frequency 
representations while enforcing physical consistency (via JSD regularization). 
Cross-domain attention allows the model to learn that noise removal in time 
corresponds to predictable changes in frequency, and vice versa.

\textbf{Sink Computation and Subtraction:}

The output of Parseval Focus (with bottleneck conditioning token removed) 
seeds a hypernetwork generating $\gamma$, $\alpha$, and $\beta$ pointwise 
modulation parameters. These are applied in the frequency domain, functioning 
as spectral convolutions whose cumulative output serves as the sink (noise 
estimate):

\begin{equation}
\mathbf{x}_{\text{skip}}^{\text{clean}} = \mathbf{x}_{\text{skip}} - \mathbf{x}_{\text{sink}}
\end{equation}

This subtractive residual formulation enables learning both direct filters 
(what to keep) and complementary filters (what to remove), providing maximal 
expressivity for all types of skip connection pollution.

\paragraph{Regularization for Target-Guided Conditioning}

To ensure the conditioning works as expected, the layer is heavily regularized:

\textbf{(1) Decorrelation Loss:} Sink is decorrelated from post-sink output (similar to encoder noise sinks):
\begin{equation}
\mathcal{L}_{\text{decorr}}^{\text{skip}} = |\text{PearsonCorr}(\mathbf{x}_{\text{sink}}, \mathbf{x}_{\text{skip}}^{\text{clean}})|
\end{equation}

This ensures the sink removes uncorrelated components rather than signal content.

\textbf{(2) Sink-Latent Decorrelation:} Sink is decorrelated from bottleneck latent guidance:
\begin{equation}
\mathcal{L}_{\text{sink-latent}} = |\text{PearsonCorr}(\mathbf{x}_{\text{sink}}, \mathbf{z})|
\end{equation}

This ensures the removed content is not aligned with the reconstruction target encoded in the latent.

\textbf{(3) Post-Sink Target Alignment Loss:} Post-sink output is encouraged to correlate with bottleneck latent guidance:
\begin{equation}
\mathcal{L}_{\text{target-align}} = 1 - \text{PearsonCorr}(\mathbf{x}_{\text{skip}}^{\text{clean}}, \mathbf{z})
\end{equation}

This ensures the cleaned signal aligns with the reconstruction target.

\textbf{(4) Dynamic Power Scaling:} Handles cases where the sink's information content is more dominant than the signal of interest, using the same power compensation mechanism as the encoder's noise sink (Section~\ref{appendix:ConvTokenization}).

\textbf{Rationale:} These regularization losses create a three-way constraint: (1) sink should be uncorrelated with cleaned output (removing noise, not signal), (2) sink should be uncorrelated with reconstruction target (not removing what we want), and (3) cleaned output should be correlated with reconstruction target (keeping what we want). Together, these ensure the skip sink performs target-guided filtering rather than arbitrary transformation.

\subsubsection{Causal Cross-Window Focus with Latent Conditioning}
\label{appendix:CausalCrossWindowDecoder}

The Causal Cross-Window Focus mechanism (Section~\ref{appendix:WindowingDualDomainTokenization}) is employed at each decoder layer to maintain phase coherence across window boundaries.

\paragraph{Per-Layer Previous Window Storage}

At each decoder layer $i$, we maintain a buffer storing the previous window's output:
\begin{equation}
\mathbf{x}_{\text{prev}}^{(i)} \in \mathbb{R}^{B \times C_i \times W}
\end{equation}

For the first window ($\text{window\_idx} = 0$), no previous context exists. For subsequent windows ($\text{window\_idx} > 0$), we compute cross-window attention.

\paragraph{Two-Stage Focus Process}

The decoder applies focus hierarchically at each layer, with domain-specific 
preprocessing to ensure physically meaningful positional encodings:

\textbf{Stage 1: Cross-Window Focus (windows 2+)}

For all windows except the first ($\text{window\_idx} > 0$), we compute 
causal cross-window focus to maintain phase coherence. The preprocessing 
differs by domain to respect physical interpretation:

\textit{Time-Domain Processing:}
\begin{enumerate}
    \item Transform current and previous windows to frequency domain (mitigates 
    aliasing during learned downsampling)
    \item Reshape to token representation and downsample (4×) via learned 
    1×1 convolution
    \item \textbf{Transform back to time domain} before concatenation
    \item Concatenate previous and current windows in time domain: 
    $\mathbf{T}_{\text{concat}} = [\mathbf{T}_{\text{prev}}^{(time)}; \mathbf{T}_{\text{curr}}^{(time)}]$
    \item Apply sinusoidal positional encoding to concatenated time-domain sequence
    \item Apply RMS normalization to split sequences independently
    \item Compute cross-focus: current window (query) attends to 
    previous window (key/value)
    \item Upsample in frequency domain, reshape, transform to time domain, 
    and add as residual
\end{enumerate}

\textit{Frequency-Domain Processing:}
\begin{enumerate}
    \item Reshape to token representation and downsample (4×) via learned 
    1×1 convolution
    \item \textbf{Remain in frequency domain} (no time-domain transformation)
    \item Apply sinusoidal positional encoding to previous and current windows 
    \textbf{independently} (no concatenation)
    \item Apply RMS normalization
    \item Compute cross-focus: current window (query) attends to 
    previous window (key/value)
    \item Upsample, reshape, and add as residual
\end{enumerate}

\textbf{Rationale for Domain-Specific Treatment:}

\textit{Time Domain:} Causal concatenation in the time domain preserves 
temporal ordering and phase relationships across window boundaries. The 
frequency-domain transformation is used only for anti-aliasing during 
downsampling—the actual positional encoding must occur in the time domain 
where position has absolute temporal meaning. This ensures the model learns 
continuous phase relationships: a feature at the end of window $i-1$ should 
smoothly connect to the beginning of window $i$.

\textit{Frequency Domain:} Each position corresponds to a spectral bin 
(e.g., the 100 Hz component) rather than a temporal instant. Concatenation 
would impose artificial temporal ordering on what is fundamentally a spectral 
covariance problem. Independent positional encoding allows the cross-attention 
mechanism to learn how corresponding frequency bins evolve across windows 
(e.g., "when the 100 Hz component is strong in the previous window, how does 
it affect the current window?") without conflating this with temporal position.

\textbf{Stage 2: Latent-Conditioned Self-Focus (all windows)}

After cross-window focus (or for the first window where Stage 1 is 
skipped), we apply self-focus conditioned on the bottleneck latent. 
Again, domain-specific preprocessing ensures physical consistency:

\textit{Time-Domain Processing:}
\begin{enumerate}
    \item Downsample tokens (4×) via learned 1×1 convolution
    \item \textbf{Transform to time domain} before latent conditioning
    \item Project bottleneck latent to token dimension and prepend: 
    $\mathbf{T}_{\text{aug}} = [\mathbf{W}_z \mathbf{z}_t; \mathbf{T}_{\text{curr}}^{(time)}]$
    \item Apply sinusoidal positional encoding and RMS normalization
    \item Compute self-focus over augmented sequence
    \item Remove latent token from output
    \item \textbf{Transform to frequency domain} before upsampling
    \item Upsample in frequency domain, reshape, transform to time domain, 
    and add as residual
\end{enumerate}

\textit{Frequency-Domain Processing:}
\begin{enumerate}
    \item Downsample tokens (4×) via learned 1×1 convolution
    \item \textbf{Remain in frequency domain}
    \item Project bottleneck latent to token dimension and prepend: 
    $\mathbf{T}_{\text{aug}} = [\mathbf{W}_z \mathbf{z}_f; \mathbf{T}_{\text{curr}}^{(freq)}]$
    \item Apply sinusoidal positional encoding and RMS normalization
    \item Compute self-focus over augmented sequence
    \item Remove latent token from output
    \item Upsample, reshape, and add as residual
\end{enumerate}

\paragraph{Bidirectional Conditioning}

This two-stage process provides:
\begin{itemize}
    \item \textbf{Temporal coherence:} Cross-window focus in time domain 
    ensures phase continuity across boundaries
    \item \textbf{Spectral coherence:} Cross-window focus in frequency 
    domain captures time-varying spectral evolution
    \item \textbf{Task-specific guidance:} Latent conditioning guides refinement 
    toward reconstruction target in both domains
    \item \textbf{Bidirectional influence:} Latent influences sequence (forward), 
    sequence influences latent via gradients (backward)
\end{itemize}

\textbf{Anti-Aliasing Strategy:} The time-domain branch performs frequency-domain 
transformations (FFT/IFFT) around the downsampling/upsampling operations to 
leverage insights motivated by frequency-preserving pooling (Section~\ref{appendix:ConvTokenization}). 
This prevents aliasing artifacts that would corrupt high-frequency information 
essential for learning frequency translation equivariance. The frequency-domain 
branch operates natively in frequency, so no additional transformations are needed.

\subsection{Global Sequence Refinement}
\label{appendix:GlobalRefinement}

After window-based processing, the decoder applies global refinement mechanisms to address sequence-level phenomena and ensure high-fidelity reconstruction.

\subsubsection{Window Aggregation and Smoothing}

\textbf{Time Domain:} Window outputs are concatenated along the sequence dimension:
\begin{equation}
\mathbf{x}_{\text{concat}} = \text{Concat}([\mathbf{x}_{\text{win}_1}, \ldots, \mathbf{x}_{\text{win}_{N_w}}], \text{dim}=-1) \in \mathbb{R}^{B \times C \times L_{\text{total}}}
\end{equation}

\textbf{Domain-Specific Final Convolutions and Fusion:}
\begin{align}
\mathbf{x}_t^{\text{final}} &= \text{Conv}_{\text{final}}^{(t)}(\mathbf{x}_{\text{concat}}) \\
\mathbf{x}_f^{\text{final}} &= \text{Conv}_{\text{final}}^{(f)}(\mathbf{x}_{\text{freq}}) \\
\mathbf{x}_f^{(t)} &= \text{IFFT}(\mathbf{x}_f^{\text{final}}) \quad \text{(frequency to time)} \\
\mathbf{x}_{\text{fused}} &= \text{GLU}([\mathbf{x}_t^{\text{final}}; \mathbf{x}_f^{(t)}])
\end{align}

\textbf{Smoothing Convolution:} A convolution with kernel size 7 is applied across the full sequence, blending features across window boundaries while preserving overall signal structure:
\begin{equation}
\mathbf{x}_{\text{smooth}} = \text{Conv}_{\text{smooth}}(\mathbf{x}_{\text{fused}})
\end{equation}

\subsubsection{Transformer-Based Frequency Refinement}

After window-based processing and concatenation, local convolutional operations 
have captured fine-grained details but may miss long-range spectral dependencies. 
We apply transformer-based attention in the frequency domain to capture global 
spectral relationships.

\textbf{Procedure:}
\begin{enumerate}
    \item Transform smoothed time-domain representation to frequency via FFT
    \item Bin spectrum into manageable tokens: $5120$ frequency bins $\rightarrow$ 
    $80$ tokens (64 bins/token)
    \item Apply single-layer transformer (8 heads, Scaled Covariance Focus) 
    to refine frequency components
    \item Transform back to time domain via IFFT and remove DC bias
\end{enumerate}

\textbf{Rationale:} Spectral relationships (harmonics, spectral envelope) 
manifest as local patterns in frequency but long-range dependencies in time. 
For example, a fundamental frequency and its harmonics (separated by thousands 
of samples in time) appear as a local pattern in frequency. The transformer 
enforces consistency across harmonically-related components, improving 
reconstruction fidelity for signals with rich harmonic structure.

\subsubsection{Sliding Window Self-Focus Activation}
\label{appendix:SlidingWindowActivation}

\textbf{Purpose:} Final layer of adaptive, equivariant refinement.

\textbf{Motivation:} The final activation function must address several dynamic symmetries essential for effective signal reconstruction: (1) translational equivariance, (2) variable output dynamic ranges, (3) relative long-range phase relationships, and (4) final refinement that earlier layers could not address.

\textbf{Challenge:} Signal reconstructions are anchored to unit power during training, but this does not produce a fixed dynamic range. Output values can vary widely depending on signal characteristics. It is unreasonable to learn fixed activation parameters that generalize to any reconstruction range—the dynamic range itself must be learnable and adaptive.

\paragraph{Solution: Attention-Based Hypernetwork for Dynamic Modulation}

We extend the Dynamic FiLM mechanism (Section~\ref{appendix:DynamicFiLM}) 
to the final activation layer, introducing an attention-based hypernetwork 
to dynamically modulate both inside and outside the nonlinearity. This creates a bidirectional relationship between the conditioning signal (bottleneck latent) and the features being modulated (reconstruction sequence).

\textbf{Computational Challenge:} Since this operates over the final reconstruction sequence, each element serves as a "token." Computing full inner-product attention across all tokens becomes computationally prohibitive for long sequences (e.g., 5120 samples = 5120 tokens).

\textbf{Solution:} We utilize our sliding window focus formulation (Section~\ref{appendix:WindowingDualDomainTokenization}) across the sequence with 20 windows. To ensure proper continuity for phase relationship learning, we leverage the causal window focus formulation. The key difference: we do not upsample back to the original signal resolution after computing causal cross-window focus. Instead, the output of cross-window focus (at reduced resolution) serves as conditioning for hyperparameter generation, which is then broadcast to the full-resolution reconstruction.

\paragraph{Modulation Parameter Generation}

\textbf{Two-Stage Process:}
\begin{enumerate}
    \item \textbf{Cross-Window Focus:} Contextualizes phase compensation by attending to the previous window, ensuring continuity across window boundaries
    \item \textbf{Bottleneck Latent-Conditioned Self-Focus:} Refines via bidirectional influence between the bottleneck latent and all tokens in the current window, providing task-specific guidance
\end{enumerate}

This ensures both local coherence (within windows) and global coherence (across windows) while maintaining computational efficiency.

\paragraph{Complex-Valued Modulation}

To avoid introducing DC bias, we only compute alpha ($\alpha$) and gamma ($\gamma$) for scaling modulations inside and outside the TanH nonlinearity (no additive beta). Since elements represent real and imaginary components of a complex-valued reconstruction (in interleaved IQ format), gamma and alpha are generated as pairs of element-wise magnitude and phase compensations.

For a complex-valued sample $z = I + jQ$ (represented as consecutive elements $[I, Q]$ in the interleaved format):

\begin{align}
|z| &= \sqrt{I^2 + Q^2} \quad \text{(magnitude)} \\
\phi &= \text{atan2}(Q, I) \quad \text{(phase)} \\
z_{\text{mod}} &= \alpha_{\text{mag}} \cdot \tanh(\gamma_{\text{mag}} \cdot |z| + |z|) \cdot e^{j(\phi + \gamma_{\text{phase}} + \alpha_{\text{phase}})}
\end{align}

This is then converted back to interleaved IQ format:
\begin{align}
I_{\text{mod}} &= |z_{\text{mod}}| \cos(\phi_{\text{mod}}) \\
Q_{\text{mod}} &= |z_{\text{mod}}| \sin(\phi_{\text{mod}})
\end{align}

\textbf{Rationale:} Complex-valued modulation provides fine-grained control over both amplitude and phase, essential for high-fidelity signal reconstruction. The magnitude modulation can adaptively adjust dynamic range, while phase modulation can correct phase discontinuities at window boundaries or compensate for phase distortions introduced by earlier processing stages.

\subsection{Complete Decoder Pipeline}
\label{appendix:DecoderPipeline}

Algorithm~\ref{alg:complete_decoder} summarizes the complete decoder forward pass.

\begin{algorithm}[H]
\caption{Complete Decoder Forward Pass}
\label{alg:complete_decoder}
\begin{algorithmic}[1]
\REQUIRE Time tokens $\mathbf{T}_t$, Frequency tokens $\mathbf{T}_f$
\REQUIRE Skip connections $\{\mathbf{S}_t^{(i)}, \mathbf{S}_f^{(i)}\}$, Bottleneck latents $\mathbf{z}_t, \mathbf{z}_f$
\ENSURE Final reconstruction $\mathbf{x}_{\text{out}}$
\STATE
\STATE \COMMENT{\textbf{Reshape tokens to convolutional format}}
\STATE $\mathbf{x}_t, \mathbf{x}_f \leftarrow \text{ReshapeToChannels}(\mathbf{T}_t, \mathbf{T}_f)$
\STATE
\FOR{$\text{window\_idx} = 1$ to $N_w$}
    \STATE \COMMENT{\textbf{Extract time window, use full frequency}}
    \STATE $\mathbf{x}_t^{\text{win}} \leftarrow \mathbf{x}_t[:, :, (\text{window\_idx}-1) \cdot W : \text{window\_idx} \cdot W]$
    \STATE
    \FOR{$i = 1$ to $N_{\text{layers}}$}
        \STATE \COMMENT{\textbf{Convolutional processing}}
        \STATE $\mathbf{x}_t^{\text{win}} \leftarrow \text{Conv}^{(i)}(\mathbf{x}_t^{\text{win}})$
        \STATE $\mathbf{x}_f \leftarrow \text{Conv}^{(i)}(\mathbf{x}_f)$
        \STATE
        \STATE \COMMENT{\textbf{Dynamic FiLM conditioning}}
        \STATE $\mathbf{x}_t^{\text{win}} \leftarrow \text{DynamicFiLM}^{(i)}(\mathbf{x}_t^{\text{win}}, \mathbf{z}_t)$
        \STATE $\mathbf{x}_f \leftarrow \text{DynamicFiLM}^{(i)}(\mathbf{x}_f, \mathbf{z}_f)$
        \STATE
        \STATE \COMMENT{\textbf{Frequency-domain upsampling}}
        \STATE $\mathbf{x}_t^{\text{win}} \leftarrow \text{IFFT}(\text{Upsample}(\text{FFT}(\mathbf{x}_t^{\text{win}})))$
        \STATE $\mathbf{x}_f \leftarrow \text{Upsample}(\mathbf{x}_f)$
        \STATE
        \IF{$i \neq N_{\text{layers}}$}
            \STATE \COMMENT{\textbf{Skip connection fusion}}
            \STATE $\mathbf{S}_t^{\text{clean}} \leftarrow \text{SkipSink}^{(i)}(\mathbf{S}_t^{(i)}, \mathbf{z}_t)$
            \STATE $\mathbf{S}_f^{\text{clean}} \leftarrow \text{SkipSink}^{(i)}(\mathbf{S}_f^{(i)}, \mathbf{z}_f)$
            \STATE $\mathbf{x}_t^{\text{win}} \leftarrow \text{GLU}([\mathbf{x}_t^{\text{win}}; \text{FFT}(\mathbf{S}_t^{\text{clean}})])$
            \STATE $\mathbf{x}_f \leftarrow \text{GLU}([\mathbf{x}_f; \mathbf{S}_f^{\text{clean}}])$
            \STATE
            \STATE \COMMENT{\textbf{Causal cross-window focus}}
            \STATE $\mathbf{x}_t^{\text{win}} \leftarrow \text{CausalCrossWindowFocus}(\mathbf{x}_t^{\text{win}}, \mathbf{x}_t^{\text{prev}}, \mathbf{z}_t)$
            \STATE $\mathbf{x}_f \leftarrow \text{CausalCrossWindowFocus}(\mathbf{x}_f, \mathbf{x}_f^{\text{prev}}, \mathbf{z}_f)$
        \ENDIF
    \ENDFOR
    \STATE
    \STATE \COMMENT{\textbf{Store for next window}}
    \STATE $\text{time\_outputs}.\text{append}(\mathbf{x}_t^{\text{win}})$
\ENDFOR
\STATE
\STATE \COMMENT{\textbf{Global refinement}}
\STATE $\mathbf{x}_{\text{concat}} \leftarrow \text{Concat}(\text{time\_outputs})$
\STATE $\mathbf{x}_t^{\text{final}} \leftarrow \text{Conv}_{\text{final}}^{(t)}(\mathbf{x}_{\text{concat}})$
\STATE $\mathbf{x}_f^{\text{final}} \leftarrow \text{Conv}_{\text{final}}^{(f)}(\mathbf{x}_f)$
\STATE $\mathbf{x}_{\text{fused}} \leftarrow \text{GLU}([\mathbf{x}_t^{\text{final}}; \text{IFFT}(\mathbf{x}_f^{\text{final}})])$
\STATE
\STATE $\mathbf{x}_{\text{smooth}} \leftarrow \text{Conv}_{\text{smooth}}(\mathbf{x}_{\text{fused}})$
\STATE $\mathbf{x}_{\text{refined}} \leftarrow \text{IFFT}(\text{TransformerRefinement}(\text{FFT}(\mathbf{x}_{\text{smooth}})))$
\STATE $\mathbf{x}_{\text{refined}} \leftarrow \mathbf{x}_{\text{refined}} - \mathbb{E}[\mathbf{x}_{\text{refined}}]$ \COMMENT{DC removal}
\STATE
\STATE \COMMENT{\textbf{Final activation}}
\STATE $\mathbf{x}_{\text{out}} \leftarrow \text{SlidingSelfFocusActivation}(\mathbf{x}_{\text{refined}}, [\mathbf{z}_t; \mathbf{z}_f])$
\STATE
\RETURN $\mathbf{x}_{\text{out}}$
\end{algorithmic}
\end{algorithm}

\subsection{Decoder Summary}
\label{appendix:DecoderSummary}

Table~\ref{tab:decoder_innovations} summarizes the key innovations in the PlanFormer Decoder.

\begin{table}[h]
\centering
\caption{PlanFormer Decoder Key Innovations}
\label{tab:decoder_innovations}
\begin{tabular}{p{5cm}p{9cm}}
\toprule
\textbf{Component} & \textbf{Innovation} \\
\midrule
Latent-Conditioned Reconstruction & Bottleneck latent injected at multiple decoder stages for instance-wise, task-specific reconstruction guidance based on physical structure \\
\midrule
Attention-Based Dynamic FiLM & Scaled Covariance Cross-Focus for bidirectional, adaptive feature modulation with pointwise inner modulation and hierarchical residual connections \\
\midrule
Frequency-Domain Upsampling & Skip connection spectrums infill high-fidelity frequency components rather than discrete time-domain samples, preserving spectral envelope \\
\midrule
Parseval Focus-Based Skip Sinks & Unified mechanism combining noise removal, cross-domain physical consistency, and conditional dynamic filter banks guided by reconstruction target \\
\midrule
Causal Cross-Window Focus & Bidirectional conditioning between bottleneck latent and window sequences for phase-coherent reconstruction \\
\midrule
Frequency-Domain Transformer & Global spectral refinement with adaptive frequency binning for long-range harmonic relationships \\
\midrule
Sliding Window Self-Focus Activation & Attention-based hypernetwork generating per-element complex-valued modulation parameters with causal coherence and task-specific guidance \\
\bottomrule
\end{tabular}
\end{table}

\subsubsection{Synergy with Encoder}

The decoder's design is deeply integrated with the encoder architecture:

\textbf{(1) Symmetric Dual-Domain Processing:} Both maintain parallel time/frequency branches with strategic fusion, ensuring consistent treatment of time-frequency duality throughout the entire architecture.

\textbf{(2) Consistent Tokenization:} Decoder respects encoder's IQ-grouped tokenization (Section~\ref{appendix:ConvTokenization}), preserving complex-valued relationships and enabling seamless FFT/IFFT operations.

\textbf{(3) Latent Conditioning:} Encoder's dual-domain latent representations (Section~\ref{appendix:DualDomainRepresentation}) condition every decoder layer, providing continuous task-specific guidance throughout reconstruction.

\textbf{(4) Skip Connection Compatibility:} Decoder's skip fusion mechanisms handle encoder's noise-sink-processed skip connections, with additional target-guided filtering to remove unwanted components based on the reconstruction objective.

\textbf{(5) Frequency Pooling/Upsampling Consistency:} Both use frequency-domain pooling (encoder) and upsampling (decoder) to preserve spectral envelopes, avoiding the spectral bias that would prevent learning high-frequency equivariances.

\textbf{(6) Shared Focus Mechanisms:} Both employ Scaled Covariance Focus and Multi-Head Parseval Focus, ensuring consistent treatment of functional relationships and cross-domain consistency throughout encoding and decoding.

\subsubsection{Design Philosophy: Physical Structure, Not Semantic Content}

Critically, the decoder operates entirely within the physical domain. The "target" encoded in the bottleneck latent is a specification of \textit{which physical structure to reconstruct}, not a semantic category:

\begin{itemize}
    \item \textbf{Denoising:} Latent encodes "reconstruct the signal component, not the noise component"
    \item \textbf{Source Separation:} Latent encodes "reconstruct speaker A's voice characteristics, not speaker B's"
    \item \textbf{Signal Restoration:} Latent encodes "reconstruct the clean signal structure, not the artifacts"
\end{itemize}

In all cases, the decoder is selecting and reconstructing physical features—spectral content, temporal dynamics, phase relationships—not learning semantic categories like "jazz" or "classical." 

\subsubsection{Enabling Zero-Shot Transfer}

The decoder's design directly enables the zero-shot cross-modal transfer 
demonstrated in the main paper (Table~\ref{tab:frozenEncoder_results}). By 
conditioning all reconstruction operations on the encoder's domain-invariant 
physical representations rather than domain-specific patterns, the decoder 
can reconstruct signals from unseen domains:

\begin{itemize}
    \item \textbf{RF $\rightarrow$ Audio:} Encoder captures temporal dynamics 
    and spectral structure; decoder reconstructs audio waveforms respecting 
    these physical constraints
    \item \textbf{RF $\rightarrow$ Images:} Encoder captures spatial frequency 
    content (via 1D unwrapping); decoder reconstructs 2D structure from 1D 
    representations
    \item \textbf{Denoising/Separation:} Skip sinks dynamically filter based 
    on latent-encoded targets, removing noise or interfering sources without 
    domain-specific training
\end{itemize}

This zero-shot capability supports our central hypothesis: by learning 
physical structure rather than semantic content, both encoder and decoder 
generalize to unseen domains sharing similar physical transformations.

\subsubsection{Training-Specific Outputs}

During training, the decoder returns additional outputs for regularization:

\textbf{Head Orthogonalization Loss:} Aggregated across all multi-head focus mechanisms:
\begin{equation}
\mathcal{L}_{\text{orth}}^{\text{decoder}} = \sum_{i} \mathcal{L}_{\text{orth}}^{(\text{FiLM}, i)} + \sum_{i} \mathcal{L}_{\text{orth}}^{(\text{skip}, i)} + \sum_{i} \mathcal{L}_{\text{orth}}^{(\text{transformer}, i)} + \mathcal{L}_{\text{orth}}^{(\text{refinement})} + \mathcal{L}_{\text{orth}}^{(\text{activation})}
\end{equation}

\textbf{Skip Sink Outputs:} For each decoder layer with skip connections, we return $\mathbf{x}_{\text{sink}}^{(i)}$ (estimated noise/interference) and $\mathbf{x}_{\text{post-sink}}^{(i)}$ (cleaned skip connection). These enable the skip sink regularization losses described in Section~\ref{appendix:SkipSinks}.

\section{Appendix C: Auxiliary Networks and Training Objectives}
\label{appendix:C_AuxiliaryNetworks}

The encoder and decoder architectures (Appendices~\ref{appendix:A_Encoder_Architecture} 
and~\ref{appendix:B_Decoder}) provide the computational substrate for 
learning signal-theoretic principles. However, these architectures alone are 
insufficient—they must be trained with co-designed loss functions and auxiliary 
networks that generate meaningful gradients for physical representation learning. 
This appendix describes the auxiliary networks and training objectives that 
enable the encoder to learn domain-invariant physical structure.

\subsection{Design Philosophy: Co-Designed Auxiliary Networks}
\label{appendix:AuxiliaryPhilosophy}

Our training methodology combines three complementary objectives, each requiring 
dedicated auxiliary networks:

\begin{enumerate}
    \item \textbf{Global Invariance Learning (IsoFICReg):} Learn what should 
    remain the same across transformations (e.g., emitter identity despite noise)
    \item \textbf{Equivariance Learning (LED):} Learn how representations should 
    change predictably under transformations (e.g., frequency shifts)
    \item \textbf{Instance-Specific Learning (Reconstruction):} Learn fine-grained 
    physical details through reconstruction fidelity
\end{enumerate}

Each objective addresses different aspects of physical representation learning:
\begin{itemize}
    \item \textbf{IsoFICReg} captures global semantic structure (what the signal is)
    \item \textbf{LED} captures transformation rules (how the signal changes)
    \item \textbf{Reconstruction} captures instance-specific details (how to 
    reconstruct this particular signal)
\end{itemize}

Together, these objectives ensure the encoder learns representations that are 
both globally informative and locally discriminative, enabling frozen encoder transfer 
to unseen domains.

\subsection{Augmentations and Transformation Encoding}
\label{appendix:Augmentations}

We apply domain-invariant transformations during training to learn equivariant 
and invariant symmetries. Each transformation is encoded as a bounded code in 
$[0, 1]$ to facilitate sigmoid-activated regression heads. These codes serve 
dual purposes: (1) conditioning tokens in the Latent Equivariant Transformer 
(Section~\ref{appendix:LatentEquivariantTransformer}), and 
(2) regression targets for equivariance losses 
(Section~\ref{appendix:EquivarianceLoss}).

\subsubsection{Augmentation Pipeline}
\label{appendix:AugmentationPipeline}

Augmentations are applied in the following sequence:
\begin{enumerate}
    \item \textbf{Frequency Shift:} Random $f_o \in [-0.33 \cdot f_s/2, 0.33 \cdot f_s/2]$
    \item \textbf{Phase Rotation:} Random $\phi \in [0, 2\pi)$
    \item \textbf{IQ Flip:} Random I/Q component flipping
    \item \textbf{Time Shift:} Random $\tau \in [-0.25L, 0.25L]$ samples
    \item \textbf{AWGN:} Random SNR $\in [-10, 100]$ dB
    \item \textbf{Unit Power Normalization:} Scale to unit power
    \item \textbf{IQ Interleaving:} Convert complex to real-valued interleaved sequence
\end{enumerate}

The transformation codes $\boldsymbol{\theta} = [\theta_{\text{freq}}, \theta_{\text{phase}}, 
\theta_{\text{h-flip}}, \theta_{\text{v-flip}}, \theta_{\text{time}}]$ are 
concatenated and used for latent conditioning and regression targets. AWGN 
parameters (noise power, signal power, SNR) are returned for regularization 
losses but not used for equivariance learning.

\subsubsection{Frequency Shifting}
\label{appendix:FrequencyShifting}

\textbf{Transformation:} Apply frequency offset $f_o \in [-f_{\text{max}}, f_{\text{max}}]$ where $f_{\text{max}} = 0.33 \cdot f_s/2$, via complex exponential multiplication followed by asymmetric anti-aliasing filtering:

\begin{equation}
\mathbf{x}_{\text{shifted}}[n] = \text{AAF}\left(\mathbf{x}[n] \cdot e^{j2\pi f_o n / f_s}, f_o, f_s\right)
\end{equation}

where the anti-aliasing filter (AAF) applies asymmetric frequency-domain masking based on the shift direction. For complex baseband signals with asymmetric spectra:

\begin{itemize}
    \item \textbf{Positive shifts} ($f_o > 0$): Suppress high positive frequencies beyond $(f_s/2 - |f_o|)$ to prevent aliasing above Nyquist
    \item \textbf{Negative shifts} ($f_o < 0$): Suppress low negative frequencies beyond $-(f_s/2 - |f_o|)$ to prevent aliasing below $-f_s/2$
\end{itemize}

This asymmetric filtering preserves the maximum signal bandwidth while preventing spectral wraparound after the frequency shift operation. The implementation uses FFT-based filtering with per-sample adaptive cutoff frequencies computed as:

\begin{equation}
f_{\text{cutoff}}^{+} = \begin{cases}
f_s/2 - |f_o| & \text{if } f_o \geq 0 \\
f_s/2 & \text{if } f_o < 0
\end{cases}, \quad
f_{\text{cutoff}}^{-} = \begin{cases}
f_s/2 & \text{if } f_o \geq 0 \\
f_s/2 - |f_o| & \text{if } f_o < 0
\end{cases}
\end{equation}

where $f_{\text{cutoff}}^{+}$ and $f_{\text{cutoff}}^{-}$ are the cutoff frequencies for positive and negative frequency components, respectively.

\textbf{Purpose:} Learn frequency translation equivariance—representations should 
transform predictably when signal frequency content shifts. 

\textbf{Transformation Code:} Shift amount normalized to $[0, 1]$:
\begin{equation}
\theta_{\text{freq}} = \frac{f_o + f_s/2}{f_s} \in [0, 1]
\end{equation}

The offset by $f_s/2$ ensures positive values for easier sigmoid regression.

\textbf{Learned Symmetry:} Frequency translation equivariance enables the model 
to recognize that a signal shifted by $\Delta f$ in input space should produce 
a predictable transformation in latent space, independent of the signal's original 
frequency content. This generalizes to arbitrary sample rates and frequency ranges.

\subsubsection{Phase Rotation}
\label{appendix:PhaseRotation}

\textbf{Transformation:} Apply random phase rotation $\phi \in [0, 2\pi)$:
\begin{equation}
x_{\text{rotated}}[n] = x[n] \cdot e^{j\phi}
\end{equation}

\textbf{Purpose:} Learn phase rotation equivariance—representations should 
transform predictably under global phase rotations. Unlike phase invariance, 
we learn how representations change with phase, enabling the model to track 
phase relationships when needed while remaining robust to arbitrary phase offsets.

\textbf{Transformation Code:} Phase normalized to $[0, 1]$:
\begin{equation}
\theta_{\text{phase}} = \frac{\phi}{2\pi} \in [0, 1]
\end{equation}

For regression, we use the circular variants of the regression losses~\ref{appendix:circularPhaseRegression}.

\textbf{Cross-Domain Generalization:} Phase rotation equivariance enables the 
model to handle signals with arbitrary time alignment (phase offset in time 
domain $\approx$ phase rotation in frequency domain), generalizing to speech 
processing where utterances have arbitrary start times.

\subsubsection{IQ Flipping (Spectral Inversion)}
\label{appendix:IQFlipping}

\textbf{Transformation:} Randomly flip real (I) and/or imaginary (Q) components:
\begin{align}
x_{\text{flipped}}[n] = \begin{cases}
x[n] & \text{no flip (25\% probability)} \\
-\text{Re}(x[n]) + j\text{Im}(x[n]) & \text{horizontal flip (25\%)} \\
\text{Re}(x[n]) - j\text{Im}(x[n]) & \text{vertical flip (25\%)} \\
-\text{Re}(x[n]) - j\text{Im}(x[n]) & \text{both flips (25\%)}
\end{cases}
\end{align}

\textbf{Purpose:} Learn equivariance to spectral inversion and conjugation. In 
RF systems, I/Q swapping can occur due to hardware configurations or sideband 
selection. This augmentation ensures representations are robust to such inversions.

\textbf{Transformation Codes:} Two binary indicators:
\begin{align}
\theta_{\text{h-flip}} &\in \{0, 1\} \quad \text{(horizontal flip applied)} \\
\theta_{\text{v-flip}} &\in \{0, 1\} \quad \text{(vertical flip applied)}
\end{align}

\textbf{Cross-Domain Generalization:} Spectral inversion equivariance generalizes 
to domains where sign conventions or orientation may vary (e.g., image flipping, 
audio polarity inversion).

\subsubsection{Time Shifting}
\label{appendix:TimeShifting}

\textbf{Transformation:} Apply random time shift $\tau \in [-\tau_{\max}, \tau_{\max}]$ 
where $\tau_{\max} = 0.25 \cdot L_{\text{seq}}$ (25\% of sequence length). Shifted 
regions are padded with complex Gaussian noise at SNR = 70 dB:
\begin{equation}
x_{\text{shifted}}[n] = \begin{cases}
x[n - \tau] & \text{if } 0 \leq n - \tau < L_{\text{seq}} \\
\mathcal{CN}(0, \sigma_{\text{noise}}^2) & \text{otherwise (padding)}
\end{cases}
\end{equation}

\textbf{Purpose:} Learn time translation equivariance—representations should 
transform predictably when signals are temporally shifted. This is fundamental 
for processing signals with arbitrary time alignment and for learning causal 
structure independent of absolute time reference.

\textbf{Transformation Code:} Shift amount normalized to $[0, 1]$:
\begin{equation}
\theta_{\text{time}} = \frac{\tau / \tau_{\max} + 1}{2} \in [0, 1]
\end{equation}

This maps $[-\tau_{\max}, \tau_{\max}]$ to $[0, 1]$ for sigmoid regression.

\textbf{Learned Symmetry:} Time translation equivariance enables the model to 
recognize temporal patterns regardless of their position in the sequence. This 
generalizes to variable-length signals and enables processing signals with 
arbitrary time alignment without retraining.

\subsubsection{Additive White Gaussian Noise (AWGN)}
\label{appendix:AWGN}

\textbf{Transformation:} Add complex Gaussian noise at target SNR $\in [-10, 100]$ dB:
\begin{align}
P_{\text{signal}} &= \mathbb{E}[|x[n]|^2] \\
P_{\text{noise}} &= P_{\text{signal}} / 10^{\text{SNR}/10} \\
x_{\text{noisy}}[n] &= x[n] + \sqrt{P_{\text{noise}}} \cdot \mathcal{CN}(0, 1)
\end{align}

where $\mathcal{CN}(0, 1)$ is complex Gaussian noise with unit variance.

\textbf{Purpose:} Learn noise invariance—representations should remain unchanged 
despite additive noise. Unlike the equivariant transformations, AWGN is a 
destructive transformation: noise does not preserve signal structure and should 
be removed rather than tracked.

\textbf{No Transformation Code:} AWGN parameters (noise power, signal power, 
SNR) are returned for use in regularization losses (e.g., Noise Sink power 
matching, Section~\ref{appendix:ConvTokenization}) but are \textbf{not} 
used for equivariance learning or latent conditioning. The model learns noise 
invariance through IsoFICReg (Section~\ref{appendix:IsoFICReg}): 
noisy versions of the same signal should produce identical representations.

\textbf{Negative-SNR Regime:} The SNR range extends to -10 dB, enabling learning 
in regimes where noise power exceeds signal power. Combined with the 
Noise Sink mechanism (Section~\ref{appendix:ConvTokenization}) and 
Latent Coherent Integration (Section~\ref{appendix:IsoFICReg}), 
this enables robust representation learning even when signal is deeply buried 
in noise.

\textbf{Cross-Domain Generalization:} Noise invariance learned on RF generalizes 
to any domain with additive noise: sensor noise in images, background noise in 
audio, measurement noise in seismology. The model learns to extract signal 
structure independent of noise level.

\subsubsection{Unit Power Normalization}
\label{appendix:UnitPowerNormalization}

After all augmentations, signals are normalized to unit power:
\begin{equation}
x_{\text{norm}}[n] = \frac{x[n]}{\sqrt{\mathbb{E}[|x[n]|^2]}}
\end{equation}

\textbf{Purpose:} Ensure consistent signal power across batches, preventing 
the model from learning spurious correlations with absolute power levels. This 
normalization is applied after AWGN injection, so the model learns representations 
at a fixed signal power but varying noise levels (controlled by SNR).

\textbf{Not an Augmentation:} This is a preprocessing step rather than an 
augmentation—it does not introduce variability but standardizes the input 
distribution.

\subsubsection{IQ Interleaving (Sample Format)}
\label{appendix:IQInterleaving}

Complex-valued signals are converted to real-valued interleaved sequences:
\begin{equation}
x_{\text{interleaved}} = [\text{Re}(x[0]), \text{Im}(x[0]), \text{Re}(x[1]), \text{Im}(x[1]), \ldots]
\end{equation}

\textbf{Purpose:} Enable processing with real-valued neural network operations 
while preserving complex-valued structure. The interleaving ensures that I and 
Q components remain paired throughout processing, facilitating seamless FFT/IFFT 
operations (Section~\ref{appendix:ConvTokenization}).

\textbf{Not an Augmentation:} This is a format conversion rather than an 
augmentation—it preserves all information while changing representation.

\subsubsection{Symmetry Learning: Equivariance vs. Invariance}
\label{appendix:SymmetryLearning}

The augmentation pipeline encodes two types of symmetries:

\textbf{Equivariant Symmetries (Structure-Preserving Transformations):}
\begin{itemize}
    \item \textbf{Frequency translation:} Shift by $\Delta f$ $\rightarrow$ 
    predictable latent transformation
    \item \textbf{Phase rotation:} Rotate by $\phi$ $\rightarrow$ predictable 
    latent transformation
    \item \textbf{Spectral inversion:} Flip I/Q $\rightarrow$ predictable latent 
    transformation
    \item \textbf{Time translation:} Shift by $\tau$ $\rightarrow$ predictable 
    latent transformation
\end{itemize}

These are learned through LED (Section~\ref{appendix:LED}): the 
model learns transformation rules by predicting how representations change under 
these operations.

\textbf{Invariant Symmetry (Destructive Transformation):}
\begin{itemize}
    \item \textbf{Noise:} Add AWGN $\rightarrow$ representation unchanged
\end{itemize}

This is learned through IsoFICReg (Section~\ref{appendix:IsoFICReg}): 
noisy versions of the same signal should produce identical representations.

\textbf{Cross-Domain Transfer Mechanism:} By learning these domain-invariant 
symmetries on RF data, the model acquires transformation rules that generalize 
to any domain respecting the same mathematical structure:
\begin{itemize}
    \item Frequency shift in RF $\approx$ frequency shift in audio
    \item Time shift in RF $\approx$ time shift in seismology
    \item Phase rotation in RF $\approx$ time alignment in speech
    \item Noise in RF $\approx$ sensor noise in images
\end{itemize}

This explains the cross-modal transfer demonstrated in Table~\ref{tab:frozenEncoder_results}: 
the model learned physical transformation rules on RF that apply generally 
to structured signals, enabling strong frozen-encoder transfer on unseen domains without 
fine-tuning the encoder on target domains.

\subsection{Latent Equivariant Differences (LED)}
\label{appendix:LED}

\subsubsection{Theoretical Foundation}
\label{appendix:LEDTheory}

Equivariant symmetry learning aims to ensure representations transform predictably 
under domain-invariant operations. Formally, for a transformation $g$ and 
representation function $f$, equivariance requires:
\begin{equation}
g(f(x)) = f(g(x))
\end{equation}

In contrast, invariant symmetry (for destructive distortions like noise) requires:
\begin{equation}
g(f(x)) = f(x)
\end{equation}

We learn both symmetries through a dual-branch Joint Embedding Predictive 
Architecture (JEPA):

\textbf{Clean Branch:} Processes raw signals with no transformations, producing 
$\mathbf{z}_{\text{clean}}$

\textbf{Augmented Branch:} Processes signals with applied transformations 
(frequency shifts, time shifts, phase rotations, AWGN), producing $\mathbf{z}_{\text{aug}}$

The key insight: transformations for which we want equivariance are explicitly 
modeled in latent space via a Latent Equivariant Transformer, while transformations 
for which we want invariance (noise) are handled implicitly through the invariance 
criterion (Section~\ref{appendix:IsoFICReg}).

\subsubsection{Latent Equivariant Transformer}
\label{appendix:LatentEquivariantTransformer}

The Latent Equivariant Transformer applies transformations directly to encoded 
token representations, enabling the model to learn how latent features should 
change under input-space transformations.

\paragraph{Architecture}

Given clean branch tokens $\mathbf{T}_{\text{clean}} \in \mathbb{R}^{B \times N \times d}$ 
and transformation parameters $\boldsymbol{\theta} \in \mathbb{R}^{B \times d_{\theta}}$ 
(e.g., frequency shift amount, time shift amount, phase rotation angle), the 
Latent Equivariant Transformer produces transformed tokens 
$\mathbf{T}_{\text{latent\_transform}} \in \mathbb{R}^{B \times N \times d}$.

\textbf{Step 1: Parameter Projection and Prepending}

Transformation parameters are linearly projected to token dimension and prepended 
as conditioning tokens:
\begin{align}
\boldsymbol{\theta}_{\text{token}} &= \mathbf{W}_{\theta} \boldsymbol{\theta} + \mathbf{b}_{\theta} \in \mathbb{R}^{B \times d} \\
\mathbf{T}_{\text{aug}} &= [\boldsymbol{\theta}_{\text{token}}; \mathbf{T}_{\text{clean}}] \in \mathbb{R}^{B \times (N+1) \times d}
\end{align}

\textbf{Rationale:} Prepending the transformation token (rather than cross-attention) 
enables bidirectional communication between the transformation and all latent 
tokens. Each token's transformation is informed by neighboring context: the 
self-attention mechanism allows tokens to understand how their transformation 
affects and is affected by other tokens in the sequence. This is critical for 
transformations like frequency shifts where the entire spectrum shifts coherently—each 
frequency bin's transformation must be coordinated with all other bins. 
Cross-attention would create one-way communication (transformation $\rightarrow$ tokens) 
without allowing tokens to coordinate their transformations with each other.

\textbf{Step 2: Parseval Transformer Processing}

The augmented sequence is processed through a Parseval Transformer block 
(Section~\ref{appendix:MHPF}) to apply the latent transformation:
\begin{equation}
\mathbf{T}_{\text{transformed}}, \mathcal{L}_{\text{orth}}, \mathcal{L}_{\text{Parseval}}, \mathcal{L}_{\text{diversity}} = \text{ParsevalTransformer}(\mathbf{T}_{\text{aug}})
\end{equation}

The conditioning token is removed after transformation:
\begin{equation}
\mathbf{T}_{\text{latent\_transform}} = \mathbf{T}_{\text{transformed}}[:, 1:, :] \in \mathbb{R}^{B \times N \times d}
\end{equation}

\textbf{Step 3: Sequence Pooling and Cross-Domain Fusion}

Transformed tokens undergo the same processing as encoder tokens 
(Section~\ref{appendix:SequencePooling}):
\begin{align}
\mathbf{z}_{\text{latent\_transform}}^{(t)} &= \text{AttentionalPool}(\mathbf{T}_{\text{latent\_transform}}^{(t)}) \\
\mathbf{z}_{\text{latent\_transform}}^{(f)} &= \text{AttentionalPool}(\mathbf{T}_{\text{latent\_transform}}^{(f)}) \\
\mathbf{z}_{\text{latent\_transform}} &= \text{CrossDomainFusion}(\mathbf{z}_{\text{latent\_transform}}^{(t)}, \mathbf{z}_{\text{latent\_transform}}^{(f)})
\end{align}

\textbf{Domain-Specific Transformers:} Separate Latent Equivariant Transformers 
are used for time and frequency domains to enable domain-specific transformation 
learning if necessary.

\subsubsection{Equivariance Learning through Differences}
\label{appendix:DifferenceLearning}

After processing through the encoder and Latent Equivariant Transformer, we 
have three representations:
\begin{align}
\mathbf{z}_{\text{clean}} &\in \mathbb{R}^{B \times d} \quad \text{(clean branch, no transformations)} \\
\mathbf{z}_{\text{latent\_transform}} &\in \mathbb{R}^{B \times d} \quad \text{(clean branch, latent-space transformation)} \\
\mathbf{z}_{\text{input\_transform}} &\in \mathbb{R}^{B \times d} \quad \text{(augmented branch, input-space transformation)}
\end{align}

To learn equivariance, we measure the transformation process rather than just 
the output. We compute differences after the sequence pooling but before expander/projection networks are used for IsoFICReg.

The differences capture "how the representation changed":
\begin{align}
\Delta_{\text{latent}} &= \mathbf{z}_{\text{latent}} - \mathbf{z}_{\text{clean}} \in \mathbb{R}^{B \times d} \\
\Delta_{\text{input}} &= \mathbf{z}_{\text{input}} - \mathbf{z}_{\text{clean}} \in \mathbb{R}^{B \times d}
\end{align}

\textbf{Comprehensive Time-Frequency Equivariance:} To learn equivariances 
across both domains, we concatenate time and frequency differences:
\begin{align}
\Delta_{\text{latent}}^{\text{full}} &= [\Delta_{\text{latent}}^{(t)}; \Delta_{\text{latent}}^{(f)}] \in \mathbb{R}^{B \times 2d} \\
\Delta_{\text{input}}^{\text{full}} &= [\Delta_{\text{input}}^{(t)}; \Delta_{\text{input}}^{(f)}] \in \mathbb{R}^{B \times 2d}
\end{align}

\subsubsection{Equivariance Loss: Focal Huber Regression}
\label{appendix:EquivarianceLoss}

The equivariance loss enforces three constraints using focal-weighted Huber 
regression:

\textbf{(1) Latent transformation matches ground truth:}
\begin{equation}
\mathcal{L}_{\text{latent}} = \text{FocalHuber}(f_{ep}(\Delta_{\text{latent}}^{\text{full}}), \boldsymbol{\theta}_{\text{true}})
\end{equation}

\textbf{(2) Input transformation matches ground truth:}
\begin{equation}
\mathcal{L}_{\text{input}} = \text{FocalHuber}(f_{ep}(\Delta_{\text{input}}^{\text{full}}), \boldsymbol{\theta}_{\text{true}})
\end{equation}

\textbf{(3) Both transformations are consistent:}
\begin{equation}
\mathcal{L}_{\text{consistency}} = \text{FocalHuber}(\Delta_{\text{latent}}^{\text{full}}, \Delta_{\text{input}}^{\text{full}})
\end{equation}

The total equivariance loss is:
\begin{equation}
\mathcal{L}_{\text{Equi}} = \mathcal{L}_{\text{latent}} + \mathcal{L}_{\text{input}} + \mathcal{L}_{\text{consistency}}
\end{equation}

\paragraph{Focal-Weighted Huber Loss}

To emphasize difficult instances (small shifts, low SNR) while maintaining 
robustness to outliers, we combine Huber loss with focal reweighting:

\begin{align}
\text{Huber}_{\delta}(y, \hat{y}) &= \begin{cases}
\frac{1}{2}(y - \hat{y})^2 & \text{if } |y - \hat{y}| \leq \delta \\
\delta \cdot (|y - \hat{y}| - \frac{1}{2}\delta) & \text{otherwise}
\end{cases} \\
d_{\text{local}} &= \text{mean}(\text{Huber}_{\delta}(y, \hat{y}), \text{dim}=-1) \\
d_{\text{norm}} &= \frac{d_{\text{local}}}{\text{mean}(d_{\text{local}}) + \epsilon} \\
\sigma_{\text{global}} &= \text{std}(d_{\text{local}}) \\
\gamma_{\text{adaptive}} &= \gamma \cdot (1 + \log(1 + \sigma_{\text{global}})) \\
w_{\text{focal}} &= \max(0.1, d_{\text{norm}}^{\gamma_{\text{adaptive}}}) \\
\text{FocalHuber}(y, \hat{y}) &= \text{mean}(w_{\text{focal}} \cdot d_{\text{local}})
\end{align}

where $\delta = 0.05$ (Huber threshold), $\gamma = 2.0$ (focal exponent), and 
$\epsilon = 10^{-8}$ (numerical stability). The focal weight is clamped to 
$[0.1, \infty)$ to prevent complete down-weighting of easy examples while 
emphasizing hard cases.

\textbf{Rationale:} Huber loss provides robustness to outliers (e.g., occasional 
large prediction errors during early training), while focal reweighting prevents 
overfitting to trivial transformations (e.g., 0 Hz shift, high SNR). The 
adaptive $\gamma$ adjusts emphasis based on batch difficulty.

\textbf{Note:} Empirically we noted this variation of a batch normalized Focal reweighting worked best for regression tasks relative to the Focal reweighting used for IsoFICReg. They are similar in nature but have minute differences in implementation that are mainly motivated to the scale differences of the respective errors.

\paragraph{Circular Phase Regression}
\label{appendix:circularPhaseRegression}

For phase rotation parameters $\phi \in [0, 2\pi)$, we account for circular 
periodicity using modulo wrapping rather than sin/cos decomposition. The 
prediction head outputs $\hat{\phi} \in [0, 1]$ (via sigmoid activation), 
which is scaled back to $[0, 2\pi)$:
\begin{equation}
\hat{\phi}_{\text{scaled}} = 2\pi \cdot \hat{\phi}
\end{equation}

The circular-aware loss wraps the error to $[-\pi, \pi]$:
\begin{align}
\Delta_{\phi} &= (\phi_{\text{true}} - \hat{\phi}_{\text{scaled}} + \pi) \mod 2\pi - \pi \\
\mathcal{L}_{\text{phase}} &= \text{FocalHuber}(\Delta_{\phi}, 0)
\end{align}

This ensures $\phi = 0$ and $\phi = 2\pi$ are treated as equivalent, avoiding 
discontinuities at the circular boundary.

\subsubsection{Summary: LED Pipeline}

Algorithm~\ref{alg:LED} summarizes the complete LED forward pass.

\begin{algorithm}[H]
\caption{Latent Equivariant Differences (LED)}
\label{alg:LED}
\begin{algorithmic}[1]
\REQUIRE Clean signal $\mathbf{x}_{\text{clean}}$, Augmented signal $\mathbf{x}_{\text{aug}}$
\REQUIRE Transformation parameters $\boldsymbol{\theta}_{\text{true}}$
\ENSURE Equivariance loss $\mathcal{L}_{\text{Equi}}$
\STATE
\STATE \COMMENT{\textbf{Encode clean and augmented branches}}
\STATE $\mathbf{z}_{\text{clean}}^{(t)}, \mathbf{z}_{\text{clean}}^{(f)} \leftarrow \text{Encoder}(\mathbf{x}_{\text{clean}})$
\STATE $\mathbf{z}_{\text{input}}^{(t)}, \mathbf{z}_{\text{input}}^{(f)} \leftarrow \text{Encoder}(\mathbf{x}_{\text{aug}})$
\STATE
\STATE \COMMENT{\textbf{Apply latent transformation}}
\STATE $\mathbf{z}_{\text{latent}}^{(t)} \leftarrow \text{LatentEquivariantTransformer}^{(t)}(\mathbf{z}_{\text{clean}}^{(t)}, \boldsymbol{\theta}_{\text{true}})$
\STATE $\mathbf{z}_{\text{latent}}^{(f)} \leftarrow \text{LatentEquivariantTransformer}^{(f)}(\mathbf{z}_{\text{clean}}^{(f)}, \boldsymbol{\theta}_{\text{true}})$
\STATE
\STATE \COMMENT{\textbf{Compute differences}}
\STATE $\Delta_{\text{latent}}^{\text{full}} \leftarrow [(\mathbf{z}_{\text{latent}}^{(t)} - \mathbf{z}_{\text{clean}}^{(t)}); (\mathbf{z}_{\text{latent}}^{(f)} - \mathbf{z}_{\text{clean}}^{(f)})]$
\STATE $\Delta_{\text{input}}^{\text{full}} \leftarrow [(\mathbf{z}_{\text{input}}^{(t)} - \mathbf{z}_{\text{clean}}^{(t)}); (\mathbf{z}_{\text{input}}^{(f)} - \mathbf{z}_{\text{clean}}^{(f)})]$
\STATE
\STATE \COMMENT{\textbf{Predict transformation parameters}}
\STATE $\hat{\boldsymbol{\theta}}_{\text{latent}} \leftarrow f_{ep}(\Delta_{\text{latent}}^{\text{full}})$
\STATE $\hat{\boldsymbol{\theta}}_{\text{input}} \leftarrow f_{ep}(\Delta_{\text{input}}^{\text{full}})$
\STATE
\STATE \COMMENT{\textbf{Compute equivariance loss}}
\STATE $\mathcal{L}_{\text{latent}} \leftarrow \text{FocalHuber}(\hat{\boldsymbol{\theta}}_{\text{latent}}, \boldsymbol{\theta}_{\text{true}})$
\STATE $\mathcal{L}_{\text{input}} \leftarrow \text{FocalHuber}(\hat{\boldsymbol{\theta}}_{\text{input}}, \boldsymbol{\theta}_{\text{true}})$
\STATE $\mathcal{L}_{\text{consistency}} \leftarrow \text{FocalHuber}(\Delta_{\text{latent}}^{\text{full}}, \Delta_{\text{input}}^{\text{full}})$
\STATE $\mathcal{L}_{\text{Equi}} \leftarrow \mathcal{L}_{\text{latent}} + \mathcal{L}_{\text{input}} + \mathcal{L}_{\text{consistency}}$
\STATE
\RETURN $\mathcal{L}_{\text{Equi}}$
\end{algorithmic}
\end{algorithm}

\subsection{Token-Level Source Separation}
\label{appendix:SourceSeparation}

\subsubsection{Motivation and Placement}

Token-level source separation provides local discriminative learning to complement 
global objectives (IsoFICReg, LED). The Latent Source Separation Transformer 
operates \textbf{after the main encoder's Parseval Transformer blocks}, processing 
tokens that have already undergone dual-domain focus and cross-domain fusion.

\textbf{Processing Flow:}
\begin{enumerate}
    \item Encoder: Convolutional tokenization $\rightarrow$ Parseval Transformer 
    $\rightarrow$ tokens $\mathbf{T}_{\text{mixture}}$
    \item \textbf{Source Separation:} $\mathbf{T}_{\text{mixture}} \rightarrow$ 
    ICA Transformer $\rightarrow$ $\mathbf{T}_A, \mathbf{T}_B$
    \item Remaining Encoder: Cross-domain fusion $\rightarrow$ sequence pooling 
    $\rightarrow$ final cross-domain fusion $\rightarrow$ $\mathbf{z}_A, \mathbf{z}_B$
    \item Decoder: Reconstruct $\mathbf{x}_A, \mathbf{x}_B$ from separated representations
\end{enumerate}

This placement ensures separation operates on physically-rich tokens rather 
than raw convolutional features, improving separation quality.

\subsubsection{Mixture Creation and SINR Control}

To avoid trivializing source separation to denoising, we create controlled 
mixtures with Signal-to-Interference-plus-Noise Ratio (SINR) ranging from 20 dB 
to 0 dB:
\begin{equation}
\mathbf{x}_{\text{mixture}} = \mathbf{x}_A + \alpha \cdot \mathbf{x}_B
\end{equation}

where $\alpha$ is computed to achieve target SINR:
\begin{equation}
\alpha = \sqrt{\frac{P_A}{P_B \cdot 10^{\text{SINR}/10}}}
\end{equation}

\textbf{Source Ambiguity Resolution:} Source A is always the dominant source 
(higher power), while Source B is the interfering source. This removes the 
permutation ambiguity inherent in blind source separation. We explicitly separate 
and compute losses on both sources to ensure the task requires true separation 
rather than simple attenuation.

\subsubsection{Latent Source Separation Transformer}
\label{appendix:LatentSourceSeparationTransformer}

The Latent Source Separation Transformer combines Parseval Focus with ICA-inspired iterative refinement for source separation.

\paragraph{Architecture}

\textbf{Step 1: High-Dimensional Projection}
\begin{align}
\mathbf{T}_{\text{proj}} &= \text{GELU}(\text{BatchNorm}(\text{Linear}_1(\mathbf{T}_{\text{mixture}}))) \in \mathbb{R}^{B \times N \times d_{\text{proj}}} \\
\mathbf{T}_{\text{proj}} &= \text{Dropout}_{0.25}(\mathbf{T}_{\text{proj}}) \\
\mathbf{T}_{\text{proj}} &= \text{GELU}(\text{BatchNorm}(\text{Linear}_2(\mathbf{T}_{\text{proj}}))) \in \mathbb{R}^{B \times N \times d_{\text{proj}}} \\
\mathbf{T}_{\text{proj}} &= \text{Dropout}_{0.25}(\mathbf{T}_{\text{proj}})
\end{align}

where $d_{\text{proj}} = 4\cdot d_{\text{input}}$.

\textbf{Step 2: Parseval Transformer for Joint Time-Frequency Analysis}
\begin{equation}
\mathbf{T}_{\text{focused}}, \mathcal{L}_{\text{orth}}, \mathcal{L}_{\text{Parseval}}, \mathcal{L}_{\text{diversity}} = \text{ParsevalTransformer}(\mathbf{T}_{\text{proj}})
\end{equation}

\textbf{Step 3: ICA-Inspired Iterative Refinement}

We apply $K=3$ iterations of ICA-inspired processing with orthogonal linear 
layers:
\begin{align}
\mathbf{T}^{(k)} &= \mathbf{W}_{\text{ICA}}^{(k)} \mathbf{T}^{(k-1)} \quad \text{where } \mathbf{W}_{\text{ICA}}^{(k)} \text{ is orthogonal} \\
\mathbf{T}^{(k)} &= \mathbf{T}^{(k)} - \mathbb{E}[\mathbf{T}^{(k)}] \quad \text{(centering)} \\
\mathbf{T}^{(k)} &= \log(\cosh(\mathbf{T}^{(k)})) \quad \text{(log-cosh nonlinearity)} \\
\mathbf{T}^{(k)} &= \frac{\mathbf{T}^{(k)}}{\|\mathbf{T}^{(k)}\|_2 + \epsilon} \quad \text{(L2 normalization)}
\end{align}

where $\mathbf{T}^{(0)} = \mathbf{T}_{\text{focused}}$ and $k \in \{1, 2, 3\}$.

The log-cosh nonlinearity approximates negentropy (non-Gaussianity), encouraging 
statistical independence between separated sources. Interestingly, the log-cosh nonlinearity should remove sign information due to its evenness but in practice we observed better training dynamics with its inclusion rather than signed alternatives like tanh or asinh. 

We attribute this to its nice differentiability properties such as a stable bounded derivative between [-1,1] from its tanh derivative. Additionally, its inclusion within the joint-embedding framework and follow-on high dimensional projections provide a rich surface to leverage its rich differentiability properties.

\textbf{Step 4: Final Unmixing}
\begin{equation}
\mathbf{T}_{\text{unmixed}} = \mathbf{W}_{\text{unmix}} \mathbf{T}^{(3)} \in \mathbb{R}^{B \times N \times 2d_{\text{input}}}
\end{equation}

where $\mathbf{W}_{\text{unmix}}$ is initialized as orthogonal. The output is 
split into two sources:
\begin{align}
\mathbf{T}_A &= \mathbf{T}_{\text{unmixed}}[:, :, :d_{\text{input}}] \\
\mathbf{T}_B &= \mathbf{T}_{\text{unmixed}}[:, :, d_{\text{input}}:]
\end{align}

\textbf{Rationale:} The iterative ICA refinement with orthogonal constraints 
encourages learning a rotation in latent space that maximizes statistical 
independence between sources. The Parseval Transformer provides joint time-frequency 
context essential for effective separation.

\paragraph{Downstream Processing}

After separation, each source is processed independently through the remaining 
encoder and decoder:
\begin{itemize}
    \item \textbf{Encoder:} Parseval Transformer, sequence pooling, cross-domain fusion
    \item \textbf{Decoder:} Reconstruction targeting the original unmixed sources
    \item \textbf{Losses:} IsoFICReg, LED, and reconstruction losses computed 
    against clean single-source counterparts
\end{itemize}

This ensures separated tokens produce representations and reconstructions 
consistent with the original unmixed signals.

\subsubsection{Source Separation Losses}

\textbf{(1) Reconstruction Loss:} Each separated source is reconstructed and 
compared to its original unmixed target:
\begin{align}
\mathcal{L}_{\text{recon}}^{(A)} &= \text{FocalHuber}(\mathbf{x}_{\text{recon}}^{(A)}, \mathbf{x}_{A-No Noise}) \\
\mathcal{L}_{\text{recon}}^{(B)} &= \text{FocalHuber}(\mathbf{x}_{\text{recon}}^{(B)}, \mathbf{x}_{B-No Noise})
\end{align}

\textbf{(2) Skip Sink Complementarity Loss:} The decoder's skip sinks 
(Section~\ref{appendix:SkipSinks}) should produce complementary 
estimates for mixed sources: what is sunk from source A should equal what is 
kept for source B, and vice versa.

For each decoder layer $i$ with skip connections:
\begin{align}
\mathcal{L}_{\text{sink\_comp}}^{(i)} &= \text{FocalHuber}(\mathbf{x}_{\text{sink}}^{(A,i)}, \mathbf{x}_{\text{post-sink}}^{(B,i)}) \\
&\quad + \text{FocalHuber}(\mathbf{x}_{\text{sink}}^{(B,i)}, \mathbf{x}_{\text{post-sink}}^{(A,i)})
\end{align}

Applied to both time and frequency domains:
\begin{equation}
\mathcal{L}_{\text{sink\_comp}} = \sum_{i} \left(\mathcal{L}_{\text{sink\_comp}}^{(t,i)} + \mathcal{L}_{\text{sink\_comp}}^{(f,i)}\right)
\end{equation}

\textbf{Rationale:} This complementarity constraint ensures the skip sinks learn 
to separate sources rather than arbitrarily attenuate. If the sink correctly 
removes source B from the mixture (leaving source A), then by definition it has 
extracted source B. This provides a strong regularization signal for learning 
effective source-specific filtering.

\textbf{Note on Scale Consistency:} The complementarity constraint is scale-consistent because both decoders process the same mixture $\mathbf{x}_{\text{mixture}} = \mathbf{x}_A + \alpha \cdot \mathbf{x}_B$. When decoder $A$ removes the interfering source ($\mathbf{x}_{\text{sink}}^{(A)} \approx \alpha \cdot \mathbf{x}_B$), this matches what decoder $B$ retains ($\mathbf{x}_{\text{post-sink}}^{(B)} \approx \alpha \cdot \mathbf{x}_B$). Similarly, what decoder $B$ removes ($\mathbf{x}_{\text{sink}}^{(B)} \approx \mathbf{x}_A$) matches what decoder $A$ retains ($\mathbf{x}_{\text{post-sink}}^{(A)} \approx \mathbf{x}_A$). The $\alpha$ scaling factor naturally appears on both sides of each comparison, ensuring scale consistency without explicit normalization.

\textbf{(3) Latent Space Losses:} IsoFICReg and LED losses are computed for 
each separated source against their clean single-source counterparts, providing 
implicit latent denoising and ensuring separated representations match the 
structure of clean signals.

\textbf{(4) Token-Level Separation Loss:} To enforce fine-grained separation 
at the token level, we apply a regression loss directly to the separated token 
representations before sequence pooling.

After the Latent Source Separation Transformer produces separated tokens $\mathbf{T}_A, \mathbf{T}_B$ 
(Section~\ref{appendix:LatentSourceSeparationTransformer}), we compute focal Huber regression against the original 
unmixed token representations:
\begin{align}
\mathcal{L}_{\text{token}}^{(t)} &= \text{FocalHuber}(\mathbf{T}_A, \mathbf{T}_A^{\text{target}}) + \text{FocalHuber}(\mathbf{T}_B, \mathbf{T}_B^{\text{target}}) \\
\mathcal{L}_{\text{token}}^{(f)} &= \text{FocalHuber}(\mathbf{T}_A^{(f)}, \mathbf{T}_A^{\text{target},(f)}) + \text{FocalHuber}(\mathbf{T}_B^{(f)}, \mathbf{T}_B^{\text{target},(f)})
\end{align}

where $\mathbf{T}_A^{\text{target}}, \mathbf{T}_B^{\text{target}}$ are the 
token representations from encoding the original unmixed sources.

\textbf{Rationale:} This token-level loss enforces separation precision at the 
finest granularity, enabling the model to learn localized separation strategies. 
For mixtures where sources occupy different time segments (e.g., 50\% duty cycle 
mixing) or different frequency bands (e.g., non-overlapping spectral content), 
the token-level loss provides dense gradients that guide the model to learn 
time-localized or frequency-selective filtering. This complements the global 
reconstruction losses (which operate on full sequences) by ensuring separation 
quality at every token position, particularly valuable for learning bandpass 
filter-like behavior when sources occupy distinct spectral regions.

\subsubsection{Limitation: Two-Source Separation}

Currently, we limit mixtures to two sources for computational efficiency and 
training stability. Extension to $N > 2$ sources is straightforward but requires 
additional computational resources and careful handling of permutation ambiguity.

\subsection{Isotropic Focal Invariance Covariance Regularization (IsoFICReg)}
\label{appendix:IsoFICReg}

IsoFICReg extends VICReg~\cite{bardes2021vicreg} with five key contributions 
that enable learning in extreme noise while maintaining fine-grained discrimination: 
(1) latent coherent integration leveraging weak supervision from dataset structure, 
(2) focal reweighting to emphasize hard examples, (3) repulsion loss for fine-grained discrimination, (4) isotropic representations 
via Z-score standardization, and (5) dual-domain consistency with latent denoising.

\subsubsection{Theoretical Foundation: VICReg}
\label{appendix:VICRegFoundation}

VICReg learns representations through three complementary objectives operating 
on projection head outputs $\mathbf{P}_a, \mathbf{P}_b \in \mathbb{R}^{B \times d}$:

\textbf{(1) Invariance:} Representations of augmented views should be similar:
\begin{equation}
\mathcal{L}_{\text{inv}} = \mathbb{E}[\|\mathbf{P}_a - \mathbf{P}_b\|^2]
\end{equation}

\textbf{(2) Variance:} Prevent collapse by maintaining variance along each dimension:
\begin{equation}
\mathcal{L}_{\text{var}} = \mathbb{E}[\text{ReLU}(1 - \sqrt{\text{Var}(\mathbf{P}_a) + \epsilon})] + \mathbb{E}[\text{ReLU}(1 - \sqrt{\text{Var}(\mathbf{P}_b) + \epsilon})]
\end{equation}

where $\epsilon = 10^{-4}$ for numerical stability.

\textbf{Note:} In our IsoFICReg formulation, we omit the variance regularization term. The Z-score standardization (described in Section~\ref{appendix:IsotropicRepresentations}) applied before computing invariance and covariance losses inherently prevents dimensional collapse by normalizing each dimension to unit variance, making an explicit variance loss redundant.

\textbf{(3) Covariance:} Decorrelate dimensions to maximize information:
\begin{equation}
\mathcal{L}_{\text{cov}} = \frac{1}{d}\sum_{i \neq j} \text{Cov}(\mathbf{P}_a)_{ij}^2 + \frac{1}{d}\sum_{i \neq j} \text{Cov}(\mathbf{P}_b)_{ij}^2
\end{equation}

The total VICReg loss is:
\begin{equation}
\mathcal{L}_{\text{VICReg}} = \lambda_{\text{inv}} \mathcal{L}_{\text{inv}} + \lambda_{\text{var}} \mathcal{L}_{\text{var}} + \lambda_{\text{cov}} \mathcal{L}_{\text{cov}}
\end{equation}

where $\lambda_{\text{inv}} = 25$, $\lambda_{\text{var}} = 25$, $\lambda_{\text{cov}} = 1$.

\paragraph{Projection Head Architecture (Expander Networks)}

Following precedent by the original VICReg and Zhang et al.~\cite{zhang2022self} we utilize dedicated projection heads per domain that expand 
latent representations to higher-dimensional space. Each are constructed as follows:
\begin{equation}
\text{ProjectionHead}: \mathbb{R}^{d} \rightarrow \mathbb{R}^{d_{\text{proj}}}
\end{equation}

where $d = 128$ (latent dimension) and $d_{\text{proj}} = 1024$ (projection dimension).

\begin{table}[h]
\centering
\caption{Projection Head Architecture (used for IsoFICReg)}
\label{tab:projection_head_arch}
\begin{tabular}{lcc}
\toprule
\textbf{Layer} & \textbf{Input Dim} & \textbf{Output Dim} \\
\midrule
Linear$_1$ & $d$ & $d_{\text{proj}}$ \\
BatchNorm$_1$ & $d_{\text{proj}}$ & $d_{\text{proj}}$ \\
GELU & - & - \\
Dropout (0.25) & - & - \\
\midrule
Linear$_2$ & $d_{\text{proj}}$ & $d_{\text{proj}}$ \\
BatchNorm$_2$ & $d_{\text{proj}}$ & $d_{\text{proj}}$ \\
GELU & - & - \\
Dropout (0.25) & - & - \\
\midrule
Linear$_3$ (no bias) & $d_{\text{proj}}$ & $d_{\text{proj}}$ \\
\bottomrule
\end{tabular}
\end{table}

\textbf{Purpose:} These projection heads serve IsoFICReg 
(Section~\ref{appendix:IsoFICReg}) and LED. By computing invariance on top of projections whose equivariant differences have already been enforced prior we force the 
representations to respect both symmetries simultaneously: the embeddings must 
remain invariant to noise while transforming predictably under frequency/time 
shifts. This coupling ensures learned representations encode physical transformation 
rules rather than arbitrary features.

\subsubsection{Contribution 1: Latent Coherent Integration}
\label{appendix:LatentCoherentIntegration}

Standard VICReg computes invariance loss between a single pair of augmented 
views per sample. We extend this by leveraging weak supervision from dataset 
structure: RF fingerprinting datasets contain multiple samples per emitter, 
each with unique hardware fingerprints but shared emitter identity.

\textbf{N-Choose-2 Pairwise Invariance:} For each emitter class $c$ with $N_c$ 
samples in the batch, we compute invariance loss for all $\binom{N_c}{2}$ pairs:
\begin{equation}
\mathcal{L}_{\text{inv}}^{(c)} = \frac{1}{\binom{N_c}{2}} \sum_{i < j} \|\mathbf{P}_a^{(i)} - \mathbf{P}_b^{(j)}\|^2
\end{equation}

where $i, j$ index samples from emitter $c$ in the augmented and latent-transformed 
branches respectively.

\textbf{Analogy to Coherent Integration:} This is analogous to coherent integration 
in classical signal processing~\cite{richards2005fundamentals}: by enforcing 
consistency across multiple noisy observations of the same emitter, noise 
(uncorrelated across samples) becomes inconsistent while signal (correlated 
emitter characteristics) becomes consistent. This effectively improves SNR in 
latent space, enabling learning with extreme AWGN injection in negative-SNR regimes.

\textbf{Distributed Training:} We train on 12 H100 GPUs with distributed data 
parallelism. Before computing invariance loss, we gather representations across 
all GPUs using \texttt{all\_gather}, increasing the effective batch size for 
N-choose-2 pairing from local batch size $B_{\text{local}}$ to global batch 
size $B_{\text{global}} = 12 \times B_{\text{local}}$. This dramatically increases 
the number of pairs per emitter, strengthening the coherent integration effect.

\subsubsection{Contribution 2: Focal Reweighting}
\label{appendix:FocalReweighting}

To prevent overfitting to trivial high-SNR samples and emphasize learning from 
difficult low-SNR samples, we apply focal reweighting to both invariance and 
covariance losses.

\paragraph{Focal Invariance Loss}

We extend the focal loss concept~\cite{lin2017focal} from classification to 
regression:
\begin{align}
d_{ij} &= \|\mathbf{P}_a^{(i)} - \mathbf{P}_b^{(j)}\|^2 \quad \text{(pairwise distance)} \\
w_{\text{focal}} &= (1 - e^{-d_{ij}})^{\gamma} \quad \text{(focal weight)} \\
\mathcal{L}_{\text{inv}}^{\text{focal}} &= \frac{1}{\binom{N_c}{2}} \sum_{i < j} w_{\text{focal}} \cdot d_{ij}\label{eq:focal_invariance}
\end{align}

where $\gamma = 2.0$. The focal weight emphasizes hard pairs (large $d_{ij}$, 
likely low-SNR) while down-weighting easy pairs (small $d_{ij}$, likely high-SNR).

\textbf{Rationale:} In negative-SNR regimes, most pairs have large distances 
due to noise. Without focal reweighting, the model would learn to minimize 
average distance without discriminating signal structure. Focal reweighting 
ensures the model focuses on the hardest pairs, which contain the most informative 
gradients for learning robust representations.

\paragraph{Focal Covariance Loss}

Similarly, we apply focal reweighting to covariance regularization:
\begin{align}
\mathbf{C}_a &= \frac{1}{B-1} \mathbf{P}_a^T \mathbf{P}_a \quad \text{(covariance matrix)} \\
c_{ij} &= \mathbf{C}_a[i, j]^2 \quad \text{for } i \neq j \text{ (off-diagonal elements)} \\
w_{\text{focal}} &= (1 - e^{-c})^{\gamma} \quad \text{(focal weight)} \\
\mathcal{L}_{\text{cov}}^{\text{focal}} &= \frac{1}{d(d-1)} \sum_{i < j} w_{\text{focal}} \cdot c_{ij}
\end{align}

\textbf{Rationale:} Focal reweighting emphasizes decorrelating dimensions with 
high covariance (harder to decorrelate) while down-weighting dimensions already 
well-decorrelated. This prevents the model from over-optimizing easy dimensions 
at the expense of hard dimensions, encouraging more isotropic representations.

\subsubsection{Contribution: Focal Repulsion Loss for Fine-Grained Discrimination}
\label{appendix:FocalRepulsionLoss}

To encourage fine-grained discrimination between different emitters, we complement the attraction-based invariance loss with a repulsion objective that pushes representations of different classes apart:

\begin{align}
d_{ij} &= \frac{\|\mathbf{P}_{\text{aug}}^{(i)} - \mathbf{P}_{\text{latent}}^{(j)}\|^2}{d_{\text{proj}}} \quad \text{(normalized squared distance)} \label{eq:repulsion_distance} \\
v_{ij} &= \text{ReLU}(\text{margin} - d_{ij}) \quad \text{(margin violation)} \label{eq:margin_violation} \\
w_{\text{focal}}^{repel} &= (1 - e^{-v_{ij}})^\gamma \quad \text{(focal weight on violations)} \label{eq:repulsion_focal} \\
\mathcal{L}_{\text{repel}} &= \frac{1}{N_{\text{neg}}} \sum_{i,j \in \mathcal{N}} w_{\text{focal}}^{repel} \cdot v_{ij} \label{eq:repulsion_loss}
\end{align}

where $\mathcal{N}$ denotes pairs from different emitter classes, $\text{margin} = 1024$, and $\gamma = 2.0$. We apply log1p scaling to the final loss: $\mathcal{L}_{\text{repel}}^{\text{final}} = \log(1 + \mathcal{L}_{\text{repel}})$.

\textbf{Rationale:} The ReLU in Eq.~\ref{eq:margin_violation} focuses computation only on pairs violating the margin (different classes that are too close). Among these violations, focal reweighting (Eq.~\ref{eq:repulsion_focal}) emphasizes the most severe cases: pairs with large violations $v_{ij}$ (very close negatives) receive weight $\approx 1$, while pairs with small violations (barely inside the margin) receive weight $\approx 0$. This concentrates gradients on the hardest negatives—different-class pairs that are erroneously similar—forcing the model to learn discriminative features that separate even subtle hardware differences.

\textbf{Note:} The large margin value (1024) is intentionally set well above the typical 
normalized distance range (~0-2) to create a constant repulsion force rather than 
a traditional margin-based dead zone. This design ensures continuous gradient flow 
that prevents trivial collapse to over-invariance, particularly for emitters from 
the same family with high redundancy. The focal weighting then modulates 
this base repulsion to emphasize hard negatives.

\textbf{Distinction from Invariance Focal Weighting:} While both use the same functional form $(1 - e^{-x})^\gamma$, they operate on different quantities:
\begin{itemize}
    \item \textbf{Invariance} (Eq.~\ref{eq:focal_invariance}): Applied to distances $d_{ij}$ between same-class pairs. Emphasizes large distances (hard positives that should be close but aren't).
    \item \textbf{Repulsion} (Eq.~\ref{eq:repulsion_focal}): Applied to margin violations $v_{ij}$ for different-class pairs. Emphasizes large violations (hard negatives that are too close).
\end{itemize}
Both mechanisms prioritize difficult cases while down-weighting trivial ones, but adapted to their respective objectives.

\textbf{Integration with Invariance:} The repulsion loss is added to the invariance 
loss with logarithmic scaling to prevent domination:
\begin{equation}
\mathcal{L}_{\text{inv}}^{\text{total}} = \mathcal{L}_{\text{inv}} + \log(1 + \mathcal{L}_{\text{repel}})
\end{equation}

The logarithmic scaling ensures repulsion provides a regularization signal without 
overwhelming the primary invariance objective.

\subsubsection{Contribution 4: Isotropic Representations via Z-Score Standardization}
\label{appendix:IsotropicRepresentations}

To encourage isotropic (spherical) representations, we apply Z-score standardization 
before computing all losses:
\begin{align}
\mathbf{P}_a &\leftarrow \frac{\mathbf{P}_a - \mathbb{E}[\mathbf{P}_a]}{\text{std}(\mathbf{P}_a)} \\
\mathbf{P}_b &\leftarrow \frac{\mathbf{P}_b - \mathbb{E}[\mathbf{P}_b]}{\text{std}(\mathbf{P}_b)}
\end{align}

where mean and standard deviation are computed per dimension across the batch. 

Unlike the original VICReg formulation, we omit the variance regularization term. The Z-score standardization applied before computing invariance and covariance losses inherently prevents dimensional collapse by normalizing each dimension to unit variance, making an explicit variance loss redundant.

The complete IsoFICReg objective combines:
\begin{align}
\mathcal{L}_{IsoFICReg} = \lambda_{inv} \mathcal{L}_{inv}^{focal} + \lambda_{cov} \mathcal{L}_{cov}^{focal}  \\
\end{align}

where $\lambda_{inv} = 25.0$ and $\lambda_{cov} = 1.0$.

\textbf{Rationale:} Z-score standardization ensures all dimensions have unit 
variance and zero mean, creating a natural isotropic prior. Combined with 
covariance-based decorrelation, this encourages representations to uniformly 
fill the unit hypersphere, maximizing information content and preventing 
dimensional collapse.

\textbf{Distributed Batch Statistics:} When training distributed, we gather 
representations across all GPUs before computing mean and standard deviation, 
ensuring batch statistics reflect the global batch rather than local mini-batches. 
This stabilizes training and improves representation quality.

\subsubsection{Contribution 5: Dual-Domain Consistency and Latent Denoising}
\label{appendix:DualDomainConsistency}

Following precedent established by Zhang et al.~\cite{zhang2022self}, we apply IsoFICReg to both time-domain and frequency-domain representations, 
with a critical design choice: we compute invariance losses between augmented 
embeddings (from input-space transformations) and latent-transformed embeddings 
(from latent-space transformations). This provides implicit latent denoising 
and equivariance enforcement.

Augmented embeddings ($\mathbf{P}_{\text{aug}}$) are derived from the augmented 
branch where transformations are applied in input space:
\begin{equation}
\mathbf{x}_{\text{aug}} \rightarrow \text{Encoder} \rightarrow \text{Projector} \rightarrow \mathbf{P}_{\text{aug}}
\end{equation}

Latent-transformed embeddings ($\mathbf{P}_{\text{latent}}$) are derived from 
the clean branch where transformations are applied in latent space via the 
Latent Equivariant Transformer (Section~\ref{appendix:LatentEquivariantTransformer}):
\begin{equation}
\mathbf{x}_{\text{clean}} \rightarrow \text{Encoder} \rightarrow \text{LatentEquivariantTransformer} \rightarrow \text{Projector} \rightarrow \mathbf{P}_{\text{latent}}
\end{equation}

The key distinction: augmented embeddings include AWGN (applied in input space), 
while latent-transformed embeddings do not (equivariant transformations applied 
in latent space after encoding clean signals).

\paragraph{In-Domain Losses}

For each domain, we compute IsoFICReg between augmented and latent-transformed 
embeddings:
\begin{align}
\mathcal{L}_{\text{IsoFICReg}}^{(t)} &= \mathcal{L}_{\text{inv}}(\mathbf{P}_{\text{aug}}^{(t)}, \mathbf{P}_{\text{latent}}^{(t)}) + \mathcal{L}_{\text{cov}}(\mathbf{P}_{\text{aug}}^{(t)}, \mathbf{P}_{\text{latent}}^{(t)}) \\
\mathcal{L}_{\text{IsoFICReg}}^{(f)} &= \mathcal{L}_{\text{inv}}(\mathbf{P}_{\text{aug}}^{(f)}, \mathbf{P}_{\text{latent}}^{(f)}) + \mathcal{L}_{\text{cov}}(\mathbf{P}_{\text{aug}}^{(f)}, \mathbf{P}_{\text{latent}}^{(f)})
\end{align}

Critically, we do \textbf{not} compute augmented-augmented or latent-latent 
invariance within domains (time-time, frequency-frequency), but do across domains. We want to still provide a cross domain gradient but do not want noise versus noise within domain to pollute the gradients that are generated via a trivial similarity. 

\paragraph{Cross-Domain Consistency}

We additionally compute four cross-domain IsoFICReg losses to enforce Parseval-like 
consistency between time and frequency representations:
\begin{align}
\mathcal{L}_{\text{cross}}^{(1)} &= \mathcal{L}_{\text{IsoFICReg}}(\mathbf{P}_{\text{aug}}^{(t)}, \mathbf{P}_{\text{aug}}^{(f)}) \\
\mathcal{L}_{\text{cross}}^{(2)} &= \mathcal{L}_{\text{IsoFICReg}}(\mathbf{P}_{\text{latent}}^{(t)}, \mathbf{P}_{\text{latent}}^{(f)}) \\
\mathcal{L}_{\text{cross}}^{(3)} &= \mathcal{L}_{\text{IsoFICReg}}(\mathbf{P}_{\text{aug}}^{(t)}, \mathbf{P}_{\text{latent}}^{(f)}) \\
\mathcal{L}_{\text{cross}}^{(4)} &= \mathcal{L}_{\text{IsoFICReg}}(\mathbf{P}_{\text{latent}}^{(t)}, \mathbf{P}_{\text{aug}}^{(f)})
\end{align}

\textbf{Clarification:} Cross-domain losses $\mathcal{L}_{\text{cross}}^{(1)}$ and $\mathcal{L}_{\text{cross}}^{(2)}$ pair augmented-augmented and latent-latent embeddings respectively across domains (time-frequency), while $\mathcal{L}_{\text{cross}}^{(3)}$ and $\mathcal{L}_{\text{cross}}^{(4)}$ pair augmented-latent embeddings across domains. This comprehensive cross-domain pairing enforces that time and frequency representations respect Parseval's theorem: energy conservation between domains should hold regardless of whether transformations are applied in input space (augmented) or latent space (latent-transformed).

The total IsoFICReg loss is:
\begin{equation}
\mathcal{L}_{\text{IsoFICReg}} = \mathcal{L}_{\text{IsoFICReg}}^{(t)} + \mathcal{L}_{\text{IsoFICReg}}^{(f)} + \sum_{i=1}^{4} \mathcal{L}_{\text{cross}}^{(i)}
\end{equation}

\paragraph{Latent Denoising via Clean Targets}

Augmented embeddings are derived from noisy input-space transformations 
($\mathbf{x} + \text{AWGN} + \text{equivariant transforms}$), while latent-transformed 
embeddings are derived from clean latent-space transformations ($\mathbf{x} + 
\text{equivariant transforms only}$). By enforcing invariance between augmented 
and latent-transformed embeddings, we provide clean targets for the noisy 
augmented embeddings. The model learns: "noisy input-space transformed representation 
should match clean latent-space transformed representation."

This is particularly powerful in negative-SNR regimes: even when noise dominates 
the input signal, the latent-transformed embeddings (from clean signals) provide 
a denoising target. The invariance loss encourages the encoder to extract signal 
structure from noisy inputs that matches the structure extracted from clean inputs.

\paragraph{Implicit Equivariance Enforcement}

Both augmented and latent-transformed embeddings undergo identical equivariant 
transformations (frequency shift, time shift, phase rotation, IQ flip)—the 
only difference is the presence of AWGN in augmented embeddings. By minimizing 
the distance between augmented and latent-transformed embeddings, we encourage 
the model to learn representations where equivariant transformations produce 
consistent effects regardless of noise level.

For example, if a frequency shift by $\Delta f$ is applied:
\begin{itemize}
    \item \textbf{Augmented:} $\mathbf{x} + \text{AWGN} \rightarrow \text{shift by } \Delta f \rightarrow \text{Encoder} \rightarrow \text{Projector} \rightarrow \mathbf{P}_{\text{aug}}$
    \item \textbf{Latent-transformed:} $\mathbf{x} \rightarrow \text{Encoder} \rightarrow \text{LatentEquivariantTransformer (shift by } \Delta f) \rightarrow \text{Projector} \rightarrow \mathbf{P}_{\text{latent}}$
\end{itemize}

The invariance loss $\mathcal{L}_{\text{inv}}(\mathbf{P}_{\text{aug}}, \mathbf{P}_{\text{latent}})$ 
encourages these to be similar, implicitly enforcing that the frequency shift 
produces consistent latent transformations whether applied in input space (with 
noise) or latent space (without noise).

\paragraph{Critical Interaction with LED}

Without LED's explicit equivariance regression (Section~\ref{appendix:LED}), 
the N-choose-2 pairwise invariance losses would drive the model toward invariance 
to \textit{all} transformations, including frequency/time shifts, phase rotations, etc. The augmented-latent 
pairing would collapse to: "all transformed versions of the same signal should 
be identical, regardless of transformation type."

LED prevents this collapse by explicitly enforcing that equivariant transformations 
produce predictable, learnable changes in the embeddings. The difference projections 
$\Delta_{\text{aug}} = \mathbf{z}_{\text{aug}} - \mathbf{z}_{\text{clean}}$ 
and $\Delta_{\text{latent}} = \mathbf{z}_{\text{latent}} - \mathbf{z}_{\text{clean}}$ 
must regress to the same transformation parameters (Section~\ref{appendix:EquivarianceLoss}), 
forcing the model to learn that our target equivariant symmetry focused transformations/augmentations produce specific, predictable 
changes rather than being only invariant.

The combination creates a balanced objective:
\begin{itemize}
    \item \textbf{IsoFICReg (augmented-latent invariance):} Encourages similarity 
    across noise levels and domains, providing denoising and cross-domain consistency
    \item \textbf{LED (difference regression):} Enforces predictable differences 
    under equivariant transformations, preventing collapse to full invariance
\end{itemize}

This dual mechanism—latent denoising via clean latent-transformed targets and 
implicit equivariance enforcement via augmented-latent pairing—enables learning 
in negative-SNR regimes while maintaining fine-grained discrimination. The 
augmented-latent pairing provides a curriculum: the model learns to extract 
signal structure from noisy inputs (augmented) that matches the structure from 
clean inputs (latent-transformed), while LED ensures the extracted structure 
respects equivariant transformation rules.

\subsection{Reconstruction Losses}
\label{appendix:ReconstructionLosses}

Reconstruction losses provide dense, instance-specific gradients throughout the 
encoder and decoder, complementing the global structure learned by IsoFICReg 
and LED. We apply reconstruction in both time and frequency domains with 
phase-aware variants to ensure high-fidelity signal recovery.

\subsubsection{Dual-Domain Reconstruction}
\label{appendix:DualDomainReconstruction}

The decoder reconstructs signals in the time domain, and we additionally compute 
reconstruction loss in the frequency domain by transforming both reconstruction 
and target via FFT.

\textbf{Time-Domain Reconstruction:}
\begin{equation}
\mathcal{L}_{\text{recon}}^{(t)} = \text{FocalHuber}(\mathbf{x}_{\text{recon}}, \mathbf{x}_{\text{target}})
\end{equation}

where $\mathbf{x}_{\text{recon}}$ is the decoder output and $\mathbf{x}_{\text{target}}$ 
is the clean target signal (before AWGN injection).

\textbf{Frequency-Domain Reconstruction:}
\begin{align}
\mathbf{X}_{\text{recon}} &= \text{FFT}(\mathbf{x}_{\text{recon}}) \\
\mathbf{X}_{\text{target}} &= \text{FFT}(\mathbf{x}_{\text{target}}) \\
\mathcal{L}_{\text{recon}}^{(f)} &= \text{FocalHuber}(\mathbf{X}_{\text{recon}}, \mathbf{X}_{\text{target}})
\end{align}

\textbf{Rationale:} Dual-domain reconstruction ensures the model learns to 
preserve both temporal dynamics (time domain) and spectral structure (frequency 
domain). Errors in one domain may not be apparent in the other—for example, 
phase errors are more visible in time domain, while spectral envelope distortions 
are more visible in frequency domain. By enforcing reconstruction fidelity in 
both domains, we ensure comprehensive signal recovery.

\subsubsection{Phase-Aware Reconstruction}
\label{appendix:PhaseReconstruction}

To encourage learning of phase relationships beyond magnitude reconstruction, 
we add explicit phase losses using circular-aware regression 
(Section~\ref{appendix:EquivarianceLoss}).

\textbf{Time-Domain Phase Loss:}
\begin{align}
\phi_{\text{recon}} &= \angle(\mathbf{x}_{\text{recon}}) \quad \text{(extract phase)} \\
\phi_{\text{target}} &= \angle(\mathbf{x}_{\text{target}}) \\
\mathcal{L}_{\text{phase}}^{(t)} &= \text{PhaseFocalHuber}(\phi_{\text{recon}}, \phi_{\text{target}})
\end{align}

\textbf{Frequency-Domain Phase Loss:}
\begin{align}
\Phi_{\text{recon}} &= \angle(\mathbf{X}_{\text{recon}}) \\
\Phi_{\text{target}} &= \angle(\mathbf{X}_{\text{target}}) \\
\mathcal{L}_{\text{phase}}^{(f)} &= \text{PhaseFocalHuber}(\Phi_{\text{recon}}, \Phi_{\text{target}})
\end{align}

The PhaseFocalHuber loss wraps phase errors to $[-\pi, \pi]$ before applying 
focal Huber regression:
\begin{align}
\Delta_{\phi} &= (\phi_{\text{target}} - \phi_{\text{recon}} + \pi) \mod 2\pi - \pi \\
\mathcal{L}_{\text{phase}} &= \text{FocalHuber}(\Delta_{\phi}, 0)
\end{align}

\textbf{Rationale:} Phase information is critical for signal reconstruction but 
often neglected by magnitude-only losses. Explicit phase losses ensure the model 
learns to preserve phase relationships, particularly important for coherent 
signals (communications, audio) where phase distortions cause audible artifacts 
or demodulation failures. The circular-aware formulation handles phase wrapping 
at $\pm\pi$.

\subsubsection{Total Reconstruction Loss}
\label{appendix:TotalReconstructionLoss}

The complete reconstruction loss combines all components:
\begin{equation}
\mathcal{L}_{\text{recon}} = \mathcal{L}_{\text{recon}}^{(t)} + \mathcal{L}_{\text{recon}}^{(f)} + \mathcal{L}_{\text{phase}}^{(t)} + \mathcal{L}_{\text{phase}}^{(f)} 
\end{equation}

where $\lambda_{\text{percept}}$ controls the perceptual loss weight.

\textbf{Application:} Reconstruction losses are applied to:
\begin{itemize}
    \item \textbf{Single-source reconstruction:} Clean signals after encoding 
    and decoding
    \item \textbf{Source-separated reconstruction:} Each separated source 
    reconstructed to its original unmixed target
    \item \textbf{Denoised reconstruction:} Noisy inputs reconstructed to clean 
    targets (implicit denoising)
\end{itemize}

All reconstruction losses use focal Huber regression 
(Section~\ref{appendix:EquivarianceLoss}) to emphasize hard examples 
(low-SNR, high-error regions) while maintaining robustness to outliers.

\subsection{Perceptual Loss with EMA Encoder}
\label{appendix:PerceptualLoss}

To encourage reconstructions that preserve high-level signal structure, we include a perceptual loss using an Exponential Moving 
Average (EMA) copy of the encoder.

\subsubsection{EMA Encoder}

The EMA encoder $f_{\text{EMA}}$ is updated as a moving average of the training 
encoder $f_{\theta}$:
\begin{equation}
\theta_{\text{EMA}} \leftarrow \alpha \cdot \theta_{\text{EMA}} + (1 - \alpha) \cdot \theta
\end{equation}

where $\alpha = 0.999$ (momentum). The EMA encoder is not trained via backpropagation 
but provides stable feature targets for perceptual loss.

\subsubsection{Perceptual Loss Computation}

For a reconstructed signal $\mathbf{x}_{\text{recon}}$ and its clean target 
$\mathbf{x}_{\text{target}}$, we extract features from both using the EMA encoder:
\begin{align}
\mathbf{z}_{\text{recon}}^{(t)}, \mathbf{z}_{\text{recon}}^{(f)} &= f_{\text{EMA}}(\mathbf{x}_{\text{recon}}) \\
\mathbf{z}_{\text{target}}^{(t)}, \mathbf{z}_{\text{target}}^{(f)} &= f_{\text{EMA}}(\mathbf{x}_{\text{target}})
\end{align}

The perceptual loss encourages feature-space similarity:

\begin{align}
\mathcal{L}_{\text{percept}} &= \text{FocalHuber}([\mathbf{z}_{\text{recon}}^{(t)}; \mathbf{z}_{\text{recon}}^{(f)}], [\mathbf{z}_{\text{target}}^{(t)}; \mathbf{z}_{\text{target}}^{(f)}]) \\
\end{align}

\textbf{Rationale:} Perceptual loss encourages reconstructions that preserve 
high-level signal structure (e.g., spectral envelope, temporal dynamics) rather 
than just minimizing point-wise error. The EMA encoder provides stable feature 
targets that evolve slowly as the encoder improves, preventing instability from 
rapidly changing feature spaces. Furthermore, this can be viewed as yet another form of consistency regularization and implicit latent denoising.

\textbf{Application:} Perceptual loss is applied to all reconstruction tasks: 
single-source denoising reconstructions and source-separated reconstructions.

\subsection{Complete Loss Formulation}
\label{appendix:CompleteLossFormulation}

We combine all training objectives into a unified loss function. Critically, 
we do not apply manual weighting between loss components—each loss operates at 
its natural scale. The only scaling applied is a factor of 100 to all reconstruction 
losses to balance their contribution relative to latent-space losses.

\subsubsection{Loss Components}
\label{appendix:LossComponents}

The complete training loss consists of five primary components:

\textbf{(1) Isotropic Focal Invariance Covariance Regularization (IsoFICReg):}
\begin{equation}
\mathcal{L}_{\text{IsoFICReg}} = \mathcal{L}_{\text{IsoFICReg}}^{(t,t)} + \mathcal{L}_{\text{IsoFICReg}}^{(f,f)} + \mathcal{L}_{\text{IsoFICReg}}^{(t,f)} + \mathcal{L}_{\text{IsoFICReg}}^{(f,t)}
\end{equation}

where each domain loss includes invariance ($\lambda_{\text{inv}} = 25$), covariance ($\lambda_{\text{cov}} = 1$), 
and repulsion (Section~\ref{appendix:IsoFICReg}).

\textbf{(2) Latent Equivariant Differences (LED):}
\begin{equation}
\mathcal{L}_{\text{LED}} = \mathcal{L}_{\text{latent}} + \mathcal{L}_{\text{input}} + \mathcal{L}_{\text{consistency}}
\end{equation}

where each term uses focal Huber regression on transformation parameter predictions 
(Section~\ref{appendix:LED}).

\textbf{(3) Reconstruction Losses:}
\begin{equation}
\mathcal{L}_{\text{recon}} = \mathcal{L}_{\text{recon}}^{(t)} + \mathcal{L}_{\text{recon}}^{(f)} + \mathcal{L}_{\text{phase}}^{(t)} + \mathcal{L}_{\text{phase}}^{(f)} 
\end{equation}

Applied per source with 100× scaling (Section~\ref{appendix:ReconstructionLosses}).

\textbf{(4) Source Separation Losses:}
\begin{equation}
\mathcal{L}_{\text{sep}} = \mathcal{L}_{\text{recon}}^{(A)} + \mathcal{L}_{\text{recon}}^{(B)} + \mathcal{L}_{\text{sink\_comp}} + \mathcal{L}_{\text{token}}
\end{equation}

where $\mathcal{L}_{\text{recon}}^{(A)}, \mathcal{L}_{\text{recon}}^{(B)}$ 
are per-source reconstruction losses (each scaled by 100×) and $\mathcal{L}_{\text{token}}$ 
is the token-level separation loss (Section~\ref{appendix:SourceSeparation}).

\textbf{(5) Perception Losses:}
\begin{equation}
\mathcal{L}_{\text{percept}} = \mathcal{L}_{\text{percept}}^{\text{input}} 
\end{equation}

\textbf{(6) Regularization Losses:}
\begin{align}
\mathcal{L}_{\text{reg}} = &\mathcal{L}_{\text{orth}} + \mathcal{L}_{\text{Parseval}} + \mathcal{L}_{\text{diversity}} + \mathcal{L}_{\text{noise\_decorr}} \\
&+ \mathcal{L}_{\text{power\_match}} + \mathcal{L}_{\text{SNR\_reg}} + \mathcal{L}_{\text{skip\_decorr}}
\end{align}

where:
\begin{itemize}
    \item $\mathcal{L}_{\text{orth}}$: Head orthogonalization across all multi-head 
    focus mechanisms (Section~\ref{appendix:head_orth})
    \item $\mathcal{L}_{\text{Parseval}}$: JSD-based cross-domain consistency 
    in Parseval Focus (Section~\ref{appendix:MHPF})
    \item $\mathcal{L}_{\text{diversity}}$: In-domain vs. cross-domain focus 
    diversity (Section~\ref{appendix:MHPF})
    \item $\mathcal{L}_{\text{noise\_decorr}}$: Pearson decorrelation between 
    noise sinks and cleaned signals (Section~\ref{appendix:ConvTokenization})
    \item $\mathcal{L}_{\text{power\_match}}$: Total sunk noise power matches 
    injected noise power (Section~\ref{appendix:ConvTokenization})
    \item $\mathcal{L}_{\text{SNR\_reg}}$: Reconstructed SNR matches target SNR 
    (Section~\ref{appendix:ConvTokenization})
    \item $\mathcal{L}_{\text{skip\_decorr}}$: Skip sink decorrelation and 
    target alignment (Section~\ref{appendix:SkipSinks})
\end{itemize}

\subsubsection{Total Loss: Single-Source Training}
\label{appendix:TotalLossSingleSource}

For single-source training (denoising), the total loss is:
\begin{equation}
\mathcal{L}_{\text{total}}^{\text{single}} = \mathcal{L}_{\text{IsoFICReg}} + \mathcal{L}_{\text{LED}} + 100 \cdot \mathcal{L}_{\text{recon}} + 
\mathcal{L}_{\text{percept}} + \mathcal{L}_{\text{reg}}
\end{equation}

The reconstruction loss is computed once per clean source and scaled by 100×.

\subsubsection{Total Loss: Source Separation Training}
\label{appendix:TotalLossSourceSeparation}

For source separation training, the total loss is:
\begin{equation}
\mathcal{L}_{\text{total}}^{\text{sep}} = \mathcal{L}_{\text{IsoFICReg}} + \mathcal{L}_{\text{LED}} + 100 \cdot (\mathcal{L}_{\text{recon}}^{(A)} + \mathcal{L}_{\text{recon}}^{(B)}) + \mathcal{L}_{\text{sink\_comp}} + \mathcal{L}_{\text{token}} + 
\mathcal{L}_{\text{percept}}^{(A)} + \mathcal{L}_{\text{percept}}^{(B)} + \mathcal{L}_{\text{reg}}
\end{equation}

where:
\begin{itemize}
    \item $\mathcal{L}_{\text{IsoFICReg}}$: Applied to separated source representations 
    against their clean single-source counterparts
    \item $\mathcal{L}_{\text{LED}}$: Applied to separated source representations 
    against their clean single-source counterparts
    \item $100 \cdot (\mathcal{L}_{\text{recon}}^{(A)} + \mathcal{L}_{\text{recon}}^{(B)})$: 
    Per-source reconstruction losses, each scaled by 100×
    \item $ (\mathcal{L}_{\text{percept}}^{(A)} + \mathcal{L}_{\text{percept}}^{(B)})$: 
    Per-source perception losses
    \item $\mathcal{L}_{\text{sink\_comp}}$: Skip sink complementarity loss 
    (Section~\ref{appendix:SkipSinks})
    \item $\mathcal{L}_{\text{token}}$: Token-level separation loss 
    (Section~\ref{appendix:SourceSeparation})
\end{itemize}

\subsubsection{Loss Scaling Rationale}
\label{appendix:LossScalingRationale}

\textbf{Why 100× for Reconstruction?} Latent-space losses (IsoFICReg) 
operate on high-dimensional projections ($d_{\text{proj}} = 1024$) with Z-score 
standardization, producing loss magnitudes on the order of $10^1$ to $10^3$. 
Reconstruction losses operate on raw signal space with focal Huber regression, 
producing loss magnitudes on the order of $10^{-3}$ to $10^{-1}$. The 100× scaling 
brings reconstruction losses to comparable magnitude, ensuring balanced gradient 
contributions without manual tuning.

\textbf{Why No Other Weighting?} Each loss component is designed with internal 
scaling mechanisms:
\begin{itemize}
    \item \textbf{IsoFICReg:} Built-in coefficients ($\lambda_{\text{inv}} = 25$, 
    $\lambda_{\text{var}} = 25$, $\lambda_{\text{cov}} = 1$) from VICReg~\cite{bardes2021vicreg}
    \item \textbf{LED:} Focal Huber regression with adaptive $\gamma$ based on 
    batch difficulty (Section~\ref{appendix:EquivarianceLoss})
    \item \textbf{Regularization:} Each term operates at natural scale (correlations, 
    power ratios, SNR values)
\end{itemize}

This design avoids the fragility of manual loss weighting, where small changes 
can destabilize training. By operating at natural scales with a single reconstruction 
scaling factor, the loss landscape remains stable across different batch compositions 
and training stages.

\subsubsection{Training Procedure}
\label{appendix:TrainingProcedure}

Training alternates between single-source and source separation batches:

\textbf{Single-Source Batches:} Clean signals with AWGN injection for denoising. 
Compute $\mathcal{L}_{\text{total}}^{\text{single}}$.

\textbf{Source Separation Batches:} Mixtures of two signals at controlled SINR 
(20-0 dB). Compute $\mathcal{L}_{\text{total}}^{\text{sep}}$.

Both batch types contribute to encoder learning (IsoFICReg, LED), but source 
separation batches additionally train the source separation transformer and 
skip sinks for mixture handling.

\subsubsection{Summary Table}
\label{appendix:LossSummaryTable}

Table~\ref{tab:complete_loss_summary} summarizes all loss components and their scaling.

\begin{table}[h]
\centering
\caption{Complete Loss Formulation Summary}
\label{tab:complete_loss_summary}
\begin{tabular}{lccl}
\toprule
\textbf{Loss Component} & \textbf{Scaling} & \textbf{Applied To} & \textbf{Section} \\
\midrule
\multicolumn{4}{l}{\textit{Latent-Space Losses}} \\
\midrule
IsoFICReg (time) & $\lambda_{\text{inv}}=25, \lambda_{\text{var}}=25, \lambda_{\text{cov}}=1$ & All batches & \ref{appendix:IsoFICReg} \\
IsoFICReg (freq) & $\lambda_{\text{inv}}=25, \lambda_{\text{var}}=25, \lambda_{\text{cov}}=1$ & All batches & \ref{appendix:IsoFICReg} \\
IsoFICReg (cross) & 1× & All batches & \ref{appendix:IsoFICReg} \\
Repulsion & $\log(1 + \cdot)$ & All batches & \ref{appendix:IsoFICReg} \\
LED & 1× & All batches & \ref{appendix:LED} \\
\midrule
\multicolumn{4}{l}{\textit{Reconstruction Losses}} \\
\midrule
Time-domain recon & 100× & Per source & \ref{appendix:ReconstructionLosses} \\
Freq-domain recon & 100× & Per source & \ref{appendix:ReconstructionLosses} \\
Phase (time) & 100× & Per source & \ref{appendix:ReconstructionLosses} \\
Phase (freq) & 100× & Per source & \ref{appendix:ReconstructionLosses} \\
\midrule
\multicolumn{4}{l}{\textit{Perception Losses}} \\
\midrule
Perceptual & 1× & Per source & \ref{appendix:ReconstructionLosses} \\
\midrule
\multicolumn{4}{l}{\textit{Source Separation Losses}} \\
\midrule
Skip sink complementarity & 1× & Sep batches & \ref{appendix:SkipSinks} \\
Token-level separation & 1× & Sep batches & \ref{appendix:SourceSeparation} \\
\midrule
\multicolumn{4}{l}{\textit{Regularization Losses}} \\
\midrule
Head orthogonalization & 1× & All batches & \ref{appendix:head_orth} \\
Parseval consistency & 1× & All batches & \ref{appendix:MHPF} \\
Focus diversity & 1× & All batches & \ref{appendix:MHPF} \\
Noise decorrelation & 1× & All batches & \ref{appendix:ConvTokenization} \\
Power matching & 1× & All batches & \ref{appendix:ConvTokenization} \\
SNR regression & 1× & All batches & \ref{appendix:ConvTokenization} \\
Skip decorrelation & 1× & All batches & \ref{appendix:SkipSinks} \\
\bottomrule
\end{tabular}
\end{table}

\section{Appendix D: Training Details}
\label{appendix:D_TrainingDetails}

\subsection{Model Parameter Counts}
\label{appendix:ParameterCounts}

Table~\ref{tab:parameter_counts} summarizes the parameter counts for all network components in the PlanFormer system.

\begin{table}[h]
\centering
\caption{PlanFormer System Parameter Counts}
\label{tab:parameter_counts}
\begin{tabular}{lrrr}
\toprule
\textbf{Component} & \textbf{Total Params} & \textbf{Trainable} & \textbf{Non-Trainable} \\
\midrule
\textbf{PlanFormer Encoder} & 1,990,478 & 1,978,190 & 12,288 \\
\midrule
\multicolumn{4}{l}{\textit{Training-Only Components:}} \\
\quad Projection Head/Expander & 2,234,368 & 2,234,368 & 0 \\
\quad Latent Equivariant Transformer & 480,771 & 480,771 & 0 \\
\quad Equivariant Prediction Head & 10,245 & 10,245 & 0 \\
\midrule
\multicolumn{4}{l}{\textit{Optional Inference Component:}} \\
\quad Latent Source Separation Transformer & 8,603,907 & 8,603,907 & 0 \\
\midrule
\textbf{PlanFormer Decoder} & 95,163,122 & 95,152,882 & 10,240 \\
\midrule
\textbf{Inference (Discriminative Tasks)} & \textbf{1,990,478} & \textbf{1,978,190} & \textbf{12,288} \\
\textbf{Inference (w/ Source Separation)} & \textbf{10,594,385} & \textbf{10,582,097} & \textbf{12,288} \\
\textbf{Total (Full Training System)} & \textbf{108,482,891} & \textbf{108,462,363} & \textbf{22,528} \\
\bottomrule
\end{tabular}
\end{table}

\textbf{Key Observations:}

\textbf{(1) Encoder Efficiency:} The encoder contains only 1.99M parameters, achieving the cross-modal transfer results reported in the main paper. This demonstrates that principled architectural design enables efficient cross-modal transfer with compact models, offering a complementary approach to large-scale architectures.

\textbf{(2) Decoder Parameter Distribution:} The decoder contains 95.2M parameters, with the majority allocated to hypernetwork-generated pointwise modulation parameters (Dynamic FiLM, Skip Sinks, Sliding Window Activation). Critically, these hypernetwork-generated parameters are not stored as weight matrices but computed dynamically during forward passes, resulting in lower computational burden than equivalent matrix multiplication operations. The decoder serves primarily as a training-time gradient substrate (Section~\ref{appendix:DecoderFoundation}) and can be discarded for inference on discriminative tasks.

\textbf{(3) Auxiliary Networks:} The equivariance learning components (Latent Equivariant Transformer, Equivariant Prediction Head) contain only 491K parameters combined, demonstrating that explicit symmetry learning does not require large capacity. The Latent Source Separation Transformer (8.6M parameters) operates on compressed token representations, enabling efficient token-level separation before reconstruction.

\textbf{(4) Deployment Flexibility:} For classification/retrieval tasks, only the encoder (1.99M parameters) is required at inference. For latent-space source separation, the encoder plus Latent Source Separation Transformer (10.6M parameters combined) are needed. The Projection Head, Equivariant Transformer, and Equivariant Prediction Head are training-only components supporting loss computation. For reconstruction tasks, the full encoder-decoder system (108.5M parameters) is used, with the understanding that decoder parameters primarily support training dynamics rather than representing stored knowledge.

\subsection{Hyperparameters}
\label{appendix:Hyperparameters}

We use a simple, fixed hyperparameter configuration throughout training, avoiding 
manual tuning by designing loss functions and architectures with internal adaptation 
mechanisms.

\textbf{Optimizer:} Adam~\cite{kingma2015adam} with learning rate $\eta = 10^{-4}$

\textbf{Batch Size:} 128 per GPU (1536 global batch size across 12 GPUs)

\textbf{Training Schedule:} No learning rate warmup or decay. The learning rate 
remains constant at $10^{-4}$ throughout training.

\textbf{Rationale:} By embedding adaptation mechanisms directly into the loss 
functions (focal reweighting, adaptive $\gamma$ in focal Huber, dynamic $K_{\text{focus}}$ 
in attention), we avoid the fragility of manual hyperparameter schedules. The 
model self-adapts to batch difficulty, eliminating the need for learning rate 
annealing or warmup.

\subsection{Cosine Annealing SNR Curriculum}
\label{appendix:SNRCurriculum}

While we avoid manual hyperparameter schedules, we employ a cosine annealing 
curriculum for AWGN augmentation to progressively challenge the model.

\textbf{SNR Range:} Target SNR $\in [\text{SNR}_{\text{lower}}(t), 100]$ dB, 
where $\text{SNR}_{\text{lower}}(t)$ varies according to a cosine schedule and 
the upper bound of 100 dB serves as a proxy for native (clean) SNR.

\textbf{Cosine Annealing Schedule:}
\begin{equation}
\text{SNR}_{\text{lower}}(t) = \text{SNR}_{\min} + \frac{1}{2}(\text{SNR}_{\max} - \text{SNR}_{\min})\left(1 + \cos\left(\frac{t}{T_{\text{period}}} \pi\right)\right)
\end{equation}

where:
\begin{itemize}
    \item $\text{SNR}_{\min} = -10$ dB (most challenging)
    \item $\text{SNR}_{\max} = 10$ dB (least challenging)
    \item $t$ is the current milestone (288 batch forward passes)
    \item $T_{\text{period}} = \frac{T_{\text{total}}}{80}$ (80 cycles over full training)
\end{itemize}

\textbf{Rationale:} The cosine schedule alternates between easier (high SNR floor) 
and harder (low SNR floor) training regimes. Early in each cycle, the model 
learns features present at moderate-to-high SNR. As the cycle progresses and 
SNR floor decreases, the model is challenged to maintain those features in 
increasingly noisy conditions. This curriculum provides:

\begin{enumerate}
    \item \textbf{Progressive difficulty:} Prevents overwhelming the model with 
    negative-SNR samples before it has learned basic signal structure
    \item \textbf{Regularization:} Cycling through difficulty levels acts as a 
    regularizer, preventing overfitting to any single SNR regime
    \item \textbf{Feature persistence:} Forces learning of features that persist 
    across SNR degradation rather than SNR-specific shortcuts
    \item \textbf{Whitening effect:} AWGN injection provides a whitening effect 
    on the input distribution, improving conditioning of the optimization landscape
\end{enumerate}

The upper bound of 100 dB ensures the model always sees clean samples, learning 
features naturally present at native SNR while simultaneously learning robustness 
to noise degradation.

\subsection{Data Resampling for Sample Rate Consistency}
\label{appendix:DataResampling}

To prevent the model from learning sample-rate-specific discriminative shortcuts, we resample 
all training data to a unified sample rate of 7.69 MHz (the native rate for 
POWDER 4G/5G data).

\textbf{Original Sample Rates:}
\begin{itemize}
    \item \textbf{ORACLE}~\cite{sankhe2019no}: 5 MHz $\rightarrow$ upsample to 7.69 MHz
    \item \textbf{POWDER}~\cite{reus2020trust}: 7.69 MHz (native, no resampling)
    \item \textbf{Modulation Recognition} (internal): 30.72 MHz (8 samples/symbol) $\rightarrow$ 
    downsample to 7.69 MHz
    \item \textbf{Commodity SDR} (internal): 10 MHz (10-15 kHz bandwidth, heavily 
    oversampled) $\rightarrow$ downsample to 7.69 MHz
    \item \textbf{S4\_FM Radio} (internal): 200 kHz $\rightarrow$ upsample to 7.69 MHz
\end{itemize}

\textbf{Rationale:} Different sample rates create different time-domain resolutions 
and frequency-domain spans. Without resampling, the model could learn to discriminate 
emitters based on sample rate rather than signal structure. Unified resampling 
ensures the model learns sample-rate-consistent representations, critical for transfer to domains with arbitrary sample rates.

\subsection{Stratified Batch Sampling}
\label{appendix:StratifiedSampling}

To support N-choose-2 pairwise losses in IsoFICReg 
(Section~\ref{appendix:LatentCoherentIntegration}), we employ 
stratified sampling to ensure balanced class representation in each batch.

\textbf{Stratified Distributed Sampler:}

Algorithm~\ref{alg:stratified_sampler} describes the stratified sampling procedure 
for distributed training.

\begin{algorithm}[H]
\caption{Stratified Distributed Sampler}
\label{alg:stratified_sampler}
\begin{algorithmic}[1]
\REQUIRE Concatenated dataset with $D$ sub-datasets
\REQUIRE Number of classes per dataset: $[C_1, C_2, \ldots, C_D]$
\REQUIRE Number of samples per class: $[N_1, N_2, \ldots, N_{\sum C_d}]$
\REQUIRE Number of GPUs: $G$, GPU rank: $r$
\ENSURE Stratified indices for GPU $r$
\STATE
\STATE \COMMENT{\textbf{Compute global parameters}}
\STATE $N_{\max} \leftarrow \max(N_1, \ldots, N_{\sum C_d})$ \COMMENT{Max samples per class}
\STATE $N_{\text{shard}} \leftarrow N_{\max} - (N_{\max} \mod G)$ \COMMENT{Samples per class per GPU}
\STATE $C_{\text{total}} \leftarrow \sum_{d=1}^{D} C_d$ \COMMENT{Total classes}
\STATE
\STATE \COMMENT{\textbf{Generate per-class shuffled indices}}
\FOR{each class $c \in [1, C_{\text{total}}]$}
    \STATE $N_c \leftarrow$ number of samples in class $c$
    \STATE $\text{indices}_c \leftarrow$ shuffle samples in class $c$ (with seed)
    \STATE \COMMENT{Repeat indices to reach $N_{\max}$ samples}
    \STATE $\text{indices}_c \leftarrow \text{repeat}(\text{indices}_c, \lceil N_{\max} / N_c \rceil)[:N_{\max}]$
    \STATE \COMMENT{Shard across GPUs}
    \STATE $\text{indices}_c \leftarrow \text{indices}_c[r : N_{\text{shard}} : G]$
\ENDFOR
\STATE
\STATE \COMMENT{\textbf{Stratified batch construction}}
\STATE $j \leftarrow 0$ \COMMENT{Sample index within each class}
\FOR{$i = 0$ to $(N_{\text{shard}} / G) \cdot C_{\text{total}} - 1$}
    \IF{$i > 0$ and $i \mod C_{\text{total}} = 0$}
        \STATE $j \leftarrow j + 1$ \COMMENT{Move to next sample in each class}
    \ENDIF
    \STATE $\text{class\_order} \leftarrow \text{random\_permutation}([1, \ldots, C_{\text{total}}])$
    \STATE $c \leftarrow \text{class\_order}[i \mod C_{\text{total}}]$
    \STATE \textbf{yield} $\text{indices}_c[j]$
\ENDFOR
\end{algorithmic}
\end{algorithm}

\textbf{Key Properties:}
\begin{enumerate}
    \item \textbf{Balanced representation:} Each batch contains exactly one sample 
    from each class (for batch size = number of classes) or balanced representation 
    (for larger batches)
    \item \textbf{Class shuffling:} Within each batch, class order is randomized 
    to prevent positional biases
    \item \textbf{Sample repetition:} Classes with fewer samples are repeated 
    (with shuffling) to match the largest class, ensuring no class is underrepresented
    \item \textbf{Distributed consistency:} Each GPU receives a disjoint shard 
    of samples while maintaining stratification
\end{enumerate}

\textbf{Rationale:} Stratified sampling maximizes the number of valid N-choose-2 
pairs per batch. With balanced class representation, each class has multiple 
samples in the batch, enabling pairwise invariance losses. Random sampling would 
produce batches where many classes have only one sample, wasting the N-choose-2 
mechanism. This is particularly important for datasets with many classes (39 
emitters in our training data), where random sampling would rarely produce 
multiple samples per class in a batch of 128.

\subsection{Training Scale and Milestones}
\label{appendix:TrainingScale}

Due to the high sample rate of RF data (7.69 MHz) and long sequence lengths 
(5120 samples), our training datasets contain an exceptionally large number of 
sequences.

\textbf{Milestone Definition:} We measure training progress in \textit{milestones}, 
where 1 milestone = 288 batch forward passes. This provides a more granular 
progress metric than epochs.

\textbf{Training Scale:}
\begin{itemize}
    \item \textbf{Estimated milestones per epoch:} 12,933
    \item \textbf{Estimated batches per epoch per GPU:} 3,724,780
    \item \textbf{Total GPUs:} 12 H100s (80GB each)
    \item \textbf{Global batches per epoch:} $3,724,780 \times 12 = 44,697,360$
\end{itemize}

\textbf{Results Reported:} All results in the main paper (Table~\ref{tab:frozenEncoder_results}) 
are from the same singular model trained for \textbf{400 milestones}, which represents 
$400 / 12,933 \approx 3.1\%$ of a single epoch. 

\textbf{Model Selection:} We select the checkpoint that achieves the highest validation 
accuracy on a linear probe trained online with frozen encoder representations. This 
online validation probe is trained on a held-out subset of the RF training data and 
evaluated every milestone, ensuring we select weights that produce maximally 
discriminative representations rather than simply training for a fixed duration.

\textbf{Rationale:} The exceptional scale of RF data (high sample rate, long 
sequences, multiple collections per dataset) creates a training regime where 
traditional epoch-based metrics are impractical. A single epoch would require 
$\sim$13,000 milestones or $\sim$3.7M batches per GPU. By reporting results at 400 milestones (3.1\% of one epoch), we demonstrate that the co-designed architecture and losses enable learning of physical principles from limited data exposure—a property that may be valuable for domains where exhaustive data coverage is impractical. The online validation 
probe ensures we select the checkpoint with the best representation quality 
rather than relying on training loss alone.

\subsection{Hardware and Training Time}
\label{appendix:HardwareTrainingTime}

\textbf{Hardware:} 12× NVIDIA H100 GPUs (80GB VRAM each) with NVLink interconnect

\textbf{Distributed Training:} PyTorch Distributed Data Parallel (DDP) with 
\texttt{all\_gather} for batch statistics in IsoFICReg

\textbf{Training Time:} Approximately 1,860 seconds per milestone (288 batch 
forward passes). For 400 milestones, total training time is approximately 
\textbf{8.6 days} (206.4 hours).

\textbf{Total Compute:} Approximately 2,477 H100-hours (12 GPUs × 206.4 hours) 
for the reported results, equivalent to ~103 H100-days.

\textbf{Time Breakdown:} The per-milestone time includes:
\begin{itemize}
    \item \textbf{Forward passes:} Encoder (dual-domain), decoder (dual-domain), 
    projection heads, auxiliary networks (LED transformers, source separation 
    transformers)
    \item \textbf{Loss computation:} IsoFICReg (N-choose-2 pairwise across 
    distributed GPUs), LED (differences and regression), reconstruction 
    (long sequences in memory), regularization losses
    \item \textbf{Backward passes:} Gradients through all networks
    \item \textbf{Monitoring:} t-SNE visualizations (every 6 milestones), 
    reconstruction visualizations, loss logging, online validation probe evaluation
\end{itemize}

The training cost is dominated by:
\begin{enumerate}
    \item \textbf{Multiple encoder/decoder passes:} Clean branch, augmented branch, 
    latent-transformed branch, source separation (when applicable)
    \item \textbf{Long sequence reconstruction:} 5120-sample sequences stored in 
    memory for dual-domain reconstruction losses
    \item \textbf{N-choose-2 pairwise losses:} Computing $\binom{N_c}{2}$ pairs 
    per class across global batch (1536 samples)
    \item \textbf{Distributed communication:} \texttt{all\_gather} operations 
    for batch statistics and representations
    \item \textbf{Online validation probe:} Linear classifier trained every 
    milestone for model selection
\end{enumerate}

\textbf{Memory Usage:} Approximately 168GB / 187GB per node (2 GPUs per node). 
Memory is primarily consumed by:
\begin{itemize}
    \item Model parameters (encoder, decoder, auxiliary networks)
    \item Optimizer states (Adam maintains first and second moments)
    \item Activations and gradients for long sequences (5120 samples)
    \item Intermediate representations for multiple branches (clean, augmented, 
    latent-transformed)
    \item Monitoring buffers (t-SNE embeddings, reconstruction samples, validation probe)
\end{itemize}

\textbf{GPU Scaling:} The 12-GPU configuration was selected to complete training 
within our project timeline. However, the framework scales linearly with GPU 
count—we have successfully trained comparable models using 4 and 8 H100 GPUs with identical 
hyperparameters and data. The 12-GPU specification ensures exact reproducibility 
of the reported checkpoint, but researchers with smaller computational budgets 
can achieve similar frozen-representation transfer performance with fewer GPUs and proportionally 
longer training times. This linear scaling demonstrates that our approach is 
accessible to institutions with varying resource constraints.

\textbf{Deployment Efficiency:} Critically, these training costs do not reflect 
deployment requirements. At inference, only the lightweight encoder (1.99M parameters) 
is needed for classification tasks, or encoder + decoder for reconstruction tasks 
(denoising, source separation). The auxiliary networks (projection heads, LED 
transformers) are discarded after training. Inference on a single sample requires 
only one forward pass through the encoder, with no distributed communication or 
pairwise loss computation.

\textbf{Optimization Opportunities:} The current implementation prioritizes 
research flexibility over computational efficiency. Substantial speedups are 
likely achievable through:
\begin{itemize}
    \item Gradient checkpointing for long sequences
    \item Optimized attention kernels (FlashAttention~\cite{dao2022flashattention})
    \item Reduced monitoring frequency (t-SNE, visualizations)
    \item Batch size tuning and gradient accumulation
\end{itemize}

We have not pursued these optimizations as training time was not a bottleneck 
for our research objectives. The 8.6-day training time for 400 milestones 
demonstrates that effective cross-modal transfer is achievable with reasonable 
computational budgets, particularly given that this represents only 3.1\% of a 
single epoch (Section~\ref{appendix:TrainingScale}).

\section{Appendix E: Dataset \& Evaluation Details}
\label{appendix:E_DatasetEvalDetails}

\subsection{Training Datasets}
\label{appendix:TrainingDatasets}

We train exclusively on RF fingerprinting datasets, which provide weak supervision 
critical for our N-choose-2 pairwise learning (Section~\ref{appendix:LatentCoherentIntegration}). 
RF fingerprinting datasets are collected such that each file contains transmissions 
from a single known emitter, providing weak labels for latent coherent integration 
without requiring sample-level annotations.

\textbf{Critical Challenge:} A key risk in RF fingerprinting is learning spurious 
correlations from time-varying channels rather than hardware characteristics. 
Each file experiences unique channel conditions (multipath, fading, interference), 
creating a potential shortcut where the model discriminates files by channel 
rather than emitter. This motivates our physically-anchored regularizations: 
reconstruction losses enforce instance-specific fidelity (preventing over-discrimination), 
LED learns channel-invariant equivariances (frequency/time shifts, phase rotations), 
and IsoFICReg with augmented-latent pairing provides implicit denoising. Together, 
these mechanisms facilitate the model learns hardware fingerprints rather than 
channel artifacts.

\subsubsection{ORACLE}
\label{appendix:ORACLE}

\textbf{Source:} \url{https://genesys-lab.org/oracle}~\cite{sankhe2019no}

\begin{itemize}
    \item \textbf{License:} CC BY 4.0 (Inferred/Standard for GENESYS Lab public datasets).
    \item \textbf{Access:} \url{https://genesys-lab.org/oracle}.
    \item \textbf{Version:} v1.0 (Initial Release, April 2019).
\end{itemize}

\textbf{Description:} 16 Ettus X310 software-defined radios transmitting identical 
data (same MAC address, protocol, modulation) at multiple distances and across 
two temporal collections.

\textbf{Training Subset:} We use distances of 14, 26, 38, and 50 feet from the 
first collection.

\textbf{Challenge:} ORACLE presents extreme fine-grained discrimination: all 
emitters share identical transmitted data, MAC addresses, protocols, and locations. 
The only distinguishing features are subtle hardware imperfections (I/Q imbalance, 
carrier frequency offset, phase noise). This forces learning of highly discriminative 
representations that capture manufacturing variations rather than protocol or 
content differences.

\textbf{Weak Label:} All samples from the same emitter across all distances and 
windows receive the same label, enabling N-choose-2 pairing across diverse channel 
conditions.

\textbf{Native Sample Rate:} 5 MHz $\rightarrow$ resampled to 7.69 MHz

\subsubsection{POWDER}
\label{appendix:POWDER}

\textbf{Source:} \url{https://genesys-lab.org/powder}~\cite{reus2020trust}

\begin{itemize}
    \item \textbf{License:} CC BY 4.0 (Inferred/Standard for GENESYS Lab public datasets).
    \item \textbf{Access:} \url{https://genesys-lab.org/powder}.
    \item \textbf{Version:} v1.0 (Initial Release, December 2020).
\end{itemize}

\textbf{Description:} 4 Ettus X310 base stations in an outdoor setting, each 
transmitting WiFi, 4G, and 5G protocols across 5 sequential collections over 
two days.

\textbf{Training Subset:} WiFi and 4G transmissions from all 4 base stations 
on Day 1 only. Day 2 and 5G transmissions are reserved for cross-protocol and 
temporal generalization testing (reported in Table~\ref{tab:frozenEncoder_results}).

\textbf{Protocol-Invariant Learning:} Critically, we assign the \textbf{same 
weak label} to both WiFi and 4G transmissions from each base station. This forces 
the model to learn emitter-specific hardware characteristics that persist across 
protocols, rather than protocol-specific features. The reconstruction losses 
and LED ensure protocol-specific structure is retained in the representations 
(enabling protocol classification if needed downstream), while IsoFICReg encourages 
protocol-invariant emitter fingerprints.

This design balances global (emitter identity) versus local (protocol details) 
and persistent (hardware fingerprints) versus transient (protocol structure) 
representations—enabling the model to support diverse downstream tasks without 
committing to a single level of abstraction.

\textbf{Native Sample Rate:} 
\begin{itemize}
    \item WiFi: 5 MHz $\rightarrow$ resampled to 7.69 MHz
    \item 4G/5G: 7.69 MHz (no resampling required)
\end{itemize}

\subsubsection{Modulation Recognition}
\label{appendix:ModulationRecognition}

\textbf{Source:} Internal dataset, to be released publicly with forthcoming publication

\textbf{Description:} 11 digital and analog modulations ('BPSK', 'QPSK', '8PSK', 
'CPFSK', 'GFSK', 'PAM4', '16QAM', '64QAM', 'SSB-AM', 'DSB-AM', 'B-FM') transmitted 
from a single Ettus X310 over a wired connection.

\textbf{Training Subset:} All 11 modulations from the training emitter.

\textbf{Modulation-Invariant Learning:} Similar to POWDER, we assign the 
\textbf{same weak label} to all modulations, forcing the model to learn 
modulation-invariant features (e.g., hardware characteristics) while reconstruction 
and LED preserve modulation-specific structure (magnitude/phase patterns). This 
tests whether the model can learn fine-grained discriminative features (different 
modulations have distinct magnitude and phase structures) while maintaining 
instance-level fidelity.

\textbf{Evaluation Subset:} 8 modulations ('BPSK', 'QPSK', '8PSK', 'CPFSK', 
'GFSK', 'PAM4', '16QAM', '64QAM') from a different radio and day, used for the modulation 
classification result in Table~\ref{tab:frozenEncoder_results}.

\textbf{Native Sample Rate:} 30.72 MHz (8 samples/symbol) $\rightarrow$ downsampled 
to 7.69 MHz

\subsubsection{Commodity SDR}
\label{appendix:CommoditySDR}

\textbf{Source:} Internal dataset, to be released upon acceptance

\textbf{Description:} 12 commodity software-defined radios from 4 families: 3× 
Ettus B210, 3× HackRF, 3× Ettus UBX160, 3× Analog Devices PlutoSDR. All transmit 
continuous phase shift keying (CPSK) with 10-15 kHz bandwidth (heavily oversampled 
at 10 MHz).

\textbf{Training Data:} Collections from two days with varying spatial configurations 
(5 ft and 10 ft transmitter-receiver separation, cluttered indoor environment). 
Using data across multiple days encourages learning of hardware fingerprints 
that persist despite temporal and spatial channel variations.

\textbf{Challenge:} The narrow bandwidth (10-15 kHz) relative to sample rate 
(10 MHz) creates a challenging scenario where most spectral content is empty. 
This tests whether the model learns to focus on signal-bearing regions rather 
than relying on full-bandwidth features.

\textbf{Native Sample Rate:} 10 MHz $\rightarrow$ downsampled to 7.69 MHz

\subsubsection{S4\_FM Radio}
\label{appendix:S4FM}

\textbf{Source:} Internal dataset, to be released upon acceptance

\textbf{Description:} 6 real-world FM radio stations collected over-the-air in 
outdoor settings.

\textbf{Training Subset:} All 6 stations.

\textbf{Challenge:} Real-world over-the-air collection introduces uncontrolled 
channel conditions, interference, and time-varying propagation. This tests 
robustness to realistic deployment scenarios.

\textbf{Native Sample Rate:} 200 kHz $\rightarrow$ upsampled to 7.69 MHz

\subsubsection{Training Dataset Summary}
\label{appendix:TrainingDatasetSummary}

\textbf{Train/Validation Split:} For all datasets, we use an 80/20 split per 
file for training and validation, ensuring both splits experience the same 
channel conditions while providing held-out samples for monitoring training progress.

Table~\ref{tab:training_datasets} summarizes the training datasets.

\begin{table}[h]
\centering
\caption{Training Dataset Summary}
\label{tab:training_datasets}
\begin{tabular}{lccccc}
\toprule
\textbf{Dataset} & \textbf{Emitters} & \textbf{Protocols/Mods} & \textbf{Environment} & \textbf{Native Rate} & \textbf{Resampled} \\
\midrule
ORACLE & 16 & WiFi & Indoor (4 dist.) & 5 MHz & 7.69 MHz \\
POWDER & 4 & WiFi, 4G & Outdoor & 7.69 MHz & - \\
Modulation Recognition & 1 & 11 modulations & Wired & 30.72 MHz & 7.69 MHz \\
Commodity SDR & 12 (4 families) & CPSK & Indoor (cluttered) & 10 MHz & 7.69 MHz \\
S4\_FM Radio & 6 & FM broadcast & Outdoor (OTA) & 200 kHz & 7.69 MHz \\
\midrule
\textbf{Total} & \textbf{39 classes} & \textbf{Diverse} & \textbf{Varied} & \textbf{-} & \textbf{7.69 MHz} \\
\bottomrule
\end{tabular}
\end{table}

\textbf{Diversity Rationale:} The training datasets span:
\begin{itemize}
    \item \textbf{Hardware:} Ettus X310, B210, UBX160, HackRF, PlutoSDR, commercial FM transmitters
    \item \textbf{Protocols:} WiFi, 4G, 11 digital/analog modulations, FM broadcast
    \item \textbf{Environments:} Indoor (controlled, cluttered), outdoor (OTA), wired
    \item \textbf{Sample rates:} 200 kHz to 30.72 MHz (154× range)
    \item \textbf{Bandwidths:} 10 kHz (Commodity SDR) to multi-MHz (POWDER)
\end{itemize}

This diversity forces learning of general signal-theoretic principles rather 
than dataset-specific shortcuts, critical for frozen-encoder representations to transfer to unseen 
domains (audio, images, seismology, text).

\subsection{Evaluation Datasets}
\label{appendix:EvaluationDatasets}

We evaluate cross-modal frozen-encoder representation transfer across 15 tasks spanning RF, audio, seismology, 
text, images, and video (Table~\ref{tab:frozenEncoder_results_appendix}). All evaluations 
use frozen encoder representations with no fine-tuning.

\subsubsection{Evaluation Protocol}
\label{appendix:EvaluationProtocol}

\textbf{Linear Classifier:} Support Vector Machine (SVM) with linear kernel 
(reported in main paper) and RBF kernel (reported in Table~\ref{tab:frozenEncoder_results_appendix}).

\textbf{Train/Test Split:} For datasets without explicit test sets, we perform 
5-fold cross-validation with stratified 80/20 splits per class. For datasets 
with explicit test sets, we perform 5-fold cross-validation on the training 
data while using the same test set for all folds.

\textbf{Variable-Length Processing:} For sequences longer than the training 
window (5120 samples), we apply the variable-length processing strategy 
(Section~\ref{appendix:VariableLengthProcessing}): segment into fixed-size 
windows, process through encoder, aggregate tokens (time: concatenate, frequency: 
average), and apply sequence pooling.

\textbf{Preprocessing:} Real-valued signals undergo Hilbert transformation to 
generate complex-valued (IQ) representations. Images are unwrapped in snake 
pattern (vertical then horizontal), with each color channel processed independently 
and representations concatenated. Video frames are unwrapped as images and 
concatenated to maintain causal structure. All image data is resampled to 5120 samples 
minimum (images, video) or processed at native length with variable-length 
extension (audio, text, seismology).

\subsubsection{Linear Probing}

\textbf{Classifier:} Linear SVM (LinearSVC from scikit-learn~\cite{scikit-learn}) with balanced class weights.

\textbf{Hyperparameter Search:} 5-fold cross-validation grid search over regularization parameter $C \in \{10^{-4}, 10^{-3}, \ldots, 10^{5}\}$. 

\textbf{Configuration:}
\begin{itemize}
    \item Regularization: $C$ selected via grid search
    \item Class weighting: Balanced (inversely proportional to class frequencies)
    \item Maximum iterations: $10^7$
    \item Solver: Default (dual optimization for $n_{\text{samples}} < n_{\text{features}}$, else primal)
\end{itemize}

\textbf{Top-k Accuracy:} For top-3 accuracy, we use the decision function scores (distance to separating hyperplane) as confidence estimates, ranking predictions accordingly.

\subsubsection{Non-Linear Probing (RBF Kernel)}

\textbf{Classifier:} SVM with Radial Basis Function (RBF) kernel.

\textbf{Hyperparameter Search:} 5-fold cross-validation grid search over:
\begin{itemize}
    \item Regularization: $C \in \{10^{-4}, 10^{-3}, \ldots, 10^{5}\}$
    \item Kernel coefficient: $\gamma = \text{scale} = \frac{1}{n_{\text{features}} \cdot \text{Var}(\mathbf{X})}$
\end{itemize}

\textbf{Configuration:}
\begin{itemize}
    \item Kernel: RBF, $K(\mathbf{x}_i, \mathbf{x}_j) = \exp(-\gamma \|\mathbf{x}_i - \mathbf{x}_j\|^2)$
    \item Class weighting: Balanced
    \item Maximum iterations: $10^7$
    \item Decision function: One-vs-rest (OVR)
\end{itemize}

\textbf{Top-k Accuracy:} For top-3 accuracy, we enable probability estimates (Platt scaling~\cite{platt1999probabilistic}) and use predicted probabilities for ranking.

\subsubsection{MFCC Baseline Features}

For audio tasks (speaker recognition, language recognition, music classification), we compare against expert-crafted Mel-Frequency Cepstral Coefficients (MFCCs) to validate that learned representations capture comparable or superior information to decades of domain-specific signal processing research.

\textbf{Feature Extraction:} We use the \texttt{python\_speech\_features} library~\cite{lyons2020python_speech_features} with default parameters:
\begin{itemize}
    \item Window length: 25ms
    \item Window step: 10ms
    \item Number of cepstral coefficients: 13
    \item Number of mel-filterbank channels: 26
    \item FFT size: 512
    \item Pre-emphasis coefficient: 0.97
\end{itemize}

\textbf{Aggregation:} For variable-length audio signals, we compute frame-level MFCCs and aggregate via mean and standard deviation across the temporal dimension:
\begin{equation}
\mathbf{f}_{\text{MFCC}} = [\mu(\text{MFCC}); \sigma(\text{MFCC})] \in \mathbb{R}^{26}
\end{equation}
where $\mu(\cdot)$ and $\sigma(\cdot)$ compute mean and standard deviation over frames, producing a fixed-size feature vector of dimension 26 (13 coefficients × 2 statistics).

\textbf{Classification:} MFCC features are classified using the same linear and non-linear SVM protocols described above, ensuring fair comparison.

\subsubsection{RF Tasks}
\label{appendix:RFTasks}

\textbf{Modulation Recognition:} 8 modulations ('BPSK', 'QPSK', '8PSK', 
'CPFSK', 'GFSK', 'PAM4', '16QAM', '64QAM') from a held-out radio (different 
from training) and channel (different day). Explicit test set. 5-fold CV on training data, fixed test set.

\textbf{RF Fingerprinting (POWDER)~\cite{reus2020trust}:} 4 base stations transmitting WiFi, 4G, \textbf{and} 5G (unseen 
protocol) on Day 2 (unseen temporal collection). Explicit test set. 5-fold CV 
on training data, fixed test set.

\subsubsection{Audio Tasks}
\label{appendix:AudioTasks}

\textbf{Bilingual Speaker Recognition (TidyVoiceX\_Dev)~\cite{farhadipour2026tidyvoice, farhadipour2026tidyvoice2026challengeevaluation}:} 808 speakers across 
40 languages. Due to scale, we randomly sample 50 speakers for evaluation. 
5-fold CV with stratified 80/20 splits. Each fold produces $\sim$29 languages 
on average. Audio files: $\sim$5 seconds at 16 kHz. Variable-length processing 
applied. Each file is one sample.
\begin{itemize}
    \item \textbf{License:} CC BY 4.0.
    \item \textbf{Access:} \url{https://mozilladatacollective.com/datasets/cmkv32i5e02tumg07j79d3c35}.
    \item \textbf{Version:} v2.0 (Official Evaluation Set, April 2, 2026).
\end{itemize}

\textbf{Language Recognition (TidyVoiceX\_Dev)~\cite{farhadipour2026tidyvoice, farhadipour2026tidyvoice2026challengeevaluation}:} Same data as speaker recognition. 
Each fold produces $\sim$29 languages from the 50 sampled speakers. Same processing 
as speaker recognition.
\begin{itemize}
    \item \textbf{License:} CC BY 4.0.
    \item \textbf{Access:} \url{https://mozilladatacollective.com/datasets/cmkv32i5e02tumg07j79d3c35}.
    \item \textbf{Version:} v2.0 (Official Evaluation Set, April 2, 2026).
\end{itemize}

\textbf{Instrument Classification (TinySOL)~\cite{cella_2020_3685331}:} 14 instruments across 4 families. 
All data used. 5-fold CV with stratified 80/20 splits. Audio files: 2-10 seconds 
at 44.1 kHz. Variable-length processing applied. Each file is one sample.
\begin{itemize}
    \item \textbf{License:} CC BY 4.0. 
    \item \textbf{Access:} \url{https://zenodo.org/record/3685331}. 
    \item \textbf{Version:} v1.0.
\end{itemize}

\textbf{Music Genre Recognition (GTZAN)~\cite{tzanetakis2002musical, gtzan_kaggle}:} 10 genres. All data used. 5-fold CV 
with stratified 80/20 splits. Audio files: 30 seconds at 22.05 kHz. Variable-length 
processing applied. Each file is one sample.
\begin{itemize}
    \item \textbf{License:} Available for research and educational purposes. The dataset is intended strictly for non-commercial academic use, as it was originally collected for research without explicit copyright permissions for all tracks.
    \item \textbf{Access:} \url{http://marsyas.info/downloads/datasets.html} 
    \item \textbf{Version:} Original release (2002). This foundational version includes 1,000 audio tracks (30 seconds each) across 10 distinct musical genres.
\end{itemize}

\subsubsection{Seismology Task}
\label{appendix:SeismologyTask}

\textbf{Seismic Event Classification (SCSN/SCEDC)~\cite{SCEDC2013}:} 3 classes (local quakes, 
noise, teleseismic events). Explicit test set with class imbalance.
\begin{itemize}
    \item \textbf{License:} Open access for research use per SCEDC data policy. 
    \item \textbf{Access:} \url{https://scedc.caltech.edu/}. 
    \item \textbf{Version:} Data accessed [03/2026].
\end{itemize}

\textbf{Data Structure:} Each sample consists of 3 channels (X, Y, Z seismometer 
measurements). We process each channel independently through the encoder and 
concatenate the resulting representations before classification.

\textbf{Sampling Strategy:} The test set has imbalanced classes. To create 
balanced evaluation, we:
\begin{enumerate}
    \item Identify the smallest test class 
    \item Randomly sample from larger test classes (local quakes, 
    noise) to match
    \item Use all available training data with 5-fold CV
\end{enumerate}

This ensures the classifier is evaluated on balanced test data while using all 
available training samples. 

\subsubsection{Text Tasks}
\label{appendix:TextTasks}

\textbf{ArXiv Paper Classification~\cite{He2019LongDC}:} 9 sub-disciplines (2 fields: math, computer 
science). Explicit test set: 28,000 training files, 2,500 test files. 5-fold 
CV on training data, fixed test set.

\begin{itemize}
    \item \textbf{License:} [Mixed] arXiv Non-exclusive License / CC BY (per original authors).
    \item \textbf{Access:} \url{https://huggingface.co/datasets/ccdv/arxiv-classification}.
    \item \textbf{Version:} v1.0 (March 2019/2021).

\end{itemize}
\textbf{Processing:} Papers converted to byte streams (UTF-8 encoding), processed 
as 1D sequences. Average length: $>$4,000 characters. Variable-length processing 
applied.

\textbf{Binary Field Classification:} Same data, binary classification (math 
vs. computer science).

\subsubsection{Image Tasks}
\label{appendix:ImageTasks}

All images resampled to 5120 samples after unwrapping (no variable-length processing 
needed).

\textbf{MNIST~\cite{lecun2010mnist}:} 10 digit classes. Explicit test set (60k train, 10k test). 
5-fold CV on training data, fixed test set. Images: 28×28 grayscale.
\begin{itemize}
    \item \textbf{License:} Public domain (CC0 1.0). 
    \item \textbf{Access:} \url{http://yann.lecun.com/exdb/mnist/}. 
    \item \textbf{Version:} Original release.
\end{itemize}

\textbf{FashionMNIST~\cite{xiao2017/online}:} 10 clothing classes. Explicit test set (60k train, 10k 
test). 5-fold CV on training data, fixed test set. Images: 28×28 grayscale.
\begin{itemize}
    \item \textbf{License:} MIT License.
    \item \textbf{Access:} \url{https://github.com/zalandoresearch/fashion-mnist}.
    \item \textbf{Version:} 1.0 (August 2017).
\end{itemize}

\textbf{PathMNIST~\cite{medmnistv2}:} 9 tissue pathology classes. Explicit test set. 5-fold CV 
on training data, fixed test set. Images: 64×64 RGB.
\begin{itemize}
    \item \textbf{License:} CC BY 4.0.
    \item \textbf{Access:} \url{https://zenodo.org/records/6496656}.
    \item \textbf{Version:} v2.0 (April 2022).
\end{itemize}

\textbf{CIFAKE~\cite{bird2023cifakeimageclassificationexplainable}:} 2 classes (real vs. AI-generated images). Explicit test set. 
5-fold CV on training data, fixed test set. Images: 32×32 RGB.
\begin{itemize}
    \item \textbf{License:} CC BY 4.0.
    \item \textbf{Access:} \url{https://www.kaggle.com/datasets/birdy654/cifake-real-and-ai-generated-synthetic-images}.
    \item \textbf{Version:} v1.0 (March 2023).
\end{itemize}

\subsubsection{Video Task}
\label{appendix:VideoTask}

\textbf{Mitosis Classification~\cite{delgado2024automatic, Delgado_Mitosis_Classification_2023}:} 2 classes (normal vs. abnormal cell division). 
Explicit test set: 317 normal + 146 abnormal (train), 47 normal + 47 abnormal 
(test). 5-fold CV on training data, fixed test set.
\begin{itemize}
    \item \textbf{License:} CC BY 4.0.
    \item \textbf{Access:} \url{https://doi.org/10.5281/zenodo.7788748}.
    \item \textbf{Version:} v1.0 (April 2023).
\end{itemize}

\textbf{Processing:} Each frame unwrapped as image, resampled to 5120 samples, successive frames concatenated to maintain causal structure.

\subsection{Evaluation Results with RBF Kernel}
\label{appendix:RBFResults}

Table~\ref{tab:frozenEncoder_results_appendix} presents results for both linear and RBF kernel SVMs. The RBF kernel provides non-linear decision boundaries, testing whether frozen representations benefit from non-linear classification. Results show modest improvements for most tasks (average: 77.7\% linear vs. 78.2\% RBF), with physical tasks maintaining strong performance (84.5\%/83.6\% both kernels) and semantic tasks showing slight benefit from non-linearity (70.0\% linear vs. 72.0\% RBF). This demonstrates that learned representations are largely linearly separable, validating representation quality.

\begin{table}[h]
\centering
\small
\caption{Frozen-encoder representation transfer results with linear and RBF kernel SVMs (5-fold CV). 
Physical tasks show strong performance with both kernels, while semantic tasks 
show graceful degradation in linear setting and slight benefit from nonlinearity.}
\label{tab:frozenEncoder_results_appendix}
\begin{tabular}{|l|c|c|c|c|}
\hline
\textbf{Task} & \textbf{Type} & \textbf{Top-1 (Linear)} & \textbf{Top-1 (RBF)} & \textbf{Top-3 (Linear)} \\
\hline
\multicolumn{5}{|c|}{\textit{RF}} \\
\hline
Modulation Rec. & Physical & 95.5$\pm$0.8 & 88.1$\pm$3.4 & 99.2$\pm$0.2 \\
Fingerprinting (POWDER) & Physical & 87.6$\pm$0.3 & 85.4$\pm$1.4 & N/A \\
\hline
\multicolumn{5}{|c|}{\textit{Audio}} \\
\hline
Speaker (TidyVoice) & Physical & 90.1$\pm$2.6 & 86.1$\pm$2.6 & 97.7$\pm$1.0 \\
Language (TidyVoice) & Semantic & 69.8$\pm$4.1 & 72.9$\pm$3.1 & 91.2$\pm$2.0 \\
Instrument Family (TinySOL) & Physical & 91.7$\pm$1.1 & 91.1$\pm$0.9 & N/A \\
Instrument (TinySOL) & Phys+Sem & 80.5$\pm$1.5 & 79.9$\pm$1.5 & 95.5$\pm$0.9 \\
Genre (GTZAN) & Semantic & 64.1$\pm$3.2 & 62.4$\pm$2.5 & 89.4$\pm$2.1 \\
\hline
\multicolumn{5}{|c|}{\textit{Seismology}} \\
\hline
Event Classification & Physical & 89.0$\pm$0.3 & 90.2$\pm$0.1 & N/A \\
\hline
\multicolumn{5}{|c|}{\textit{Text}} \\
\hline
ArXiv Sub-Discipline & Semantic & 36.9$\pm$0.4 & 37.9$\pm$0.3 & 71.8$\pm$0.3 \\
ArXiv Field & Struct+Sem & 82.7$\pm$0.2 & 84$\pm$0.4 & N/A \\
\hline
\multicolumn{5}{|c|}{\textit{Images}} \\
\hline
MNIST & Phys+Sem & 79.2$\pm$0.1 & 85.5$\pm$0.1 & 94.5$\pm$0.05 \\
FashionMNIST & Phys+Sem & 77.0$\pm$0.01 & 81.6$\pm$0.2 & 96.1$\pm$0.04 \\
PathMNIST & Physical & 71.3$\pm$0.2 & 73.2$\pm$0.4 & 91.9$\pm$0.2 \\
CIFAKE & Physical & 82.0$\pm$0.1 & 87.2$\pm$0.1 & N/A \\
\hline
\multicolumn{5}{|c|}{\textit{Video}} \\
\hline
Mitosis (Full Video) & Physical & 68.7$\pm$2.8 & 67.4$\pm$1.0 & N/A \\
\hline
\multicolumn{2}{|r|}{\textbf{Average (All)}} & 77.7\% & 78.2\% & 91.9\% \\
\multicolumn{2}{|r|}{\textbf{Average (Physical)}} & 84.5\% & 83.6\% & 96.3\% \\
\multicolumn{2}{|r|}{\textbf{Average (Semantic/Mixed)}} & 70.0\% & 72.0\% & 89.8\% \\
\hline
\end{tabular}
\end{table}

\begin{figure*}[p]
\centering
\includegraphics[width=\textwidth]{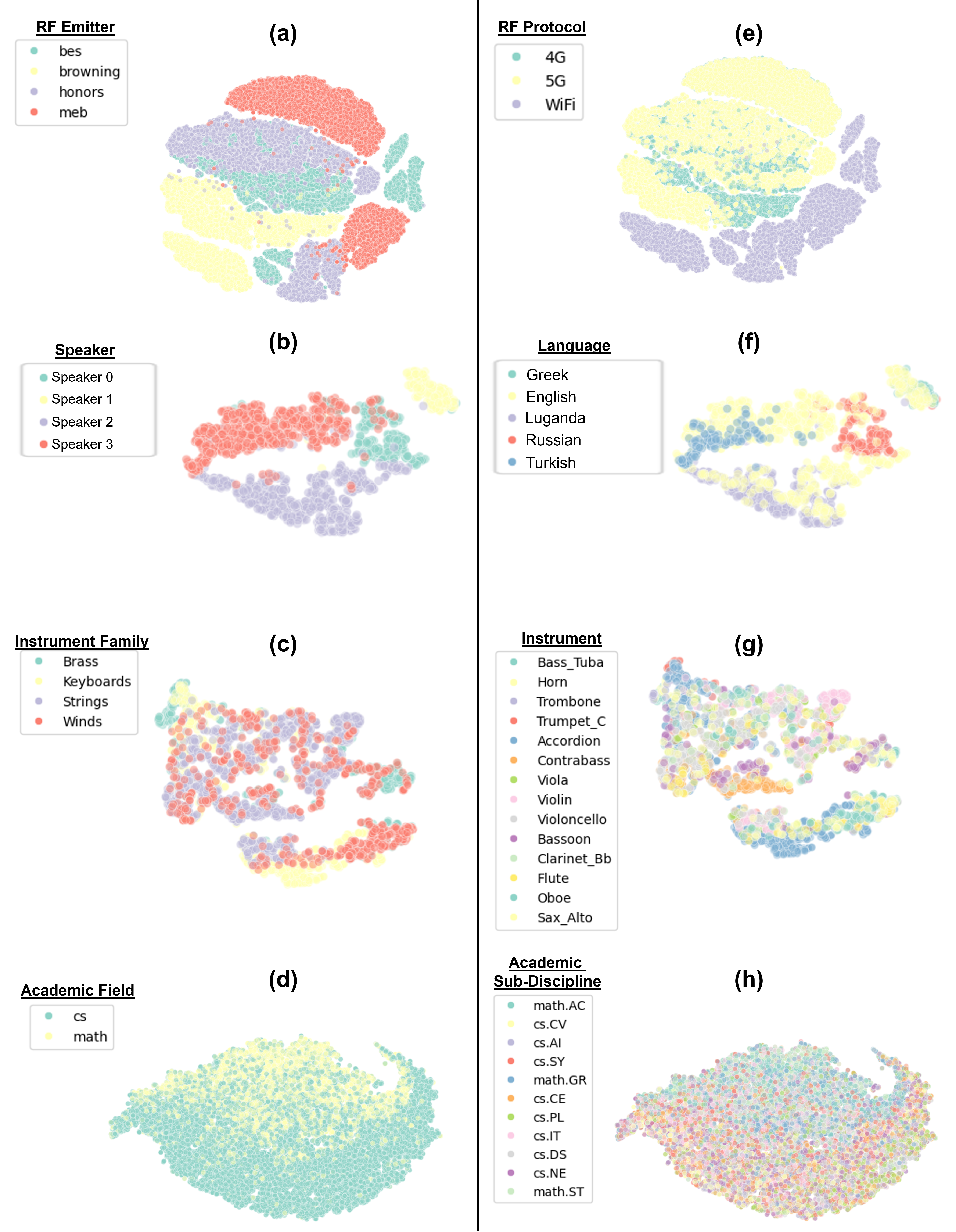}
\caption{\textbf{Learned representation structure: Physical vs. Semantic tasks.} t-SNE visualizations of frozen encoder representations across six representative tasks. \textbf{Left Column (Physical tasks):} (a) RF Fingerprinting shows well structured emitter representations (b) Bilingual speaker recognition demonstrates clear speaker-specific clusters consistent across languages. (c) Instrument family classification shows distinct family groupings based on physical sound production mechanisms. (d) The academic fields present within the ArXiv corpus show a clear structural distinction between fields. \textbf{Right Column (Semantic tasks):} (e) Protocol recognition shows the representations respect protocol uniqueness but retain emitter level structure. (f) Language recognition shows more overlap but maintains structure, with some clustering by linguistic families. (g) We see structural overlap across instruments that exhibit similar human level excitation patterns. (h) ArXiv sub-discipline classification shows the most diffuse structure, yet remains non-random, demonstrating that even raw text contains structural cues. This systematic progression from tight physical clusters to structured semantic overlap helps support our hypothesis that signal-theoretic learning captures physical structure but semantic content requires further capacity. See Section~\ref{appendix:EvaluationDatasets} for quantitative results.}
\label{fig:tsne_physical_semantic}
\end{figure*}

\subsection{Representation Structure Visualization}
\label{appendix:RepresentationVisualization}

We visualize frozen encoder representations using t-SNE and confusion matrices to qualitatively assess clustering quality and error patterns across the physical-semantic spectrum.

\subsubsection{Clustering Quality: Physical vs. Semantic Tasks}

Figure~\ref{fig:tsne_physical_semantic} shows t-SNE projections across eight tasks. Physical tasks (left column) exhibit tight, well-separated clusters: RF fingerprinting shows distinct emitter representations, speaker recognition demonstrates clear speaker-specific clusters consistent across languages, instrument families separate by sound production mechanisms, and ArXiv fields show structural distinction. Semantic tasks (right column) show more diffuse but structured representations: protocol recognition retains emitter-level structure, language recognition shows overlap with linguistic family clustering, individual instruments overlap based on similar excitation patterns, and ArXiv sub-disciplines remain non-random despite being the most diffuse. This progression from tight physical clusters to structured semantic overlap supports our hypothesis that signal-theoretic learning captures physical structure but semantic content needs further abstraction.

\begin{figure*}[p]
\centering
\includegraphics[width=\textwidth]{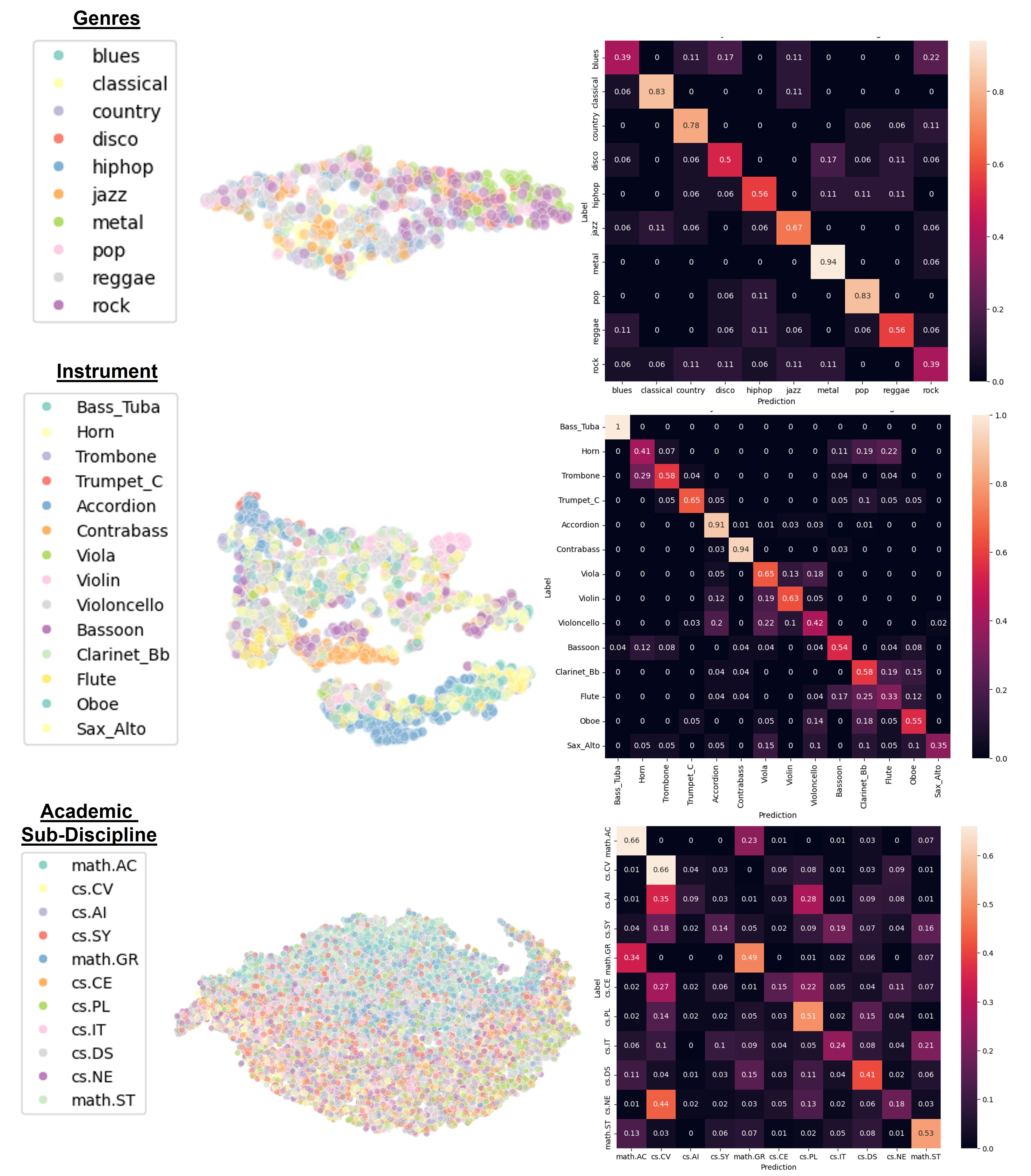}
\caption{\textbf{Interpretable confusion patterns across semantic spectrum.} Confusion matrices for three representative 1D tasks. \textbf{Top:} Music genre classification shows confusion between physically similar genres (rock/metal, classical/jazz) that share instrumentation but differ in cultural context. \textbf{Middle:} Individual instrument classification shows within-family confusion (brass instruments cluster, strings cluster), demonstrating learned physical sound production mechanisms. \textbf{Bottom:} ArXiv sub-discipline classification shows structured confusion within fields (math sub-disciplines confuse with each other, CS sub-disciplines confuse with each other) despite being the most semantic task. Errors reflect genuine structural similarity rather than random misclassification, validating that learned representations capture meaningful physical features.}
\label{fig:confusion_physical_semantic}
\end{figure*}

\subsubsection{Error Patterns: Interpretable Confusion}

Figure~\ref{fig:confusion_physical_semantic} shows confusion matrices for three representative 1D tasks spanning the semantic spectrum. \textbf{Music genre classification} (top) shows interpretable confusion between related genres: rock/metal share instrumentation and tempo, classical/jazz share harmonic complexity. Errors reflect shared physical structure despite semantic differences. \textbf{Individual instrument classification} (middle) shows within-family confusion: brass instruments (French Horn, Trombone, Trumpet) cluster together, as do strings (Cello, Double Bass). This demonstrates the model learned physical sound production mechanisms, with confusion arising from similar timbral characteristics. \textbf{ArXiv sub-discipline classification} (bottom) shows the most diffuse confusion pattern, yet errors remain structured: math sub-disciplines (Differential Geometry, Algebraic Geometry) confuse with each other, as do CS sub-disciplines (Machine Learning, Computational Complexity). Even in this extreme semantic test, confusion reflects genuine structural similarity in notation and equation density rather than random misclassification.

\subsubsection{Interpretation}

These visualizations provide qualitative evidence that: (1) physical tasks yield tight clusters with discriminative features, (2) semantic tasks show structured overlap reflecting shared physical foundations, (3) errors are interpretable—confusion arises from genuine similarity rather than arbitrary misclassification, and (4) representations remain meaningful across all tasks, even the most semantic. Together with quantitative results (Table~\ref{tab:frozenEncoder_results}), this demonstrates that the physical-semantic performance gap reflects fundamental differences in task nature rather than representation quality.

\subsection{Qualitative Reconstruction Analysis}
\label{appendix:ReconstructionQuality}


Following the quantitative evaluation via linear and non-linear probing, we assess representation 
quality through a complementary qualitative lens: reconstruction fidelity. Our decoder combines 
bottleneck latent guidance with two skip connections from compressed encoder features (4x and 16x 
downsampled, Section~\ref{appendix:DecoderRationale}): the bottleneck latent provides global 
structural information learned through discriminative and equivariant objectives, while skip 
connections preserve high-resolution details from already-compressed representations. Critically, 
we omit top-level skip connections (input and first conv layer) to prevent trivial information 
leakage, forcing the bottleneck to learn meaningful compressed representations. 

To validate that learned representations (not architectural biases) drive reconstruction quality, 
we evaluate the same architecture with random weight initialization (Figure~\ref{fig:random_reconstruction}). 
Random weights produce near-complete reconstruction failure—uniform gray fields with no structure—demonstrating 
that skip connections alone cannot reconstruct without learned bottleneck representations. This 
design enables zero-shot reconstruction while ensuring the encoder learns transferable representations 
validated by linear probing (Table~\ref{tab:frozenEncoder_results}), where only the bottleneck latent 
is used—skip connections are not involved in transfer learning evaluation.

\paragraph{Reconstruction as Representation Quality Metric}

Recent work has demonstrated that reconstruction quality from SSL bottleneck latents serves as a reliable proxy for learned feature representation quality~\cite{ericsson2021well}. The intuition is straightforward: if learned features are meaningful, they should enable reconstruction that preserves the input's structural properties. Random or poorly learned features would produce incoherent reconstructions.

\paragraph{Zero-Shot Image Reconstruction}

\begin{figure}
    \centering
    \includegraphics[width=\textwidth]{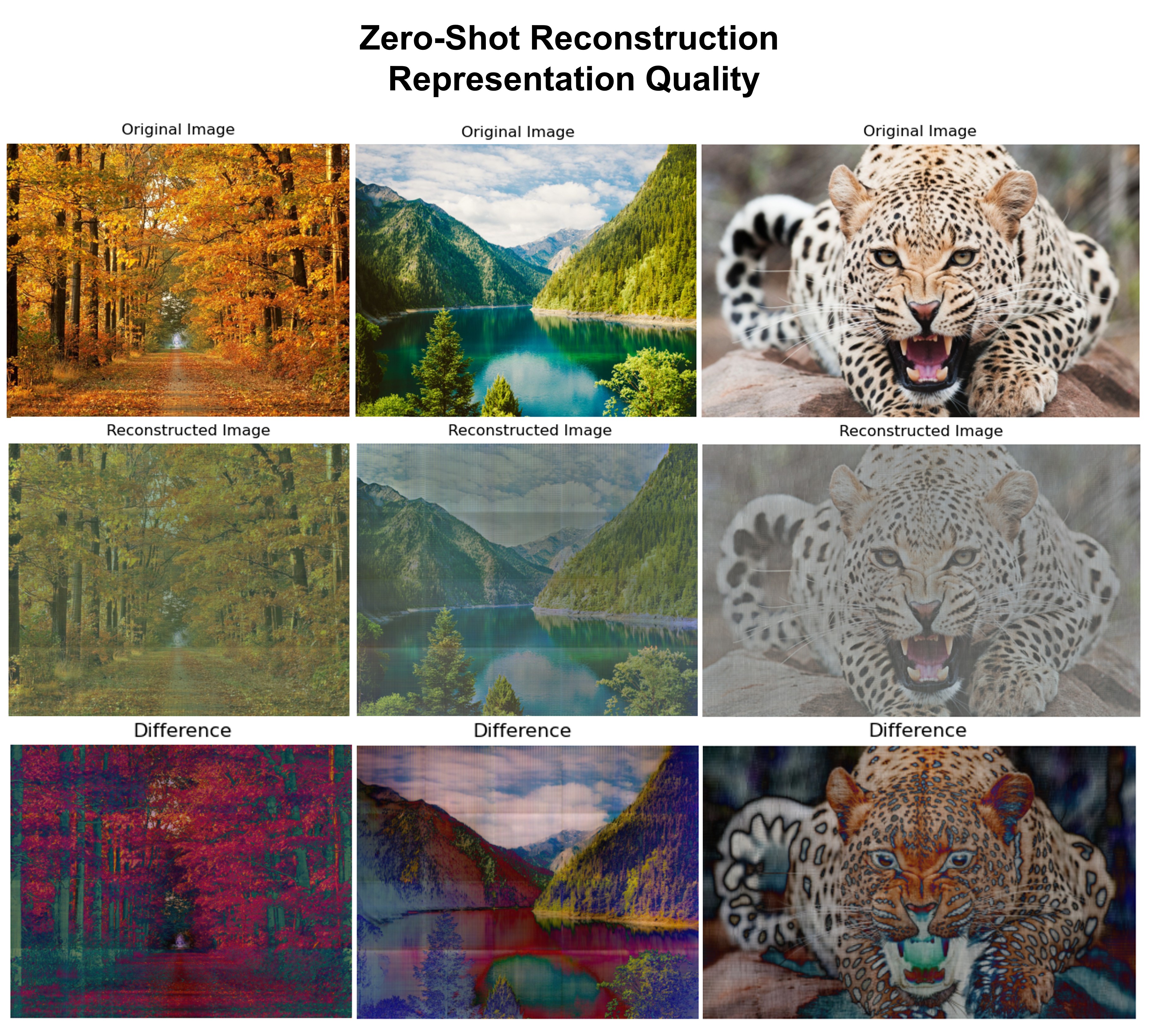}

\caption{Zero-shot image reconstructions from encoder-decoder system trained exclusively on RF data. 
\textbf{Top row:} Original images. \textbf{Middle row:} Reconstructions combining bottleneck latent 
guidance (global structure from transformer-processed tokens) with two skip connections from compressed 
encoder features (4x and 16x downsampled, providing local details). \textbf{Bottom row:} Absolute 
difference. Despite 1D unwrapping and no exposure to natural images during training, reconstructions 
preserve structural coherence and spatial relationships. Compare to Figure~\ref{fig:random_reconstruction} 
showing complete failure with random weights, validating that learned representations (not architectural 
biases) drive reconstruction quality. Linear probing (Table~\ref{tab:frozenEncoder_results}) validates 
that the bottleneck latent alone (without skip connections) captures sufficient structure for 
classification tasks.}
\label{fig:reconstruction}
\label{fig:image_reconstructions}
\end{figure}

Figure~\ref{fig:image_reconstructions} shows reconstructions of natural images from three categories: wildlife (cheetah), natural scenes (forest), and landscapes. Critically, these are **zero-shot reconstructions**—the model was trained exclusively on RF data and has never encountered natural images during training. Moreover, our 1D unwrapping strategy (snake pattern) discards 2D spatial correlations, placing the model at a significant disadvantage for image reconstruction.

\paragraph{Observations}

Despite the architectural disadvantage and complete domain mismatch, reconstructions exhibit several notable properties:

\textbf{Structural Preservation:} Reconstructions maintain overall composition, layout, and spatial relationships. In the cheetah image, the animal's body shape, posture, and separation from the background are preserved. In the forest scene, vertical tree structures, depth layers, and color gradients are maintained. The nature scene preserves sky-ground boundaries, object placements, and color zones.

\textbf{Edge and Boundary Detection:} Foreground-background separation and object boundaries are clearly delineated, indicating the encoder learned edge detection as a fundamental signal processing operation transferable across domains.

\textbf{Semantic Coherence:} Reconstructions are not random or noisy—they are clearly structured representations of the input images. This demonstrates that the encoder learned meaningful physical features (edges, textures, other frequency oriented details) that transfer zero-shot to visual data.

\textbf{Expected Limitations:} Full color details are lost (expected with 1D processing of color channels independently of one another). Spatial correlations have minor artifacts due to snake unwrapping, but the preservation of global structure demonstrates that the encoder captures sufficient information for coherent reconstruction.

\paragraph{Interpretation}

These reconstructions provide qualitative evidence that our encoder learns transferable physical structure rather than domain-specific features. The ability to reconstruct recognizable images from RF-trained representations demonstrates that:

\textbf{(1) Signal-theoretic principles generalize:} Features learned from RF signals (frequency content, temporal dynamics, edge detection) apply to visual signals (color spectra, spatial frequencies, boundaries).

\textbf{(2) Representations are information-rich:} The bottleneck latent retains sufficient information about physical structure to enable coherent reconstruction, not just classification.

\textbf{(3) Architectural mismatch is surmountable:} Even with 1D unwrapping destroying 2D correlations, the learned features capture enough structural information for meaningful reconstruction. Native 2D processing would likely improve fidelity substantially.

This qualitative analysis complements our quantitative results (Table~\ref{tab:frozenEncoder_results}), demonstrating that strong classification accuracy is not merely due to linear separability of arbitrary features, but reflects genuine learning of physical signal structure that transfers across modalities. These reconstructions also highlight a key advantage of principle-driven approaches: the learned features are interpretable as physical signal properties (edges, frequencies, textures) rather than opaque statistical correlations.

\subsection{Comparison to CLIP Foundation Model}
\label{appendix:clip_comparison}

We compare PlanFormer to CLIP ViT-B/32~\cite{radford2021learning}, a widely-used vision-language foundation model trained on 400 million image-text pairs from the internet. CLIP consists of dual encoders (image and text) trained via contrastive learning to align visual and linguistic representations.

\subsubsection{Model Configuration and Evaluation Protocol}

\textbf{Model:} We use the pretrained CLIP ViT-B/32 image encoder (151.28M parameters, 14,780 MFLOPs) from the official OpenAI release. The encoder processes 224$\times$224 RGB images and produces 512-dimensional embeddings.

\textbf{Evaluation Protocol:} Identical to PlanFormer (Section~\ref{Section:Experiments}):
\begin{itemize}
    \item Freeze CLIP image encoder (no fine-tuning)
    \item Extract features from final layer before classification head
    \item Train linear SVM classifier via 5-fold cross-validation
    \item Report mean top-1 accuracy $\pm$ standard deviation
\end{itemize}

\textbf{Computational Complexity and Model Size}: to compute the number of parameters of PlanFormer and the amount of FLOPs required for a forward pass, we used Meta's \textbf{fvcore}\footnote{\href{https://github.com/facebookresearch/fvcore}{https://github.com/facebookresearch/fvcore}}. For CLIP we referred to the CLIP repository~\cite{ilharco_gabriel_2021_5143773}.

\subsubsection{Domain-Specific Preprocessing}

To enable CLIP evaluation on 1-D signals, we convert time-series data to 2-D spectrograms. All spectrograms are normalized to [0, 1], converted to RGB via the Viridis colormap, and resized to 224$\times$224 using bicubic interpolation before applying CLIP's standard preprocessing.

\paragraph{Audio Signals (Speaker, Language, Instrument, Genre)}

Audio signals are processed in 5120-sample windows with the following pipeline:
\begin{enumerate}
    \item \textbf{Mel Spectrogram:} Compute mel-scaled power spectrogram using 2048-point FFT, 512-sample hop length, and 128 mel bins spanning 0-8 kHz
    \item \textbf{Logarithmic Scaling:} Convert to decibels via amplitude-to-dB transformation to compress dynamic range
    \item \textbf{Normalization:} Min-max normalize to [0, 1] per spectrogram
    \item \textbf{Colorization:} Apply Viridis colormap to convert grayscale to RGB
    \item \textbf{Resizing:} Resize to 224$\times$224 via bicubic interpolation
    \item Standard CLIP Preprocessing Function
\end{enumerate}

For audio files longer than 5120 samples, we extract non-overlapping windows, encode each spectrogram independently, and compute the mean-pooled embedding across all windows to produce the final sample representation.

\paragraph{RF and Seismology Signals}
Seismology signals (resampled to exactly 5120 samples consistent with Planformer evaluation) and RF (Modulation Recognition, POWDER Fingerprinting) are processed as follows:
\begin{enumerate}
    \item \textbf{Power Spectrogram:} Compute power spectrogram using 1024-point FFT with 22-sample hop length, producing approximately 513$\times$233 frequency-time representation
    \item \textbf{Logarithmic Scaling:} Apply $\log_{10}(\text{spec} + 10^{-6})$ to compress dynamic range
    \item \textbf{Normalization:} Min-max normalize to [0, 1]
    \item \textbf{Colorization:} Apply Viridis colormap to convert to RGB
    \item \textbf{Resizing:} Resize to 224$\times$224 via bicubic interpolation
    \item Standard CLIP Preprocessing Function
\end{enumerate}

No mean-pooling is required as each sample is exactly 5120 samples.

\paragraph{Images (MNIST, FashionMNIST, PathMNIST, CIFAKE)}

Images are processed using CLIP's standard preprocessing pipeline:
\begin{enumerate}
    \item \textbf{Resizing:} Resize to 224$\times$224 via bicubic interpolation
    \item \textbf{Channel Conversion:} Convert grayscale images to RGB by replicating the single channel three times
    \item \textbf{Normalization:} Apply CLIP's standard normalization (mean=[0.48145466, 0.45782750, 0.40821073], std=[0.26862954, 0.26130258, 0.27577711])
\end{enumerate}

Notably, we do \textit{not} apply PlanFormer's domain-specific preprocessing (Hilbert transform, unwrapping), allowing CLIP to process images in its native format.

\paragraph{Video (Mitosis Classification)}

Videos are processed frame-by-frame:
\begin{enumerate}
    \item \textbf{Frame Extraction:} Extract all frames from video
    \item \textbf{Frame Encoding:} Process each frame as a standard image (resize to 224$\times$224, convert to RGB, apply CLIP normalization)
    \item \textbf{Temporal Aggregation:} Encode each frame independently with frozen CLIP encoder and compute mean-pooled embedding across all frames
    \item \textbf{Classification:} Train linear SVM on mean-pooled video embeddings
\end{enumerate}

\paragraph{Text (ArXiv Classification)}

We do \textit{not} evaluate CLIP on text classification tasks. CLIP's text encoder uses a separate Transformer architecture optimized for linguistic processing, making direct comparison to PlanFormer's frozen image encoder inappropriate. Our inclusion of text tasks for PlanFormer serves to demonstrate a stress test for signal-theoretic principles on highly semantic data, not to claim superiority over language-specialized models.

\subsubsection{Results Analysis}

Table~\ref{tab:frozenEncoder_results} shows CLIP achieves higher overall accuracy (83.8\% vs 77.7\%), but performance gaps vary systematically by task type, validating our physical-semantic boundary hypothesis.

\paragraph{Physical Tasks: Competitive Performance with Massive Efficiency Gains}

On physical/structural tasks (Modulation Recognition, POWDER, Speaker ID, Instrument Family, Seismology, PathMNIST, CIFAKE, Mitosis), PlanFormer achieves 84.5\% average accuracy compared to CLIP's 87.7\%—a gap of only 3.2\%. This competitive performance is achieved with:
\begin{itemize}
    \item \textbf{76$\times$ fewer parameters} (1.99M vs 151M)
    \item \textbf{158$\times$ lower computational cost} (93.6 vs 14,780 MFLOPs)
    \item \textbf{Single-modality pretraining} (RF signals only vs 400M image-text pairs)
\end{itemize}

Notably, PlanFormer \textit{outperforms} CLIP on RF fingerprinting (87.6\% vs 55.0\%), demonstrating that physics-driven pretraining provides substantial advantages when the task domain aligns with the pretraining modality. CLIP's lower performance on RF tasks highlights the limitations of vision-language pretraining for signal data that lacks natural visual or linguistic structure.

\paragraph{Semantic Tasks: Language Grounding Advantage}

On semantic tasks (Language Recognition, Music Genre, Individual Instrument) and mixed physical-semantic tasks (MNIST, FashionMNIST), CLIP achieves 91.2\% compared to PlanFormer's 70.0\%—a gap of 21.2\%. This substantial difference supports our hypothesis that semantic understanding requires language grounding beyond physical principles.

CLIP's training on 400 million image-text pairs provides explicit supervision for semantic concepts (e.g., "jazz music," "English language," "digit 7"), while PlanFormer learns only the physical structure of signals. The graceful degradation on semantic tasks demonstrates that PlanFormer captures fundamental signal properties but lacks the linguistic grounding necessary for human-defined semantic categories.

\paragraph{Efficiency-Performance Trade-off}

These results position PlanFormer and CLIP on the accuracy-efficiency Pareto frontier. PlanFormer achieves 92\% of CLIP's performance on physical tasks while requiring only 0.66\% of the parameters and 0.63\% of the computational cost. This demonstrates that physics-driven design enables a favorable trade-off for applications where:
\begin{itemize}
    \item Physical/structural pattern recognition is the primary objective
    \item Computational resources are constrained (edge devices, embedded systems)
    \item Training data is limited or expensive to acquire
    \item Deployment requires low latency and energy efficiency
\end{itemize}

\paragraph{Implications for Foundation Model Design}

The complementary strengths of PlanFormer and CLIP suggest a hierarchical architecture for future foundation models:
\begin{enumerate}
    \item \textbf{Physical Layer (PlanFormer-like):} Lightweight, physics-driven encoder captures causal structure and domain-invariant signal properties with minimal resources
    \item \textbf{Semantic Layer (CLIP-like):} Language-grounded model maps physical representations to human-defined semantic concepts
\end{enumerate}

This division of labor enables resource-efficient deployment where the physical layer runs on-device (edge inference) and the semantic layer is queried only when high-level reasoning is required. Our results demonstrate that physics-driven pretraining is not a replacement for scale-driven approaches, but rather a complementary paradigm that democratizes foundation model development for resource-constrained applications.

\subsection{Comparison to DinoV3}
\label{appendix:dino_comparison}

We additionally compare to DinoV3 ViT-S~\cite{Simeoni2025DINOv3}, a self-supervised vision foundation model trained on ImageNet (21M parameters, 12,000 MFLOPs). Preprocessing is identical to CLIP (Section~\ref{appendix:clip_comparison}) except using DinoV3's standard ImageNet normalization.

\textbf{Results:} Table~\ref{tab:frozenEncoder_results} shows DinoV3 achieves 83.7\% overall accuracy, nearly identical to CLIP (83.8\%). Performance stratifies by task type:
\begin{itemize}
    \item \textbf{Physical Tasks:} PlanFormer is competitive with DinoV3 (84.5\% vs 83.6\%, +0.9\%) with 11$\times$ fewer parameters and 128$\times$ lower FLOPs
    \item \textbf{Semantic Tasks:} DinoV3 outperforms PlanFormer (84.0\% vs 70.0\%, +14.0\%), though less than CLIP's advantage (+21.2\%)
    \item \textbf{RF Fingerprinting:} PlanFormer substantially outperforms DinoV3 (87.6\% vs 66.3\%, +21.3\%)
\end{itemize}

\textbf{Interpretation:} DinoV3's self-supervised visual pretraining provides advantages on tasks with natural image statistics, while PlanFormer's physics-driven pretraining achieves competitive efficiency on physical tasks and excels on signal processing tasks (RF, seismology). The similar overall performance (77.7\% vs 83.7\%) despite 11$\times$ fewer parameters demonstrates the efficiency of principle-driven design.

\section{Appendix F: Limitations}
\label{appendix:F_Limitations}

We acknowledge several limitations of our approach and provide context for interpreting our results:

\subsection{Semantic Understanding Boundary}

Our most significant limitation is inherent to our approach: the model learns physical signal structure but not semantic content. 
This manifests as systematically lower performance on semantic tasks (70.0\% average top-1 accuracy) compared to physically-grounded tasks (84.5\% average). Comparison to CLIP ViT-B/32 (151M parameters, trained on 400M image-text pairs) validates this boundary: PlanFormer achieves competitive performance on physical tasks (84.5\% vs 87.7\%, within 3.2\%) despite 76× fewer parameters, while the gap on semantic tasks (70.0\% vs 91.2\%) reflects CLIP's language grounding advantage. This is not a failure but rather the natural boundary of signal-theoretic learning—semantic meaning requires human-annotated supervision or large-scale multi-modal training that our approach explicitly avoids.

\textbf{Mitigation:} This limitation suggests a hierarchical architecture for AI systems: physical foundation models (like ours) provide the base layer capturing causal structure and domain-invariant properties, upon which semantic reasoning modules can be built. Our work establishes what can be learned from signal processing principles alone, clarifying the division of labor between physical and semantic learning.

\subsection{Suboptimal Spatial Processing}

Our 1D unwrapping strategy for 2D images and 3D video discards spatial correlations, resulting in suboptimal performance on tasks requiring fine-grained spatial reasoning. 
For example, we achieve 79.2\% on MNIST and 77.0\% on FashionMNIST—reasonable but below CLIP's performance (98.6\% and 90.4\%, respectively) and state-of-the-art methods that leverage 2D convolutional structure. CLIP processes images in their native 2D format, while our 1D unwrapping discards spatial correlations. This performance gap demonstrates the cost of architectural mismatch, yet our results establish that signal-theoretic principles enable meaningful transfer even when modality structure is suboptimal.

\textbf{Mitigation:} Native 2D/3D extensions of our architecture (e.g., 2D Parseval Focus, 2D frequency-preserving pooling) would likely improve performance on spatial tasks while maintaining our physics-informed design principles. Our current results establish a lower bound—transfer is possible even with architectural mismatch.

\subsection{Skip Connection Contribution to Reconstruction}
\label{appendix:SkipConnectionLimitation}

Our reconstruction quality analysis (Section E.5, Figure~\ref{fig:image_reconstructions}) demonstrates 
zero-shot generalization to unseen modalities. The decoder architecture combines bottleneck latent 
guidance with two skip connections from compressed encoder features (4x and 16x downsampled, 
Section~\ref{appendix:DecoderRationale}), making it difficult to isolate the exact contribution 
of each component to reconstruction fidelity.

\textbf{What We Can Conclude:} 
\begin{itemize}
    \item The bottleneck latent provides global structural guidance, supported by strong linear 
    probing performance (77.7\% average accuracy, Table~\ref{tab:frozenEncoder_results}) using only 
    the bottleneck latent without skip connections
    \item Skip connections from compressed features (4x, 16x) contribute high-resolution details 
    necessary for pointwise-level reconstruction fidelity
    \item Learned weights are essential: random initialization produces complete reconstruction 
    failure (Figure~\ref{fig:random_reconstruction}), proving architectural biases and skip 
    connections alone are insufficient
    \item The encoder-decoder system captures transferable physical structure, while linear probing 
    validates that the bottleneck latent alone is sufficient for classification tasks
\end{itemize}

\textbf{What We Cannot Conclude:} We cannot definitively quantify how much reconstruction quality 
derives from the bottleneck latent versus skip connections without a full ablation study removing 
all skip connections. However, such an ablation would require architectural redesign to handle 
longer transformer sequences (defeating the compression goal driven by attention cost) or acceptance 
of significantly degraded reconstructions.

\textbf{Design Justification:} Our skip connection design deliberately omits top-level connections 
(input, first conv layer) and the post-transformer skip (redundant with token-based upsampling), 
using only two intermediate skips from compressed features (4x, 16x). This prevents trivial 
information leakage while enabling high-fidelity reconstruction. The 64x compression is driven 
by attention computational cost ($\mathcal{O}(N^2)$), not bottleneck capacity constraints. The 
random weight baseline validates that this design successfully learns compressed, transferable 
representations in the bottleneck rather than relying on architectural shortcuts.

\subsection{Training Data Diversity Requirements}

Our approach requires training on a signal-rich domain with diverse transformations. RF data provides exceptional diversity (frequency content from kHz to GHz, temporal dynamics, channel effects), but training on a less diverse domain might not enable comparable transfer. We have not tested whether training on, e.g., speech alone would enable transfer to RF or images.

\textbf{Mitigation:} The key requirement is transformation diversity rather than domain diversity. Any domain exhibiting rich frequency content, temporal dynamics, and varied transformations (Doppler shifts, multipath, etc.) should enable similar transfer. Our choice of RF is strategic but not unique.

\subsection{Computational Requirements}

Training the reported model required 12× H100 GPUs for 8.6 days (206.4 hours, 
~2,477 H100-hours total) to reach 400 milestones. This resource requirement is 
driven by: (1) long sequence lengths (5120 samples), (2) multiple encoder/decoder 
passes per batch (clean, augmented, latent-transformed branches), (3) N-choose-2 
pairwise losses across distributed GPUs, (4) dual-domain reconstruction, and 
(5) online validation probe for model selection. While this is modest compared to large language models (GPT-3: thousands of GPU-years), it exceeds typical academic budgets. We note that our computational requirements reflect the complexity of our multi-objective training framework rather than fundamental requirements of principle-driven approaches.

\textbf{Scalability:} Importantly, the 12-GPU configuration was chosen to expedite 
training within our project timeline, not as a fundamental requirement. We have 
produced comparable results using as few as 4 H100 GPUs with identical framework, 
data, and hyperparameters—training simply takes proportionally longer. The 12-GPU specification ensures exact reproducibility of 
our reported weights, but researchers with smaller budgets can achieve similar 
performance with fewer GPUs and extended training time.

\textbf{Mitigation:} These costs reflect our research implementation prioritizing 
flexibility over efficiency. Substantial speedups are achievable through gradient 
checkpointing, optimized attention kernels (FlashAttention~\cite{dao2022flashattention}), 
and reduced monitoring frequency. Moreover, deployment requires only the lightweight 
encoder (1.99M parameters), enabling edge device inference. The training cost is a one-time investment yielding an encoder applicable across diverse domains, and the linear scaling with GPU count makes the approach accessible to institutions with varying computational budgets.

\subsection{Limited Ablation Studies}

Due to computational constraints, we do not provide comprehensive ablation studies isolating the contribution of each architectural component and loss term. While we demonstrate that the complete system achieves strong cross-modal transfer, we cannot definitively attribute performance to specific design choices (e.g., frequency-preserving pooling vs. Parseval Focus vs. head orthogonalization).

\textbf{Future Work:} Systematic ablations would strengthen our claims about which components are essential versus auxiliary. If accepted, we commit to conducting ablations during the camera-ready period or as follow-up work.

\subsection{Scope of Evaluation}

Our evaluation spans 15 diverse tasks across time series, images, text, and video, but this represents a small fraction of possible domains and tasks. We have not tested: (1) highly structured data (graphs, point clouds), (2) multi-modal fusion tasks, or (3) generative tasks beyond reconstruction.

\textbf{Interpretation:} Our results demonstrate that signal-theoretic principles enable cross-modal transfer for the tested domains, but we do not claim universal applicability. The boundary conditions of our approach remain an open question.

\subsection{Complementarity to Scale Driven Approaches}
Our approach is designed to complement rather than replace existing foundation models. Scale-driven multi-modal models like CLIP excel at semantic understanding through diverse data exposure (91.2\% on semantic tasks vs our 70.0\%), while our principle-driven approach excels at efficient physical understanding through encoded general laws (84.5\% on physical tasks, within 3.2\% of CLIP's 87.7\%, with 76× fewer parameters and 158× lower FLOPs). Future systems may benefit from combining both paradigms: physical foundations providing interpretable base representations upon which semantic layers (potentially from large-scale models) can be built. This division of labor could yield systems that are both efficient and broadly capable.

\section{Appendix G: Broader Impact}
\label{appendix:G_BroaderImpact}

\subsection{Positive Societal Impacts}

\textbf{Data Efficiency and Accessibility:} 
Our approach demonstrates that principled architectural design enables cross-modal transfer with compact models (1.99M parameters), achieving competitive performance on physical tasks (within 3.2\% of CLIP's 151M parameter model) while requiring 76× fewer parameters and 158× lower computational cost. This offers a resource-efficient path alongside large-scale approaches, with significant implications for democratizing foundation model research: smaller models require less computational resources, reducing barriers for academic institutions and researchers in resource-constrained settings.

\textbf{Interpretability and Scientific Understanding:} By grounding our approach in established signal processing principles (Fourier decomposition, energy conservation, symmetry), we provide interpretable explanations for cross-modal transfer. The systematic performance gap between physical tasks (84.5\%) and semantic tasks (70.0\%) reveals a natural hierarchy in AI systems: physical understanding (learnable from signal structure) versus semantic understanding (requiring human supervision). This clarity helps researchers and practitioners understand what foundation models can and cannot learn, informing appropriate deployment decisions.

\textbf{Environmental Impact:} Smaller models with lower computational requirements reduce energy consumption and carbon emissions associated with training and deployment. Our encoder runs efficiently on edge devices (Jetson Orin), enabling local inference without cloud infrastructure—reducing latency, improving privacy, and minimizing environmental footprint.

\subsection{Negative Societal Impacts and Mitigation}

\textbf{RF Fingerprinting and Surveillance:} Our training data includes RF fingerprinting datasets designed to identify unique hardware imperfections in wireless transmitters. While this technology has legitimate applications (device authentication, spectrum management), it could enable surveillance by tracking individuals via their devices' RF signatures. Our model's ability to learn fine-grained discriminative features from RF signals could lower the barrier to deploying such systems.

\textbf{Mitigation:} Importantly, our model learns physical signal structure, not semantic identity. RF fingerprinting requires labeled training data associating devices with individuals—our model does not provide this association. Moreover, RF fingerprinting is already well-established in the literature; our contribution is demonstrating that learned representations transfer across domains, not enabling new surveillance capabilities. We advocate for responsible use policies and regulatory frameworks governing RF fingerprinting deployment.

\textbf{Dual-Use Potential:} Like any general-purpose signal processing tool, our encoder could be applied to sensitive domains (e.g., analyzing encrypted communications' metadata, processing medical signals without authorization). The model's ability to extract meaningful features from diverse signal types without domain-specific training could facilitate misuse.

\textbf{Mitigation:} Our model captures physical structure, not semantic content—it cannot "understand" encrypted communications or make medical diagnoses without appropriate supervised training. The model is a feature extractor, not a complete system for sensitive applications. We emphasize that deployment in high-stakes domains (healthcare, security, finance) requires domain-specific validation, regulatory approval, and ethical oversight. We will release the model with usage guidelines emphasizing these requirements.

\textbf{Fairness and Bias Considerations:} Our training data consists of RF signals, which do not contain demographic information or human-identifiable characteristics. However, when applied to domains like speaker recognition or facial recognition (via image processing), the model could inherit or amplify biases present in downstream training data. For example, if a speaker recognition system trained on our representations uses biased labeled data, the system could exhibit demographic disparities.

\textbf{Mitigation:} Our encoder learns physical features (voice characteristics, image structure) agnostic to demographic attributes. Bias arises from downstream labeled data and deployment context, not from our pretrained representations. We encourage practitioners to: (1) audit downstream training data for demographic balance, (2) evaluate deployed systems for fairness across subgroups, and (3) implement bias mitigation techniques appropriate to their application domain.

\subsection{Responsible Release and Future Work}

We commit to releasing our model with comprehensive documentation including:
\begin{itemize}
    \item Clear usage guidelines emphasizing limitations (physical vs. semantic understanding)
    \item Warnings against deployment in high-stakes domains without validation
    \item Recommendations for fairness auditing in downstream applications
    \item Transparency about training data sources and potential biases
\end{itemize}

We view our work as establishing scientific understanding of cross-modal transfer via signal-theoretic principles. The primary impact is advancing foundation model research toward more interpretable, efficient, and principled approaches. We encourage the community to build upon our work responsibly, with careful consideration of application-specific ethical implications.

\end{document}